\documentclass[10pt,twocolumn,letterpaper]{article}

\usepackage[pagenumbers]{cvpr} 
\usepackage{algorithm}
\usepackage{algorithmic}
\usepackage{multirow}
\usepackage{pgffor} 
\usepackage{graphicx}
\usepackage{xstring}
\usepackage{lineno}
\usepackage{wrapfig}

\usepackage{tikz}

\newcommand{\boldsubsection}[1]{\vspace{0.5em}\noindent\textbf{#1}}

\definecolor{cvprblue}{rgb}{0.21,0.49,0.74}
\usepackage[pagebackref,breaklinks,colorlinks,allcolors=cvprblue]{hyperref}

\def\paperID{xxxxx} 
\def\confName{CVPR}
\def\confYear{2026}

\title{Diffusion Mental Averages}
 
\author{Phonphrm Thawatdamrongkit \qquad
Sukit Seripanitkarn \qquad
Supasorn Suwajanakorn\vspace{5pt}\\
VISTEC, Thailand\\
}

\begin{document}

\twocolumn[{%
\renewcommand\twocolumn[1][]{#1}%
\maketitle

\begin{center}
      \centering
    \captionsetup{type=figure}
    \vspace{-2em}
    \includegraphics[width=0.979\textwidth]{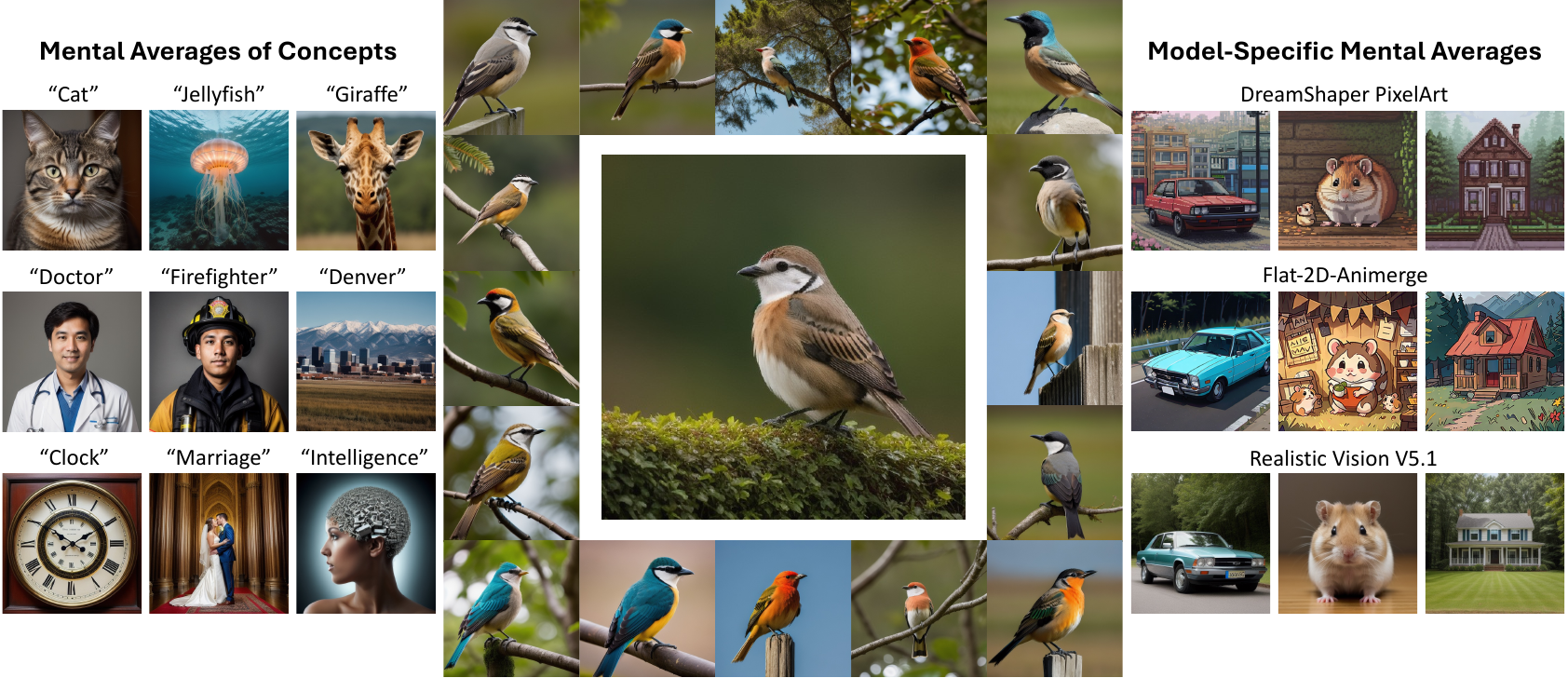}
    \vspace{-5pt}
    \captionof{figure}{
          Our method unveils the \emph{Mental Averages} encoded by pre-trained diffusion models across diverse concepts (left) and generalizes across different model variants (right), offering new tools for analyzing model biases and interpreting learned concept representations.
     } \label{fig:teaser}
\end{center}%
}]

\begin{abstract}
Can a diffusion model produce its own “mental average” of a concept—one that is as sharp and realistic as a typical sample? We introduce Diffusion Mental Averages (DMA), a model-centric answer to this question. While prior methods aim to average image collections, they produce blurry results when applied to diffusion samples from the same prompt.  These data-centric techniques operate outside the model, ignoring the generative process. In contrast, DMA averages within the diffusion model’s semantic space, as discovered by recent studies.  Since this space evolves across timesteps and lacks a direct decoder, we cast averaging as trajectory alignment: optimize multiple noise latents so their denoising trajectories progressively converge toward shared coarse-to-fine semantics, yielding a single sharp prototype. We extend our approach to multimodal concepts (e.g., dogs with many breeds) by clustering samples in semantically-rich spaces such as CLIP and applying Textual Inversion or LoRA to bridge CLIP clusters into diffusion space. This is, to our knowledge, the first approach that delivers consistent, realistic averages, even for abstract concepts, serving as a concrete visual summary and a lens into model biases and concept representation. Please visit our project page: \url{https://diffusion-mental-averages.github.io}

\end{abstract}    
\section{Introduction}
When most people imagine a bird, they picture something small and ordinary, perhaps a sparrow rather than a large ostrich or the exotic Nene. That mental prototype reflects our exposure, preferences, and cultural context. Diffusion models can generate countless variations of a single concept, yet one might ask whether they, too, hold a single ``mental average'' that defines what they consider typical. Visualizing such prototypes provides a tangible summary of a prompt, reveals biases in the model and its training data, and offers new ways to analyze or steer its generative behavior.


To create a concept prototype learned by diffusion models like StableDiffusion~\cite{rombach2022high}, we can begin by generating many samples of the same concept.
Prior work visually summarizes such collections by selecting representative samples according to some criteria~\cite{simon2007scene,crandall2009mapping,kennedy2008generating}, or by identifying discriminative patches~\cite{doersch2015makes,siglidis2024diffusion,lee2015linking,jae2013style}.
Data Distillation~\cite{su2024d,cazenavette2023generalizing,chan2025mgd} instead synthesizes a compact set of prototypes by distilling information from the entire image collection. 
Our goal, however, is to synthesize a single prototype that reconciles cross-sample variations, which selection-based methods cannot achieve. Unlike Data Distillation, which targets downstream objectives and often yields unnatural or inconsistent prototypes, we target concept summarization via prototypes that match the sharpness and realism of typical samples from the probe model.
A common way to synthesize such prototypes is by averaging pixels across \emph{spatially aligned} samples~\cite{zhu2014averageexplorer,peebles2022gan,luccioni2023stable, kemelmacher2014illumination,ginosar2015century}.
However, spatial alignment is rarely feasible beyond constrained domains like faces~\cite{deng2020retinaface,peebles2022gan, suwajanakorn2015makes} or animals~\cite{peebles2022gan}, and virtually impossible for abstract concepts like ``poverty.'' Even with perfect alignment, fine details tend to average out at the pixel level, yielding unrealistic appearance~\cite{luccioni2023stable,peebles2022gan}.
A promising direction is to perform averaging in a semantic space, where distances reflect conceptual similarity~\cite{karras2019style}. This can be done by encoding images into the semantic space, averaging their embeddings, then decoding---a straightforward process in autoencoders~\cite{baldi2012autoencoders,bank2023autoencoders}, VAEs~\cite{kingma2013auto}, or GANs~\cite{goodfellow2020generative,karras2019style} equipped with inversion mechanisms~\cite{zhu2016generative}.
Diffusion models, however, lack such an explicit bottleneck representation. Although some studies have identified ``semantic layers'' within diffusion models~\cite{kwon2022diffusion}, the semantic information is distributed across timesteps, with no direct way to decode a time-averaged embedding into a realistic average.

To address this, we recast averaging as trajectory alignment within the diffusion model. 
Instead of generating samples then averaging them post hoc, 
we jointly optimize multiple noise latents so that their denoising trajectories progressively converge toward shared coarse-to-fine semantics. Conceptually, the process first aligns high-level structures, such as global shape and layout, toward a semantic mean, then gradually refines alignment to local patterns, enabling spatial averaging of fine details without sacrificing realism. During optimization, each sample follows its own denoising trajectory while being constrained to match the mean semantic representation of all samples at each timestep.
The resulting trajectory forms a semantic consensus across samples, which can then be decoded into a single sharp and realistic prototype---effectively the model’s own ``mental average'' of the concept.

While averaging captures a concept's shared structure, many real-world concepts are visually multimodal or ambiguous: ``dog'' spans many breeds, while ``bike'' may mean a bicycle or a motorbike. In such cases, a single average may blur distinct modes and obscure diversity.
To capture this, we create mode-specific prototypes by clustering samples in a semantic space and averaging within each cluster.
We find that CLIP~\cite{radford2021learning} serves as a stable and semantically rich space, leading to clearer mode separation than diffusion-based feature spaces. Clustering can also be grounded using models like BLIP~\cite{li2022blip}, enabling user-defined partitioning (e.g., by ethnicity for human faces). However, the latent space used for clustering (e.g., CLIP) differs from the one used for averaging within the diffusion model, causing inconsistencies between clusters and their prototypes. To bridge this gap, we adapt the conditioning of the diffusion model to be cluster-specific: for each cluster, we learn a textual inversion embedding~\cite{gal2022image} or LoRA~\cite{hu2022lora} that captures its semantic subspace and guides the denoising process toward that region, yielding more faithful averages.

We evaluate our method for generating concept averages across diverse categories, including animals, humans, objects, and abstract concepts, and compare it against several baselines, including GANgealing~\cite{peebles2022gan} and adapted dataset distillation methods such as D$^4$M~\cite{su2024d} and MGD$^3$~\cite{chan2025mgd}. Our method produces results that are both visually realistic and consistent across different initial random seeds, whereas the baselines tend to be either consistent but blurry or realistic but inconsistent. We further demonstrate that our mode discovery method can effectively generate averages for multiple modes in both unsupervised and grounded setups, and that our approach generalizes to other diffusion architectures, such as DiT~\cite{peebles2023scalable}. In summary, our contributions are:
\begin{itemize}
  \item 
  We propose the first method to produce consistent and realistic averages directly from pre-trained diffusion models by formulating the task as trajectory alignment.
  \item We demonstrate that DMA generalizes to multimodal concepts with light-weight model conditioning, as well as across diffusion model variants and architectures.
\end{itemize}

\section{Related Work}
\boldsubsection{Prototype image.}
A prototype image aims to represent the visual essence of a concept or dataset. Early approaches decompose images into patches, group them by visual similarity, and use these clusters as prototypical elements~\cite{doersch2015makes,lee2015linking,siglidis2024diffusion,jae2013style,singh2012unsuperviseddiscoverymidleveldiscriminative,doersch2013mid,rematas2015dataset,li2017mining,goel2017visualhashtags}. While effective for analyzing local features, patch-based prototypes fail to capture holistic relationships between visual components. Other works instead select entire images to serve as prototypes~\cite{simon2007scene,crandall2009mapping,kennedy2008generating,camargo2016multimodal}. However, this approach is limited by the finite sample set, potentially overlooking variations or information contained in other samples.

Beyond selecting existing samples, another line of work focuses on synthesizing prototype images. Dataset distillation methods~\cite{zhao2020DC,lee2022DCC,zhao2021siamese,wang2022cafe,zhao2023dm,cazenavette2022dataset,cui2023scaling,yin2023sre2l,cazenavette2023generalizing} generate synthetic images that encapsulate the knowledge of an entire class, but their prototypes often lack recognizable structure and visual coherence. Recent methods in dataset distillation employ diffusion models to distill class-level representations into a set of high-quality prototypes~\cite{su2024d,chan2025mgd,gu2024minimax}. However, they primarily target downstream tasks over conceptual summarization and can exhibit inconsistent generations. Pixel-space averaging on spatially aligned images provides an intuitive way to summarize visual information~\cite{zhu2014averageexplorer,luccioni2023stable,ginosar2015century}. Although alignment methods~\cite{peebles2022gan,barel2024spacejam,miller2000learning,ofri2023neural,learned2005data} preserve the overall structure of the average image, fine details are averaged out and blurred because pixel-space averaging operates on low-level image features, such as texture and illumination, which vary significantly across samples. A more effective strategy is to perform averaging in a semantic latent space, where high-level concept information is encoded. This can be done by encoding samples into the latent space, computing their mean representation, and decoding it back into the image domain~\cite{karras2019style,goodfellow2020generative,baldi2012autoencoders,bank2023autoencoders,kingma2013auto}. However, this remains unexplored in the context of diffusion models, which lack an explicit semantic latent space and a dedicated decoder for reconstruction.



Building on this idea, we aim to construct high-quality average images of any concept by leveraging the semantic space of pre-trained diffusion models, without requiring access to the underlying dataset. Closest to our work, Feng et al.~\cite{feng2025gps} use compositional generation~\cite{liu2022compositional} in a diffusion model fine-tuned with GPS signals to produce average representations for specific geographic areas, relying on multiple location-based conditions. In contrast, we operate in text-to-image diffusion models, where such multi-conditional inputs (e.g., GPS coordinates) are impractical, as each concept is defined solely by a single text prompt.

\begin{figure*}[ht]
    \centering
    \includegraphics[width=0.95\linewidth]{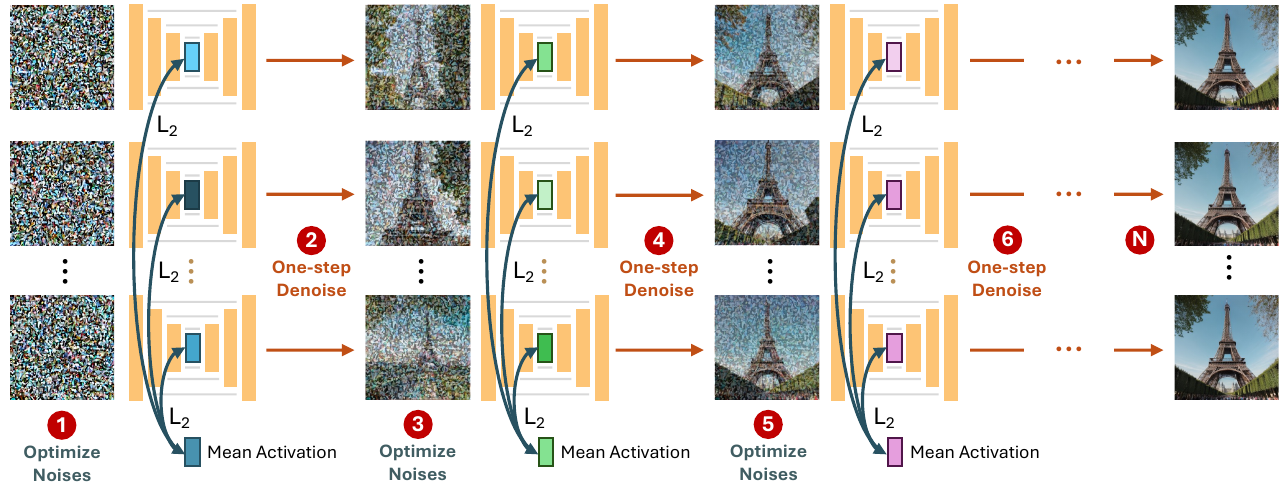}
    \caption{\textbf{Overview.} Multiple noise latents are jointly optimized so that their denoising trajectories converge toward shared semantics. At each timestep $t$, their $h$-space activations are averaged to form a semantic target, and each latent is optimized to match it before denoising to the next step. Repeating this process across timesteps aligns coarse-to-fine semantics, yielding a single ``mental average'' of the concept.}
    \label{fig:pipeline}
    \vspace{-8pt}
\end{figure*}

\boldsubsection{Semantic space in diffusion models.}
Our method leverages the semantic spaces of UNet–based diffusion models to generate average images. Some studies have explored these spaces for tasks such as segmentation~\cite{meng2024not,baranchuk2021label,tian2024diffuse,li2023open,ma2023diffusionseg,pnvr2023ld,wu2023diffumask,namekata2024emerdiff,samuel2024s}, visual correspondence~\cite{tang2023emergent,meng2024not,luo2023diffusion,zhang2023tale}, and concept ablation~\cite{basu2023localizing,basu2024mechanistic}. However, these semantic spaces typically retain spatial information, suitable for their tasks but suboptimal for ours, as they can introduce the same blurring issues seen in pixel-space averaging. Other works attempt to create high-level semantic spaces, as in autoencoders or VAEs~\cite{kingma2013auto}, by using an encoder to compress images into semantic latent vectors that guide the denoising process~\cite{wang2023infodiffusion,preechakul2022diffusion,leng2023diffusegae,hudson2024soda}. However, this requires retraining the entire network and is unsuitable for our goal of finding the mental average within a pre-trained diffusion model. Kwon et al.~\cite{kwon2022diffusion} show that the U-Net bottleneck layer ($h$-space) exhibits properties, such as linearity, that make it an effective semantic representation for our task. 
Semantic information in diffusion models is also found to spread across timesteps, with early steps capturing coarse structure and later steps encoding fine details~\cite{park2023understanding,sclocchi2025phase,choi2022perception,daras2022multiresolution}. We leverage the $h$-space for its low spatial specificity and rich semantic content.


\boldsubsection{Noise optimization in diffusion models.}
Once the average semantic latent is obtained, decoding it into an average image is non-trivial. A naive approach is to directly substitute the $h$-space latent with the average during denoising, but this often produces inconsistent outputs because skip connections allow unconstrained information from other layers to influence the output. To constrain activations in other layers without modifying model parameters, noise optimization has proven effective. Prior works have employed this approach for enhancing image quality~\cite{eyring2024reno,guo2024initno,qi2024not,kim2025diverse}, generating rare concepts~\cite{samuel2024generating}, and motion editing~\cite{karunratanakul2024optimizing,ota2025pino,liu2024programmable}. 

We adopt noise optimization to decode the average semantic latent into an average image by applying it sequentially across diffusion timesteps.

\newcommand{\zb}{\mathbf{z}}

\section{Background on T2I Diffusion Models}
 Modern text-to-image (T2I) diffusion models~\cite{song2020denoising,ho2020denoising} can be viewed as discrete-time instances within the broader family of score-based~\cite{song2020score} and flow-matching~\cite{lipman2022flow} generative models, which learn to transform Gaussian noise into data through a reverse stochastic or deterministic process. 
These models generate high-resolution images efficiently by operating in a latent space rather than directly in pixel space~\cite{rombach2022high}.
A Variational Autoencoder (VAE)~\cite{kingma2013auto} first maps images into a compact latent space, where the diffusion process is applied.

For text-conditioned generation, a prompt $c$ is encoded by a transformer-based text encoder to guide the denoising trajectory. At inference, generation begins from a Gaussian latent $\mathbf{z}_T \sim \mathcal{N}(0, I)$ and progressively denoises it into a clean latent $\mathbf{z}_0$, which is decoded by the VAE decoder into the final image. At each timestep $t$, the model predicts the noise $\epsilon_t=\epsilon_\theta(\mathbf{z}_t,\, t,\, c)$ and the next latent $\mathbf{z}_{t-1}$ is computed using, e.g., the DDIM update~\cite{song2020denoising}:
\begin{equation}
\mathbf{z}_{t-1}
=
\sqrt{\alpha_{t-1}}
\left(
\frac{\mathbf{z}_t - \sqrt{1-\alpha_t}\,\epsilon_t}{\sqrt{\alpha_t}}
\right)
+
\sqrt{1-\alpha_{t-1}}\,\epsilon_t.
\label{eq:ddim}
\end{equation}

To improve image-prompt fidelity, Ho and Salimans~\cite{ho2022classifier} introduced classifier-free guidance (CFG), which blends conditional and unconditional predictions:
\begin{equation}
\tilde{\epsilon}_\theta(\zb_t, t, c)
=
(1 - w)\,\epsilon_\theta(\zb_t, t, c)
-
w\,\epsilon_\theta(\zb_t, t),
\label{eq:cfg_factorized}
\end{equation}
where $w$ is the guidance scale controlling the trade-off between text faithfulness and sample diversity.

\section{Method}

Given a concept prompt, we aim to synthesize an average image that captures the concept’s shared semantics under a probe diffusion model while preserving visual realism.

Averaging across diffusion samples is conceptually simple yet technically challenging: pixel-space averaging destroys realism, and feature-space averaging is ill-defined because diffusion models lack a semantic bottleneck and decoder. Semantic information is instead distributed along the denoising trajectory, evolving from coarse layout to fine detail. This motivates performing averaging not in image or latent space confined to a single layer, but along the model's denoising process itself. We therefore recast the problem as aligning multiple denoising trajectories so that they converge toward a shared semantic consensus. Figure~\ref{fig:pipeline} provides an overview.

To achieve this, we leverage a semantically meaningful latent layer, referred to as the $h$-space, across \emph{multiple} diffusion timesteps. This layer, located near the middle of the U-Net denoiser, has been shown to encode interpretable and approximately linear semantics~\cite{li2024self,kwon2022diffusion,shi2025dissecting,parihar2024balancing}. Rather than collapsing $h$-space representations from all timesteps into a single latent vector, which would be non-decodable, we progressively align the $h$-space representations of multiple noise latents to their means along the denoising process. This mirrors the diffusion model’s inherent generation from high-level to low-level attributes, enabling the model to first reach agreement on global structure, such as pose and composition, before successively aligning lower-level attributes.

Specifically, we start with $K$ noisy latents $\{ \zb_k^{(0)} \}_{k=1}^{K}$, each drawn from $\mathcal{N}(0, \mathbf{I})$. At each diffusion timestep $t$, we perform the following steps:

\textbf{1. Compute the average $h$-space activation.}
For each noisy latent $\zb^{(t)}_k$, we feed it into the diffusion model conditioned on timestep $t$ and the prompt to extract its $h$-space activation $H(\zb^{(t)}_k)$, and compute the average activation across all $K$ latents, denoted as $\bar{A}^{(t)} = \frac{1}{K}\sum_{k=1}^K H(\zb^{(t)}_k)$.


\textbf{2. Noise optimization.}
To align each noisy latent with the average activation $\bar{A}^{(t)}$, we optimize each $\zb^{(t)}_i$ using the following objective:
\begin{equation}
\min_{\zb^{(t)}_i} \left\| H(\zb^{(t)}_i) - \bar{A}^{(t)} \right\|_2^2.
\end{equation}
Minimizing this loss gradually shifts each latent toward the shared semantic feature at timestep $t$. The optimization is performed for $N$ iterations for each latent.

\textbf{3. Denoising.}
After aligning the noisy latents with the mean semantic features, each optimized latent is denoised to the next timestep using DDIM sampling~\cite{song2020denoising}.

This process is repeated across diffusion timesteps until a cutoff $t_\text{stop} < T$, which helps reduce computation cost. Then, we continue denoising with standard DDIM sampling until completion and decode any final latent $\zb^{(T)}_k$ with the VAE decoder to produce the prototype image, as outlined in Algorithm~\ref{alg:psa}. We examine how different noise latents converge under varying cutoff values $t_\text{stop}$ in Appendix~\ref{sec:c4_tstop}.

\begin{algorithm}[t]
\caption{Diffusion Mental Averaging}
\label{alg:psa}
\begin{algorithmic}[1]
\REQUIRE
    Diffusion model $\mathcal{D}$, 
    concept prompt $c$,
    optimization cutoff $t_{\text{stop}}$,
    diffusion timestep schedule $\{\tau_t\}_{t=1}^{T}$,
    number of sampled latents $K$,
    number of optimization steps $N$,
    learning rate $\eta$,
    $h$-space activation $H(\cdot)$
\ENSURE
Prototype Image $I$

\STATE Initialize a set of $K$ noisy latents
$\mathbf{Z}^{(0)} = \{ \zb_k^{(0)} \}_{k=1}^{K}$ with $\zb_k^{(0)} \sim \mathcal{N}(0, \mathbf{I})$
\vspace{0.3em}

\FOR{$t = 0$ \TO $T-1$}
    \IF{$t \le t_{\text{stop}}$}
        \STATE $\displaystyle
        \bar{A}_t \leftarrow \frac{1}{K} \sum_{\zb \in \mathbf{Z}^{(t)}}
        H\!\big(\zb; \mathcal{D}, \tau_t, c\big)$
        \FOR{$k = 1$ \TO $K$}
            \FOR{$i = 1$ \TO $N$}
                \STATE $\mathcal{L} \leftarrow \left\| H\left(\zb^{(t)}_k; \mathcal{D}, \tau_t, c\right)- \bar{A}_t \right\|_2^2$
                \STATE Update $\zb_k^{(t)}$ using
                $\text{Adam}(\nabla_{\zb^{(t)}_k} \mathcal{L}, \eta)$
            \ENDFOR
        \ENDFOR

    \ENDIF
    \STATE $\mathbf{Z}^{(t+1)} \leftarrow
    \text{DDIM-Sampling}\big(\mathbf{Z}^{(t)}; \mathcal{D}, \tau_t, c\big)$

\ENDFOR

\STATE $I \leftarrow \text{VAE-Decode}(\zb^{(T)}_k)$ for any $k$

\STATE \textbf{Return:} Prototype Image $I$

\end{algorithmic}
\end{algorithm}

\subsection{Mode Discovery}\label{sec:mode_discov}
While our method aligns multiple trajectories toward a single semantic consensus, this process may oversimplify concepts that the diffusion model represents through several distinct semantic modes. For example, ``crane'' can refer to a bird or a construction machine---two modes that are difficult to meaningfully blend into a single prototype.

We extend our method with a simple mode separation step: we first cluster the noise latents and then compute an average image per cluster.
Effective mode discovery requires a feature space that captures high-level semantics.
Although the internal $h$-space encodes semantic content, its representations are stochastic and vary across timesteps, making it unreliable for consistent mode separation.
Instead, we use the CLIP embedding space~\cite{radford2021learning}, which offers a stable, high-level semantic representation~\cite{shen2023clip,sammani2024interpreting,bhalla2024interpreting,gandelsman2023interpreting,barraco2022unreasonable}. 
To support user-specified clustering (e.g., by ethnicity for ``doctor''), we alternatively employ a grounded approach using BLIP-VQA~\cite{li2022blip} to obtain attribute-focused embeddings~\cite{luccioni2023stable} for targeted mode discovery. 
Noise latents are clustered by generating samples through standard denoising, embedding them with CLIP or BLIP-VQA, and applying a clustering method such as GMM; cluster assignments are then mapped back to the latents.

After clustering, a straightforward approach is to apply DMA independently within each cluster. 
However, we find that this often yields prototypes that do not faithfully reflect their cluster's semantics.
We hypothesize that this inconsistency stems from a mismatch between the semantic space used for clustering and the diffusion model’s $h$-space. 
Since these spaces encode semantics differently, samples that form a coherent mode in one space may scatter across the other, causing the average to lose semantic coherence.

To mitigate this, we adjust the model’s conditioning to steer denoising toward each cluster's semantic subregion. 
This can be done using techniques such as Textual Inversion~\cite{gal2022image}, which learns a compact textual embedding from a set of images, or by training a LoRA~\cite{hu2022lora} to fine-tune the model’s conditioning pathway.
While Textual Inversion can handle simple or well-defined modes, we find that its limited capacity struggles with complex modes~\cite{zhang2024compositional}.
LoRA, being more expressive, captures richer intra-mode variation and provides better alignment between semantic modes and their corresponding prototypes.
We compare the two approaches in our experiments. 
With these lightweight conditioning methods, our approach is able to summarize each concept's mode with greater semantic fidelity.




\section{Experiments}

\boldsubsection{Implementation details.}
Unless stated otherwise, we use a variant of Stable Diffusion called Realistic Vision v5.1~\cite{realisticvision2023} due to its higher image quality.
All experiments use classifier-free guidance (CFG) scale of 7.0 and 20 DDIM sampling steps.
We set the number of latents $K=1000$. At each diffusion timestep, each latent is optimized using Adam~\cite{kingma2014adam} with a learning rate of $2\times10^{-2}$ for 300 iterations. We set the cutoff $t_{\text{stop}} = 10$. The optimization process took about 10 hours on an NVIDIA RTX 4080. Additional details are provided in Appendix~\hyperlink{my:a_implement_details}{\ref*{sec:a_implement_details}}.


   
\subsection{Baselines}
As no method currently exists for averaging diffusion models, we compare DMA to adapted versions of two related approaches: congealing and dataset distillation.
All methods use the same 1,000 noise latents or their samples.

\vspace{-0.2em}
\boldsubsection{Average VAE.}
Starting from noisy latents, we denoise them, compute their mean in VAE latent space, and decode it once into pixel space.

\vspace{-0.2em}
\boldsubsection{GANgealing~\cite{peebles2022gan}.}
We generate samples from the same noise latents and use GANgealing’s pretrained Spatial Transformer Network to align the images. The aligned images are then averaged in pixel space.

\vspace{-0.2em}
\boldsubsection{D$^4$M~\cite{su2024d}}
uses cluster centers in VAE space as prototypes and refines them by adding noise and denoising through a diffusion model.
In our adaptation, we set IPC = 1, so the prototype becomes the average of all clean VAE latents.
We inject noise into this averaged latent at timestep $t = 6$ (equivalent to the default SDEdit strength 0.7) and denoise once to obtain the prototype image.

\vspace{-0.2em}
\boldsubsection{MGD$^3$~\cite{chan2025mgd}} also uses cluster centers as prototypes but introduces Mode Guidance to steer predicted noise toward the prototype.
In our adaptation, we also set IPC = 1, and apply Mode Guidance for the first 10 steps (default 50\% of total steps) with the default mode guidance of 0.1, followed by standard denoising to obtain the prototype image. See our Appendix~\hyperlink{my:d_hyperparams}{\ref*{sec:d_hyperparams}} for results using other hyperparameter values.

\subsection{Prototypical Image Evaluation}
\begin{figure*}[t]
    \centering
    \setlength{\tabcolsep}{2pt}
    \renewcommand{\arraystretch}{1.5}
    \begin{tabular}{@{}c|ccccc@{}}
        \textbf{Samples} & 
        \textbf{GANgealing}~\cite{peebles2022gan} &
        \textbf{Avg VAE} &
        \textbf{D$^4$M}~\cite{su2024d} & 
        \textbf{MGD$^3$}~\cite{chan2025mgd} & 
        \textbf{Ours} \\[0em]
        
        \includegraphics[width=0.16\linewidth]{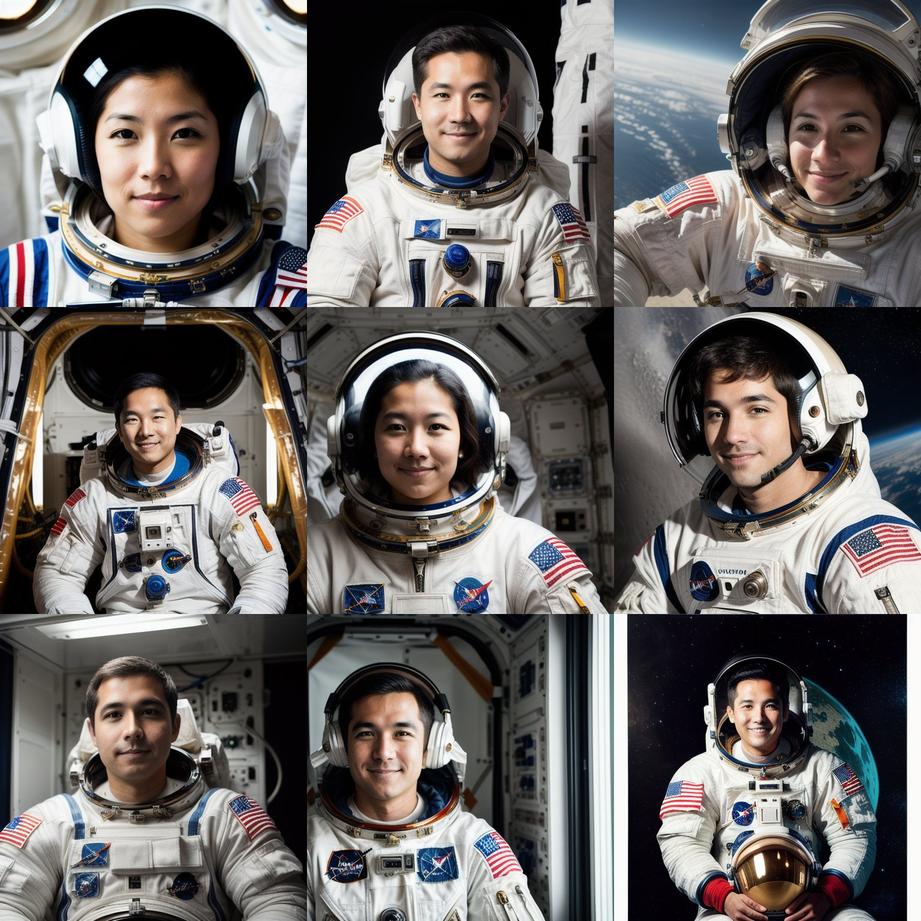} &
        \includegraphics[width=0.16\linewidth]{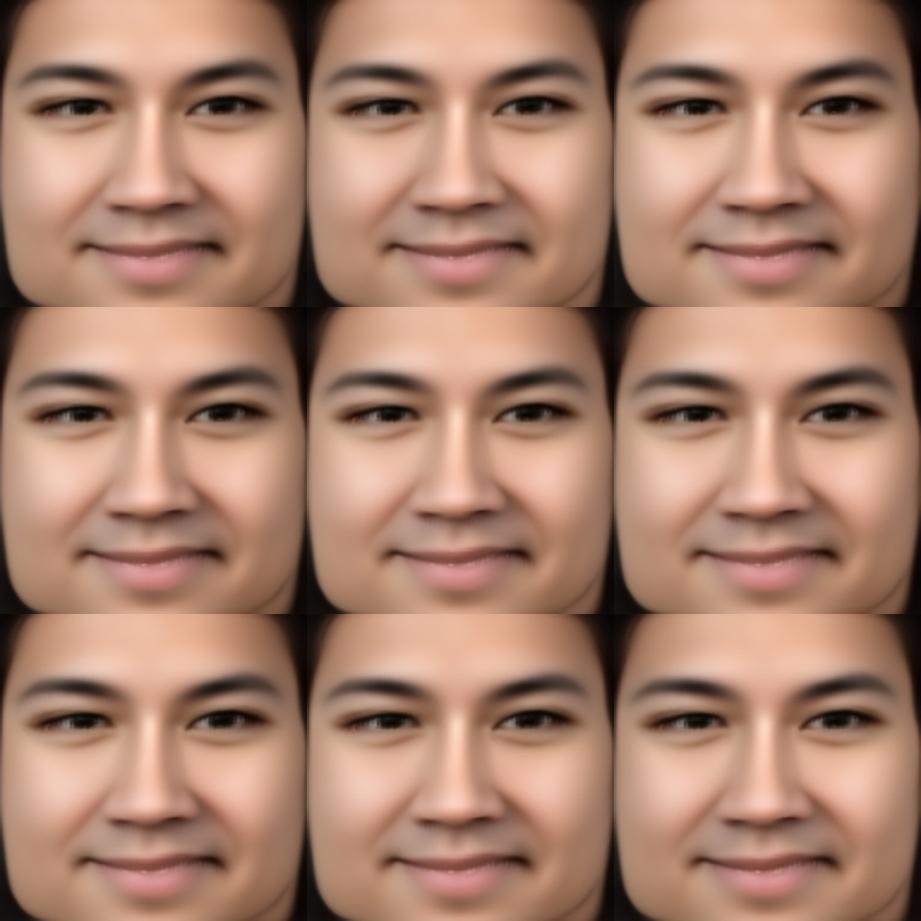} &
        \includegraphics[width=0.16\linewidth]{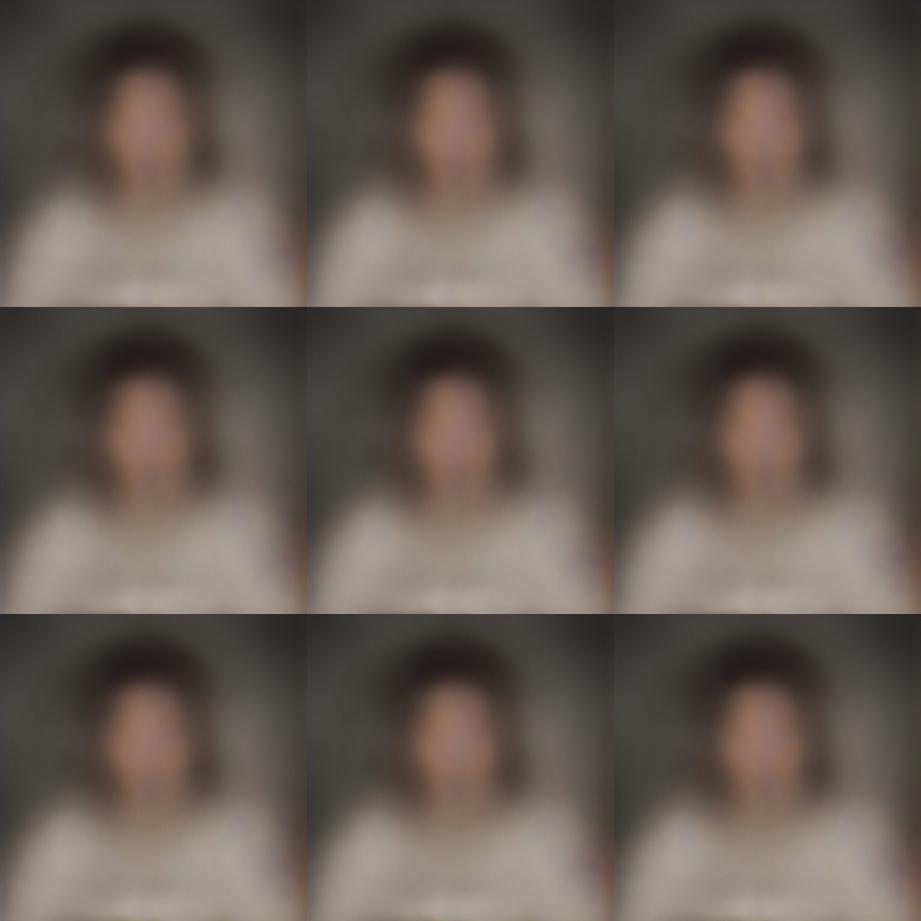} &
        \includegraphics[width=0.16\linewidth]{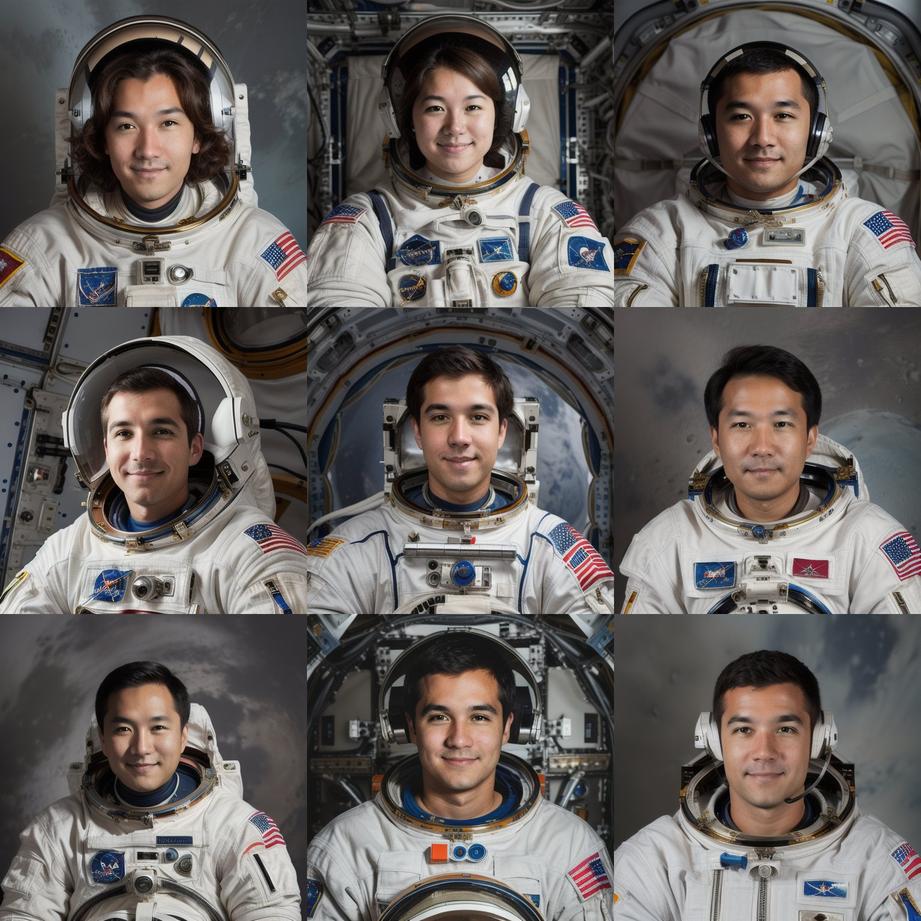} &
        \includegraphics[width=0.16\linewidth]{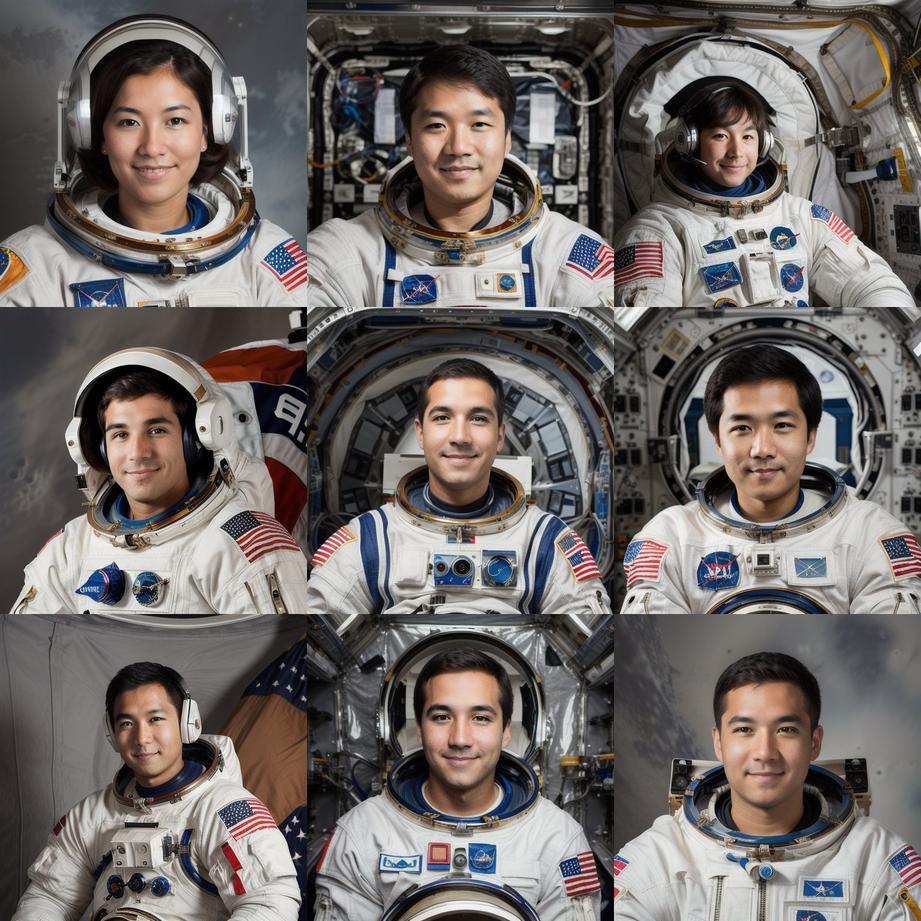} &
        \includegraphics[width=0.16\linewidth]{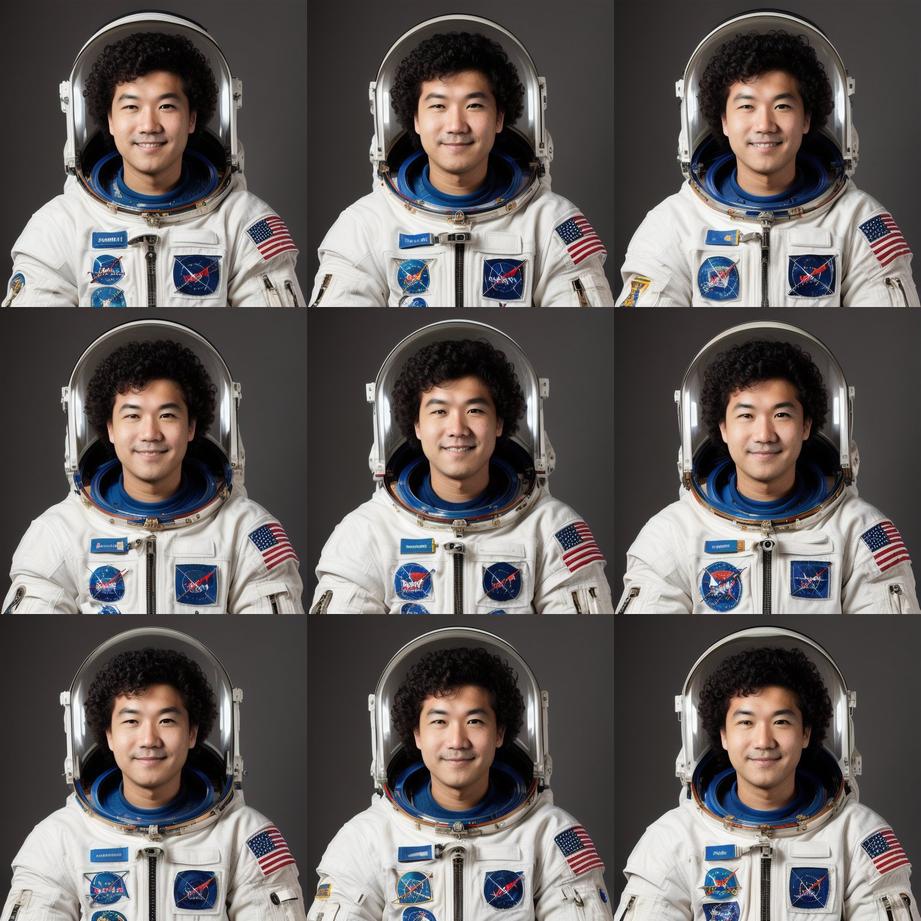} \\[-0.1em]
        
        \includegraphics[width=0.16\linewidth]{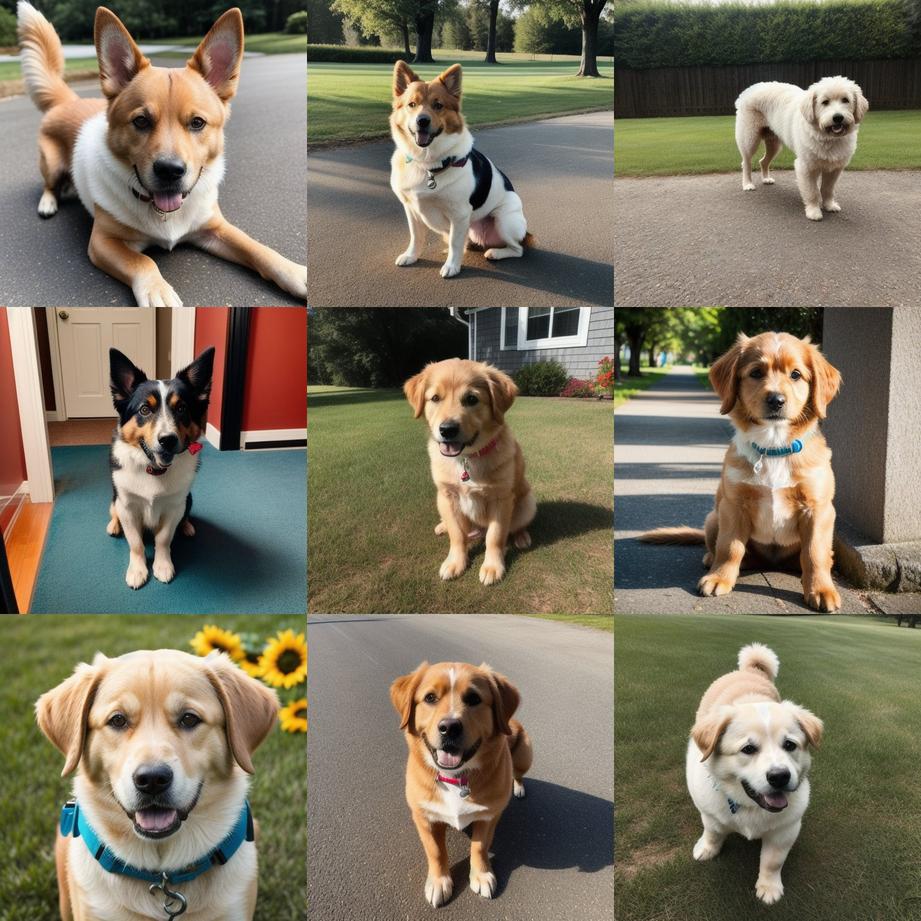} &
        \includegraphics[width=0.16\linewidth]{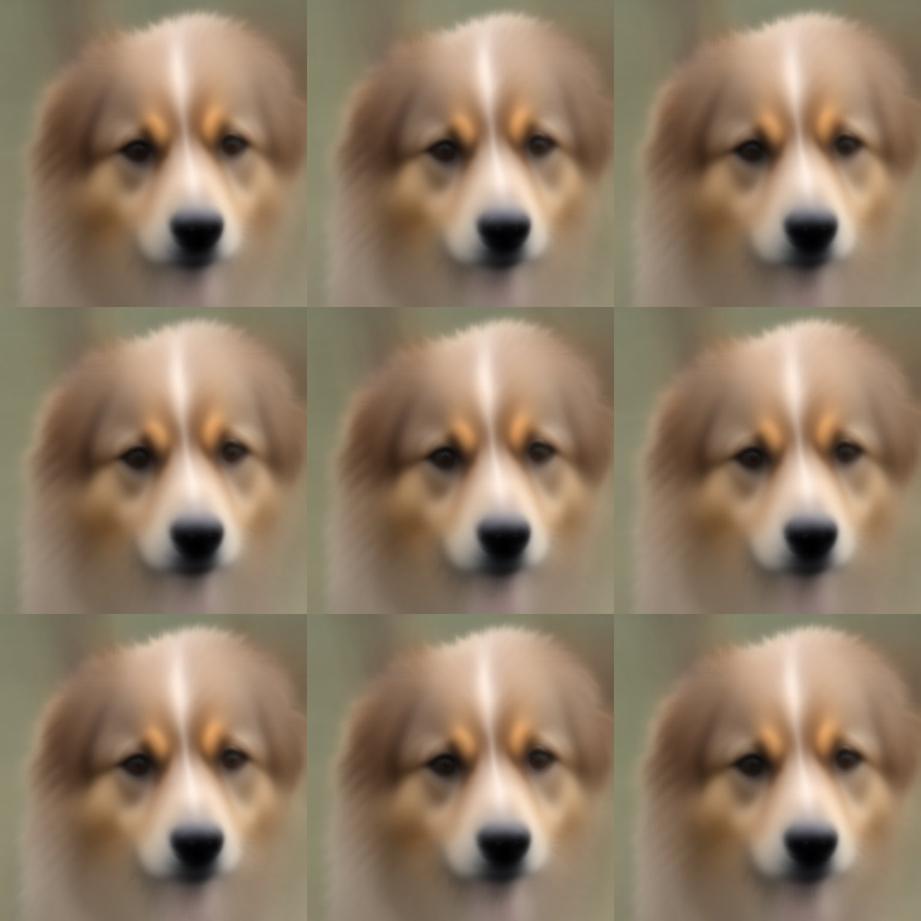} &
        \includegraphics[width=0.16\linewidth]{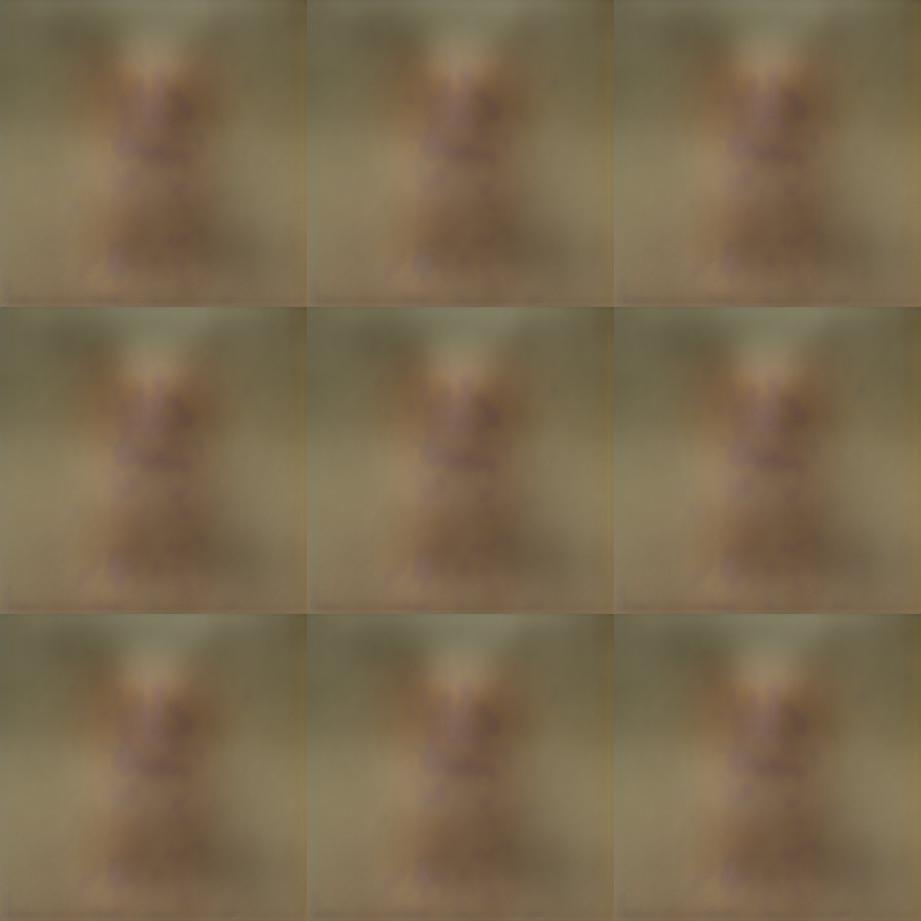} &
        \includegraphics[width=0.16\linewidth]{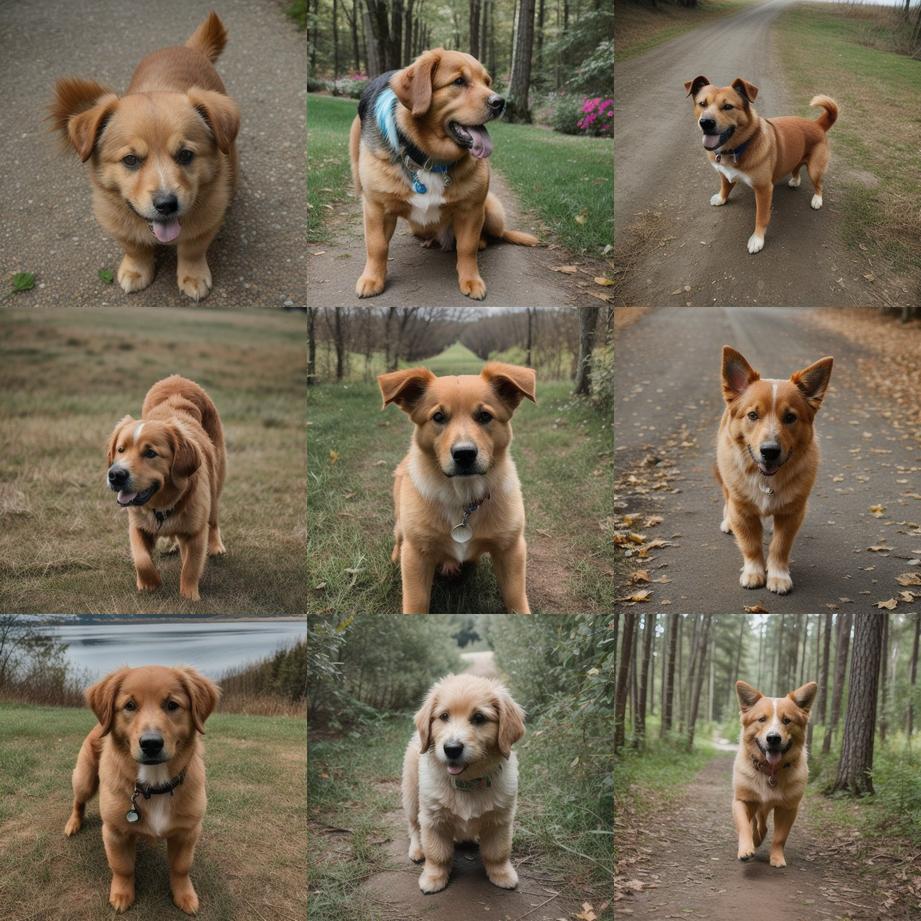} &
        \includegraphics[width=0.16\linewidth]{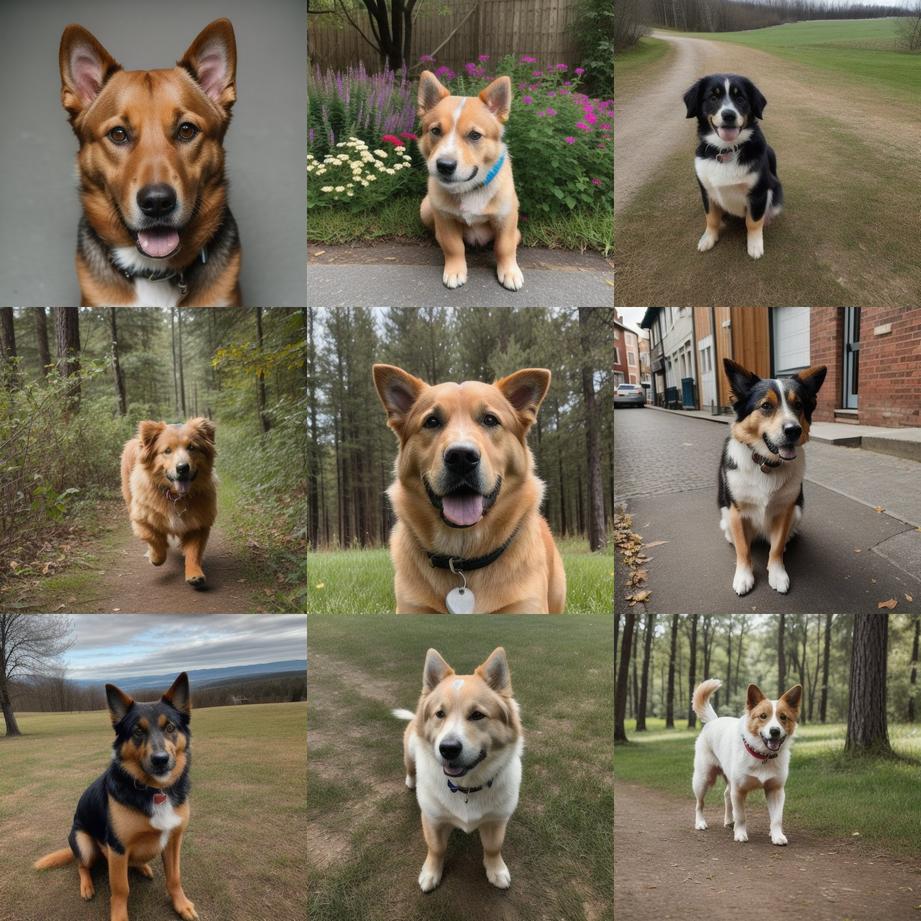} &
        \includegraphics[width=0.16\linewidth]{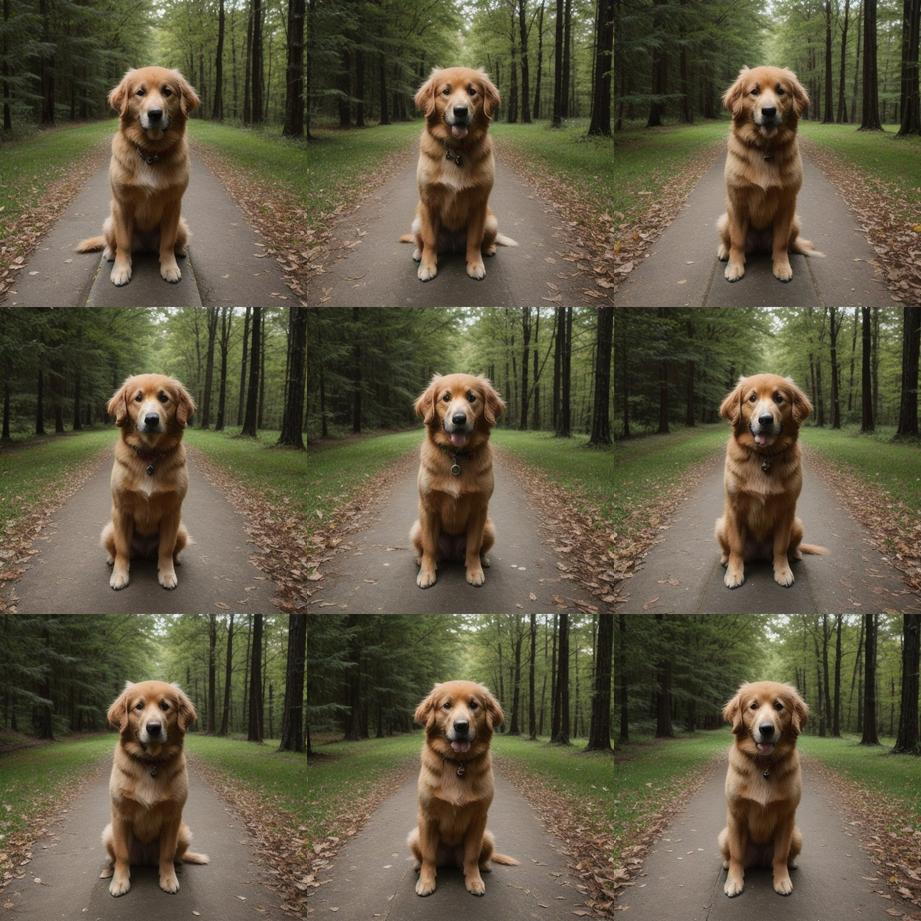} \\[-0.1em]
        
        \includegraphics[width=0.16\linewidth]{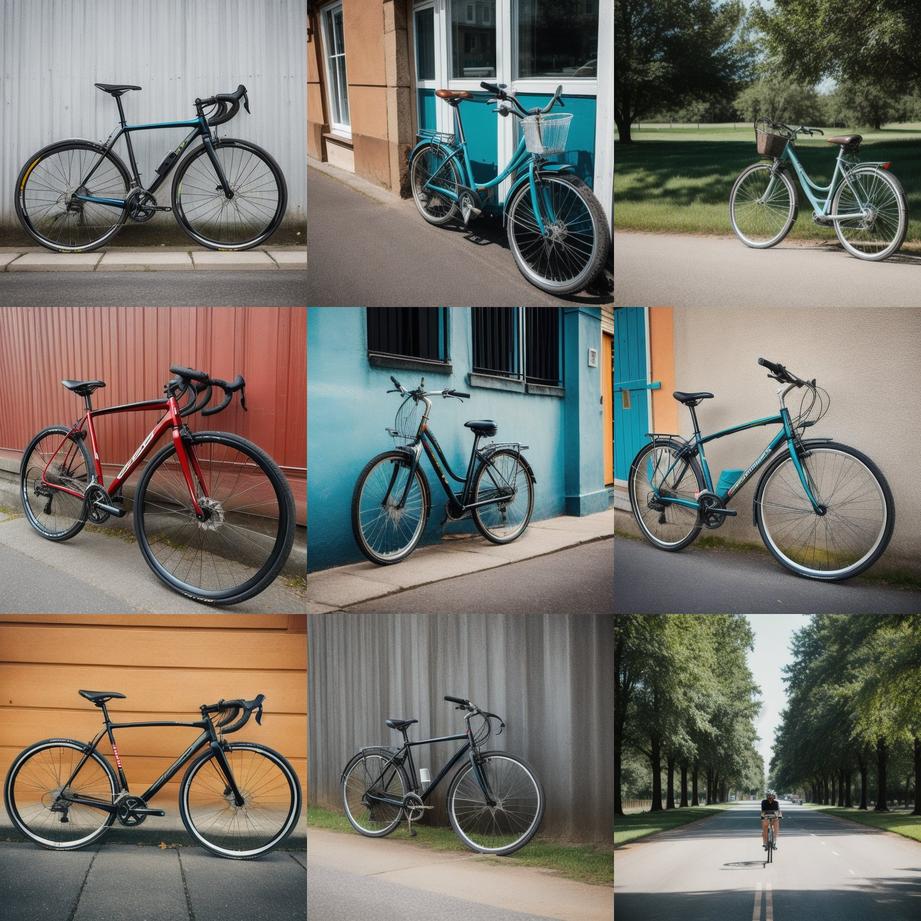} &
        \includegraphics[width=0.16\linewidth]{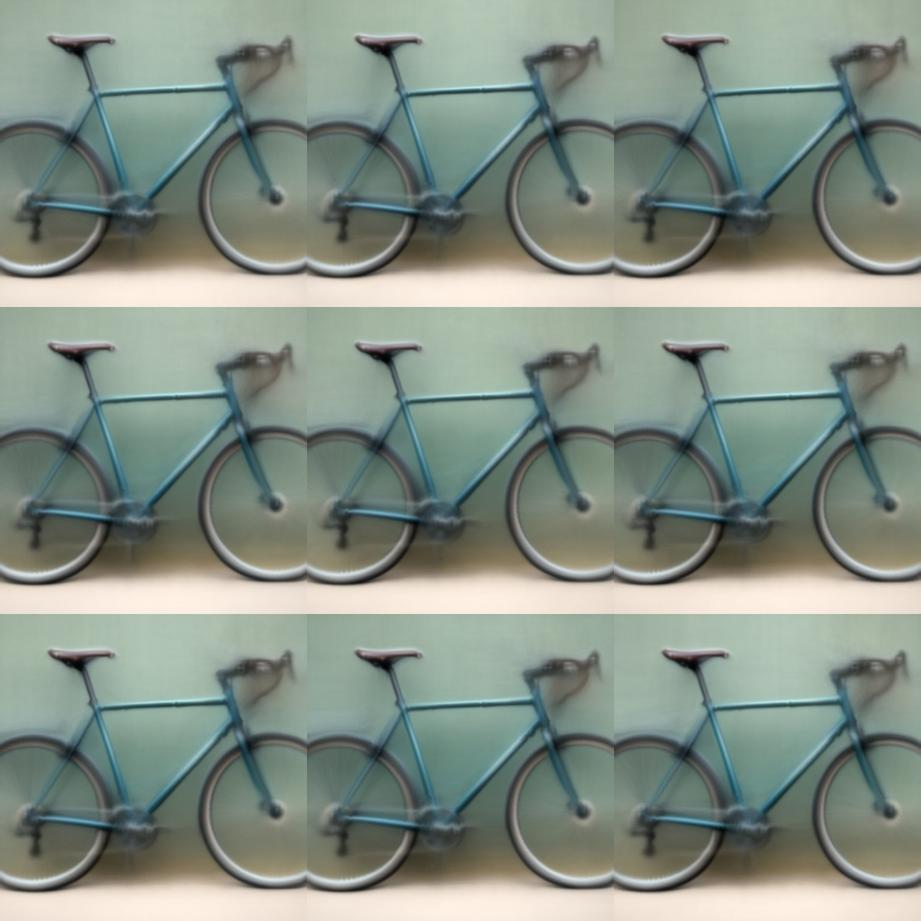} &
        \includegraphics[width=0.16\linewidth]{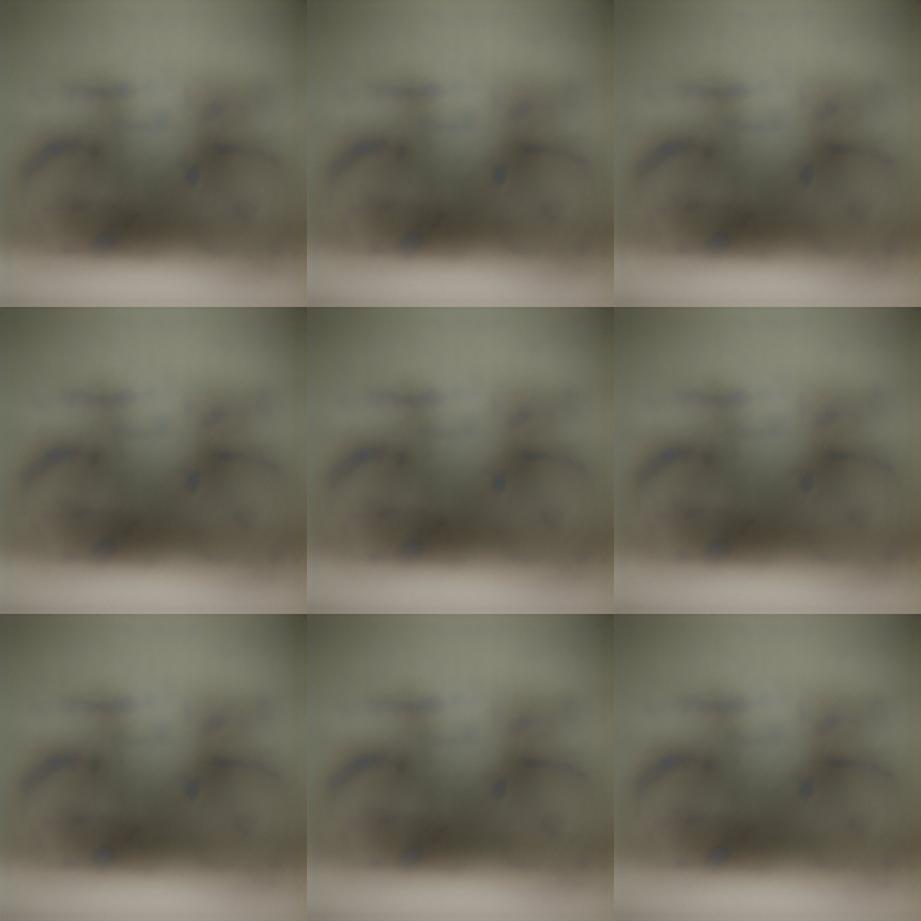} &
        \includegraphics[width=0.16\linewidth]{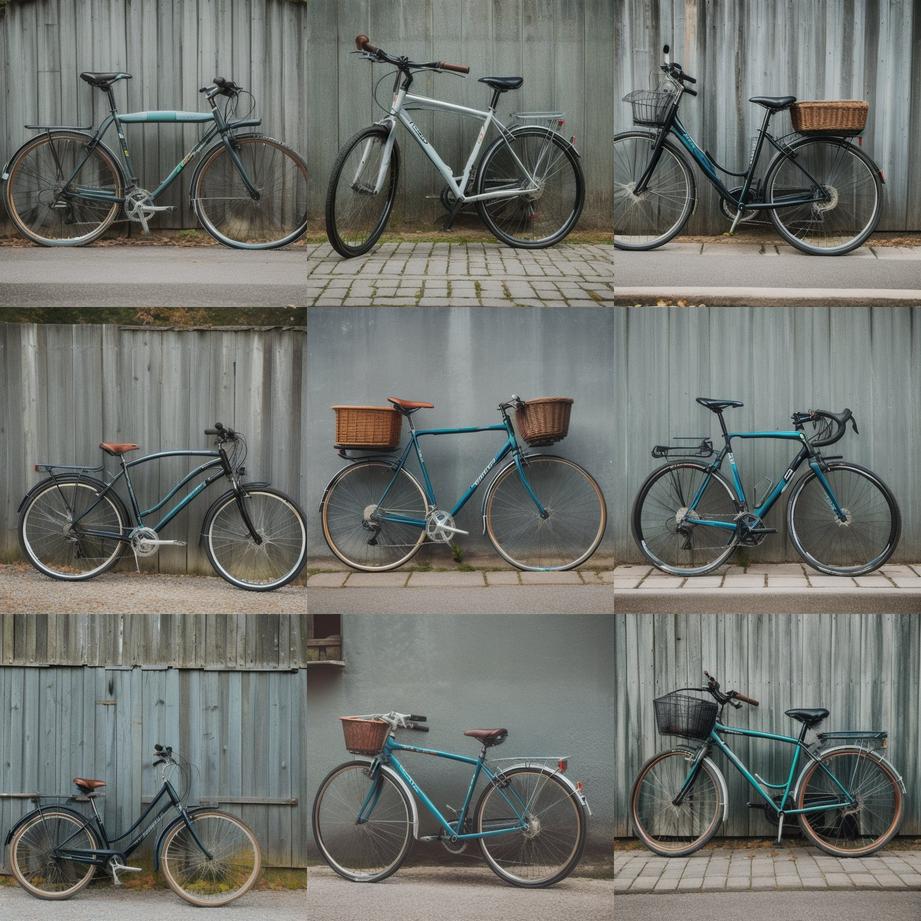} &
        \includegraphics[width=0.16\linewidth]{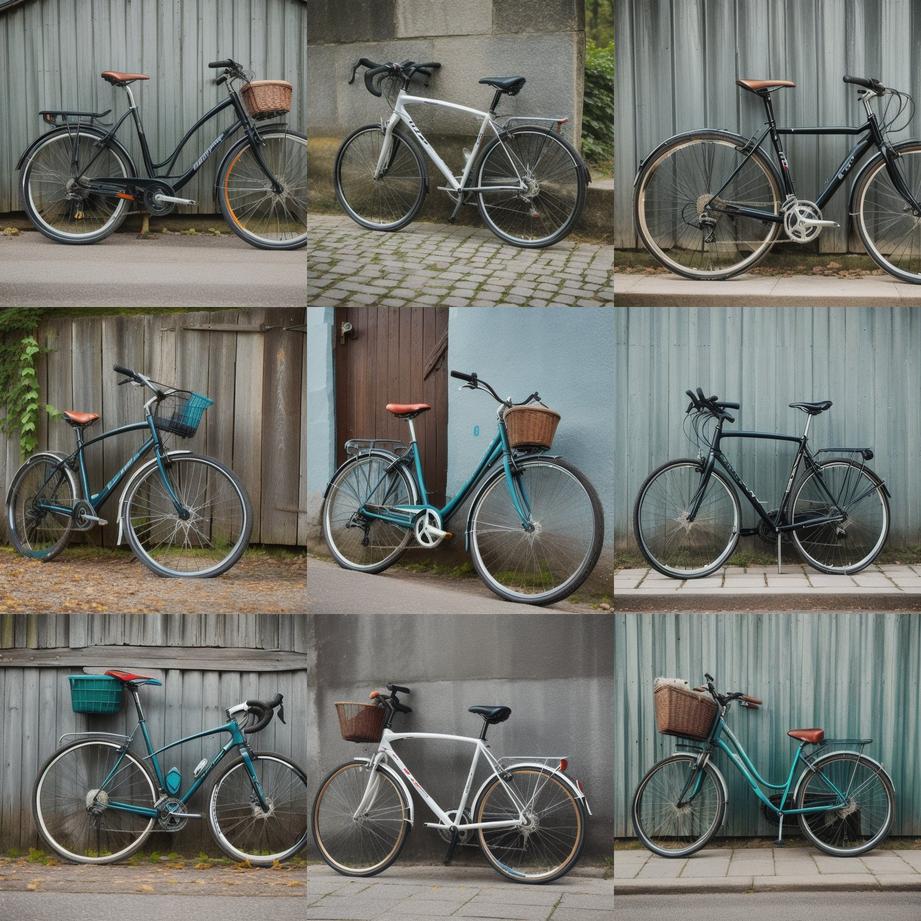} &
        \includegraphics[width=0.16\linewidth]{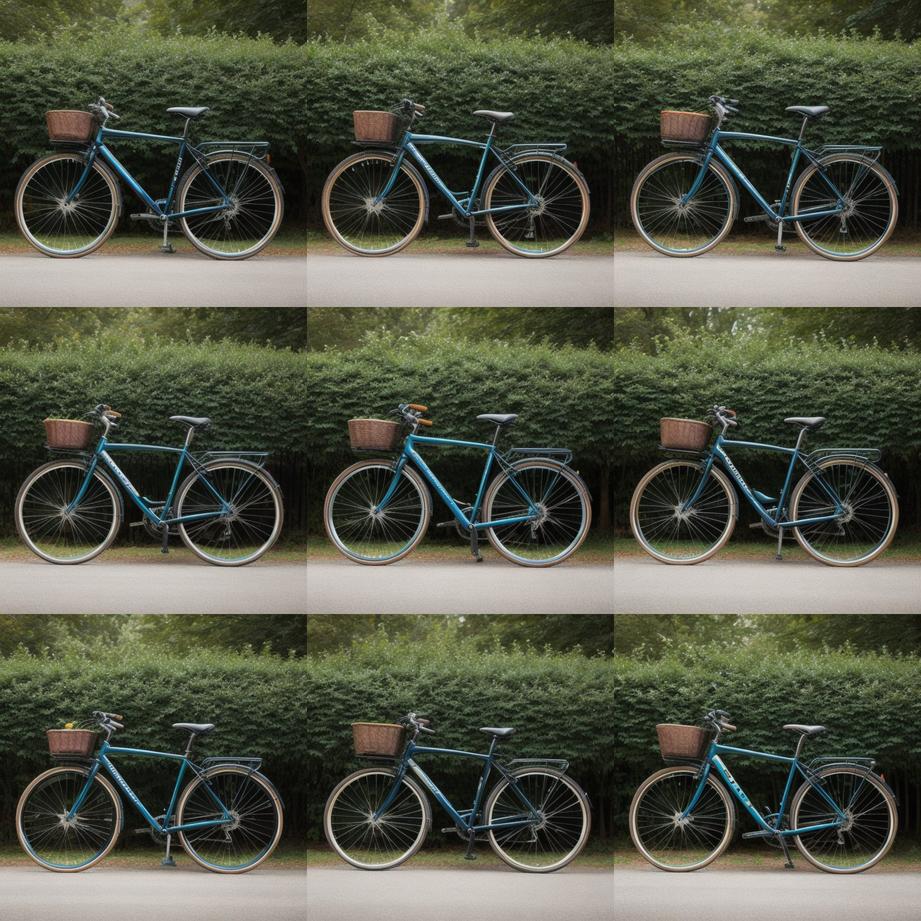} \\[-0.1em]
        
        \includegraphics[width=0.16\linewidth]{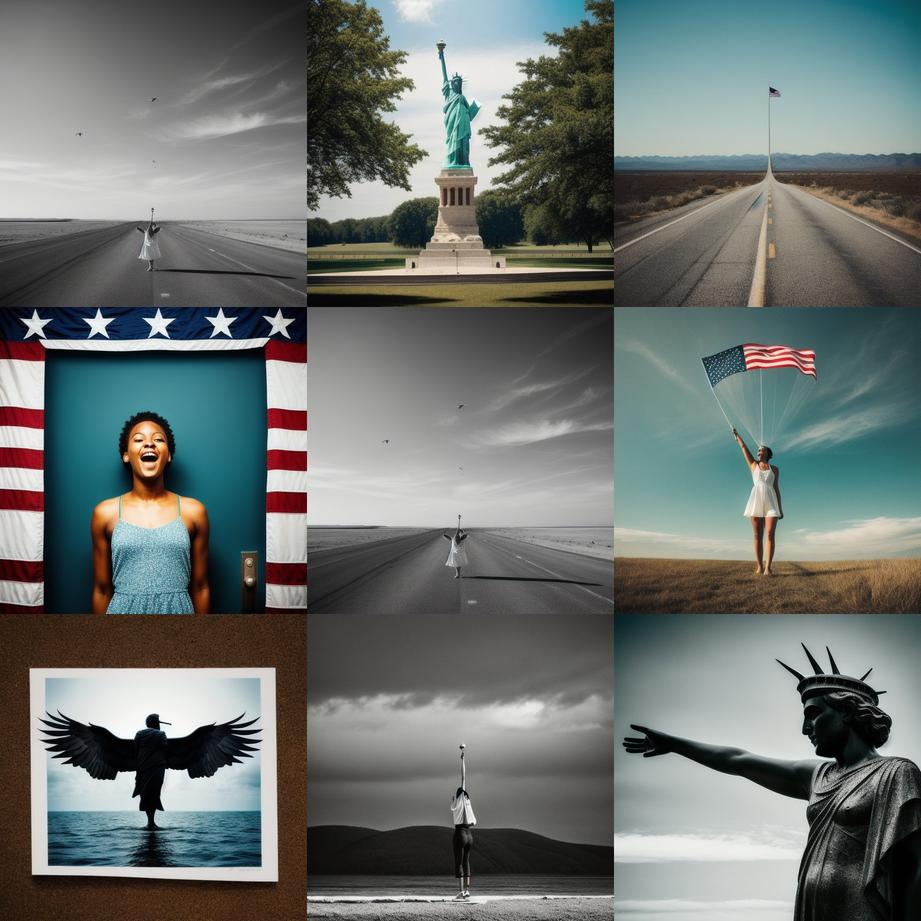} &
        \begin{subfigure}[b]{0.16\linewidth}
            \centering
            \makebox[\linewidth][c]{\textbf{N/A}} 
            \makebox[\linewidth][c]{\rule{0pt}{0.4in}} 
        \end{subfigure} &
        \includegraphics[width=0.16\linewidth]{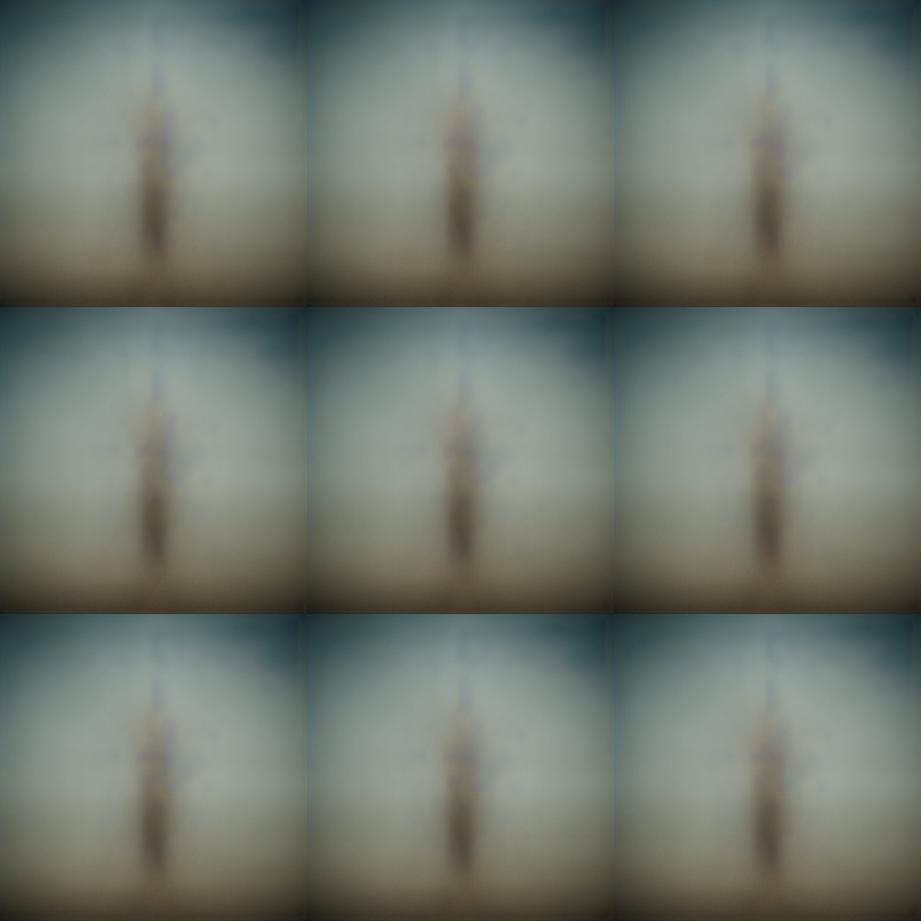} &
        \includegraphics[width=0.16\linewidth]{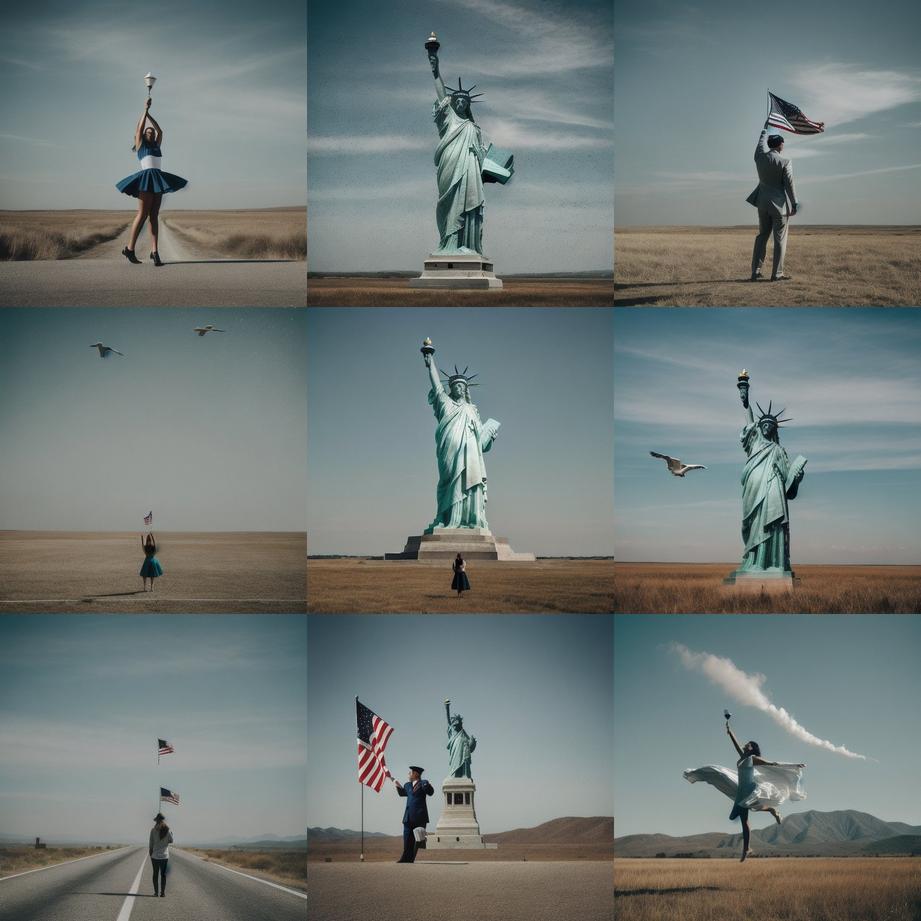} &
        \includegraphics[width=0.16\linewidth]{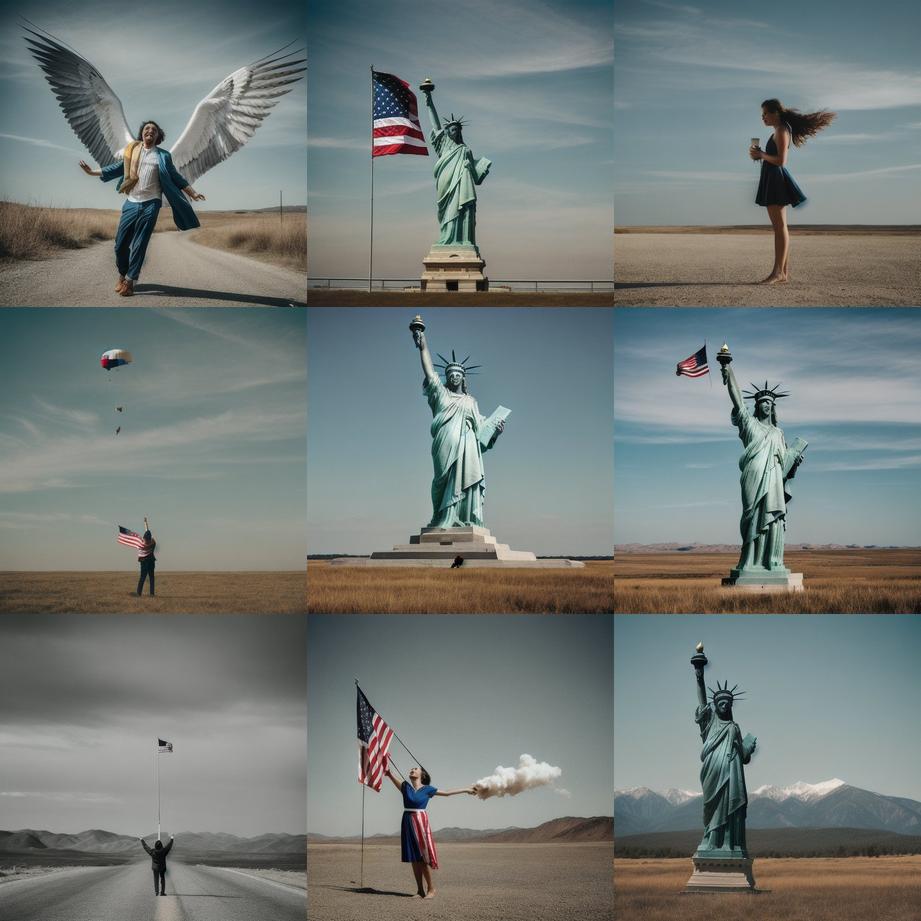} &
        \includegraphics[width=0.16\linewidth]{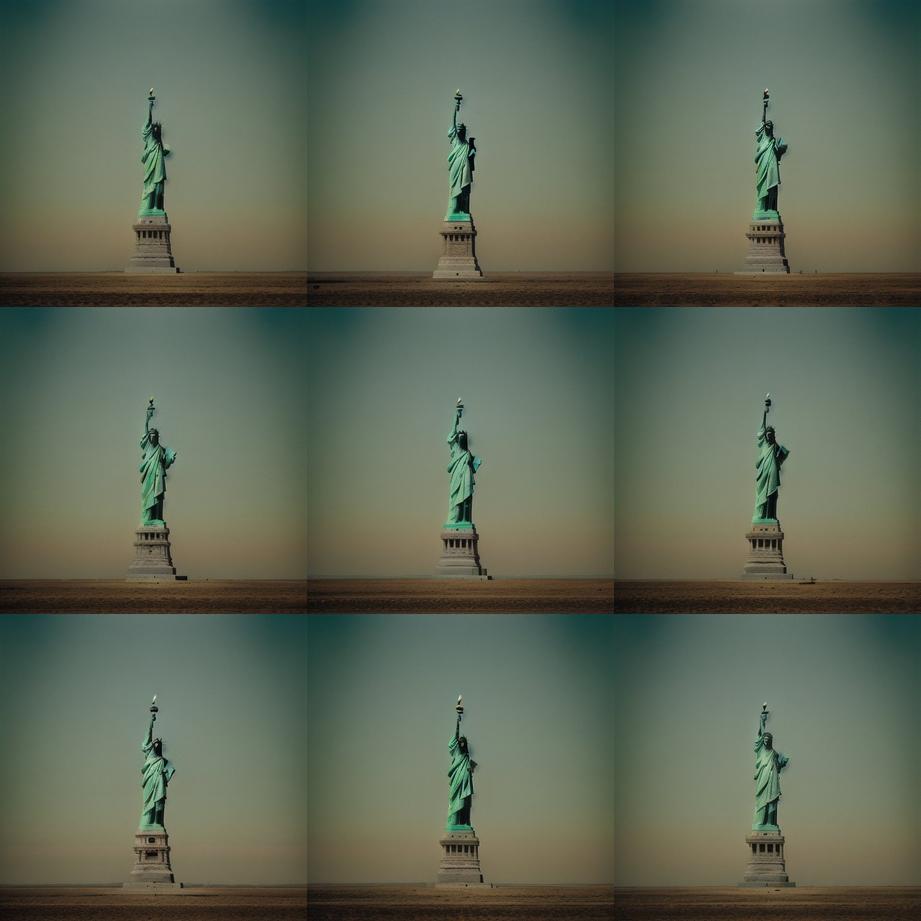} \\
    \end{tabular}
    \vspace{-5pt}
    \caption{
        Qualitative comparison across methods:
        \textbf{GANgealing}~\cite{peebles2022gan},
        \textbf{Avg. VAE},
        \textbf{D$^4$M}~\cite{su2024d},
        \textbf{MGD$^3$}~\cite{chan2025mgd},
        and \textbf{DMA (Ours)}.
        Rows show concepts: \textit{astronaut}, \textit{dog}, \textit{bicycle}, \textit{freedom}.
        GANgealing cannot process abstract concepts, so its \textit{freedom} cell is intentionally left blank.
    }
    \label{fig:baseline_comparison}
\end{figure*}




We evaluate on 12 concepts spanning four categories (Animal, Person, Object, Abstract).
For each concept, we generate 10 sets of 1,000 samples, and each method produces 10 prototypes per set, yielding 100 prototypes per concept per method. This evaluation does not perform clustering as in the multimodal setting (Sec.~\ref{sec:mode_discov}). We use three metrics:

\boldsubsection{1. Consistency (↓)}  
measures how reliably a method converges to the same prototype when starting from different initializations. For each set, we compute the average pairwise distance among the 10 generated prototypes using CLIP~\cite{radford2021learning} cosine distance, DreamSim~\cite{fu2023dreamsim}, and LPIPS~\cite{zhang2018unreasonable}.
We report consistency scores averaged over all sets in each category, where lower is better.

\boldsubsection{2. Representativeness (↓)}  
assesses how well each prototype captures the semantic center of its concept distribution. For each prototype, the Representativeness score is the average distance to the 1,000 samples in its set. We use the same three distance functions as in the Consistency metric and report scores averaged across all sets in each category, where lower is better.

\boldsubsection{3. ImageReward~\cite{xu2023imagereward} (↑)} is a learned reward model trained on large-scale human preference data. This metric reflects human preference by jointly evaluating image quality and text–image alignment. For each concept set, we compute the average score across the 10 generated prototypes, and then report the average ImageReward over all sets within each category, where higher is better.


\begin{table*}[t]
\centering
\small
\setlength{\tabcolsep}{6pt}
\caption{
\textbf{Average scores} across four categories
(\textit{Animal}, \textit{Person}, \textit{Object}, \textit{Abstract}) on 
Consistency (↓), Representativeness (↓), and ImageReward (↑).
GANgealing$^\dagger$ is evaluated only on three non-abstract categories, as it lacks pretrained GANs on abstract concepts 
}
\begin{tabular}{lccc ccc c}
\toprule
& \multicolumn{3}{c}{\textbf{Consistency (↓)}} 
& \multicolumn{3}{c}{\textbf{Representativeness (↓)}} 
& \multirow{2}{*}{\textbf{ImageReward (↑)}} \\
\cmidrule(lr){2-4} \cmidrule(lr){5-7}
\textbf{Method} 
& CLIP & DreamSim & LPIPS
& CLIP & DreamSim & LPIPS
&  \\
\midrule
GANgealing$^\dagger$~\cite{peebles2022gan}
& 0 & 0 & 0
& 0.386 & 0.477 & 0.851
& $-0.684$ \\
Avg~VAE 
& 0 & 0 & 0
& 0.473 & 0.806 & 0.805
& $-2.262$ \\
\midrule
D$^4$M~\cite{su2024d} 
& 0.168 & 0.274 & 0.572
& 0.197 & 0.363 & 0.672
& 0.823 \\
MGD$^3$~\cite{chan2025mgd} 
& 0.180 & 0.319 & 0.643
& 0.195 & 0.364 & 0.687
& 0.755 \\
\midrule
\textbf{DMA (Ours)} 
& \textbf{0.031} & \textbf{0.032} & \textbf{0.129}
& \textbf{0.179} & \textbf{0.341} & \textbf{0.655}
& \textbf{1.002} \\
\bottomrule
\end{tabular}
\label{tab:combined_metrics}
\end{table*}

\begin{figure}[t]
    \centering
    \setlength{\tabcolsep}{1pt}
    \begin{tabular}{cc}
        \includegraphics[width=0.49\linewidth]{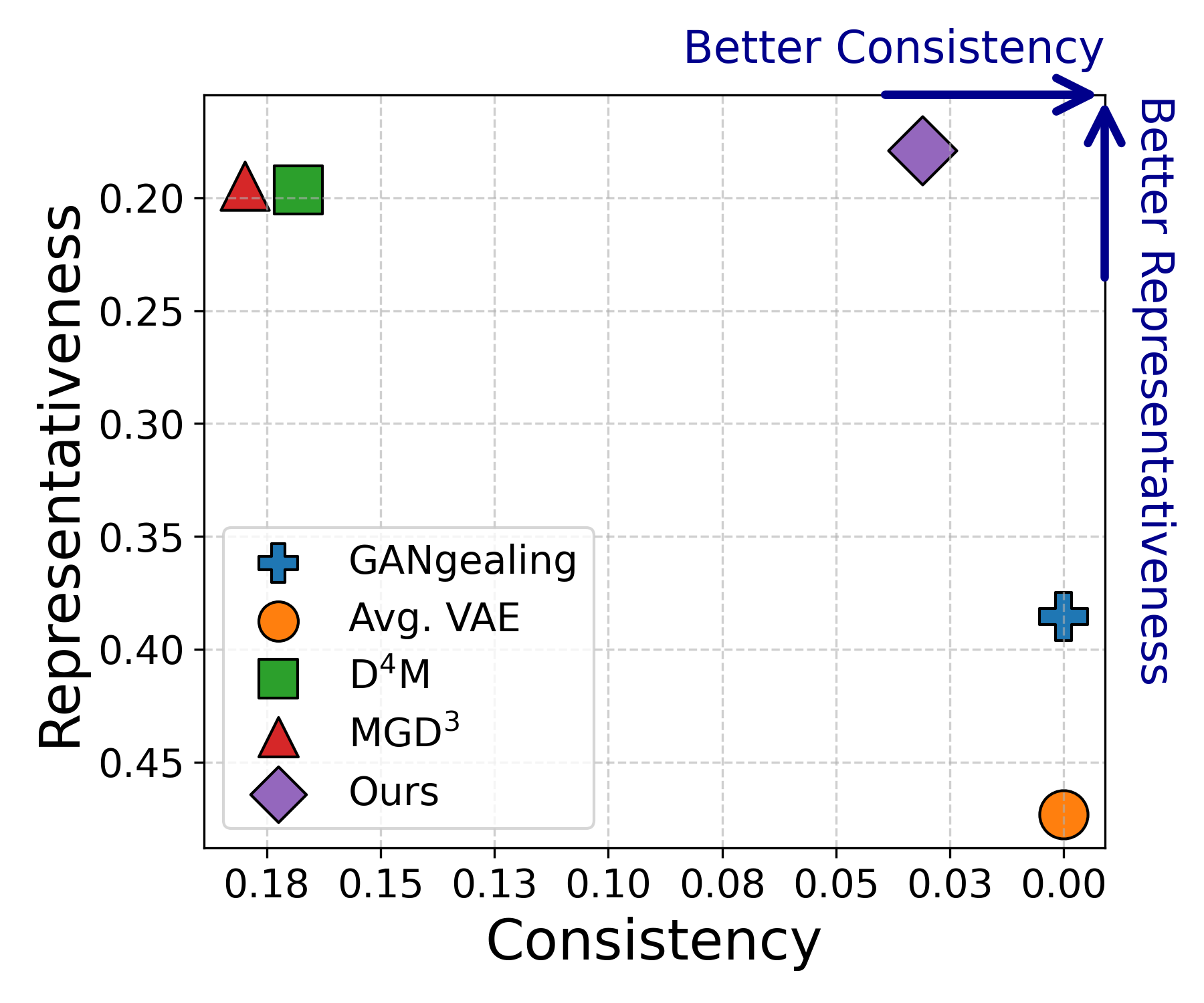} &
        \includegraphics[width=0.49\linewidth]{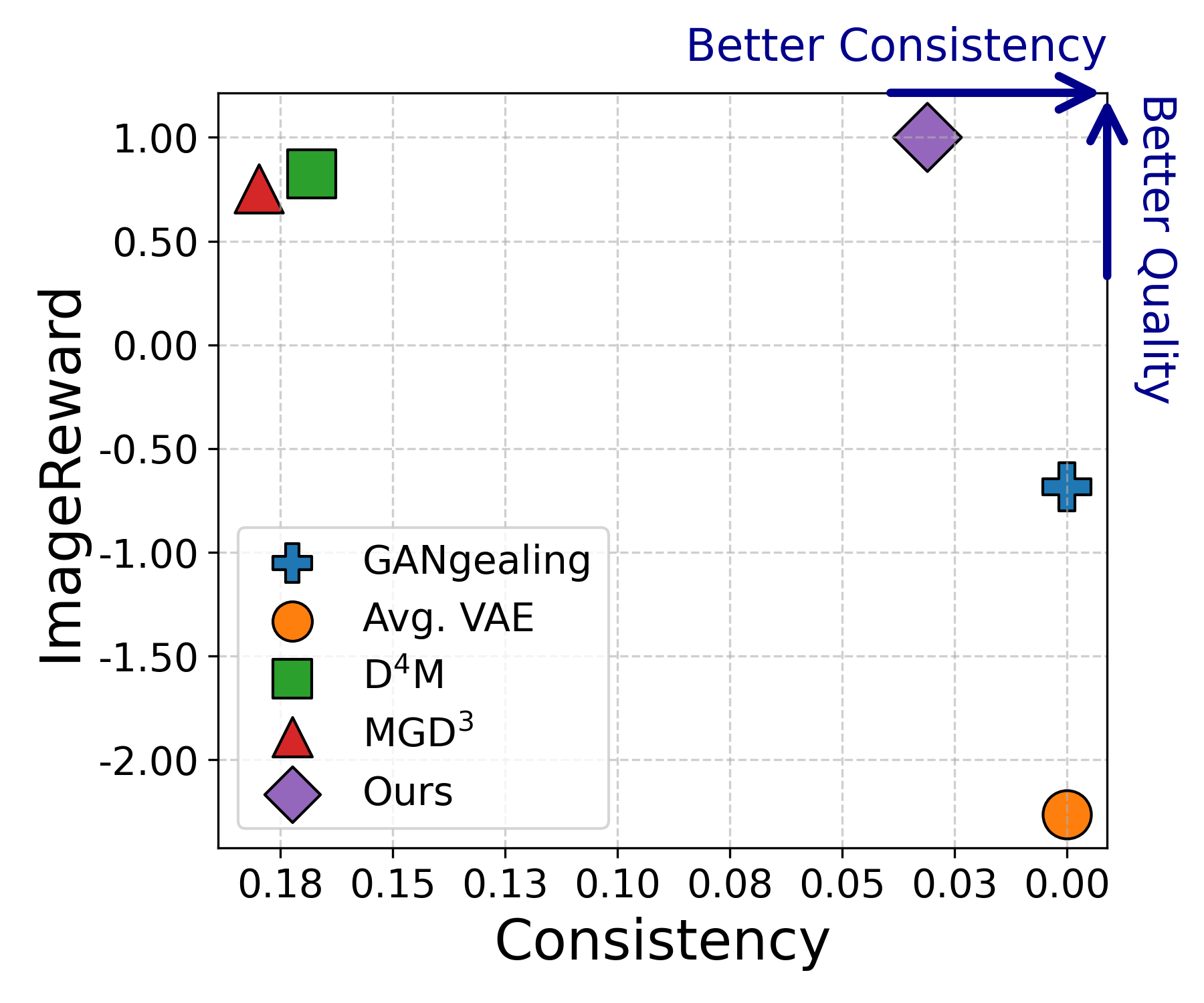}
    \end{tabular}
    \vspace{-10pt}
    \caption{
        \textbf{Quality Trade-off. } DMA achieves a superior balance of representativeness, consistency, and perceptual quality. 
    }
    \vspace{-5pt}
    \label{fig:tradeoff_plots}
\end{figure}

Quantitative scores are shown in Table ~\ref{tab:combined_metrics}, with qualitative results in Figure~\ref{fig:baseline_comparison}. Avg. VAE and GANgealing achieve perfect consistency scores of zero by design, as they always produce identical outputs given the same image set. However, their averages are blurry and lack realism. DMA is significantly more consistent than D$^4$M and MGD$^3$ with better consistency scores across all three distance metrics. DMA also achieves the best ImageReward and representativeness scores across all metrics, demonstrating more faithful and visually preferred concept summaries than all baselines.

Qualitatively, DMA generates clear, structured averages, for example, a detailed astronaut in uniform with averaged attributes, such as centered composition. By contrast, the baselines only capture vague helmet shapes and omit key components like the helmet's visor or uniform altogether.
For abstract concepts, like ``freedom,'' DMA consistently depicts the Statue of Liberty, whereas D$^4$M and MGD$^3$ generate inconsistent or distorted versions of the statue. GANgealing, meanwhile, cannot handle such concepts because no pretrained GAN exists for them. Furthermore, as shown in the trade-off plots in Figure~\ref{fig:tradeoff_plots}, DMA delivers the most balanced performance, combining high consistency, strong representativeness, and perceptual realism.

\subsection{Mode Averages}
In this section, we present results of applying DMA to clustered images. To refine the model's conditioning for each mode, we use a CFG scale of 3.0 during inference and train a rank-1 LoRA for 2,000 steps with a learning rate of $10^{-4}$. For Textual Inversion, we use a CFG scale of 7.0 during inference and train for 3,000 steps with a learning rate of $10^{-2}$. For mode separation, we first reduce the computed features of CLIP-ViT-B/32 or BLIP-VQA-base~\cite{luccioni2023stable} using PCA (to 2 dimensions for unsupervised clustering and 10 for grounded clustering), then cluster them using a Gaussian Mixture Model to obtain mode assignments. 

\begin{figure}[t]
    \centering
    \renewcommand{\arraystretch}{1.2}
    \setlength{\tabcolsep}{1pt}
    \footnotesize
    \begin{tabular}{c c c c c} 
        & \textbf{Fan} & \textbf{Crane} & \textbf{Pumpkin} & \textbf{Boat} \\

        \raisebox{1.2\height}{\rotatebox[origin=c]{90}{\text{Average}}} &
        \includegraphics[width=0.21\linewidth]{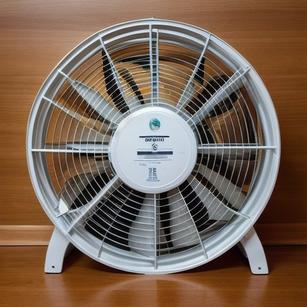} &
        \includegraphics[width=0.21\linewidth]{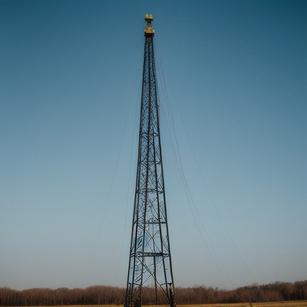} &
        \includegraphics[width=0.21\linewidth]{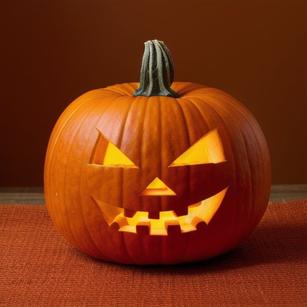} &
        \includegraphics[width=0.21\linewidth]{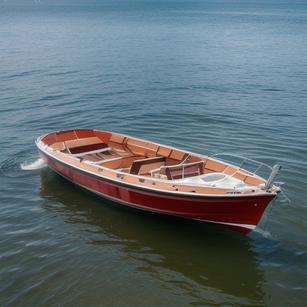} \\
        
        \raisebox{1.2\height}{\rotatebox[origin=c]{90}{\text{Mode 1}}} &
        \includegraphics[width=0.21\linewidth]{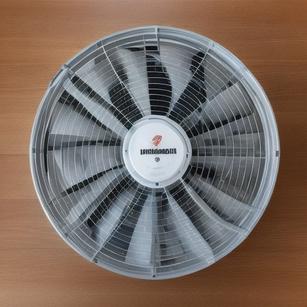} &
        \includegraphics[width=0.21\linewidth]{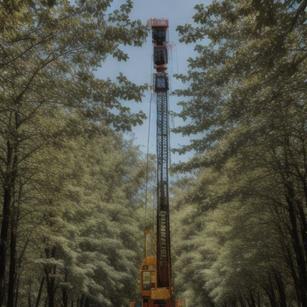} &
        \includegraphics[width=0.21\linewidth]{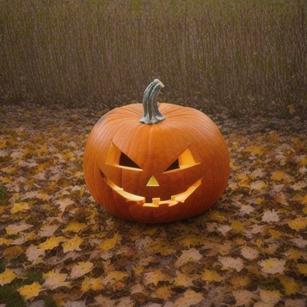} &
        \includegraphics[width=0.21\linewidth]{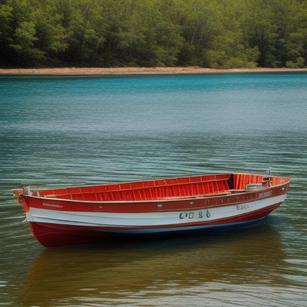} \\

        \raisebox{1.2\height}{\rotatebox[origin=c]{90}{\text{Mode 2}}} &
        \includegraphics[width=0.21\linewidth]{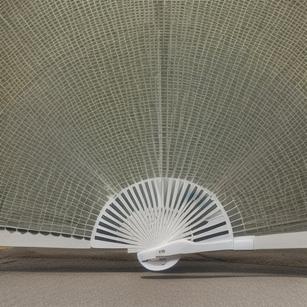} &
        \includegraphics[width=0.21\linewidth]{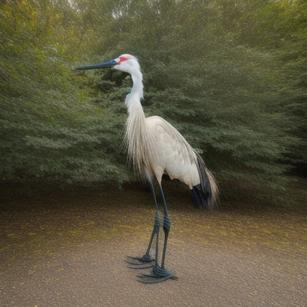} &
        \includegraphics[width=0.21\linewidth]{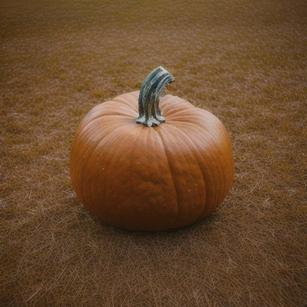} &
        \includegraphics[width=0.21\linewidth]{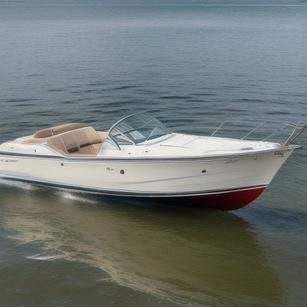} \\
    \end{tabular}
    \vspace{-10pt}
    \caption{\textbf{DMA prototypes of unsupervised modes.} Top row: overall average. Bottom rows: averages of discovered modes.
}
    \label{fig:mode_unsup}
\end{figure}

\boldsubsection{Unsupervised clustering.}
Figure~\ref{fig:mode_unsup} presents our cluster averages discovered through unsupervised clustering, along with a single overall average, for four concepts. We observe that the single average is biased toward the dominant mode, thereby suppressing less frequent ones (e.g., the \textit{fan} average only depicts an electric fan). In contrast, our cluster averages successfully reveal the distinct semantic modes within each concept.

\begin{figure}[t]
    \centering
    \renewcommand{\arraystretch}{1.2}
    \setlength{\tabcolsep}{1pt}
    \footnotesize
    \begin{tabular}{c c c c c} 
        & \textbf{Cluster 1} & \textbf{Cluster 2} & \textbf{Cluster 3} & \textbf{Cluster 4} \\
        
        \raisebox{0.8\height}{\rotatebox[origin=c]{90}{\shortstack{Dog \\ (``Breed'')}}} &
        \includegraphics[width=0.21\linewidth]{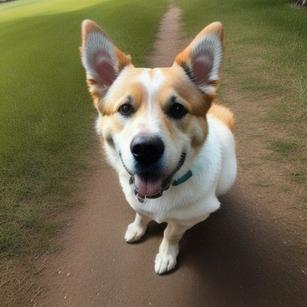} &
        \includegraphics[width=0.21\linewidth]{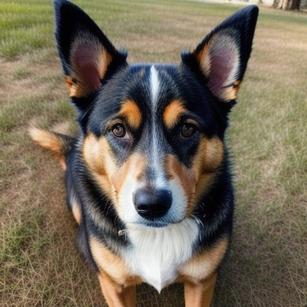} &
        \includegraphics[width=0.21\linewidth]{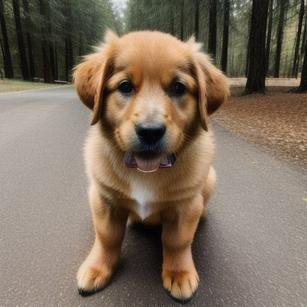} &
        \includegraphics[width=0.21\linewidth]{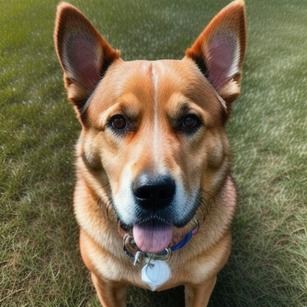} \\

        \raisebox{0.7\height}{\rotatebox[origin=c]{90}{\shortstack{Astronaut \\ (``Ethnicity'')}}} &
        \includegraphics[width=0.21\linewidth]{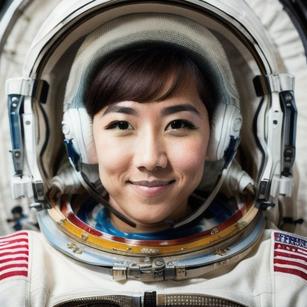} &
        \includegraphics[width=0.21\linewidth]{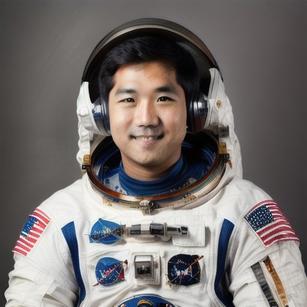} &
        \includegraphics[width=0.21\linewidth]{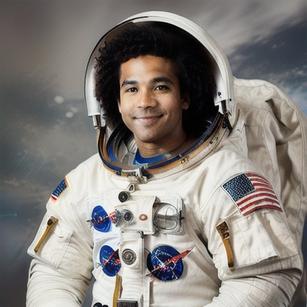} &
        \includegraphics[width=0.21\linewidth]{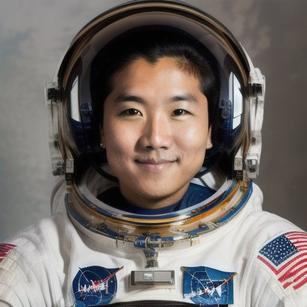} \\

        \raisebox{0.9\height}{\rotatebox[origin=c]{90}{\shortstack{Car \\ (``Model'')}}} &
        \includegraphics[width=0.21\linewidth]{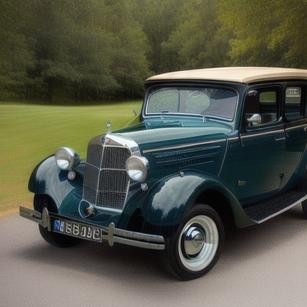} &
        \includegraphics[width=0.21\linewidth]{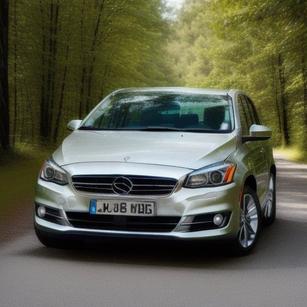} &
        \includegraphics[width=0.21\linewidth]{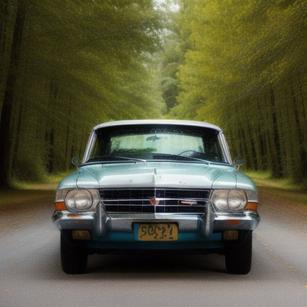} &
        \includegraphics[width=0.21\linewidth]{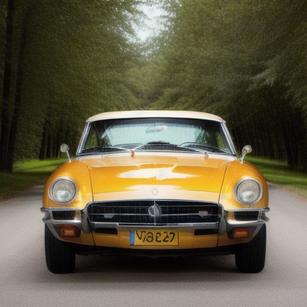} \\

        \raisebox{0.9\height}{\rotatebox[origin=c]{90}{\shortstack{Car \\ (``Color'')}}} &
        \includegraphics[width=0.21\linewidth]{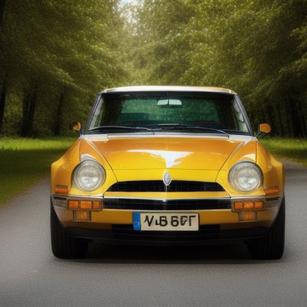} &
        \includegraphics[width=0.21\linewidth]{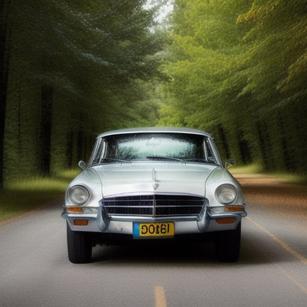} &
        \includegraphics[width=0.21\linewidth]{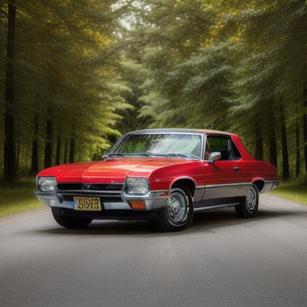} &
        \includegraphics[width=0.21\linewidth]{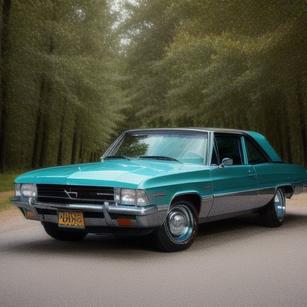} \\
    \end{tabular}
    \vspace{-10pt}
    \caption{\textbf{DMA prototypes of grounded modes.} We cluster \textit{astronaut} by ethnicity, \textit{dog} by breed, and \textit{car} by model or color.}
    \label{fig:mode_ground}
    \vspace{-10pt}
\end{figure}

\boldsubsection{Grounded clustering.}
Figure~\ref{fig:mode_ground} presents our cluster averages from grounded clustering using BLIP-VQA. The results show that DMA can generate accurate average images across different grounded criteria. For the \textit{car} concept grounded by color, the averages show that yellow and silver cars look similar, while red and blue cars form another consistent group. This demonstrates how DMA can be used as a probing tool to reveal how models organize and associate attributes within a concept.

\begin{figure}[t]
    \centering
    \footnotesize
    \renewcommand{\arraystretch}{1.2}
    \setlength{\tabcolsep}{1pt}
    \begin{tabular}{c cccc}
        \textbf{Cluster Images} & \textbf{DMA} & \textbf{+TI} & \textbf{+LoRA}\\
        \includegraphics[width=0.23\linewidth]{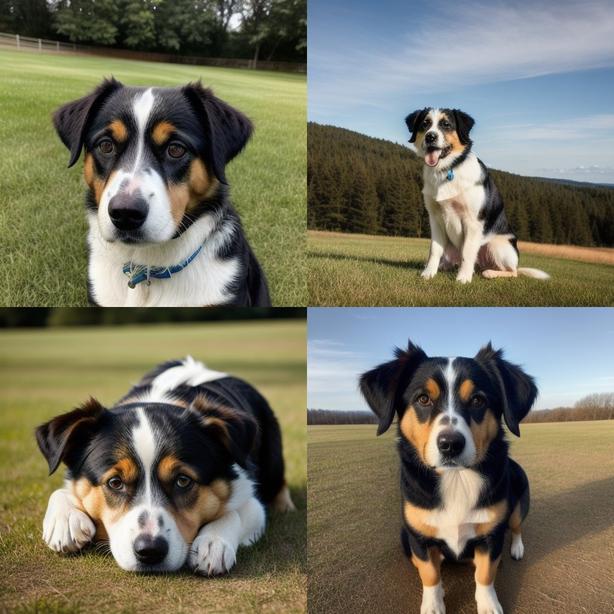} &
        \includegraphics[width=0.23\linewidth]{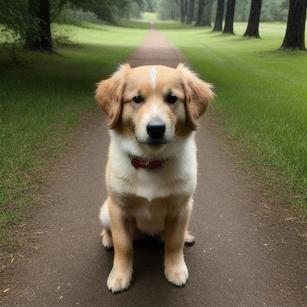} &
        \includegraphics[width=0.23\linewidth]{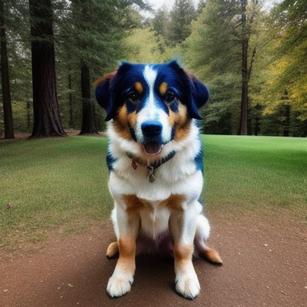} &
        \includegraphics[width=0.23\linewidth]{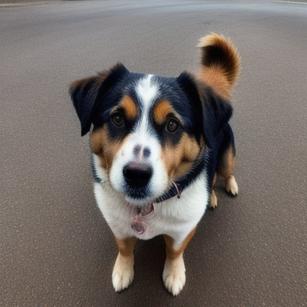} \\

        \includegraphics[width=0.23\linewidth]{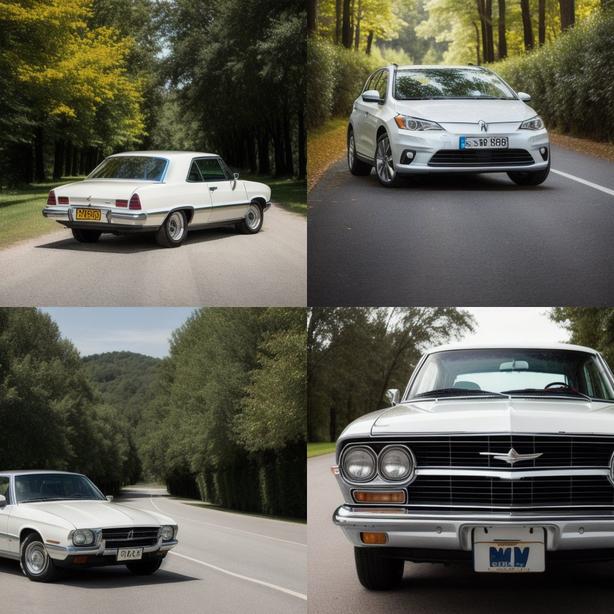} &
        \includegraphics[width=0.23\linewidth]{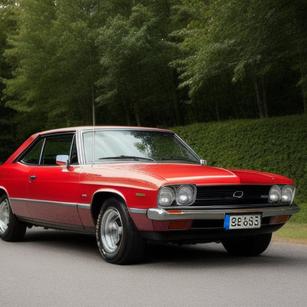} &
        \includegraphics[width=0.23\linewidth]{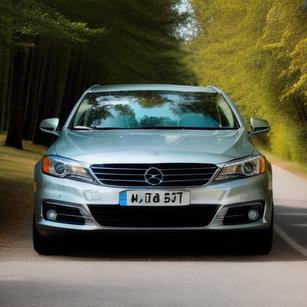} &
        \includegraphics[width=0.23\linewidth]{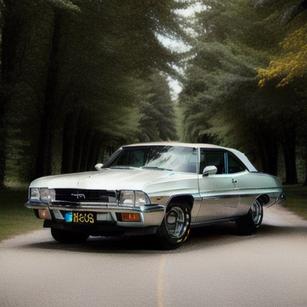} \\
    \end{tabular}
    \vspace{-10pt}
    \caption{\textbf{LoRA \& Textual Inversion}
bridge the gap between the clustering and the $h$-spaces, enabling cluster-specific prototypes.}
    \label{fig:ab3_lora}
    \vspace{-6pt}
\end{figure}

\boldsubsection{LoRA and Textual Inversion.}
Figure~\ref{fig:ab3_lora} shows that using LoRA or Textual Inversion helps guide the model toward cluster-specific regions, producing averages that better reflect the modes' semantics, with LoRA better preserving color and car shape than Textual Inversion.
This suggests that both methods help capture semantics beyond the $h$-space, including aspects typically encoded in the noisy latent, such as color and image layout~\cite{wen2024detecting,morita2025tkg,zhang2024diffmorpher}.


\vspace{-6pt}
\subsection{Generalization Across SD Variants}
\begin{figure}[t]
    \centering
    \renewcommand{\arraystretch}{1.2}
    \setlength{\tabcolsep}{1pt}
    \footnotesize
    \begin{tabular}{c c c c c} 
        & \shortstack{SD1.5} & \shortstack{Realistic \\ Vision V5} & \shortstack{Dreamshaper \\ PixelArt} & \shortstack{Flat 2d \\ Animerge} \\
        
        \raisebox{1.5\height}{\rotatebox[origin=c]{90}{\text{Soldier}}} &
        \includegraphics[width=0.21\linewidth]{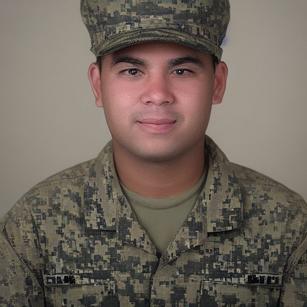} &
        \includegraphics[width=0.21\linewidth]{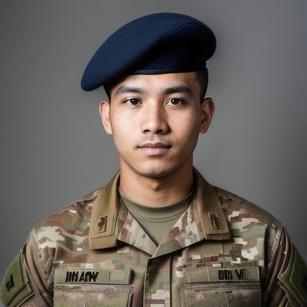} &
        \includegraphics[width=0.21\linewidth]{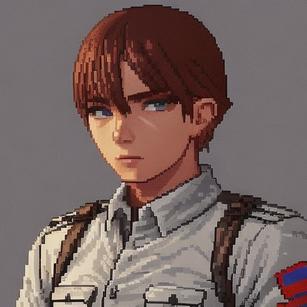} &
        \includegraphics[width=0.21\linewidth]{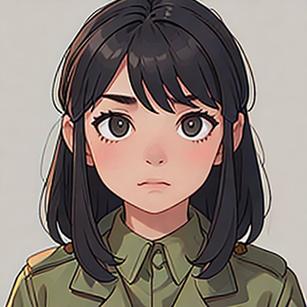} \\

        \raisebox{1.8\height}{\rotatebox[origin=c]{90}{\text{Dolphin}}} &
        \includegraphics[width=0.21\linewidth]{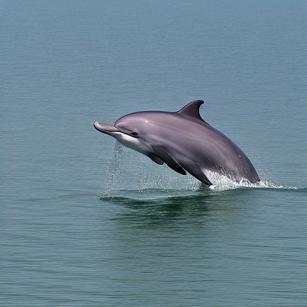} &
        \includegraphics[width=0.21\linewidth]{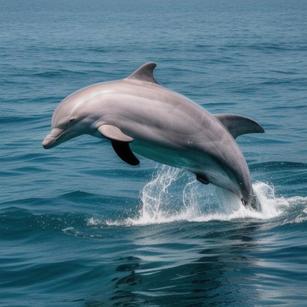} &
        \includegraphics[width=0.21\linewidth]{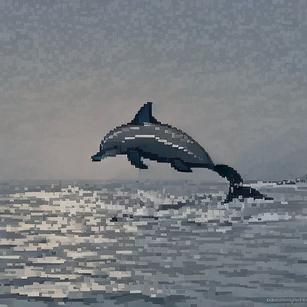} &
        \includegraphics[width=0.21\linewidth]{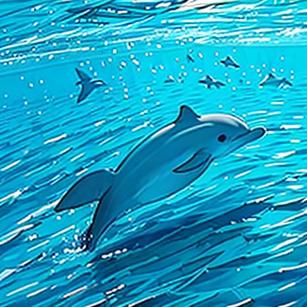} \\

        \raisebox{2.4\height}{\rotatebox[origin=c]{90}{\text{Italy}}} &
        \includegraphics[width=0.21\linewidth]{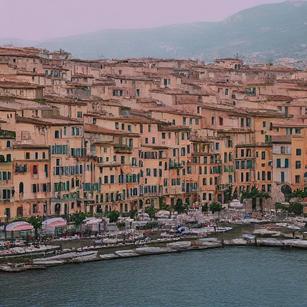} &
        \includegraphics[width=0.21\linewidth]{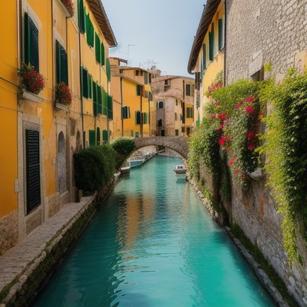} &
        \includegraphics[width=0.21\linewidth]{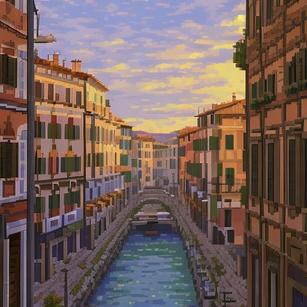} &
        \includegraphics[width=0.21\linewidth]{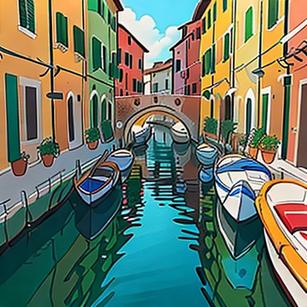} \\
    \end{tabular}
    \vspace{-10pt}
    \caption{
    \textbf{DMA prototypes of various SD Variants.} \textit{Soldier} inherits model-specific gender bias, while \textit{Italy} exhibits a persistent static bias toward a Venice-like scene across all variants.
    } 
    \label{fig:cross-models}
\end{figure}

We evaluate DMA across several Stable Diffusion variants: SD1.5, Realistic Vision~v5.1, Dreamshaper PixelArt, and Flat-2D Animerge.
In Figure~\ref{fig:cross-models}, DMA can produce averages that are specific to each variant, revealing distinct stylistic or demographic interpretations.
The concept \textit{soldier} exhibits strong identity bias: SD1.5 and Realistic Vision consistently generate male soldiers, whereas PixelArt appears more gender-neutral, and Flat-2D Animerge renders a female anime-style cadet. 
For \textit{Italy}, all models converge to a similar ``Venice canal'' scene despite stylistic differences, indicating a shared dominant bias.
This highlights that DMA generalizes across models and helps diagnose how fine-tuning alters or preserves concept semantics.

\subsection{Generalization Across Architectures}
We extend DMA to DiT-XL~\cite{peebles2023scalable}, a class-conditional, transformer-based diffusion model. Since DiT lacks a U-Net-style semantic bottleneck like the $h$-space, we instead use the output of the final transformer block, which yields the most consistent results (Figure~\ref{fig:DiT}). See Appendix~\ref{sec:a_dit} and~\ref{sec:dit_layer} for implementation details and block comparisons.
\begin{figure}[t]
    \centering
    \renewcommand{\arraystretch}{1.2}
    \setlength{\tabcolsep}{1pt}
    \begin{tabular}{c ccccc} 
        \includegraphics[width=0.18\linewidth]{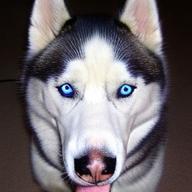} &
        \includegraphics[width=0.18\linewidth]{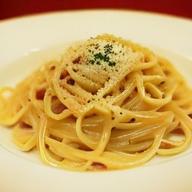} &
        \includegraphics[width=0.18\linewidth]{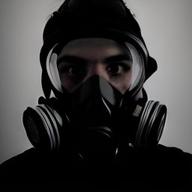} &
        \includegraphics[width=0.18\linewidth]{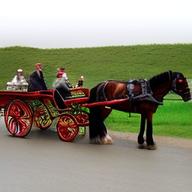} &
        \includegraphics[width=0.18\linewidth]{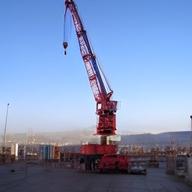}\\

        \includegraphics[width=0.18\linewidth]{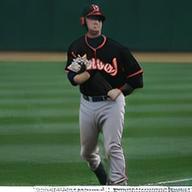} &
        \includegraphics[width=0.18\linewidth]{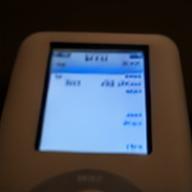} &
        \includegraphics[width=0.18\linewidth]{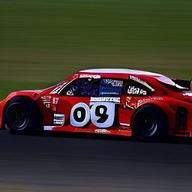} &
        \includegraphics[width=0.18\linewidth]{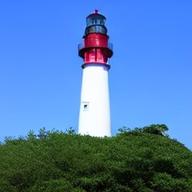} &
        \includegraphics[width=0.18\linewidth]{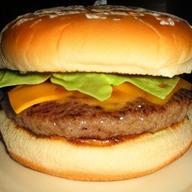}\\
    \end{tabular}
    \vspace{-10pt}
    \caption{\textbf{DMA generalizes to the DiT architecture~\cite{peebles2023scalable}.}}
    \label{fig:DiT}
    \vspace{-10pt}
\end{figure}

\section{Limitations and Discussion}
As with sample generation, the concept and hyperparameters, like CFG, affect the quality and consistency of the average image. High-variation concepts require more samples for stable averages, while higher CFG improves prompt alignment but reduces diversity, influencing the final result.


Our clustering relies on external encoders and thus inherits their biases~\cite{hamidieh2024identifying,baherwani2024racial,luo2024fairclip}; however, the averaging process remains unbiased once clusters are defined. Evaluating the representativeness score remains challenging as it depends on the specific embedding space, which can be subjective. We discuss this further in Appendix~\ref{sec:b_metric}. Extending DMA to other diffusion architectures requires identifying a semantic layer analogous to the $h$-space, which remains an open direction for future work. Currently, our method requires high computational costs, which are detailed in Appendix~\ref{sec:b_efficeincy}. Further discussions regarding potential applications and limitations of our mode discovery are provided in Appendix~\ref{sec:b_application} and~\ref{sec:b_mode}.



\boldsubsection{Conclusion.}
We present a simple yet effective method for generating high-quality average images from pretrained diffusion models. 
By progressively aligning denoising trajectories across multiple noise latents, DMA produces consistent, representative results and extends to concepts with multiple meanings and modes. DMA outperforms baselines in consistency, image quality, and representativeness, and generalizes across diffusion models and architectures.

\boldsubsection{Acknowledgment.} 
This research was supported by Thailand Science Research and Innovation (TSRI) Fundamental Fund (Grant number: FRB690039/0457; Project number: 4823418), SCB Public Company Limited, and PTT Public Company Limited.
{
    \small
    \bibliographystyle{ieeenat_fullname}
    \bibliography{main}
}

\clearpage
\setcounter{page}{1}
\maketitlesupplementary

\setcounter{section}{0}
\renewcommand{\thesection}{\Alph{section}}
\section{Additional Implementation Details}\label{sec:a_implement_details}
\hypertarget{my:a_implement_details}{}
\subsection{Prompt Templates}\label{sec:a_prompt_template}
Our method uses category-specific prompt templates as input to the Stable Diffusion model:
\begin{itemize}
    \item Animal: \textit{A photo of a \{concept name}\}
    \item Person: \textit{Photo portrait of a \{concept name}\}
    \item Object: \textit{A photo of a \{concept name\}}
    \item Abstract: \textit{A conceptual photo representing \{concept name\}}
\end{itemize}
For Dreamshaper PixelArt, we append ``pixel art style, detailed'' to the end of the prompt.
\subsection{Negative Prompt}\label{sec:a_negative_prompt}
For each variant, we use the negative prompt recommended by the respective author. All weighted negative prompts were encoded using \textbf{Compel}\footnote{https://github.com/damian0815/compel}.

\boldsubsection{Realistic Vision v5.1’s negative prompt}:
\textit{(deformed iris, deformed pupils, semi-realistic, cgi, 3d, render, sketch, cartoon, drawing, anime:1.4), text, close up, cropped, out of frame, worst quality, low quality, jpeg artifacts, ugly, duplicate, morbid, mutilated, extra fingers, mutated hands, poorly drawn hands, poorly drawn face, mutation, deformed, blurry, dehydrated, bad anatomy, bad proportions, extra limbs, cloned face, disfigured, gross proportions, malformed limbs, missing arms, missing legs, extra arms, extra legs, fused fingers, too many fingers, long neck"}

\boldsubsection{Dreamshaper PixelArt's negative prompt}: \textit{``worst quality"}

\boldsubsection{Flat 2D Animerge's negative prompt}: \textit{
(worst quality:0.8), verybadimagenegative\_v1.3, (surreal:0.8), (modernism:0.8), (art deco:0.8), (art nouveau:0.8)"}

\subsection{BLIP-VQA for Grounded Clustering}\label{sec:a_blip_vqa}
We follow the grounded clustering method introduced in Stable Bias~\cite{luccioni2023stable}, where it is used as one of several tools for evaluating bias in diffusion models. This method clusters image embeddings produced by BLIP-VQA to group images that share similar attributes. In their work, these embeddings are conditioned on a question prompt, such as “What word best describes this person’s ethnicity?”, which ensures that the resulting embedding focuses on the specified attribute (in this example, ethnicity). We adapt this work for our task using the following prompt: “What word best describes this \{concept name\}’s \{attribute\}?”.

\subsection{DiT Implementation Details}\label{sec:a_dit}
We use the DiT-XL/2-256 model~\cite{peebles2023scalable}, a large-scale Diffusion Transformer trained for class-conditional image generation. Unless otherwise noted, all DiT results are generated using a classifier-free guidance scale of 10.0, a learning rate of $5\times10^{-4}$, 20 inference steps, and a DMA stopping timestep of $t_{\text{stop}}=20$. We use the output of the last transformer block in place of the Stable Diffusion's $h$-space to perform averaging in Algorithm 1. Experiment~\ref{sec:dit_layer} evaluates the use of different transformer blocks in DMA.

\section{Additional Discussion}
\subsection{Practical Applications of Mental Averages}\label{sec:b_application}
Mental averages offer a concrete visual summary of a learned concept, enabling analysis beyond individual samples and elucidating the model's internal representation.

\begin{wrapfigure}{r}{0.4\linewidth}
    \vspace{-10pt}
    \centering
    \captionsetup{font=scriptsize}
    \tiny
    \setlength{\tabcolsep}{0pt}
    \renewcommand{\arraystretch}{0.5}

    \begin{tabular}{@{}c@{\hspace{1pt}}c@{}}
        French & English \\
        \includegraphics[width=0.5\linewidth]{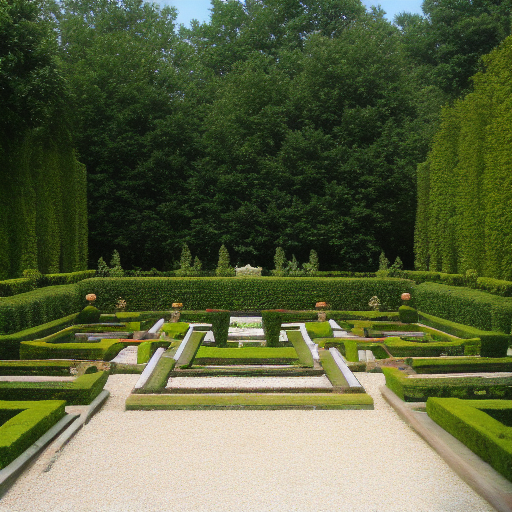} &
        \includegraphics[width=0.5\linewidth]{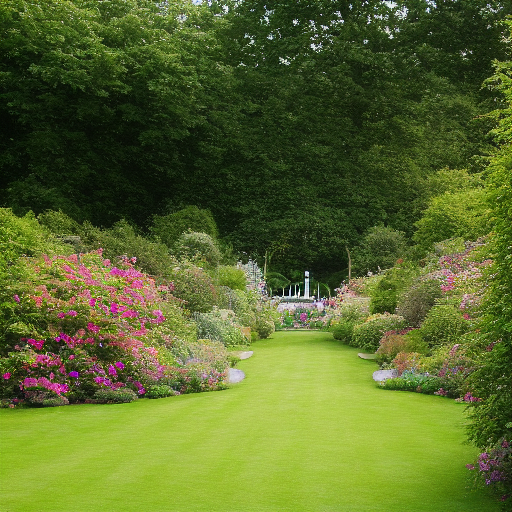} \\
    \end{tabular}

    \vspace{-10pt}
\end{wrapfigure}
Questions like ``What distinguishes French from English gardens?'' are hard to answer from samples alone, as people often lack clear visual expectations and do not know what to look for. Inspecting thousands of samples for statistical significance is inefficient and still not as conclusive, as it relies on the viewer’s ability to detect both between- and within-class patterns. Averages can readily reveal French as more geometric and English as more organic---direct visual evidence that individual samples rarely make obvious (please see our website for more results). Running attribute classifiers for ``geometricness'' could offer summaries, but it is infeasible when relevant attributes are not known in advance or easily described by words.

This summary can assist in auditing model bias and diagnosing how different model variants alter a specific concept as shown in Figure~\ref{fig:cross-models}. Beyond visualization, mental averages could serve as a tool to compress data into compact statistics for faster interpretation or privacy-preserving downstream training (e.g., \textit{dataset distillation}~\cite{su2024d,chan2025mgd,gu2024minimax}). Furthermore, they may act as regularizers and facilitate model debiasing. We leave these directions for future work.
\subsection{Computational Costs}\label{sec:b_efficeincy}

The current optimization process takes around 10 hours per concept on a single NVIDIA RTX 4080, with $K=1000$, $t_{\text{stop}} = 10$, and $N=300$. This remains our primary limitation. However, the three hyperparameters---the DMA number of optimization steps $N$, the stopping timestep $t_{\text{stop}}$, and the number of sampled latents $K$---enable a trade off between consistency and efficiency. 
The computation cost of DMA scales linearly with these hyperparameters.
As shown in Figure~\ref{fig:efficiency}, reducing $N$ from 300 to 5 speeds up DMA by approximately 60 times (taking only 10 minutes) and produces averages that already capture the same layout and composition. Increasing $N$ to 10 improves color fidelity, and $N=300$ captures fine details like leaf veins. Similarly, $t_{\text{stop}}$ acts as an early stopping point for noise optimization; without it, all diffusion steps must be optimized. As shown in Figure~\ref{fig:efficiency}, setting $t_{\text{stop}}$ to 10 doubles the optimization speed with only minor degradation, while lowering $K$ also helps accelerate DMA but may result in inconsistency across disjoint sets (Sec.~\ref{sec:effect_num_samples}). 

Ultimately, the choices for these hyperparameters depend on the specific task and the extent to which consistency can be traded for efficiency. 
For instance, a coarse-grained alignment of textures might be sufficient for general visualization (e.g., dog visualization can tolerate fur misalignment), whereas tasks like plant species recognition may require fine alignment of intricate features like leaf veins.



\setlength{\fboxsep}{0pt}      
\setlength{\fboxrule}{0.6pt}  
\newcommand{\boxedimg}[2]{
\begin{tikzpicture}
  \node[anchor=south west, inner sep=0] (img) at (0,0)
    {\includegraphics[width=#1]{#2}};
  \draw[red, thick] (img.south west) rectangle (img.north east);
\end{tikzpicture}
}
\begin{figure*}[t]
    \centering
    \renewcommand{\arraystretch}{1.2}
    \setlength{\tabcolsep}{1pt}
    \footnotesize
    \begin{tabular}{cccccc}
    
        $t_{\text{stop}}$=1 & $t_{\text{stop}}$=2 & $t_{\text{stop}}$=3 & $t_{\text{stop}}$=4 & $t_{\text{stop}}$=5 & $t_{\text{stop}}$=10 \\
        \includegraphics[width=0.15\linewidth]{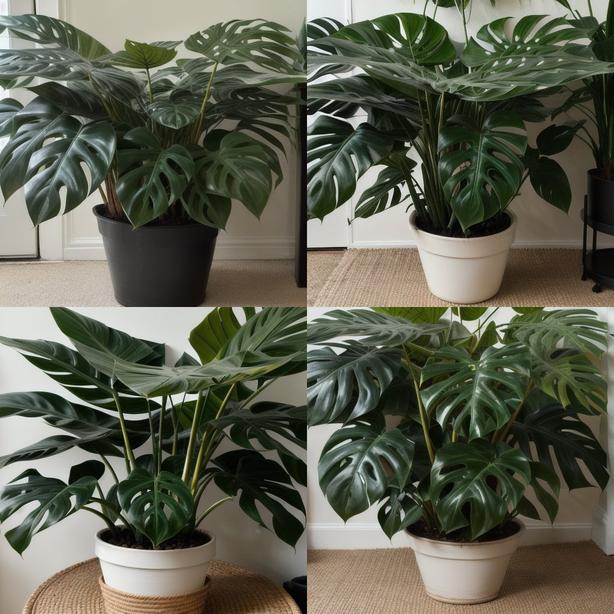} &
        \includegraphics[width=0.15\linewidth]{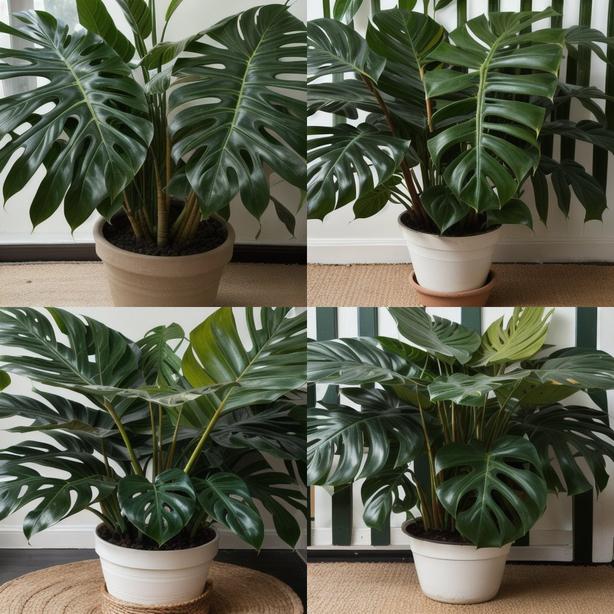} &
        \includegraphics[width=0.15\linewidth]{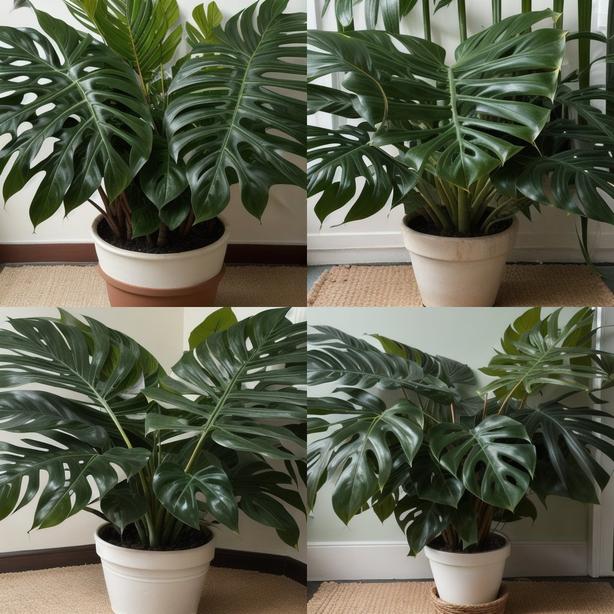} &
        \includegraphics[width=0.15\linewidth]{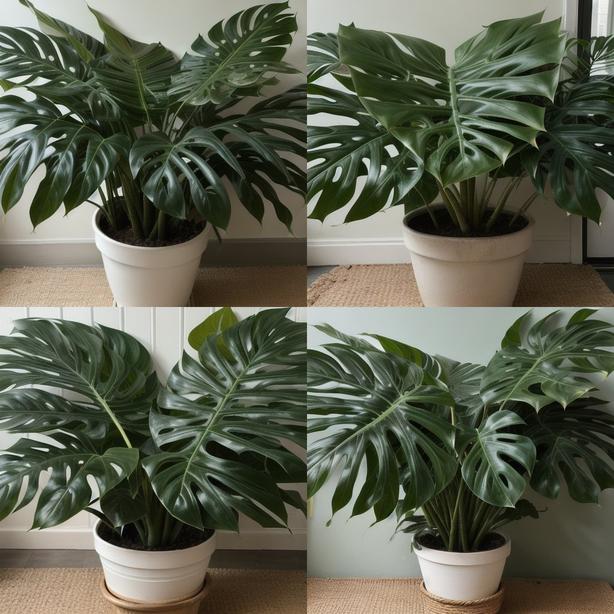} &
        \includegraphics[width=0.15\linewidth]{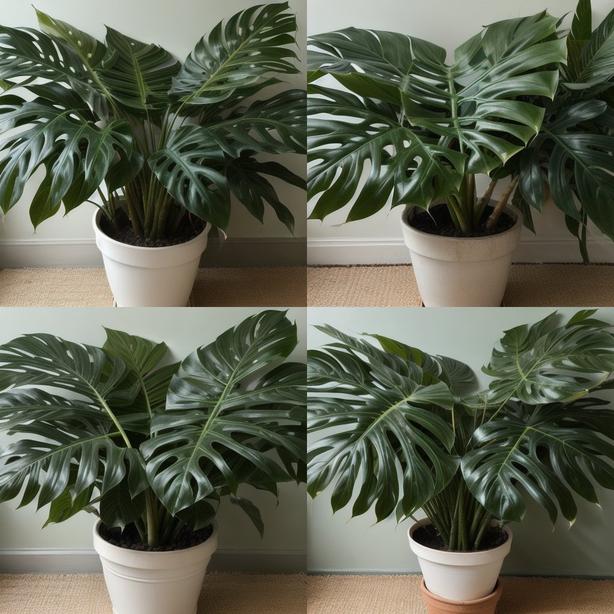} &
        \boxedimg{0.155\linewidth}{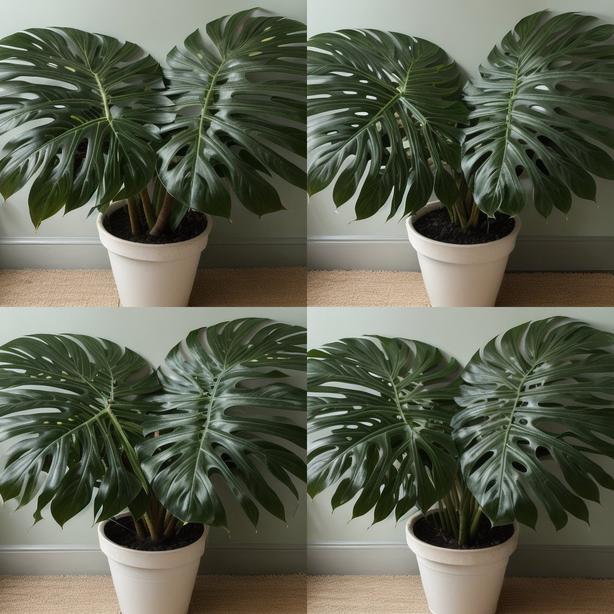}\\[-3pt]
        0.053 & 0.036 & 0.033 & 0.026 & 0.021 &\textbf{0.009}\\[5pt]
        
        $N=1$ & $N=5$ & $N=10$ & $N=100$ & $N=300$ & $N=500$ \\
        \includegraphics[width=0.15\linewidth]{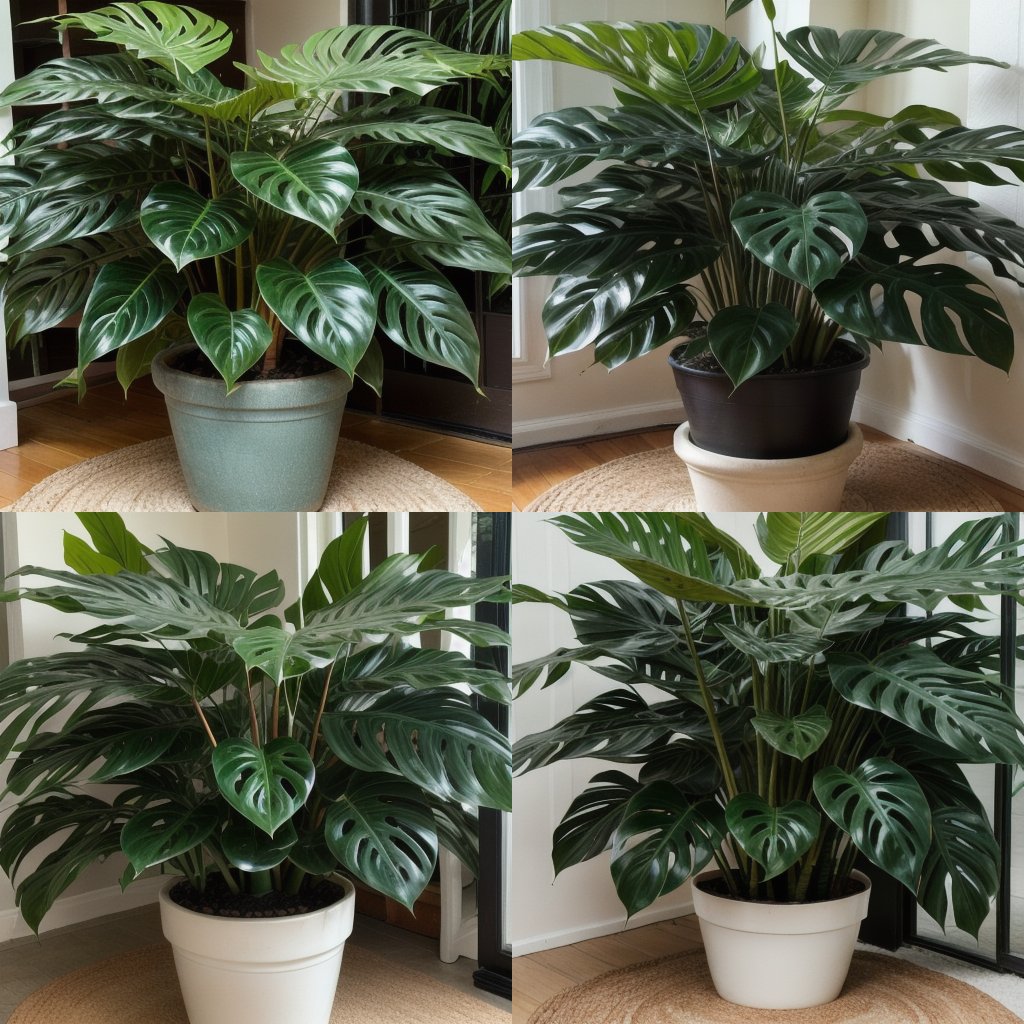} &
        \includegraphics[width=0.15\linewidth]{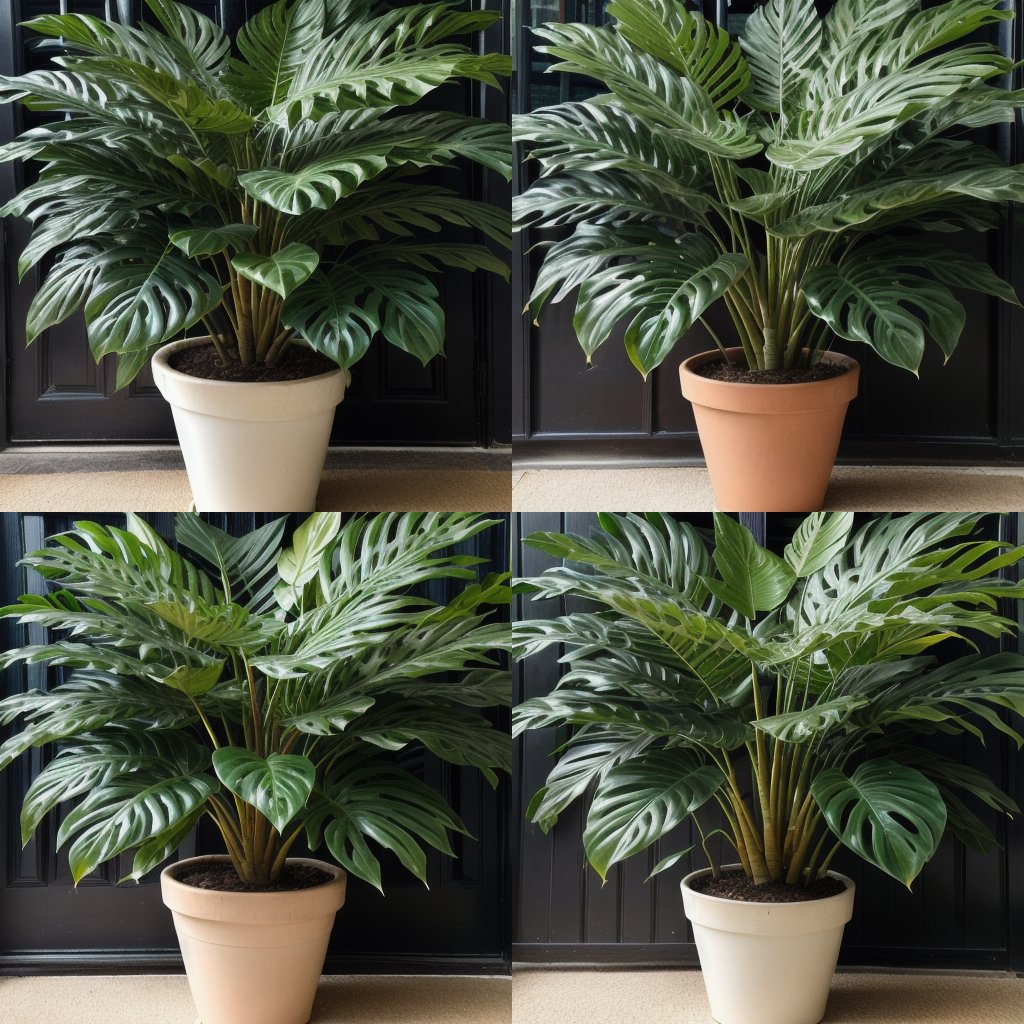} &
        \includegraphics[width=0.15\linewidth]{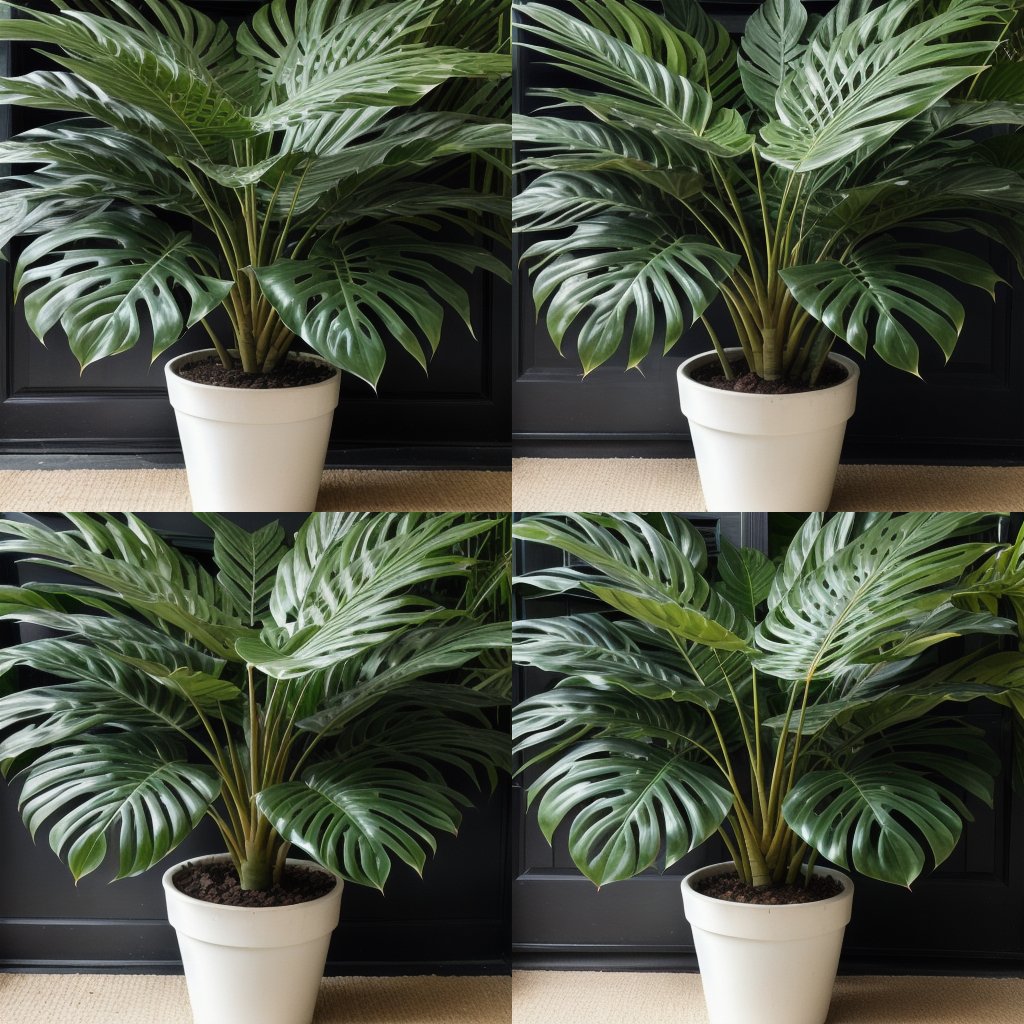} &
        \includegraphics[width=0.15\linewidth]{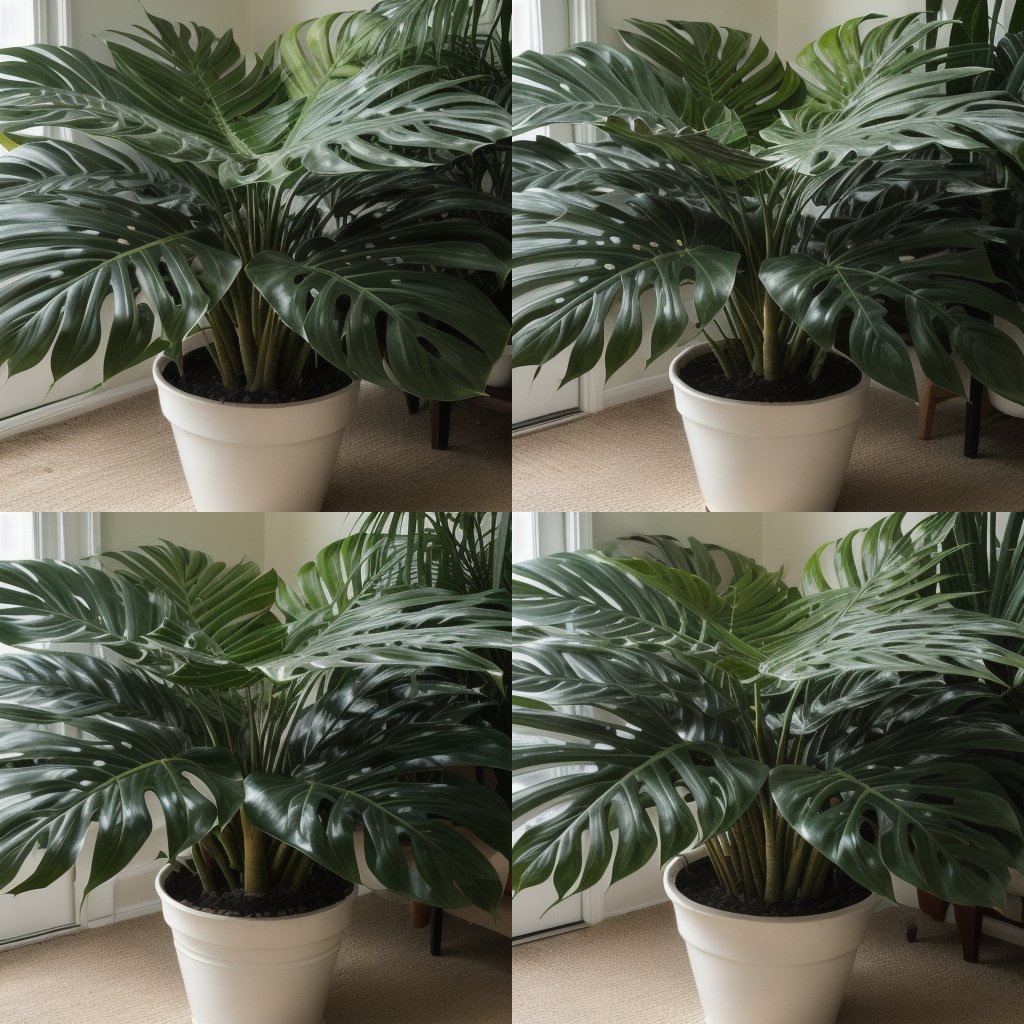} &
        \boxedimg{0.155\linewidth}{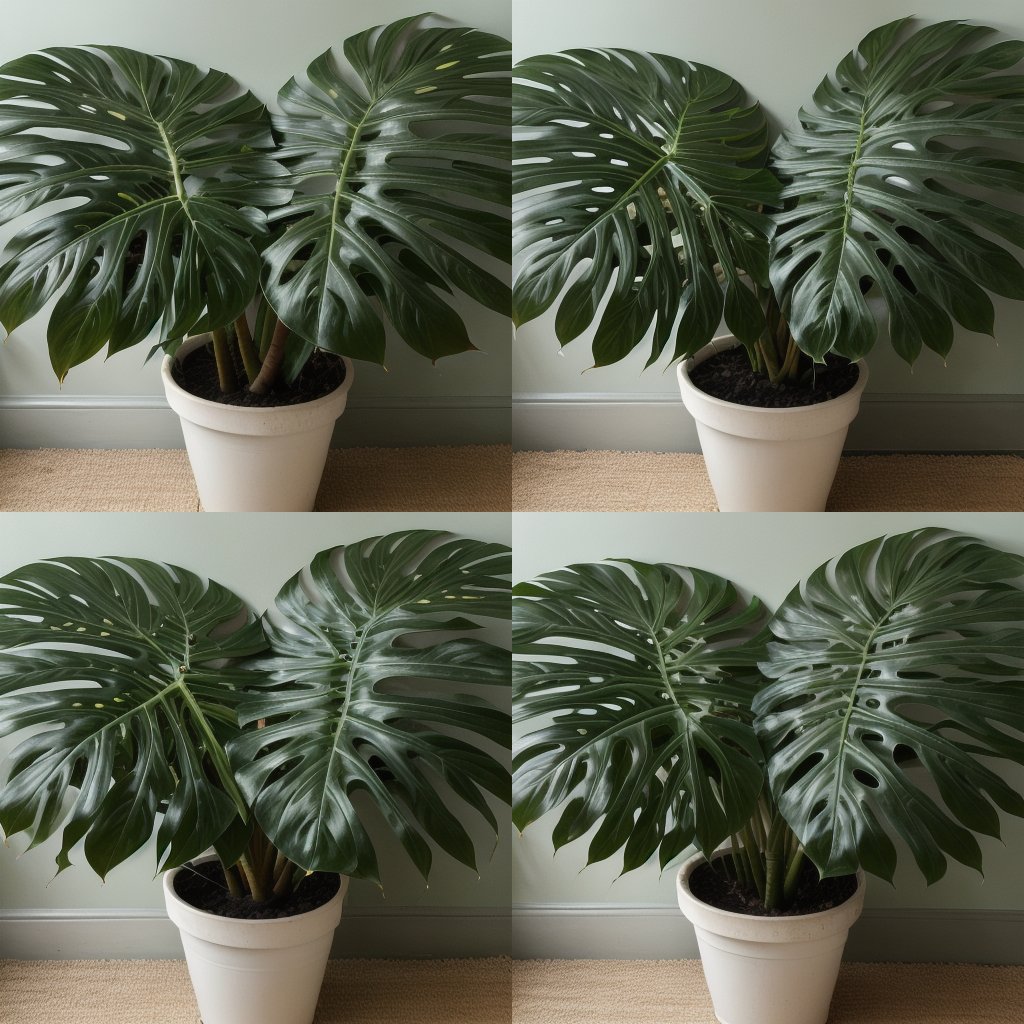} &
        \includegraphics[width=0.15\linewidth]{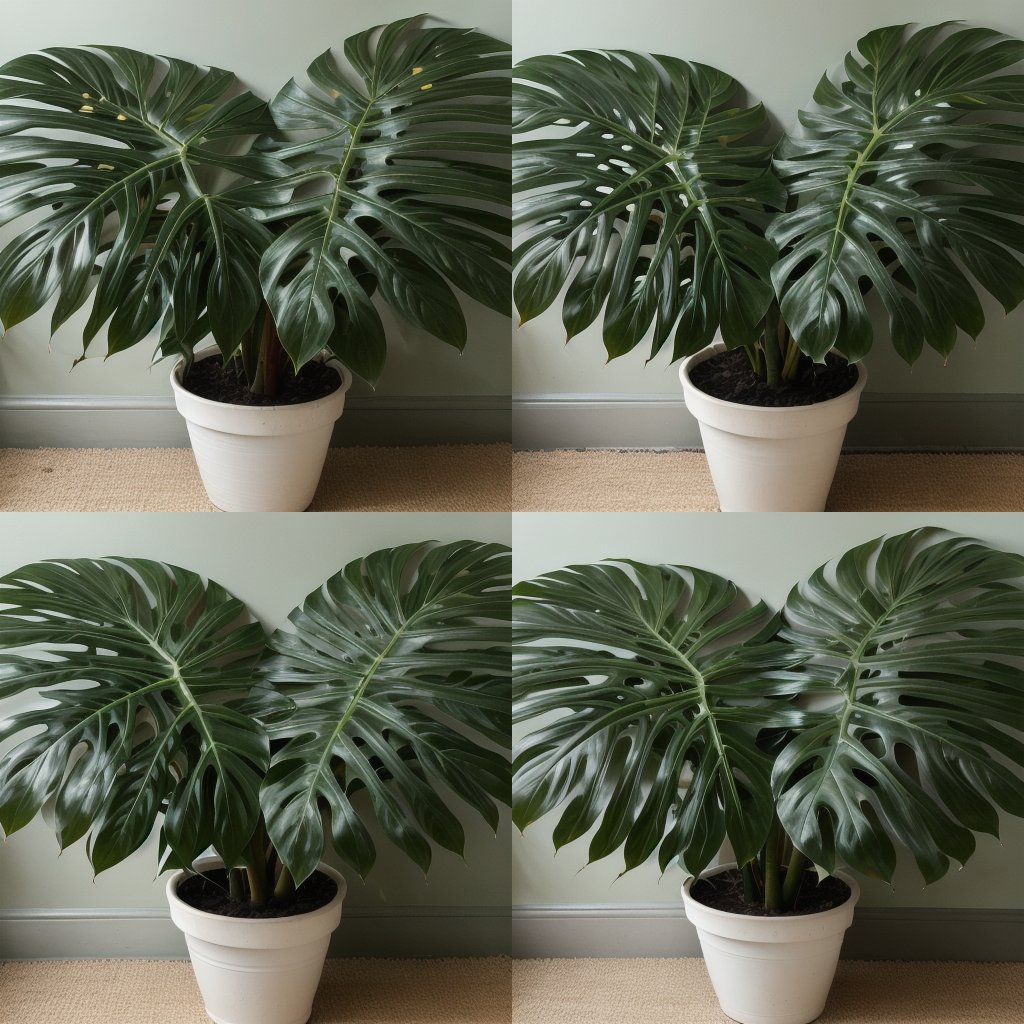}\\[-3pt]
        0.045 & 0.021 & 0.027 & 0.015 & \textbf{0.009} & 0.010 \\

    \end{tabular}
    \vspace{0pt}
    \caption{\textbf{Efficiency trade-off.} We vary $t_{\text{stop}}$ and $N$, and show consistency scores (↓). \setlength{\fboxsep}{0.5pt} \fcolorbox{red}{white}{Red} is our default setting. Lower $t_{\text{stop}}$ and $N$ trade consistency for speed and are up to the user.}
    \label{fig:efficiency}
\end{figure*}

\subsection{Limitations of Mode Discovery and Averages}\label{sec:b_mode}





In mode discovery (Section~\ref{sec:mode_discov}), our method focuses on computing the average of each cluster once they are established. This clustering is treated as a modular preprocessing step, currently using standard techniques like GMM on CLIP features, that can be independently replaced or improved. We observe that because different algorithms partition the space differently, the resulting mode averages are inherently dependent on the specific clustering method.

While applying LoRA on each cluster (Section~\ref{sec:mode_discov}) helps improve the model's conditioning of specific subconcepts, it modifies the original model being probed and may be less ideal as a diagnostic tool for the original model. Nonetheless, since our LoRA is trained exclusively on generated samples from the original model, it avoids introducing biases from external data.

\subsection{Discussion on Representativeness Score}\label{sec:b_metric}

Formally, the average of a set of images is defined as the point that minimizes the mean squared distance to all images in the set. Based on this principle, we compute distances within semantic latent spaces such as CLIP, DreamSim, and LPIPS to quantify the optimality of the average, or the Representativeness Score. However, a good score in these spaces could only reflect the quality of the average within those specific spaces. We suggest that identifying the most suitable latent space for evaluating the perceptual and semantic fidelity of average images remains an interesting direction for future work.




\section{Ablation Studies}
\subsection{Progressive Mean Alignment}
In DMA, the mean $h$-space activation at each timestep is computed progressively using the current set of latents optimized from the previous timestep. As a baseline, we consider an alternative approach that precomputes the means for all timesteps using $K$ \emph{independent} full denoising processes: we completely denoise each of the $K$ latents, extract their $h$-space activations at every timestep, and average them to obtain the per-timestep mean. DMA is then rerun from the start using the original $K$ latents from $\mathcal{N}(0, \mathbf{I})$, but with the precomputed means used as alignment targets during noise optimization.

\begin{figure*}[t]
    \centering
    \renewcommand{\arraystretch}{1.2}
    \setlength{\tabcolsep}{1pt}
    \footnotesize
    \begin{tabular}{c cccc} 
        & \textbf{Timestep 5} & \textbf{Timestep 10} & \textbf{Timestep 15} & \textbf{Timestep 20} \\
        
        \raisebox{3.4\height}{\rotatebox[origin=c]{90}{\text{Baseline}}} &
        \includegraphics[width=0.23\linewidth]{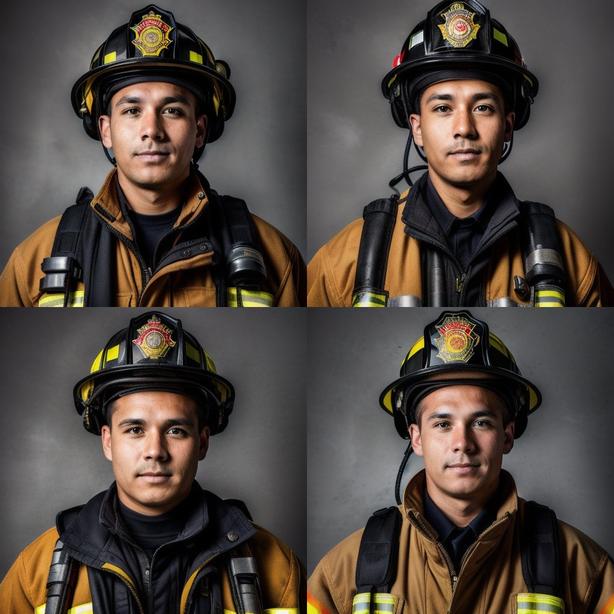} &
        \includegraphics[width=0.23\linewidth]{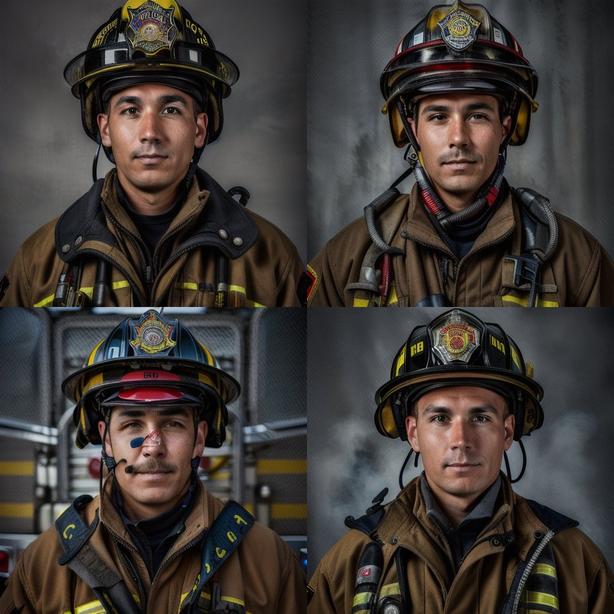} &
        \includegraphics[width=0.23\linewidth]{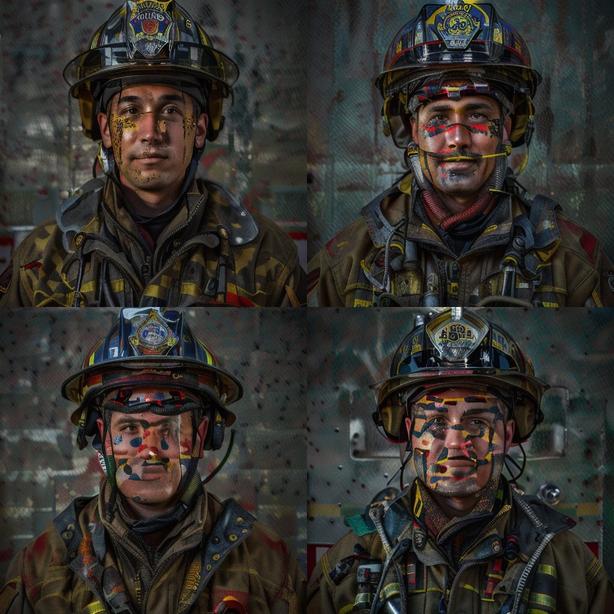} &
        \includegraphics[width=0.23\linewidth]{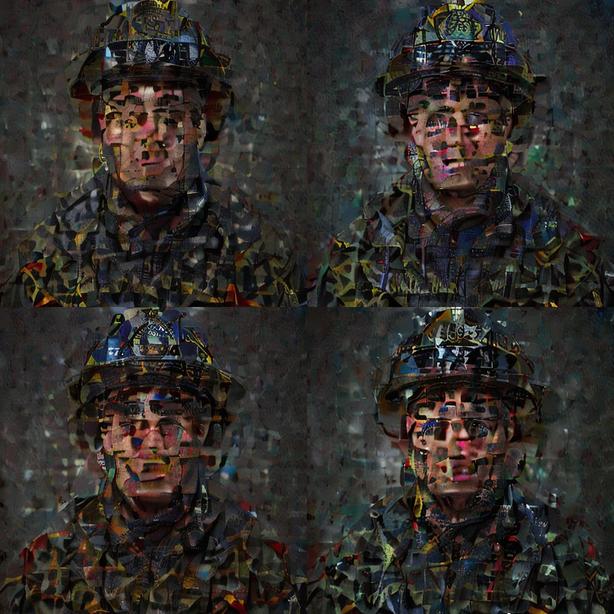} \\

        \raisebox{5.2\height}{\rotatebox[origin=c]{90}{\text{Ours}}} &
        \includegraphics[width=0.23\linewidth]{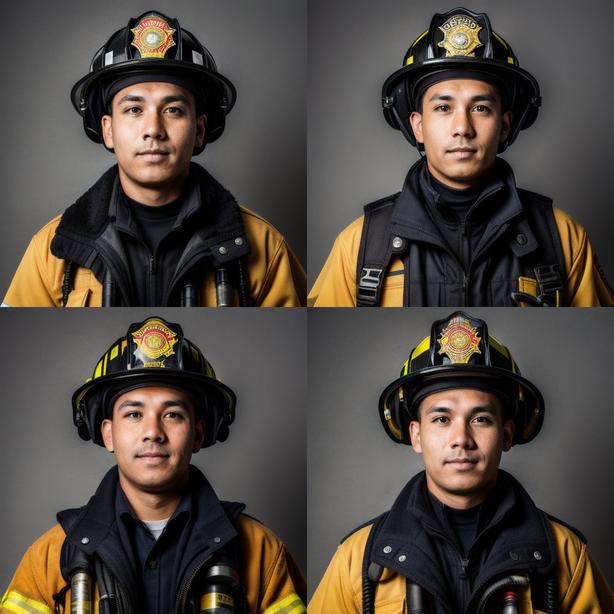} &
        \includegraphics[width=0.23\linewidth]{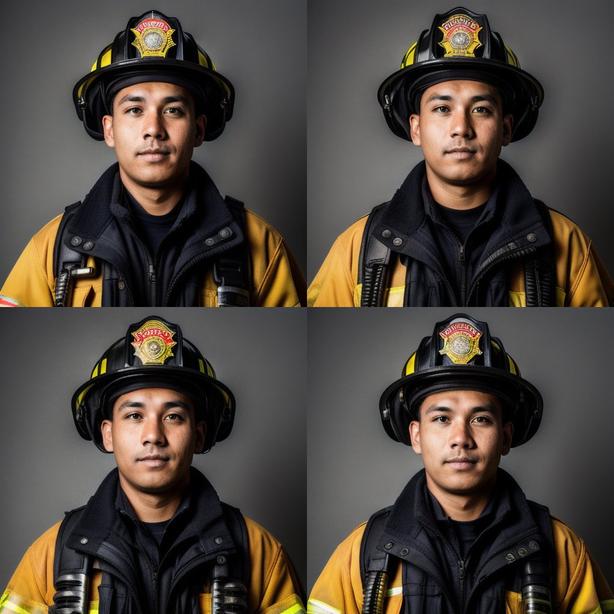} &
        \includegraphics[width=0.23\linewidth]{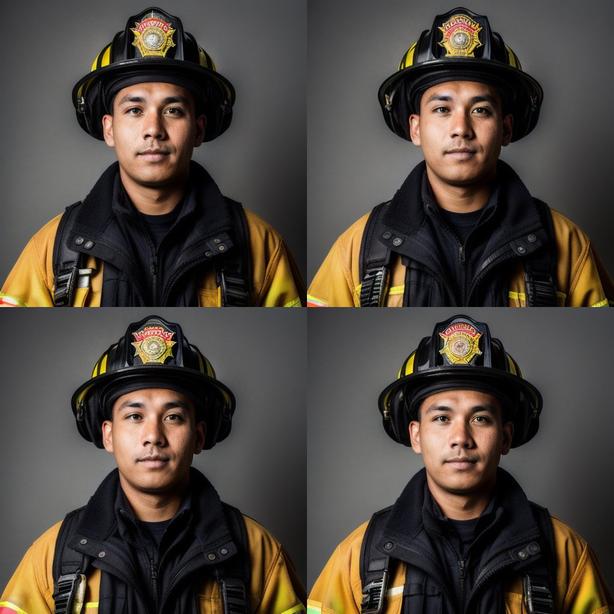} &
        \includegraphics[width=0.23\linewidth]{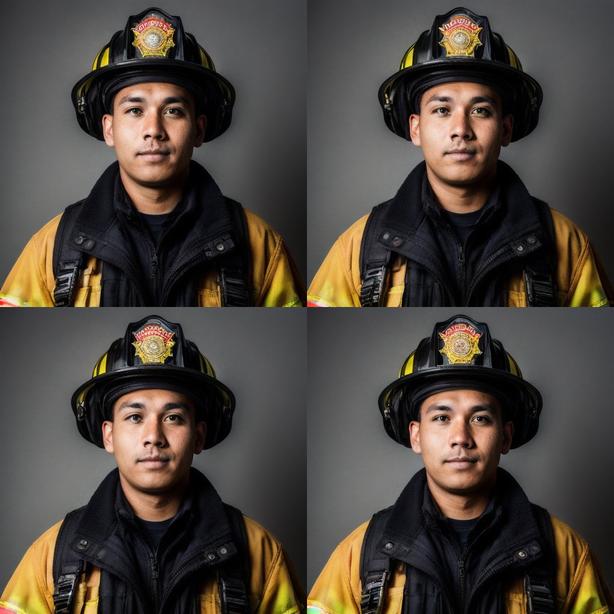} \\
    \end{tabular}
    \vspace{0pt}
    \caption{\textbf{Comparison with the pre-computed activation baseline.} The pre-computed baseline (top) degrades over timesteps, while DMA (bottom) remains stable and consistent.}
    \label{fig:ab1_precomputed_mean}
\end{figure*}

As shown in Figure~\ref{fig:ab1_precomputed_mean}, this baseline produces artifacts and increasingly corrupted results as optimization progresses toward later timesteps (i.e., closer to the final output). In contrast, our method yields high-quality images, highlighting its effectiveness.


\subsection{Multi-Timestep Optimization}
\begin{figure*}[t]
    \centering
    \footnotesize
    \makebox[0.24\linewidth][c]{\textbf{Timestep 1}}\hfill
    \makebox[0.24\linewidth][c]{\textbf{Timestep 10}}\hfill
    \makebox[0.24\linewidth][c]{\textbf{Timestep 20}}\hfill
    \makebox[0.24\linewidth][c]{\textbf{Ours}}
    \begin{subfigure}{0.24\linewidth}
        \centering
        \includegraphics[width=\linewidth]{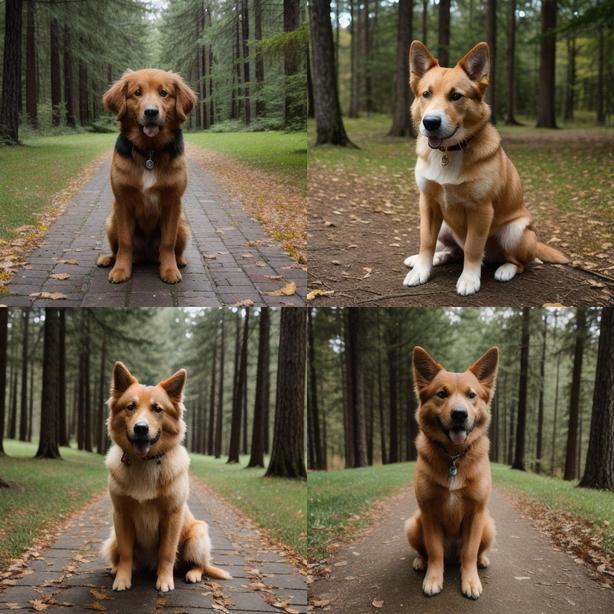}
    \end{subfigure}\hfill
    \begin{subfigure}{0.24\linewidth}
        \centering
        \includegraphics[width=\linewidth]{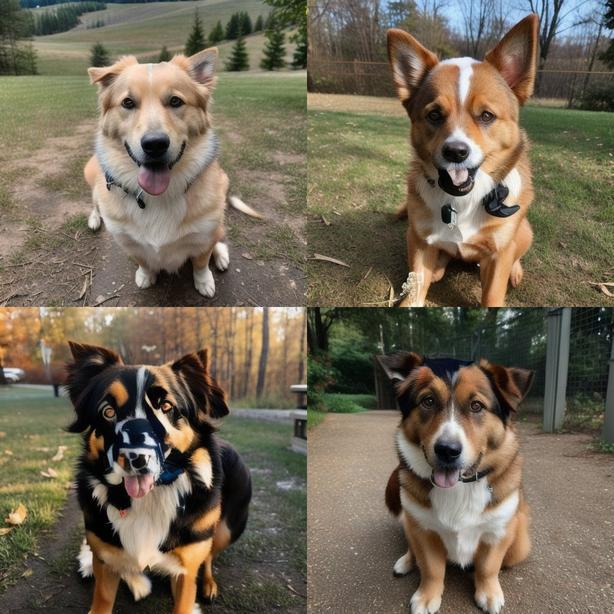}
    \end{subfigure}\hfill
    \begin{subfigure}{0.24\linewidth}
        \centering
        \includegraphics[width=\linewidth]{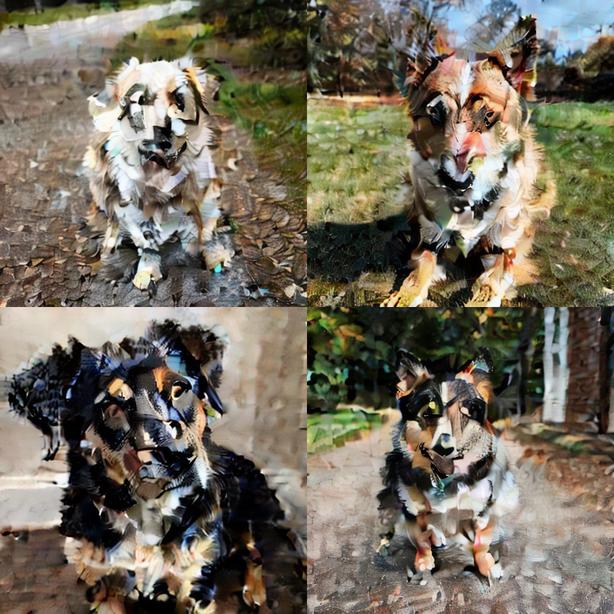}
    \end{subfigure}\hfill
    \begin{subfigure}{0.24\linewidth}
        \centering
        \includegraphics[width=\linewidth]{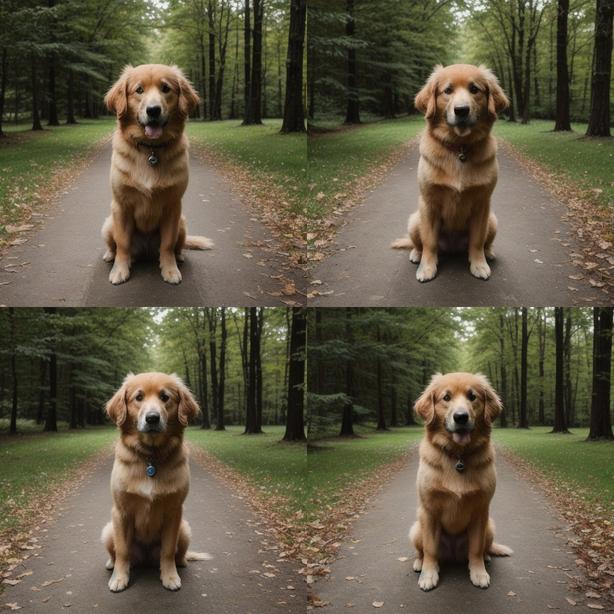}
    \end{subfigure}

    \vspace{4pt}
    \caption{\textbf{Comparison with the single-timestep optimization baseline.} This baseline produces inconsistent results and introduces noticeable artifacts if optimization is done in later timesteps. 
    In contrast, our method maintains both semantic consistency and visual quality.}
    \label{fig:ab2_single_opt}
\end{figure*}

We consider a baseline that optimizes noise latents only at a single timestep, while all other timesteps follow standard inference. As shown in Fig.~\ref{fig:ab2_single_opt}, single-timestep optimization produces inconsistent results regardless of the chosen timestep, and optimizing at later timesteps introduces artifacts and degrades output quality.
In contrast, DMA yields more consistent and high-quality results.
\subsection{Noise Optimization}
\begin{figure*}[t]
    \centering
    \renewcommand{\arraystretch}{1.2}
    \setlength{\tabcolsep}{1pt}
    \footnotesize
    \begin{tabular}{c ccccc} 
        & \shortstack{\scriptsize \textbf{Standard Denoising Samples} \\ ($t_\text{stop}=0$)} & $t_\text{stop}=5$ & $t_\text{stop}=10$ & $t_\text{stop}=15$ & $t_\text{stop}=20$ \\
        
        \raisebox{5.7\height}{\rotatebox[origin=c]{90}{\text{car}}} &
        \includegraphics[width=0.175\linewidth]{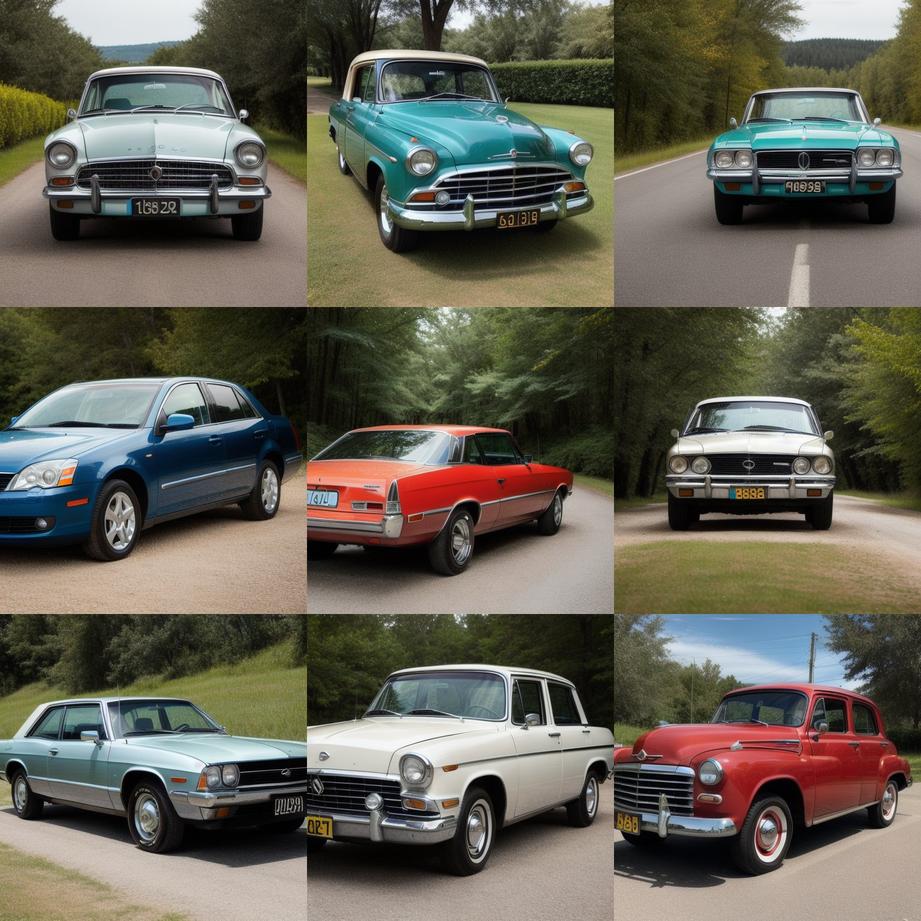} &
        \includegraphics[width=0.175\linewidth]{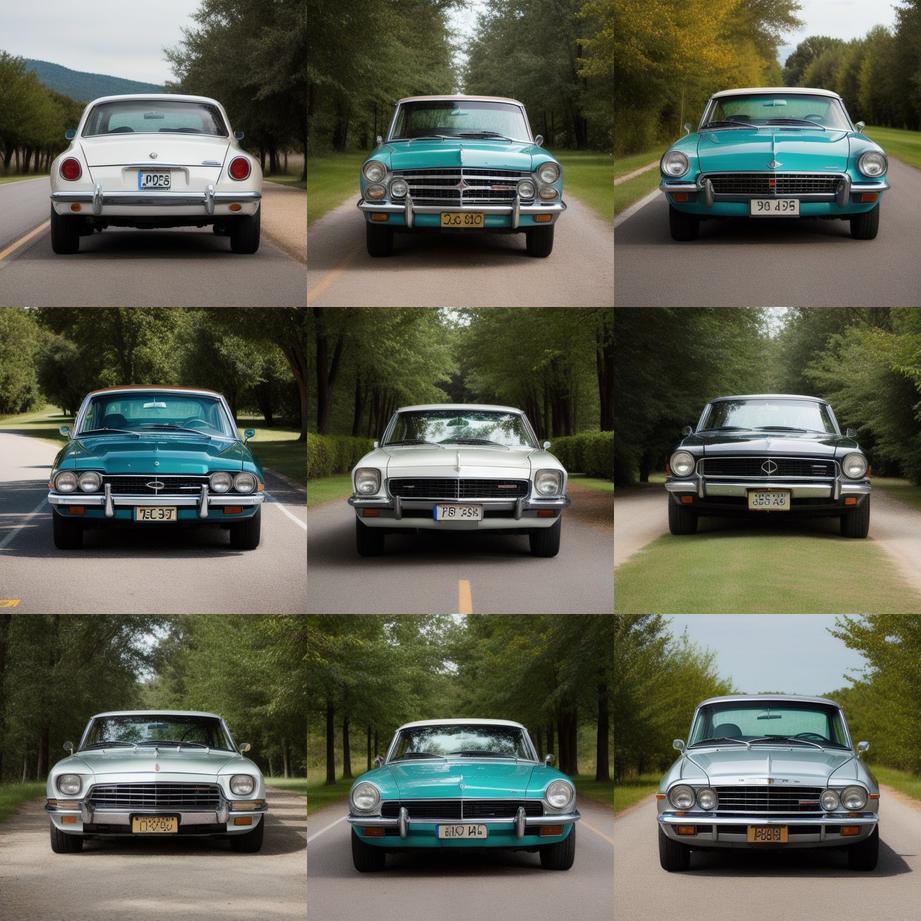} &
        \includegraphics[width=0.175\linewidth]{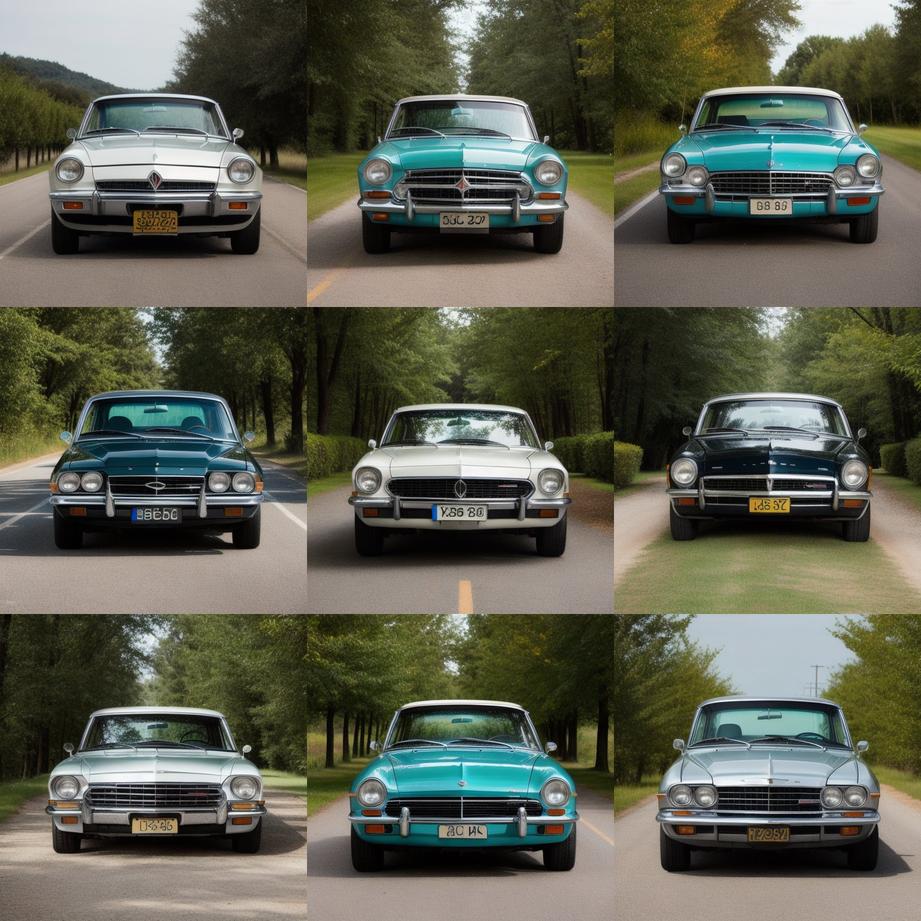} &
        \includegraphics[width=0.175\linewidth]{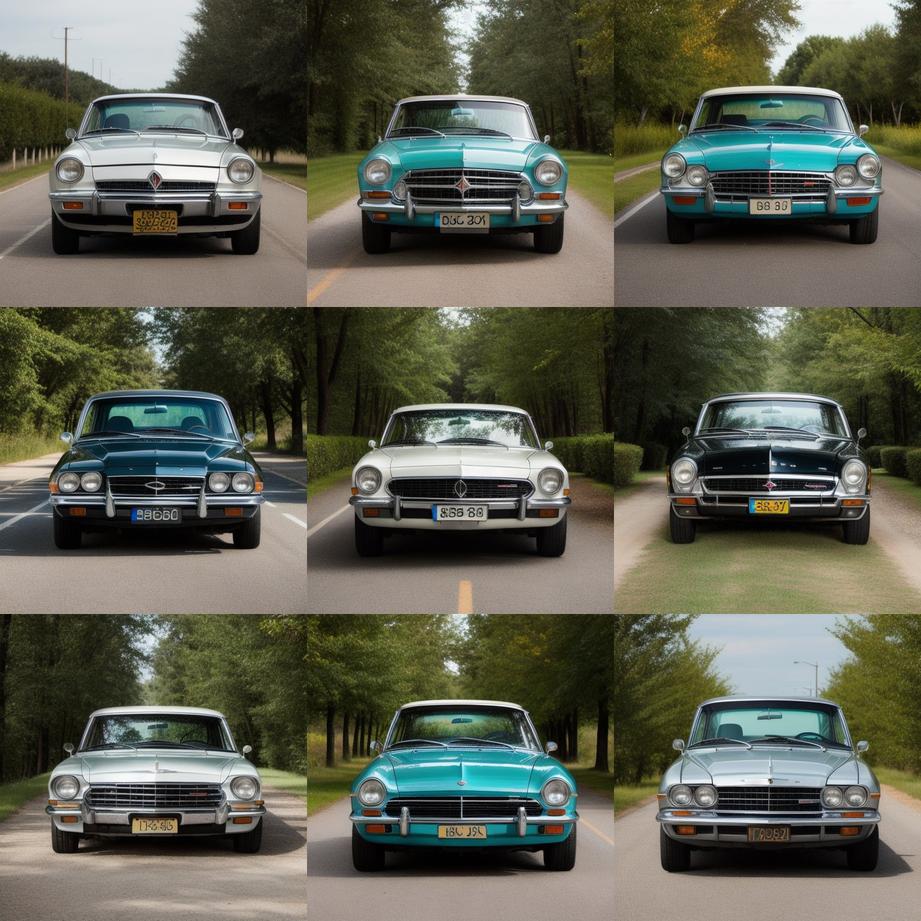} &
        \includegraphics[width=0.175\linewidth]{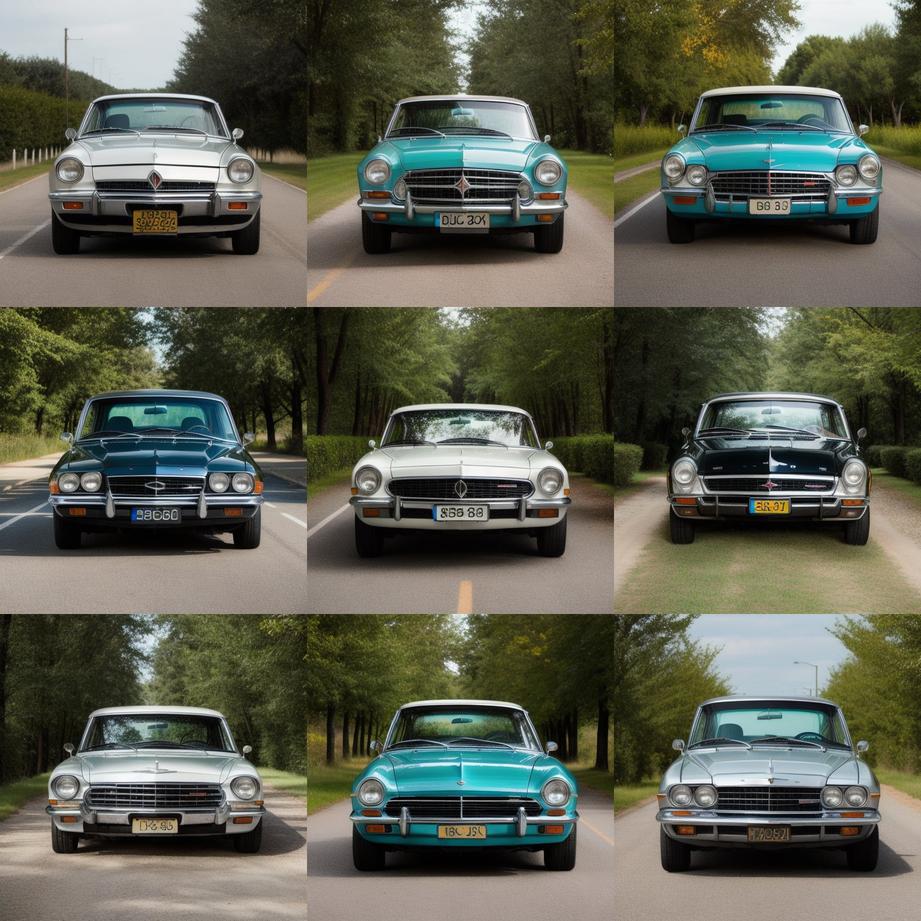} \\

        \raisebox{3\height}{\rotatebox[origin=c]{90}{\text{teacher}}} &
        \includegraphics[width=0.175\linewidth]{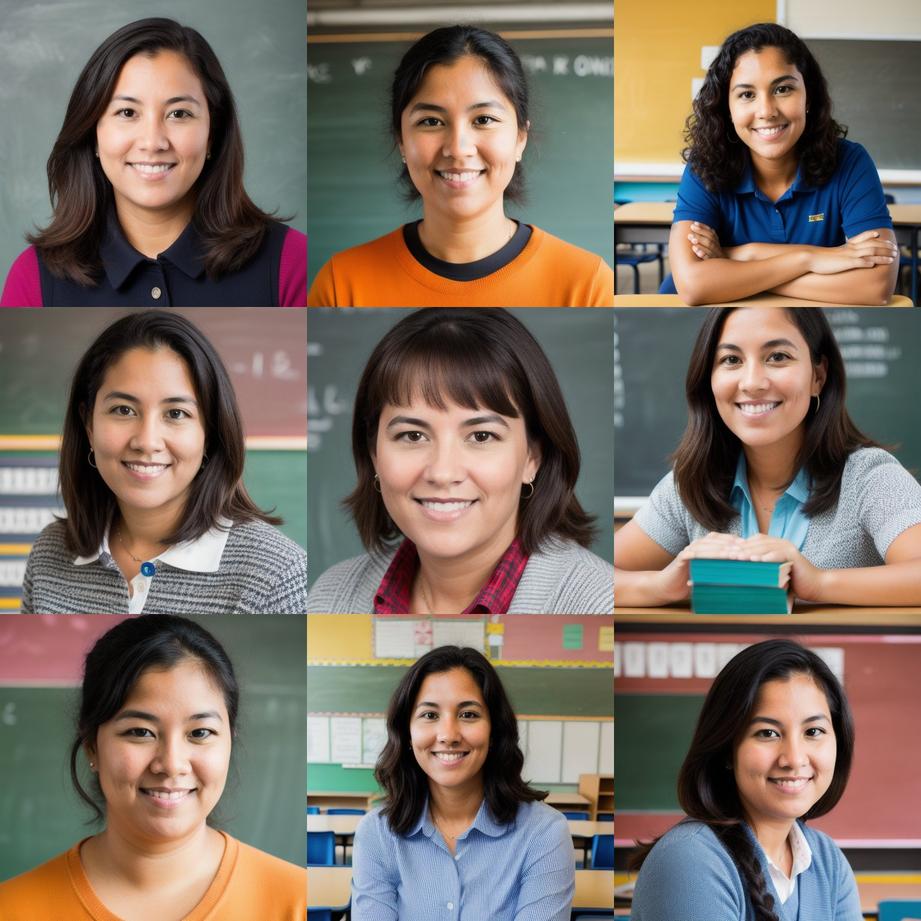} &
        \includegraphics[width=0.175\linewidth]{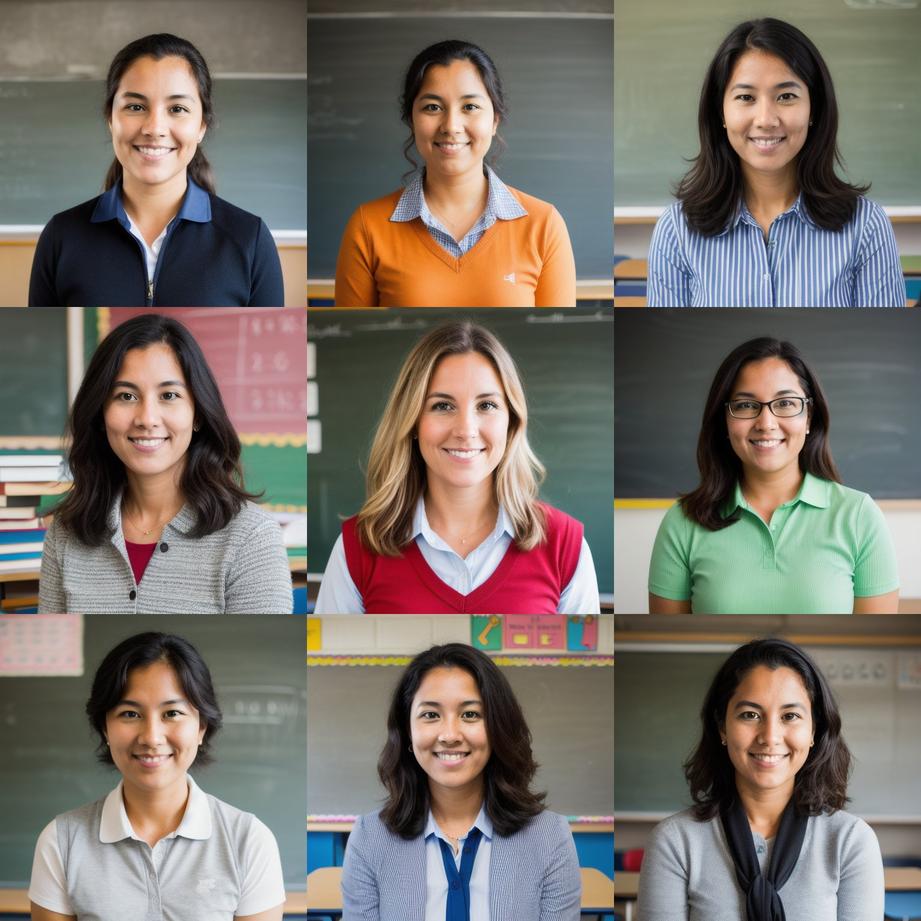} &
        \includegraphics[width=0.175\linewidth]{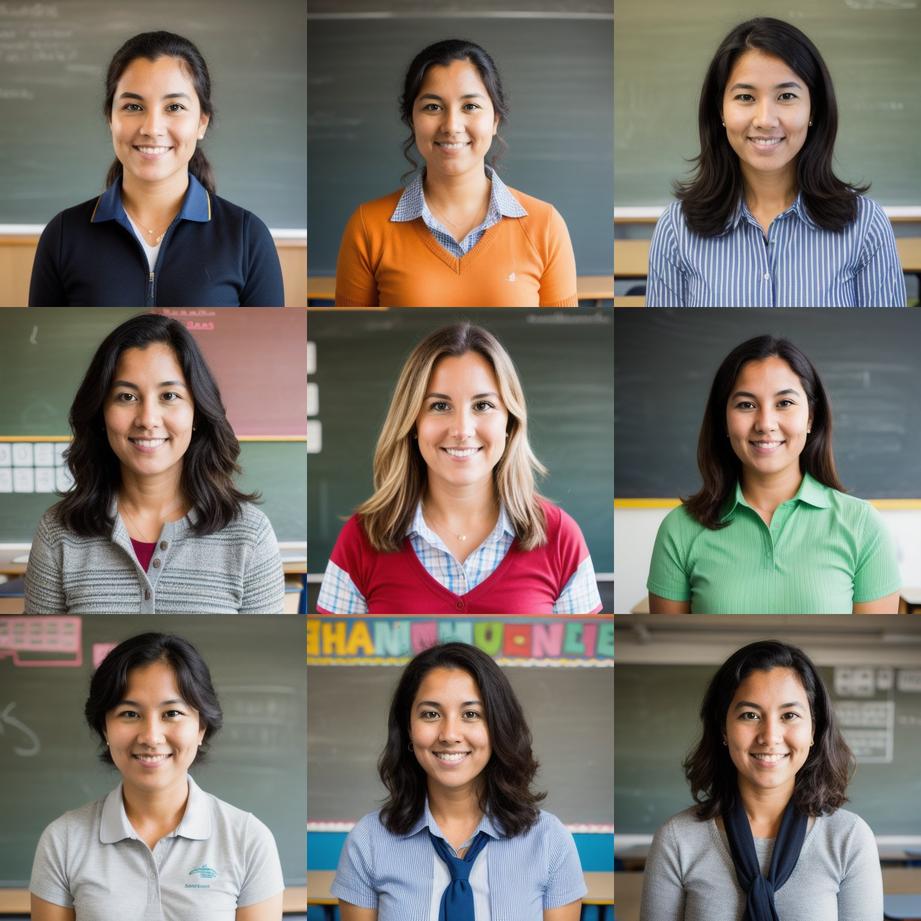} &
        \includegraphics[width=0.175\linewidth]{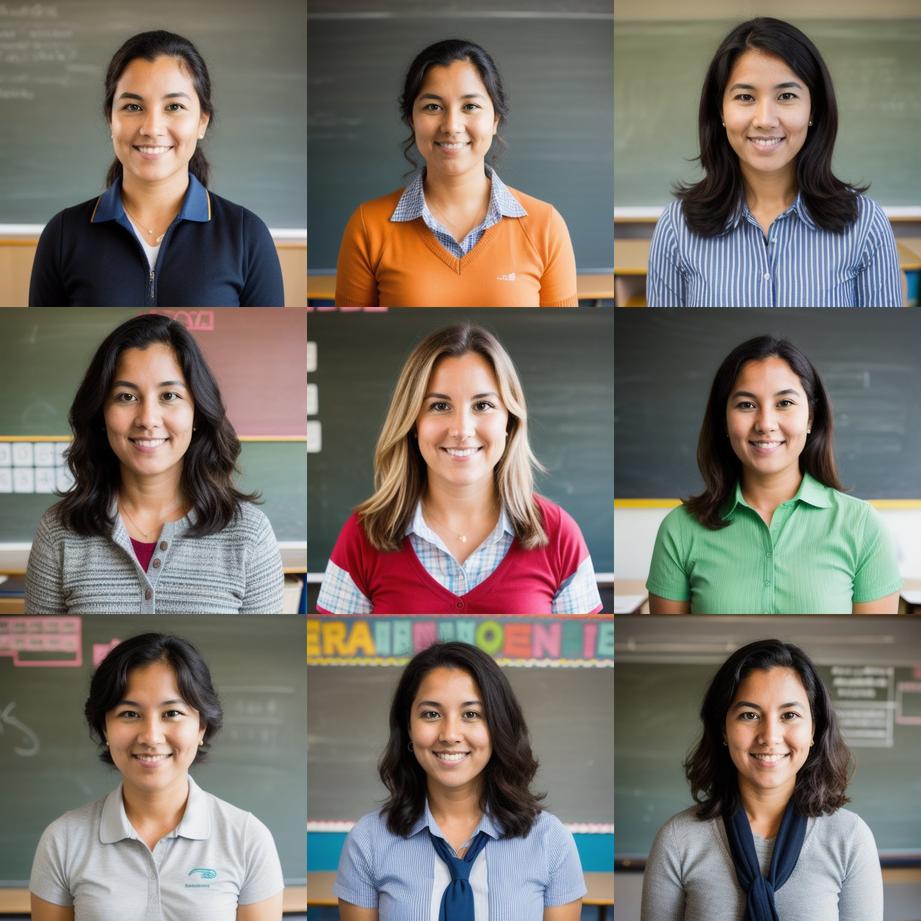} &
        \includegraphics[width=0.175\linewidth]{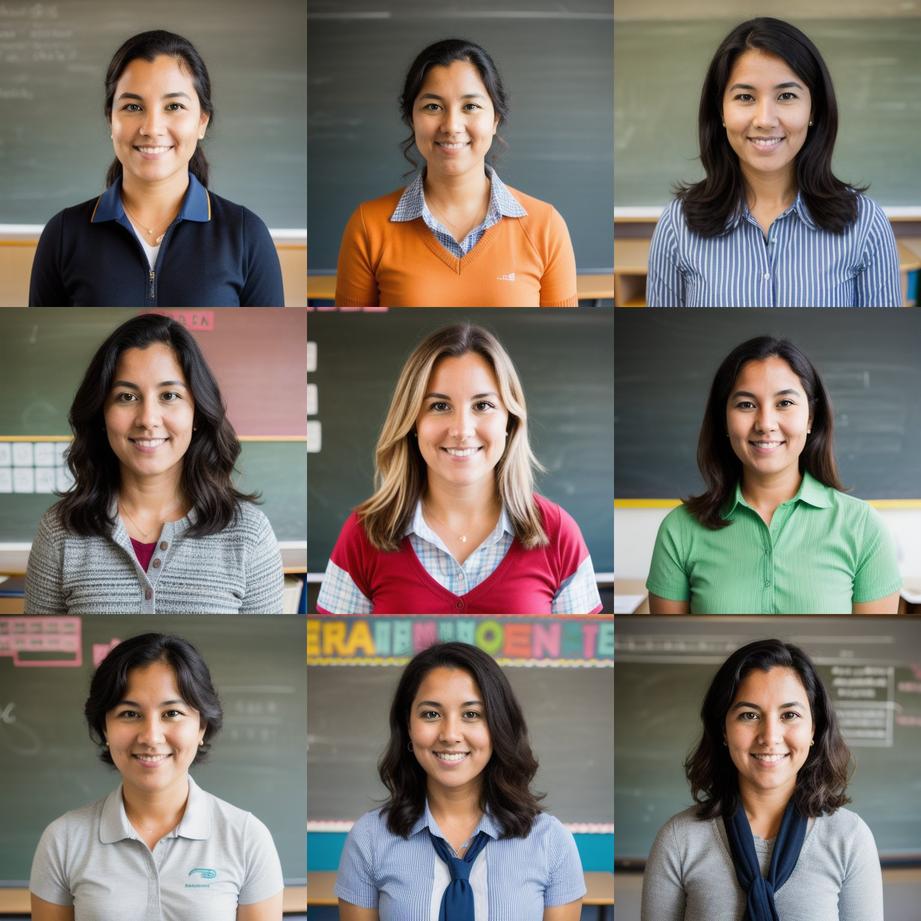} \\

        \raisebox{5.5\height}{\rotatebox[origin=c]{90}{\text{dog}}} &
        \includegraphics[width=0.175\linewidth]{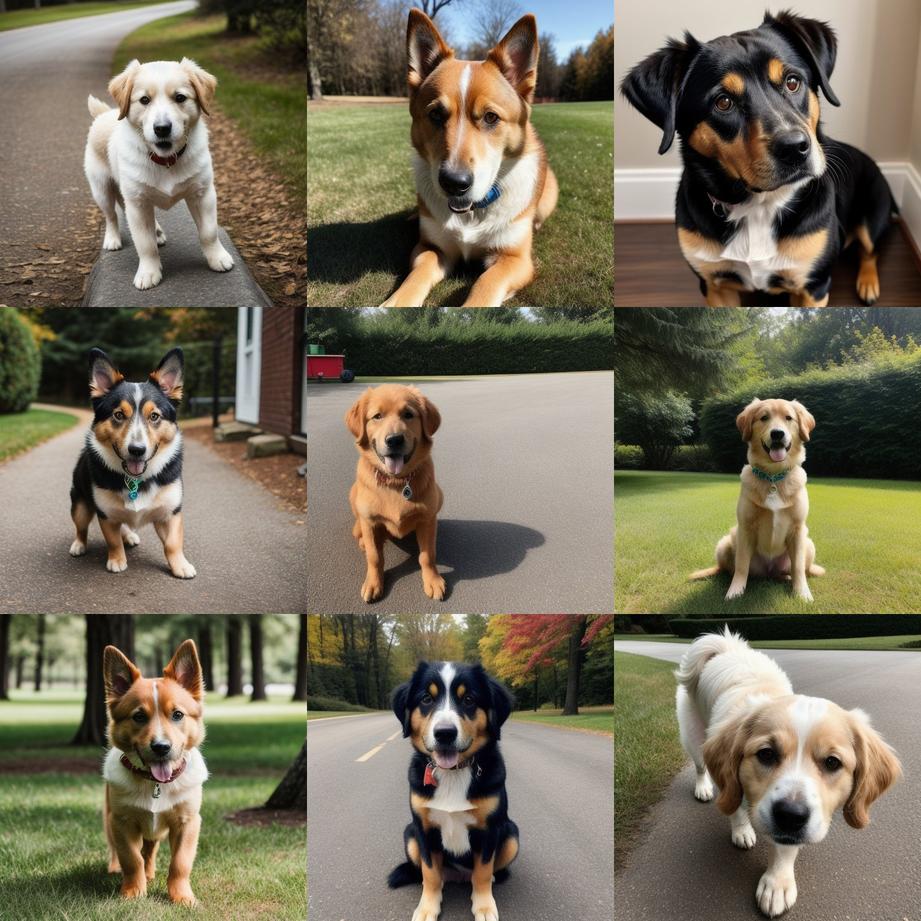} &
        \includegraphics[width=0.175\linewidth]{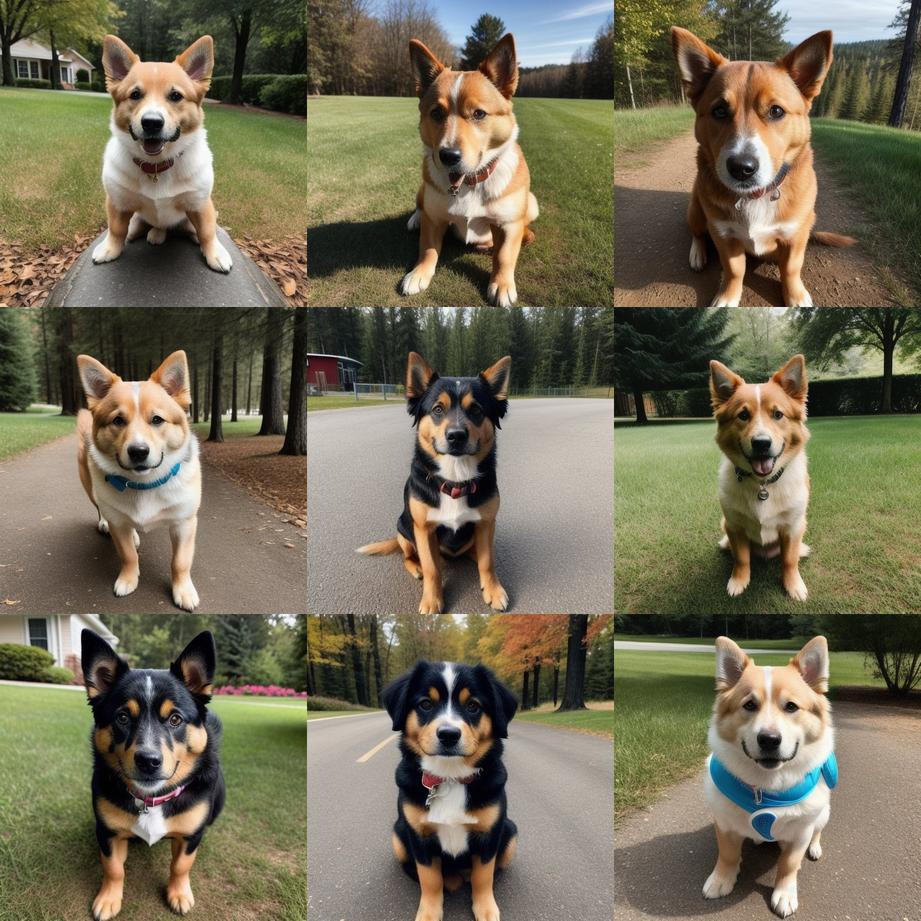} &
        \includegraphics[width=0.175\linewidth]{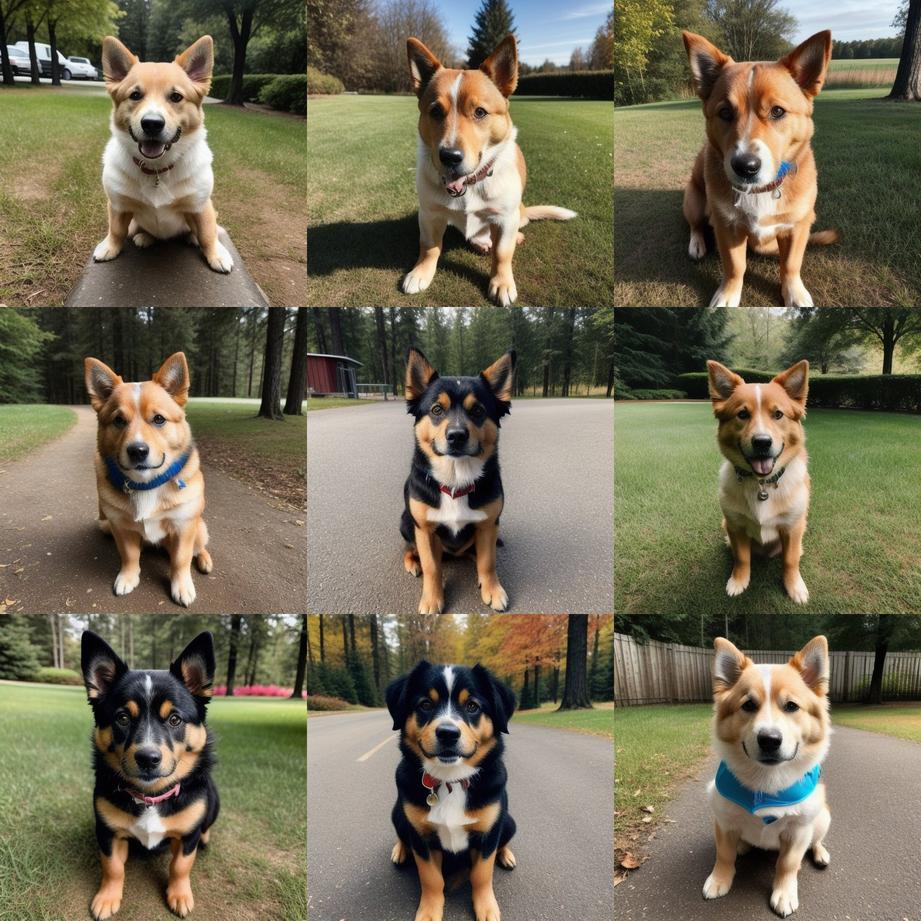} &
        \includegraphics[width=0.175\linewidth]{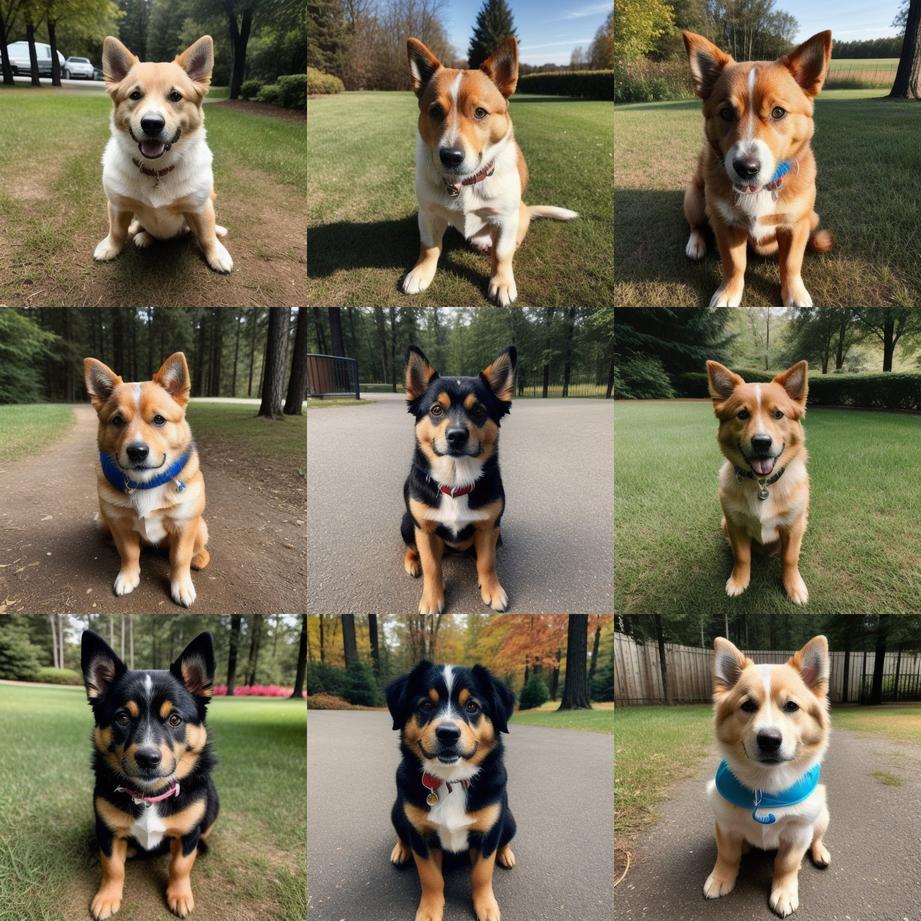} &
        \includegraphics[width=0.175\linewidth]{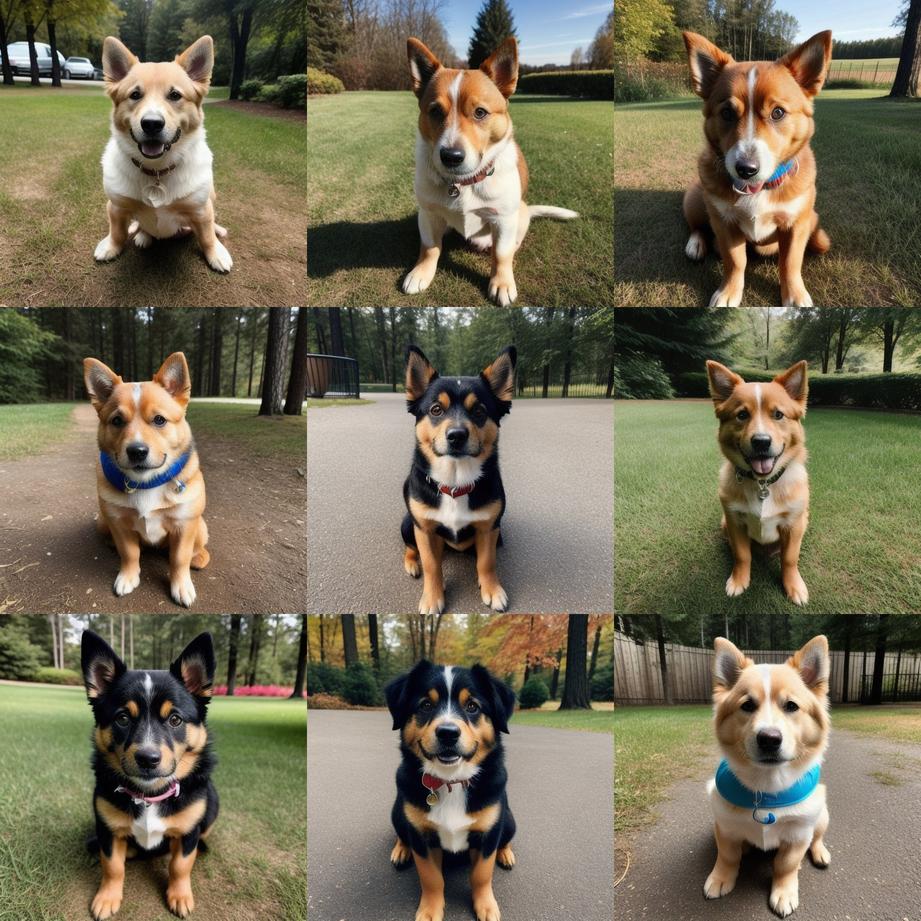} \\

        \raisebox{3.4\height}{\rotatebox[origin=c]{90}{\text{castle}}} &
        \includegraphics[width=0.175\linewidth]{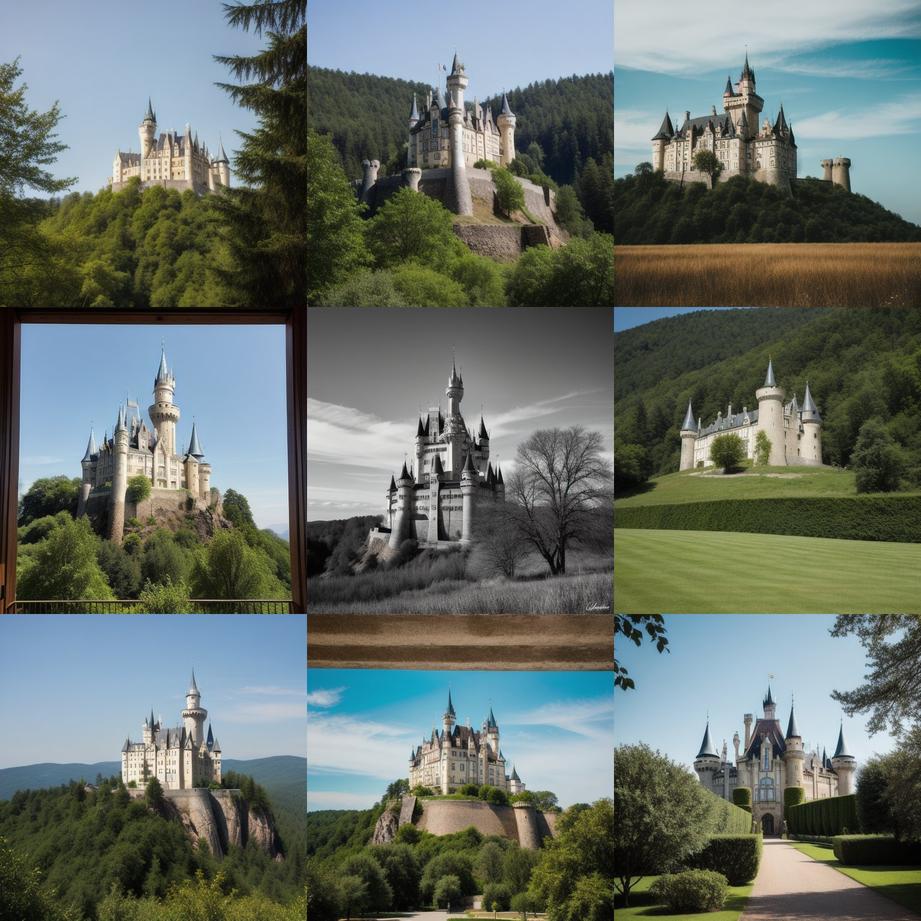} &
        \includegraphics[width=0.175\linewidth]{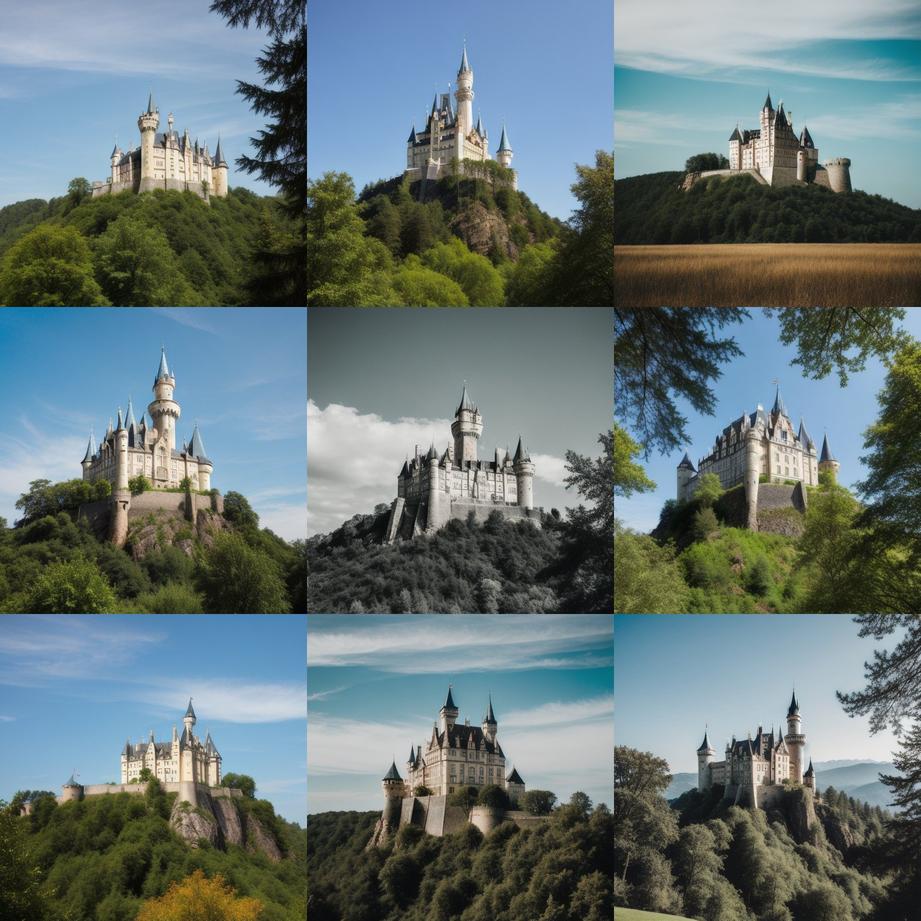} &
        \includegraphics[width=0.175\linewidth]{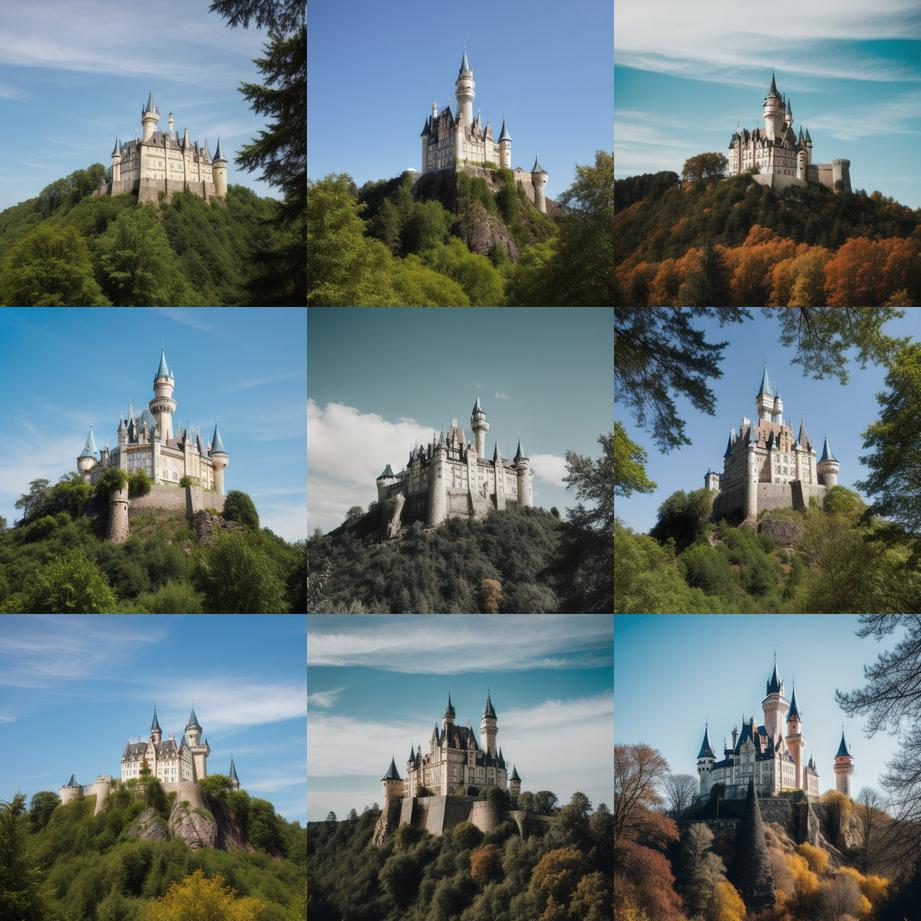} &
        \includegraphics[width=0.175\linewidth]{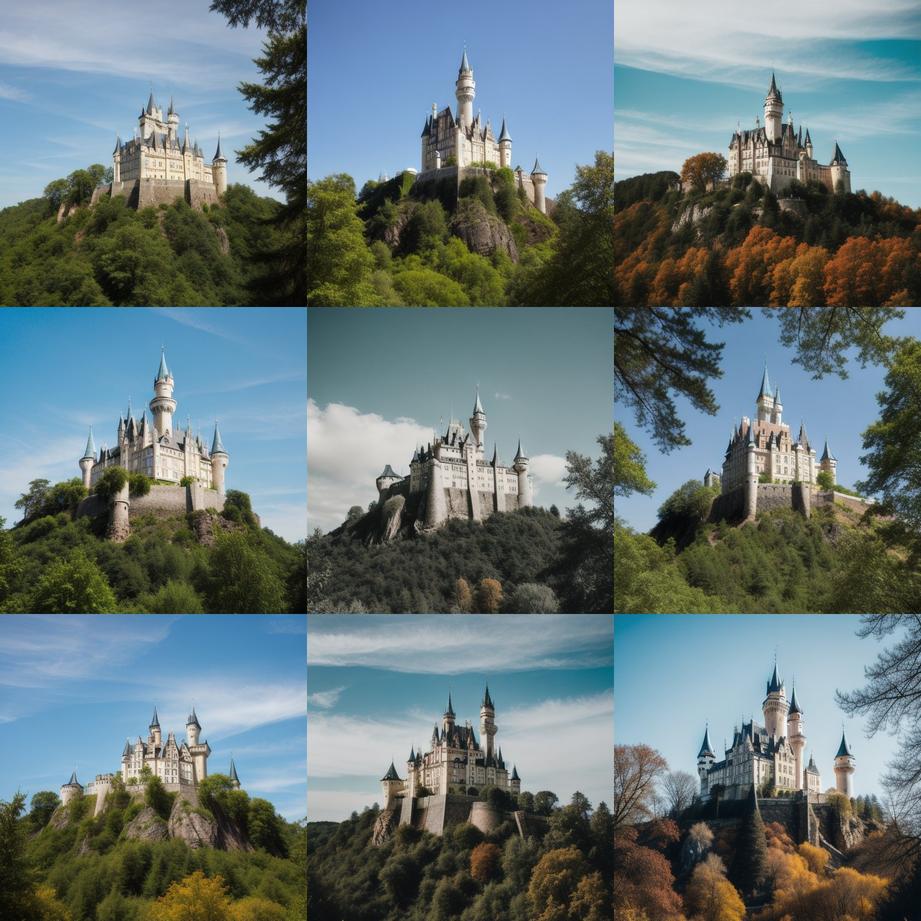} &
        \includegraphics[width=0.175\linewidth]{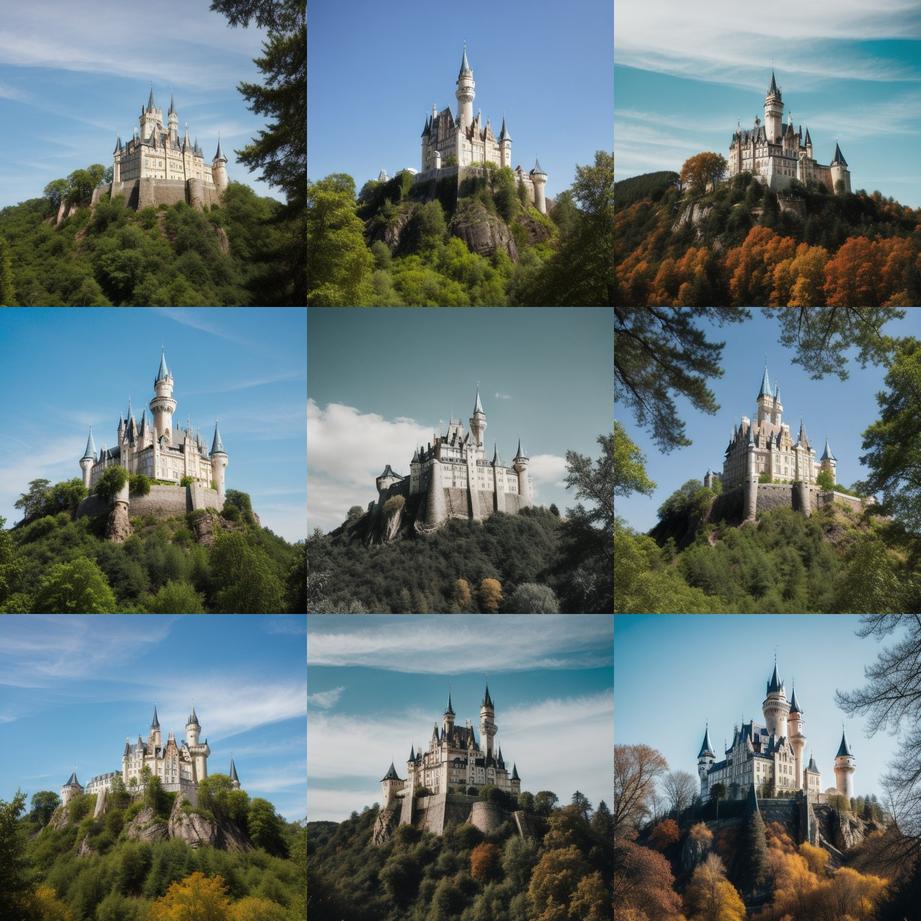} \\
    \end{tabular}
    \vspace{0pt}
    \caption{\textbf{Results of the \textit{replacement baseline} with different $t_{\text{stop}}$.} This baseline yields inconsistent images, even with high $t_{\text{stop}}$ values.}
    \label{fig:ab_replace}
\end{figure*}

We present a naive baseline, dubbed the \textit{replacement baseline}, that directly substitutes the averaged $h$-space at each timestep during denoising instead of performing noise optimization. Here, the averaged $h$-space is computed in the same way as in DMA, and this substitution is applied across diffusion timesteps similarly until a cutoff $t_\text{stop} < T$. As shown in Figure~\ref{fig:ab_replace}, the baseline produces inconsistent results for all $t_\text{stop}$ values. This inconsistency arises from skip connections, which can introduce stochastic information from individual noise latents, highlighting the need for noise optimization to constrain activations in other layers.
\subsection{Effects of $t_{\text{stop}}$}\label{sec:c4_tstop}
\begin{figure*}[t]
    \centering
    \renewcommand{\arraystretch}{1.2}
    \setlength{\tabcolsep}{1pt}
    \footnotesize
    \begin{tabular}{c ccccc} 
        & \shortstack{\scriptsize \textbf{Standard Denoising Samples} \\ ($t_\text{stop}=0$)} & $t_{\text{stop}}=5$ & $t_{\text{stop}}=10$ & $t_{\text{stop}}=15$ & $t_{\text{stop}}=20$ \\
        
        \raisebox{3.2\height}{\rotatebox[origin=c]{90}{\text{castle}}} &
        \includegraphics[width=0.175\linewidth]{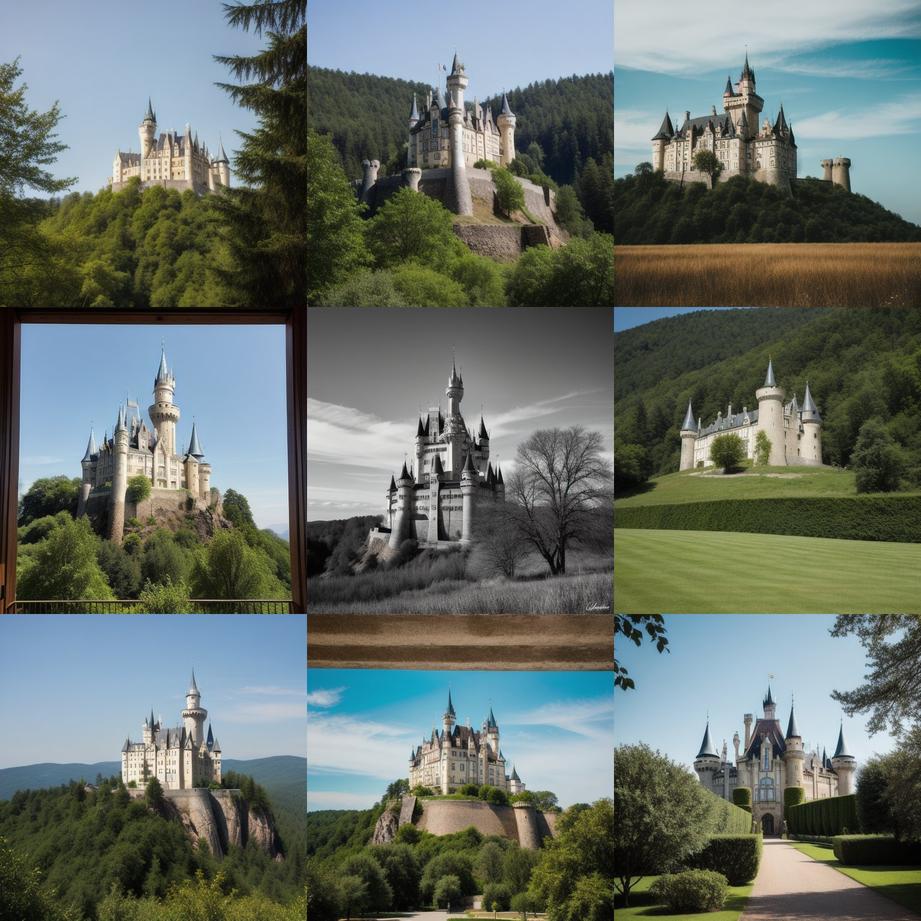} &
        \includegraphics[width=0.175\linewidth]{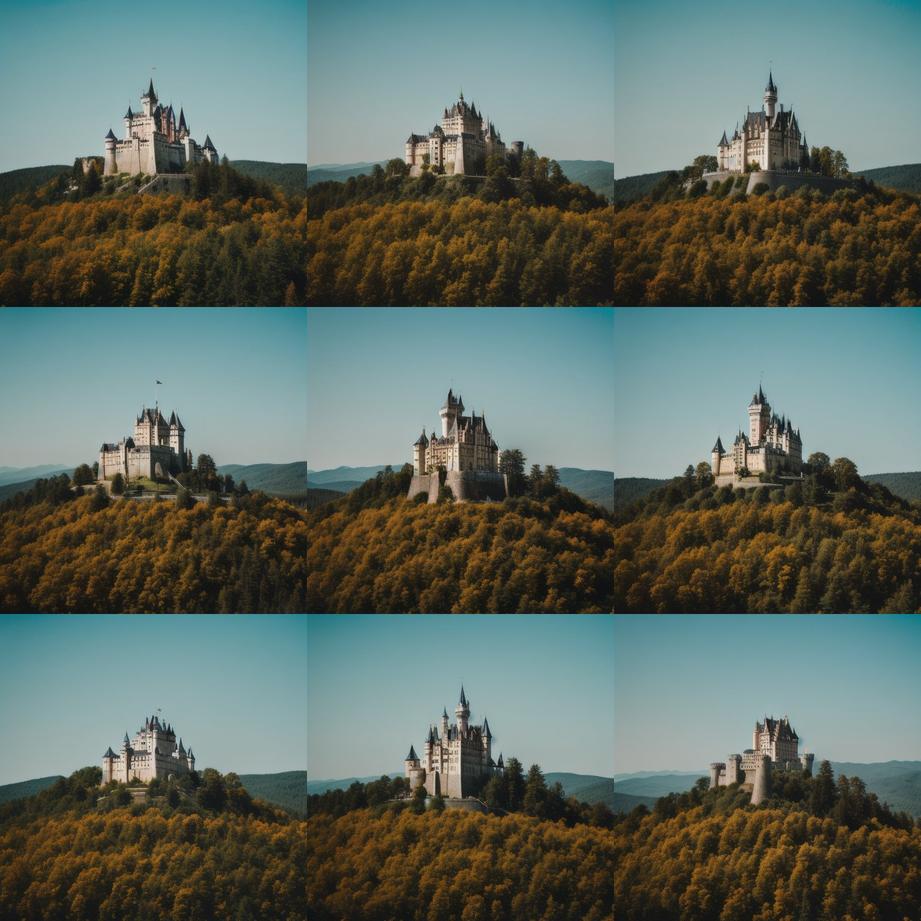} &
        \includegraphics[width=0.175\linewidth]{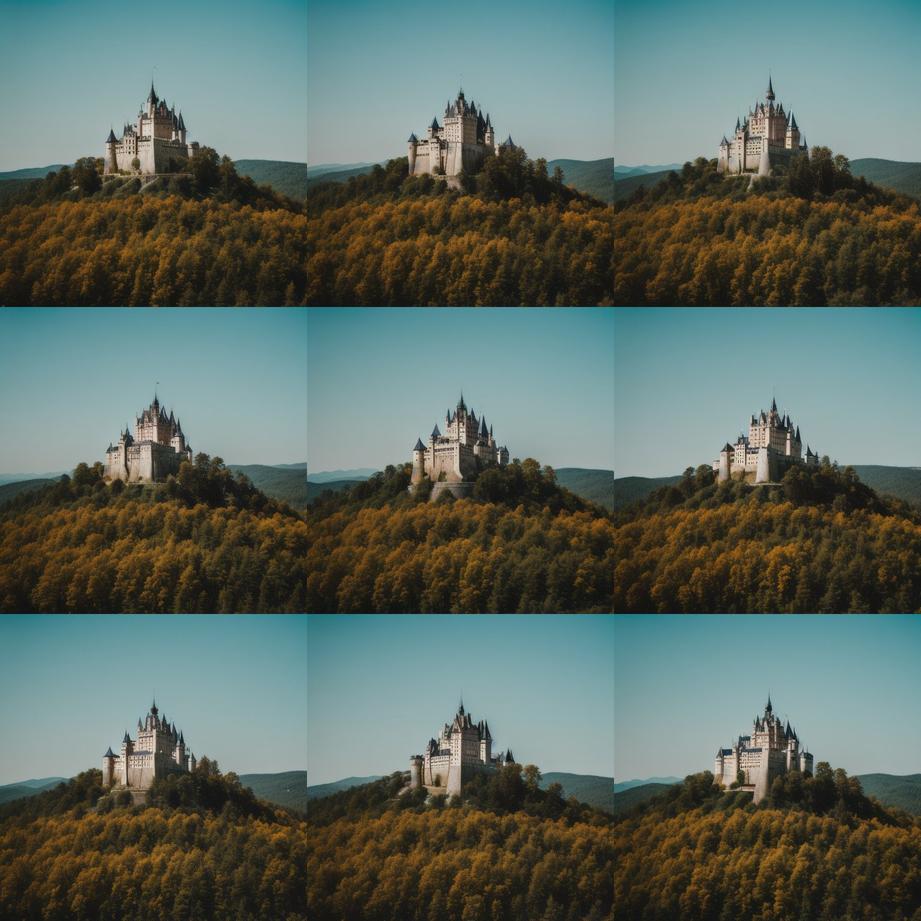} &
        \includegraphics[width=0.175\linewidth]{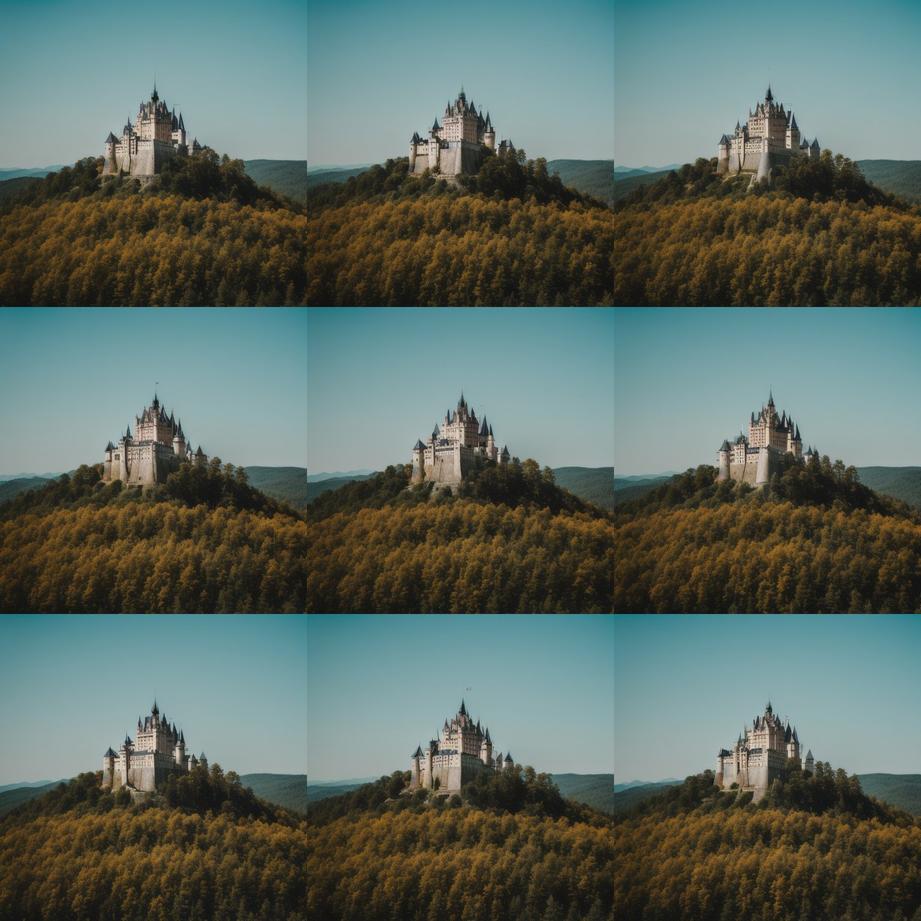} &
        \includegraphics[width=0.175\linewidth]{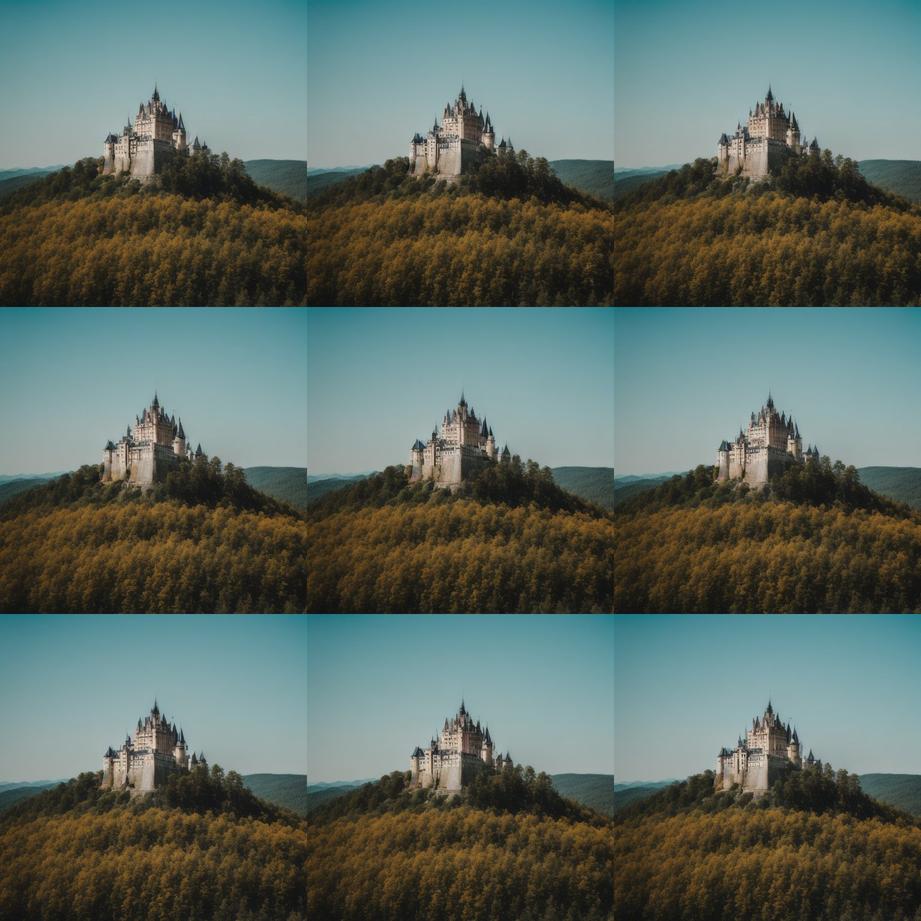} \\

        \raisebox{5.5\height}{\rotatebox[origin=c]{90}{\text{dog}}} &
        \includegraphics[width=0.175\linewidth]{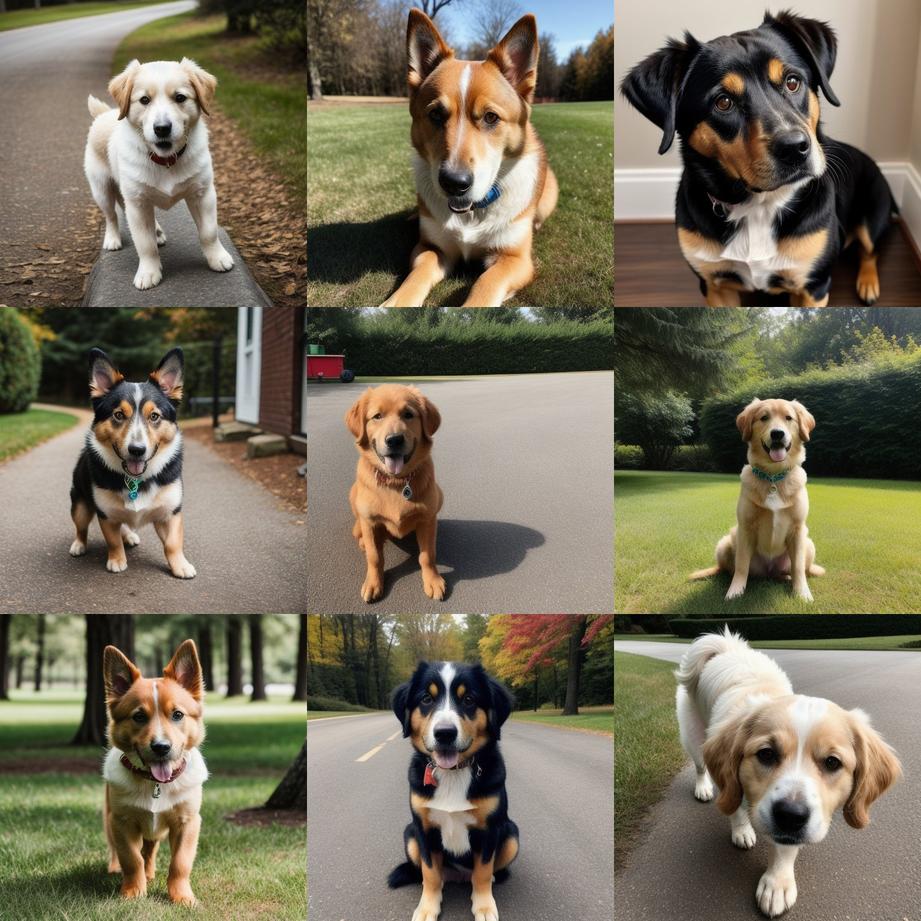} &
        \includegraphics[width=0.175\linewidth]{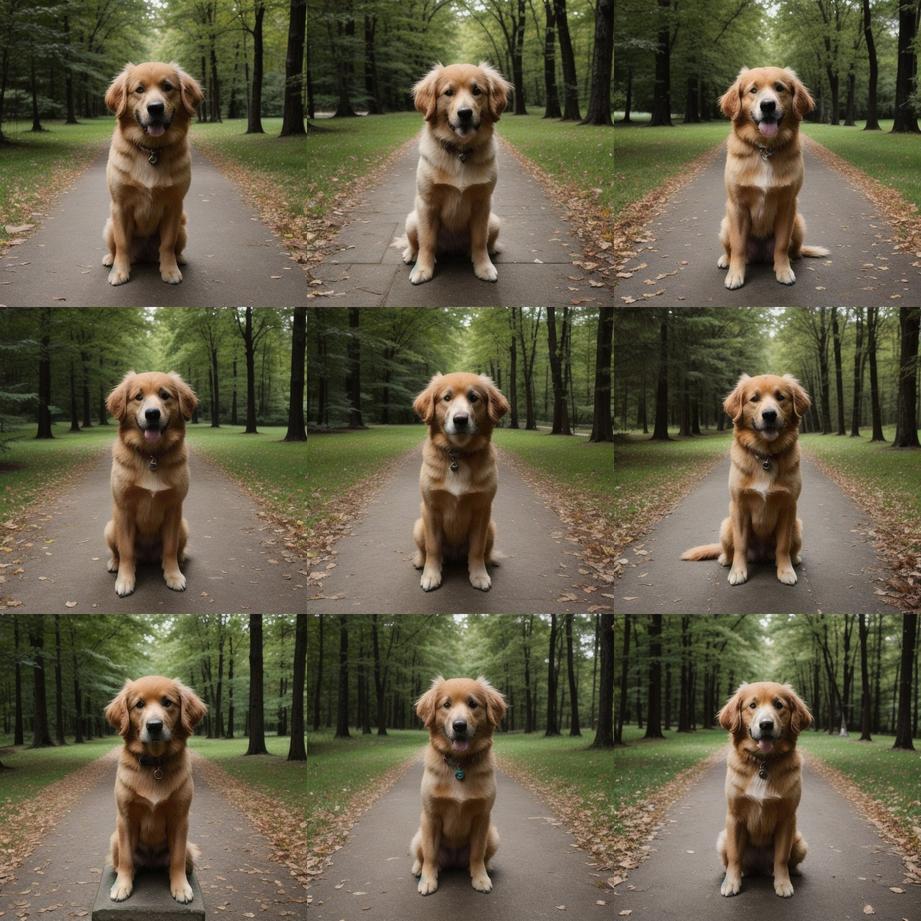} &
        \includegraphics[width=0.175\linewidth]{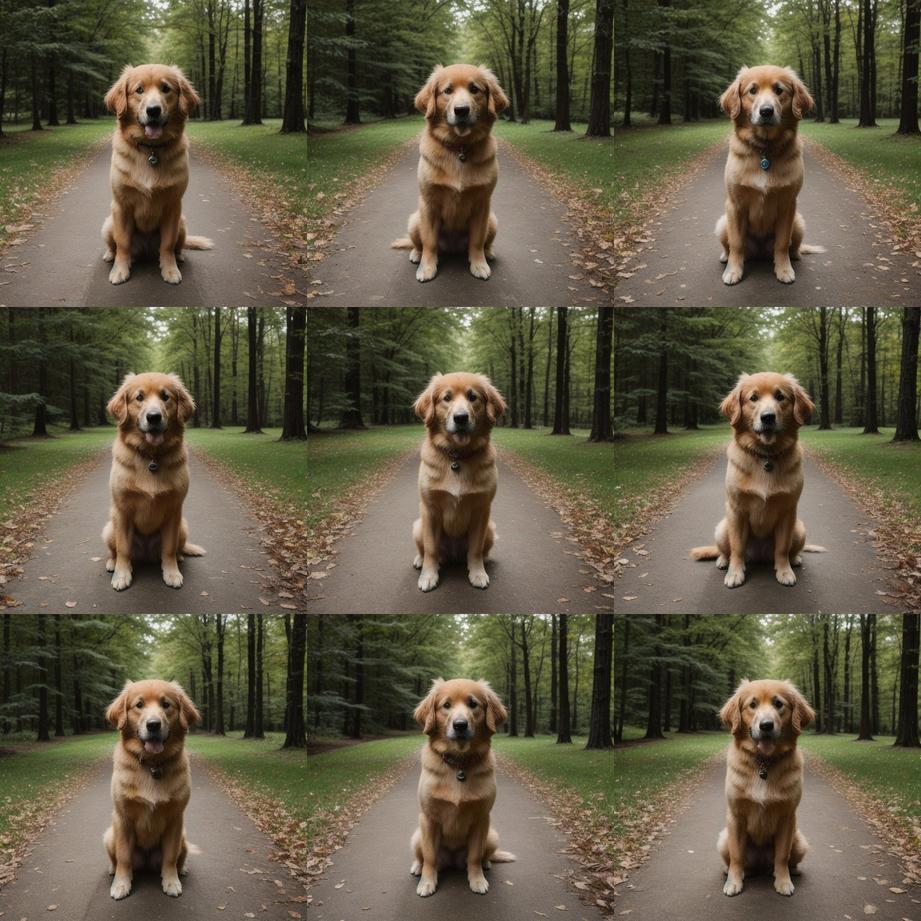} &
        \includegraphics[width=0.175\linewidth]{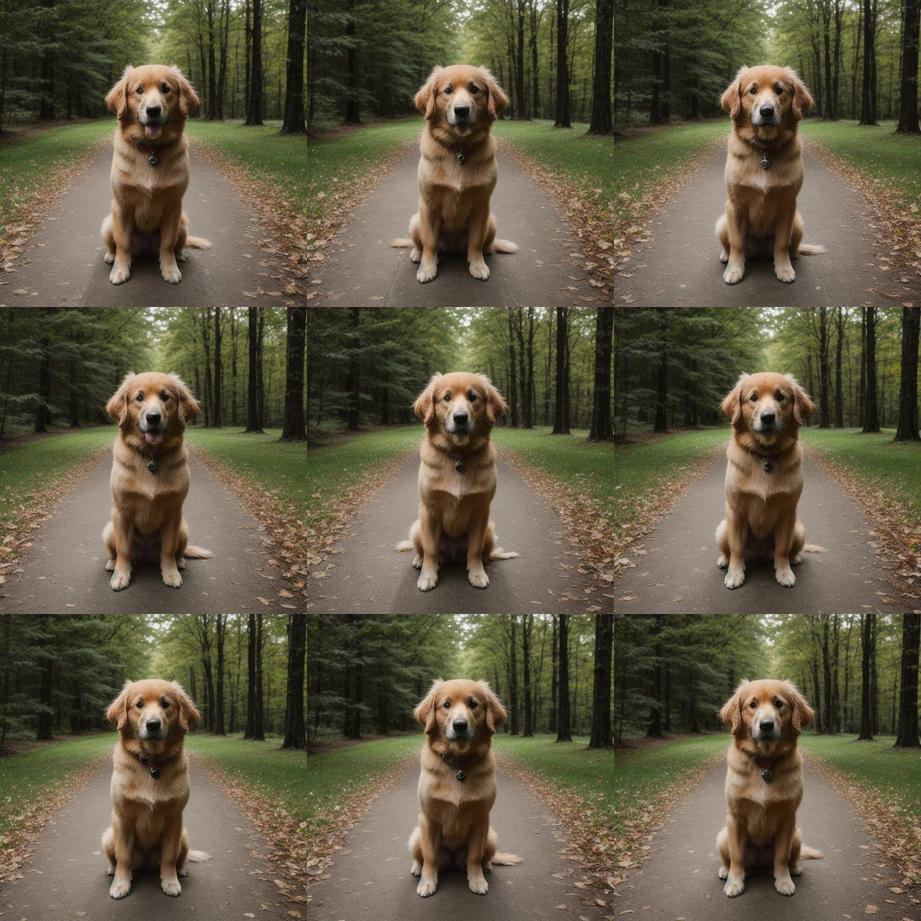} &
        \includegraphics[width=0.175\linewidth]{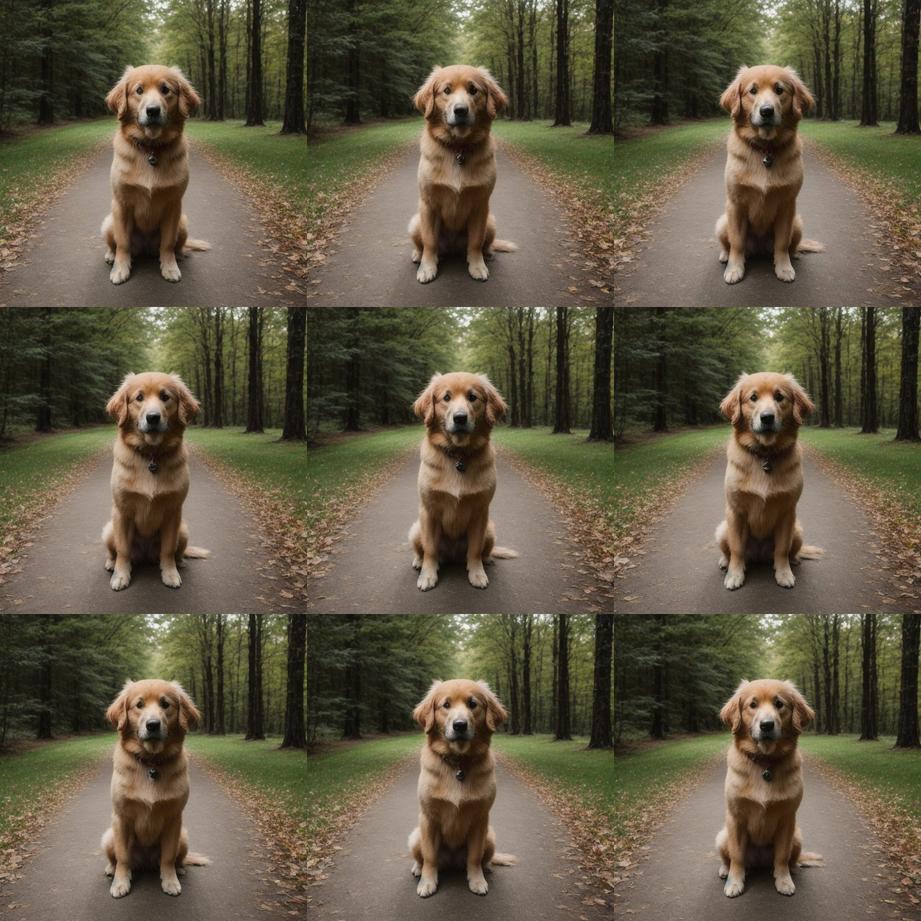} \\

        \raisebox{2.5\height}{\rotatebox[origin=c]{90}{\text{firefighter}}} &
        \includegraphics[width=0.175\linewidth]{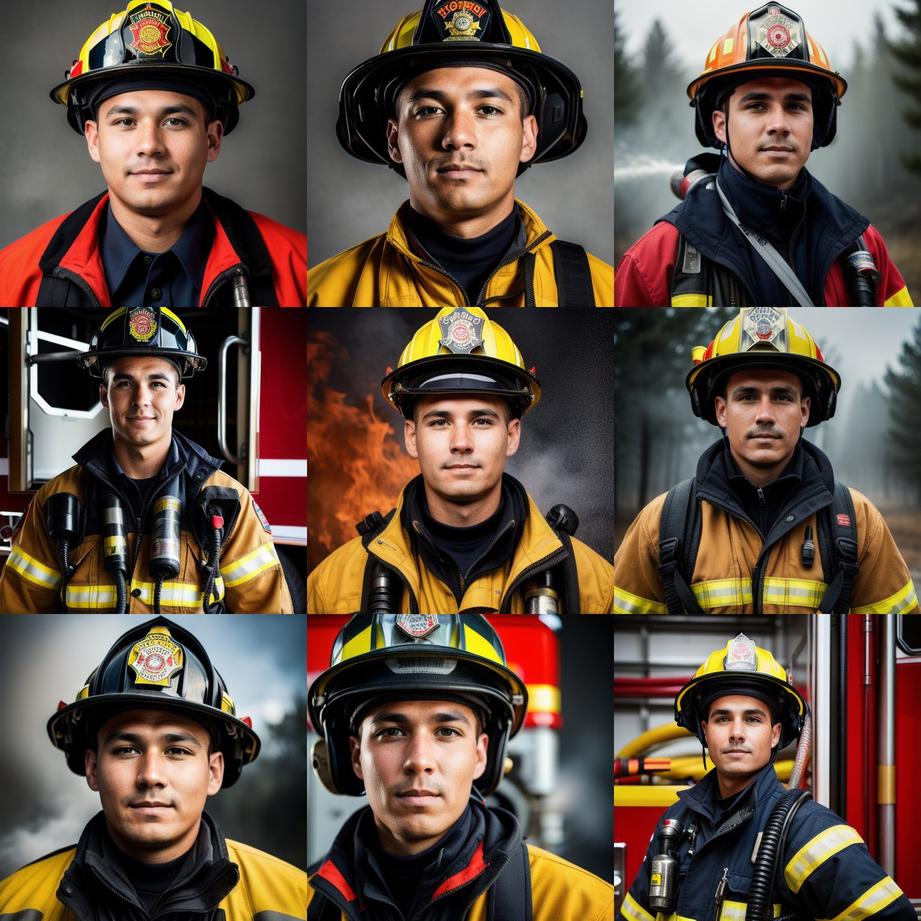} &
        \includegraphics[width=0.175\linewidth]{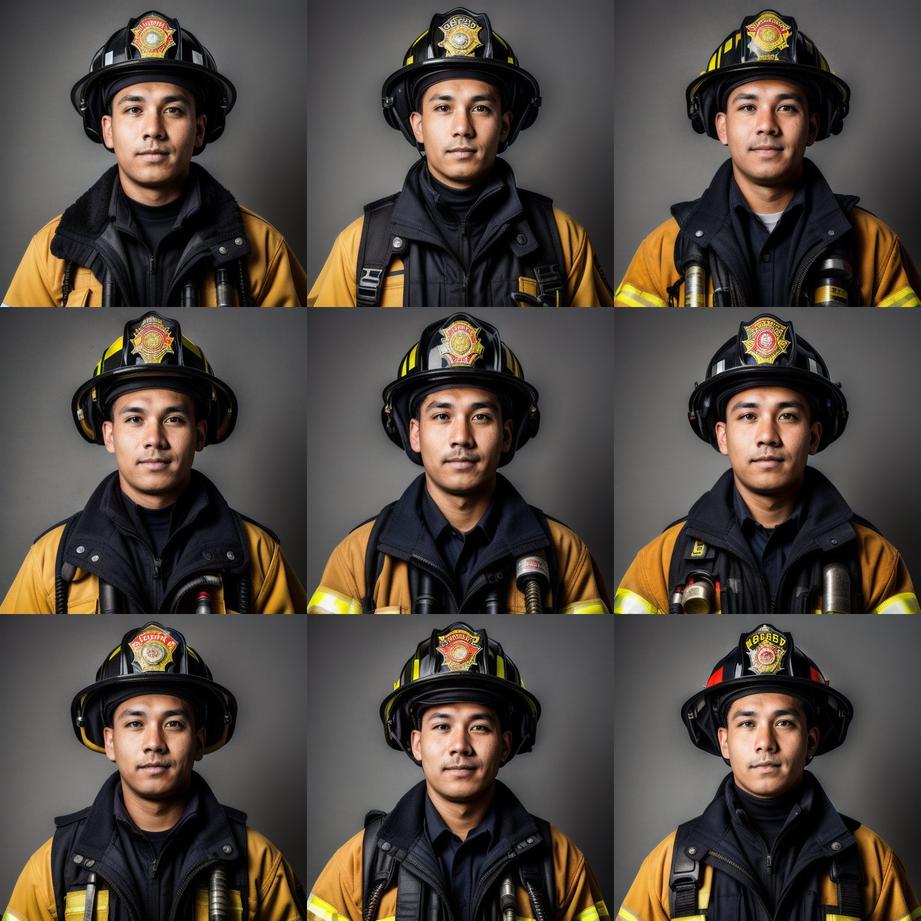} &
        \includegraphics[width=0.175\linewidth]{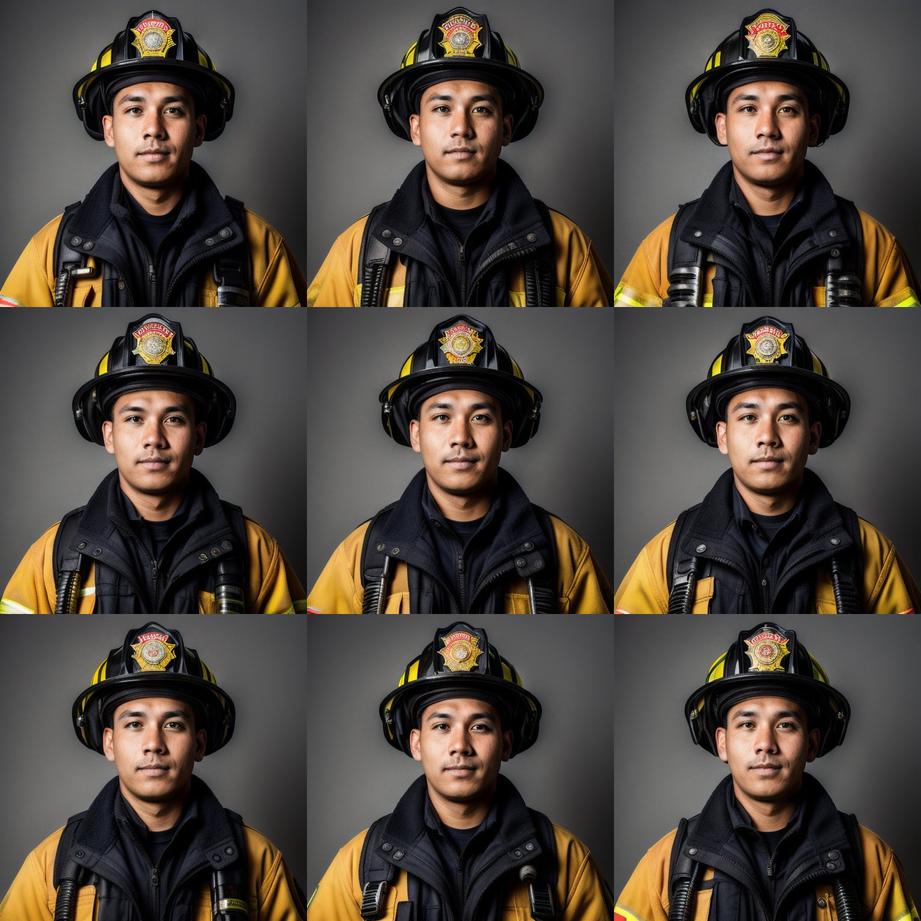} &
        \includegraphics[width=0.175\linewidth]{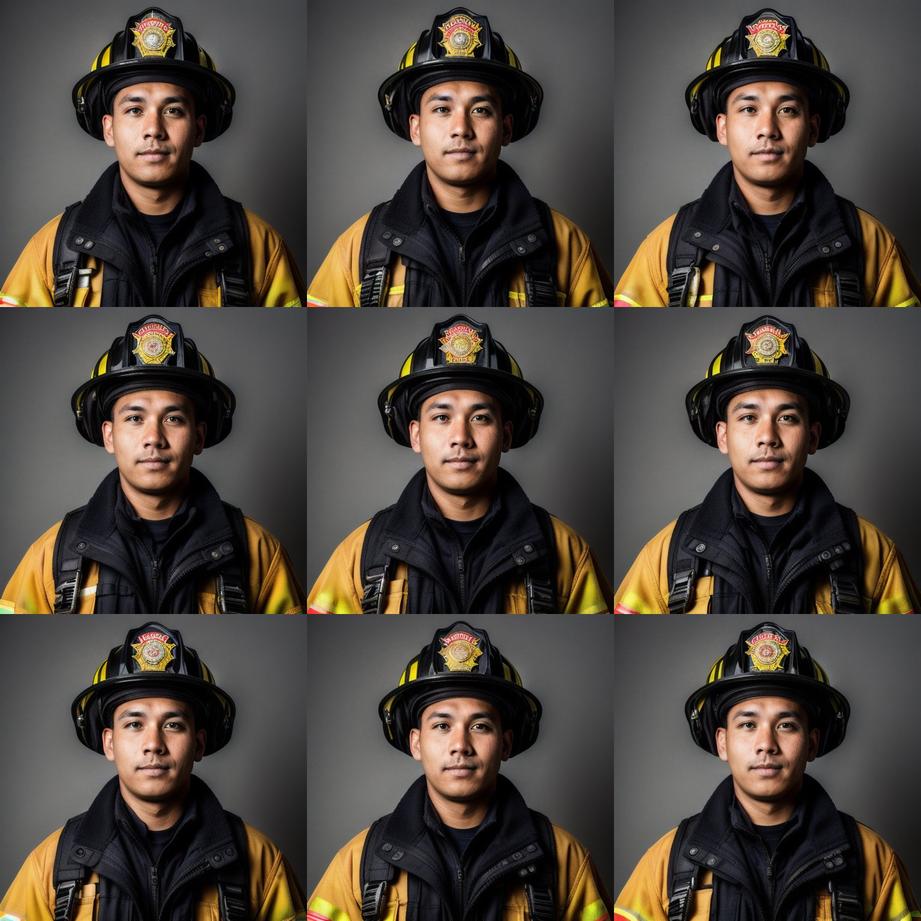} &
        \includegraphics[width=0.175\linewidth]{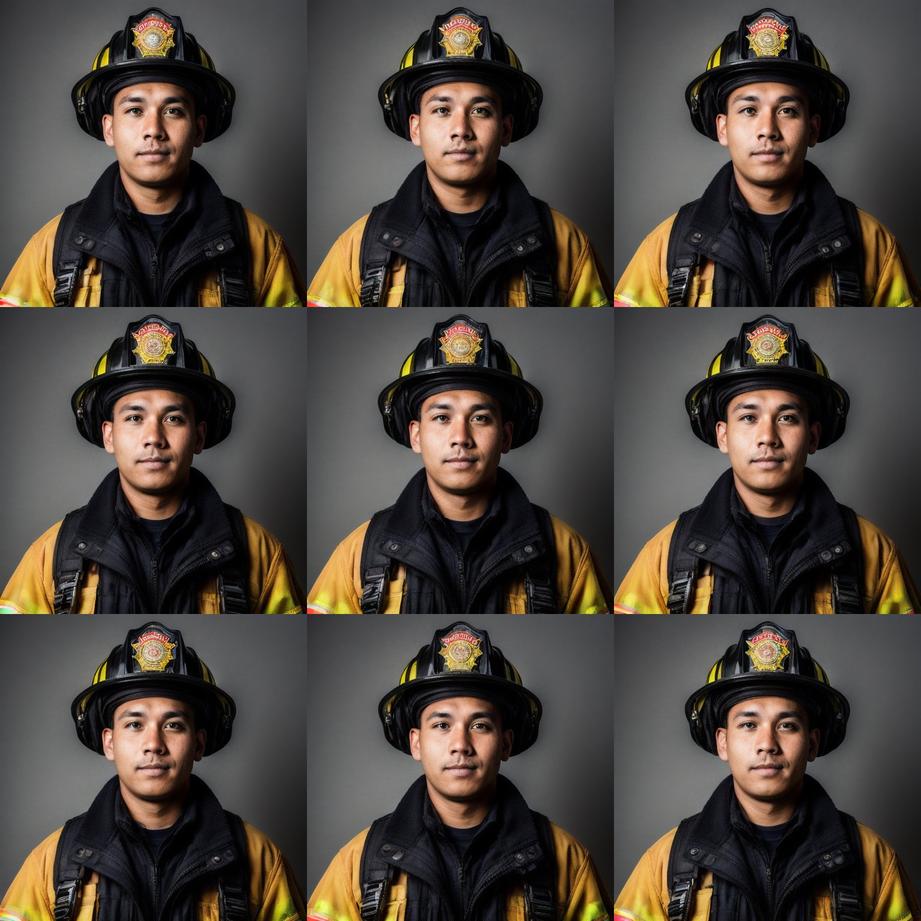} \\

        \raisebox{3.6\height}{\rotatebox[origin=c]{90}{\text{zebra}}} &
        \includegraphics[width=0.175\linewidth]{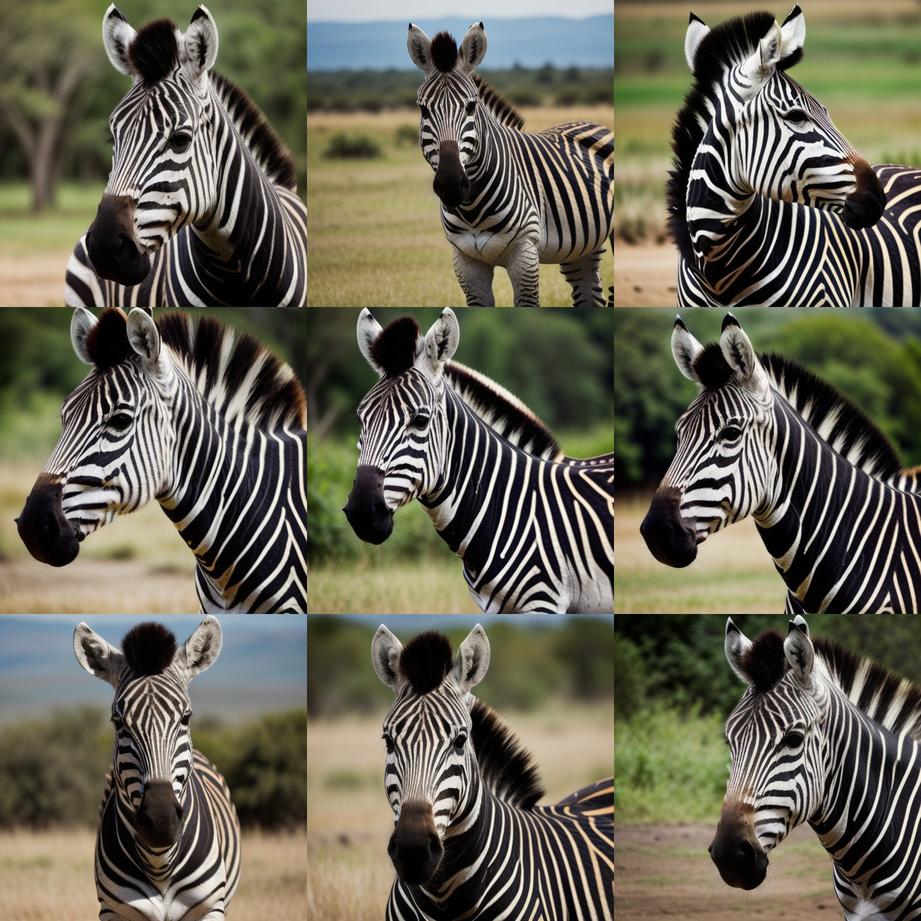} &
        \includegraphics[width=0.175\linewidth]{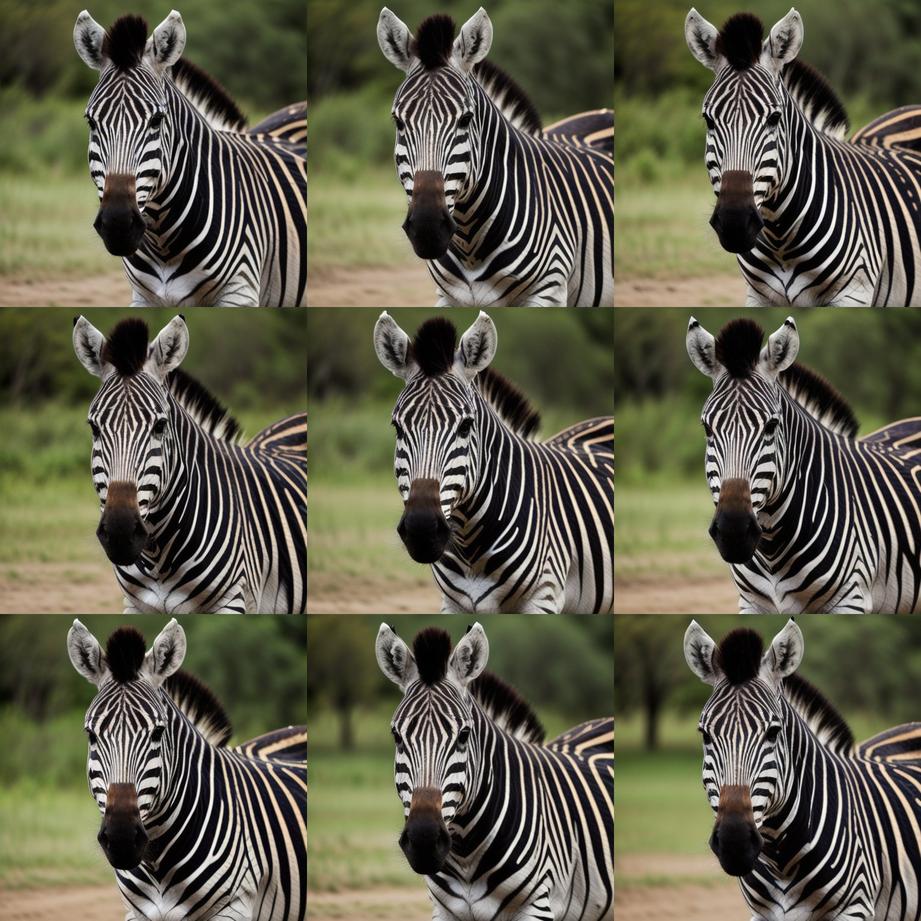} &
        \includegraphics[width=0.175\linewidth]{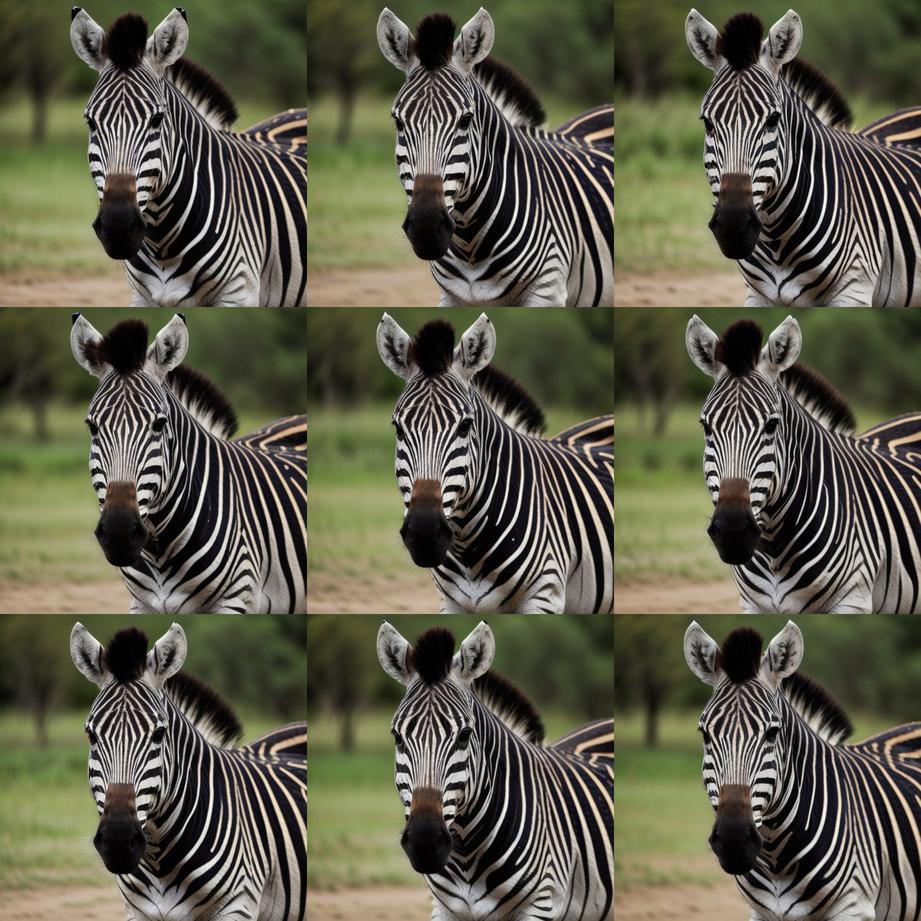} &
        \includegraphics[width=0.175\linewidth]{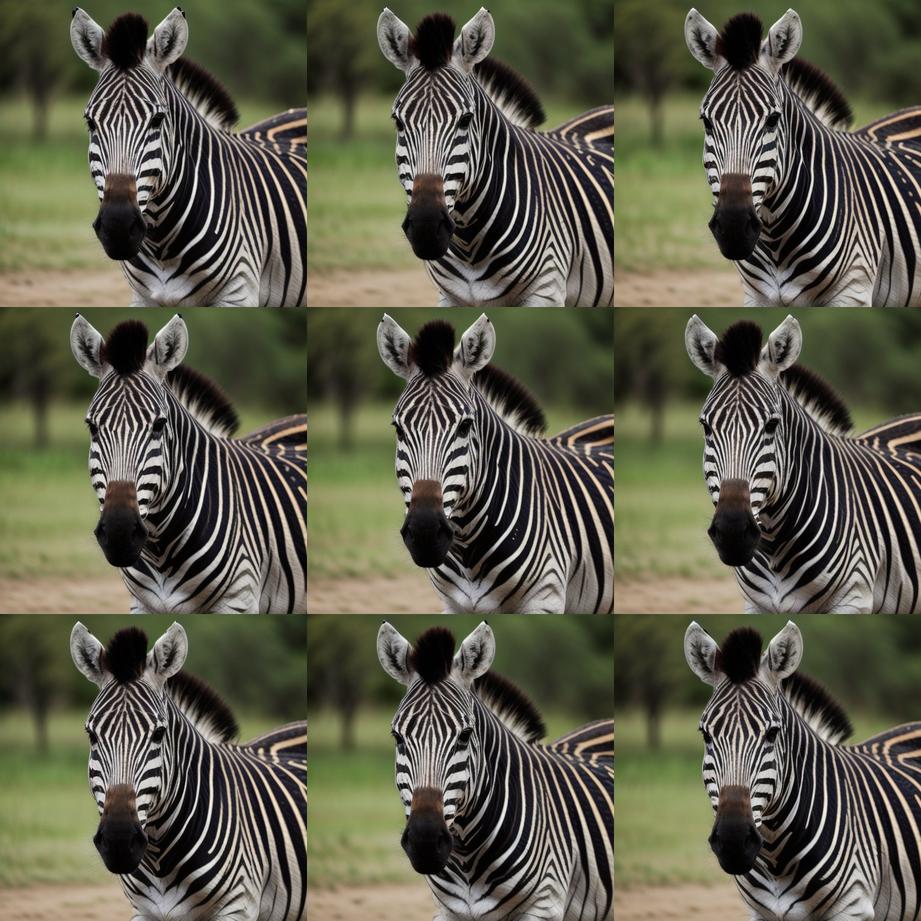} &
        \includegraphics[width=0.175\linewidth]{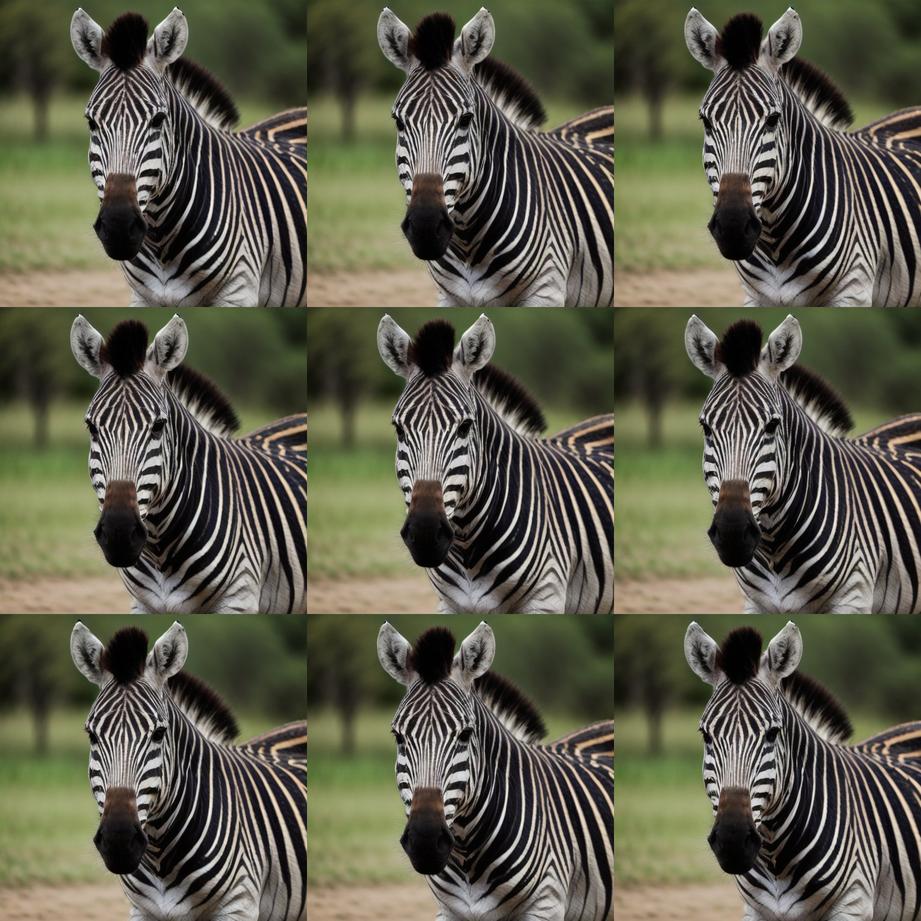} \\
    \end{tabular}
    \vspace{0pt}
    \caption{\textbf{Effects of different $t_{\text{stop}}$ values on visual consistency.} $t_{\text{stop}}=10$ yields consistent images with reduced computational cost.}
    \label{fig:ab_tstop}
\end{figure*}

We analyze the effects of varying the cutoff timestep $t_{\text{stop}}$ in Figure~\ref{fig:ab_tstop}. The results show that a low $t_{\text{stop}}=5$ leads to inconsistent average images, such as architectural incoherence in the \textit{castle}, ground-plane variations in the \textit{dog}, and costume inconsistencies in the \textit{firefighter}. Increasing $t_{\text{stop}}$ beyond 10 provides only minor improvements in visual consistency for most concepts. We therefore set $t_{\text{stop}}=10$ as a balance between computational efficiency and cross-sample consistency.

\subsection{Effects of Classifier-Free Guidance Scale}
\begin{figure*}[t]
    \centering
    \renewcommand{\arraystretch}{1.2}
    \setlength{\tabcolsep}{1pt}
    \footnotesize
    \begin{tabular}{c ccccc}
        & \textbf{CFG 1.0} & \textbf{CFG 3.0} & \textbf{CFG 5.0} & \textbf{CFG 7.0} & \textbf{CFG 12.0} \\
        
        \raisebox{4\height}{\rotatebox[origin=c]{90}{\text{artist}}} &
        \includegraphics[width=0.18\linewidth]{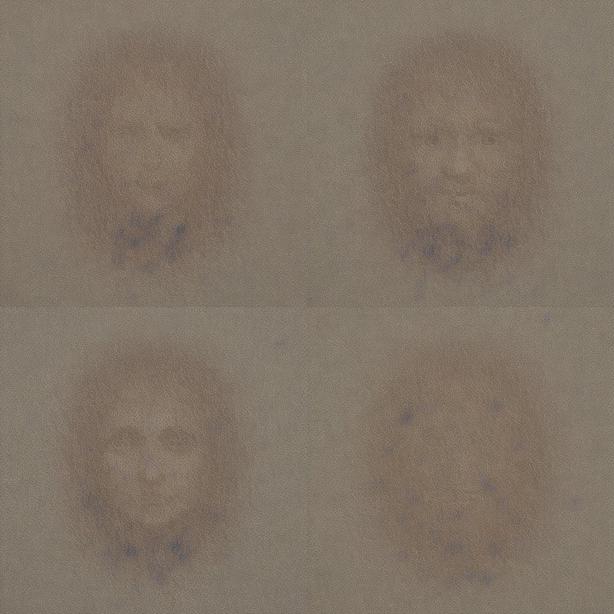} &
        \includegraphics[width=0.18\linewidth]{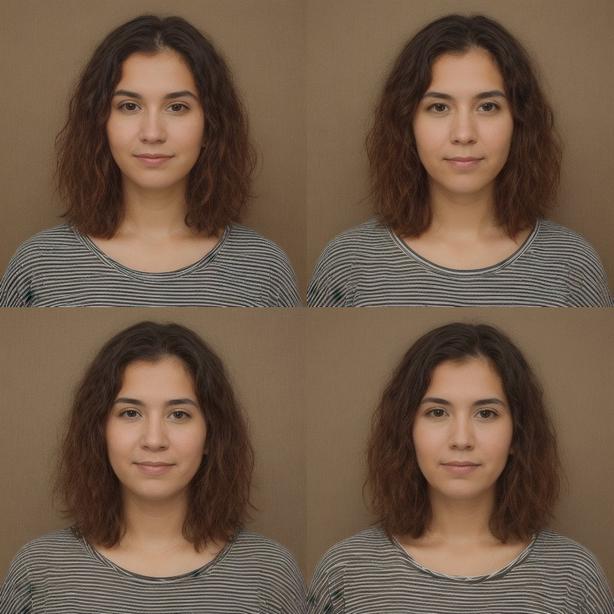} &
        \includegraphics[width=0.18\linewidth]{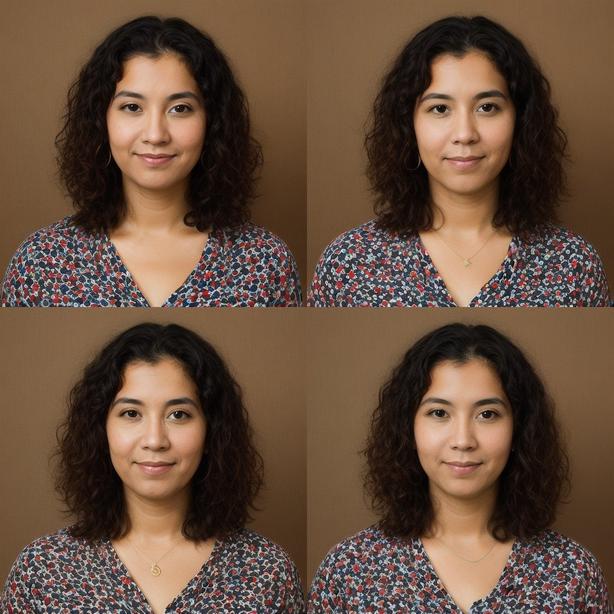} &
        \includegraphics[width=0.18\linewidth]{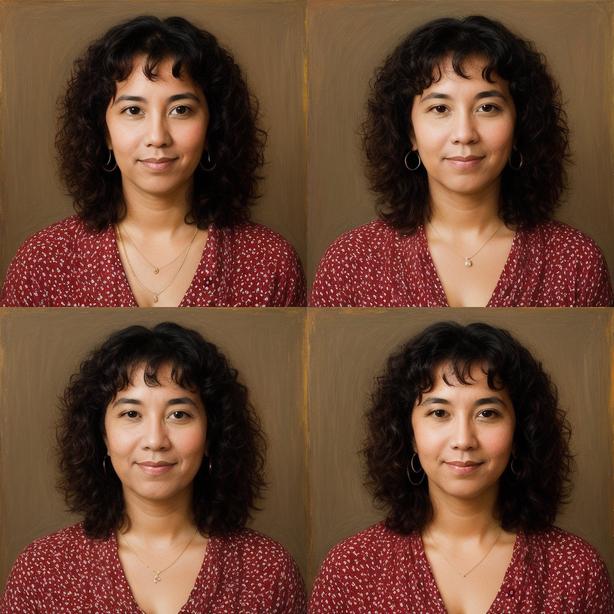} &
        \includegraphics[width=0.18\linewidth]{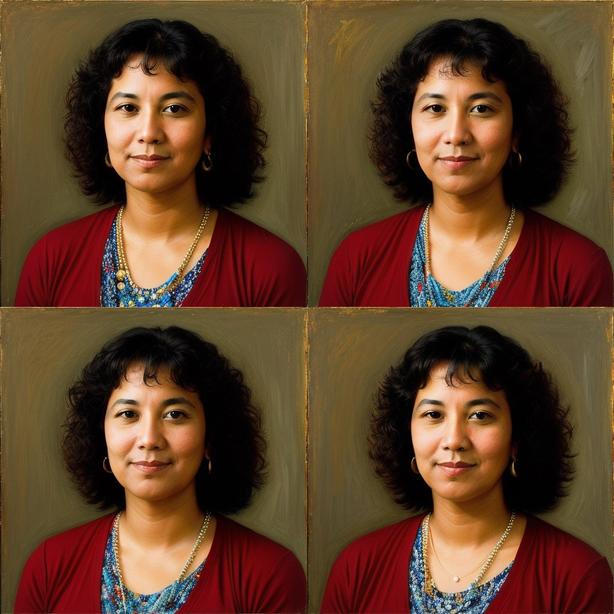}\\

        \raisebox{6\height}{\rotatebox[origin=c]{90}{\text{cat}}} &
        \includegraphics[width=0.18\linewidth]{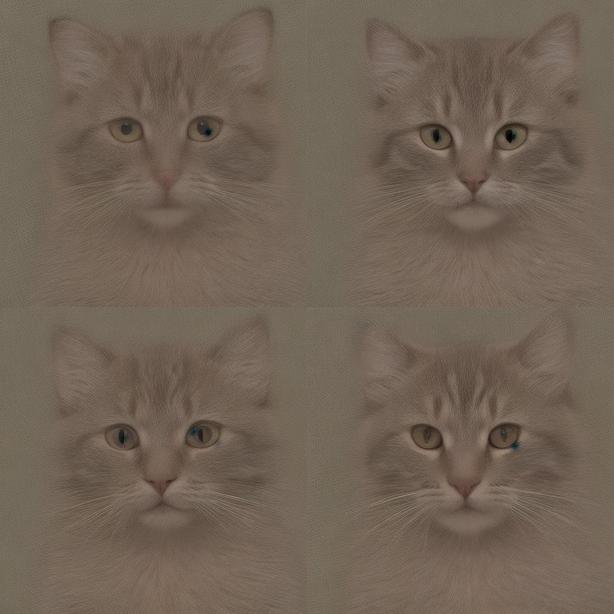} &
        \includegraphics[width=0.18\linewidth]{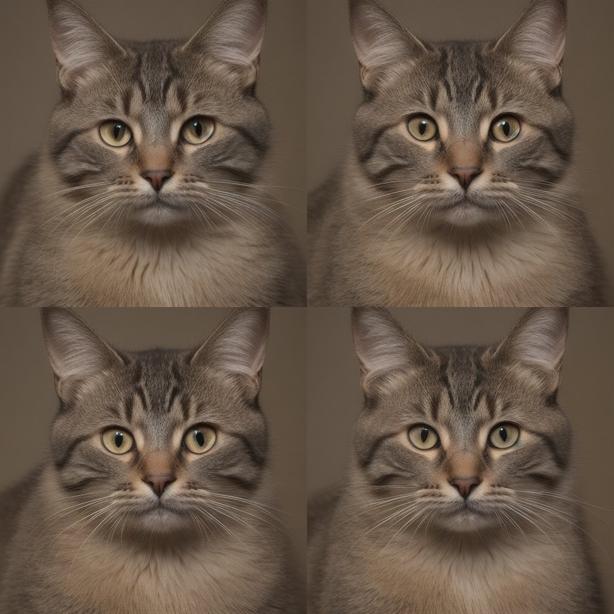} &
        \includegraphics[width=0.18\linewidth]{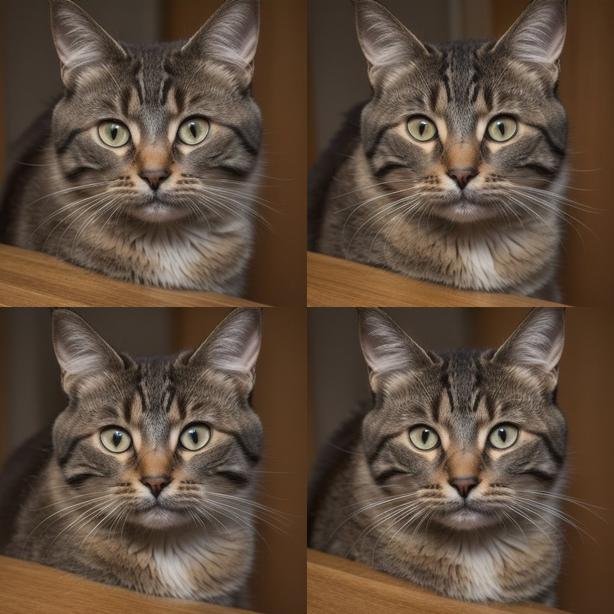} &
        \includegraphics[width=0.18\linewidth]{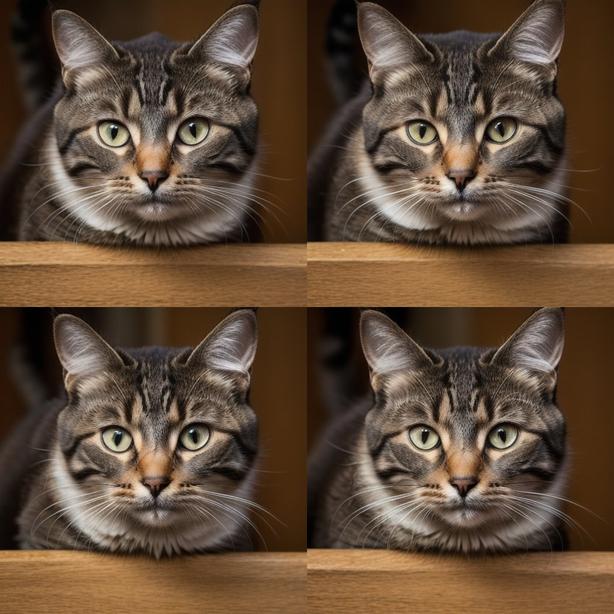} &
        \includegraphics[width=0.18\linewidth]{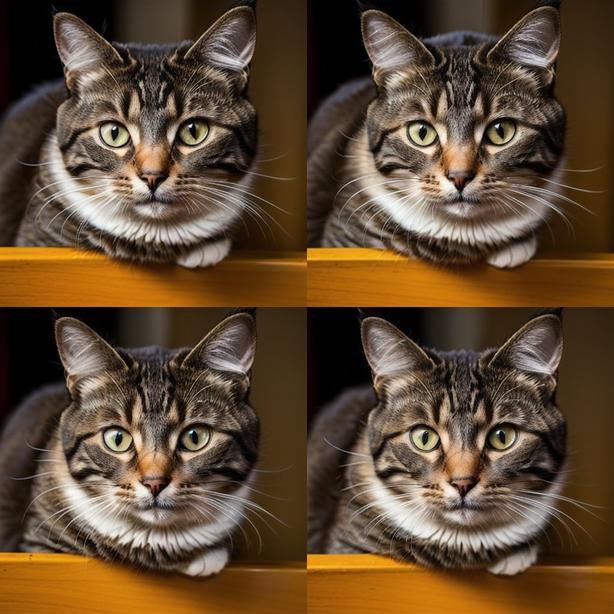}\\

        \raisebox{1\height}{\rotatebox[origin=c]{90}{\shortstack{lady sitting on bench \\ with handbag}}} &
        \includegraphics[width=0.18\linewidth]{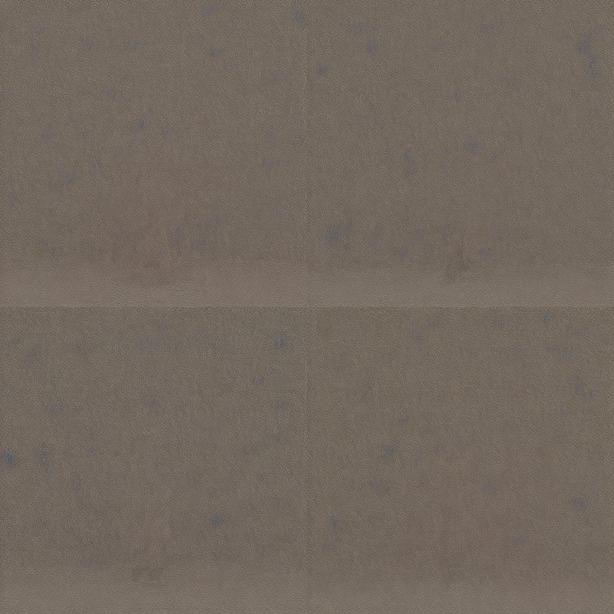} &
        \includegraphics[width=0.18\linewidth]{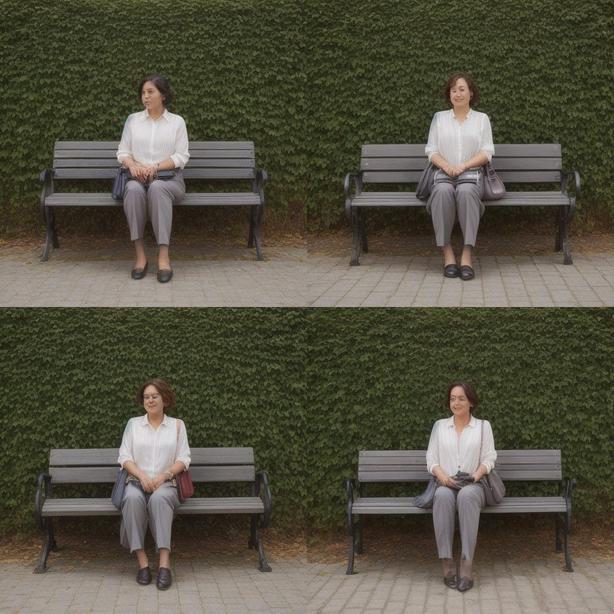} &
        \includegraphics[width=0.18\linewidth]{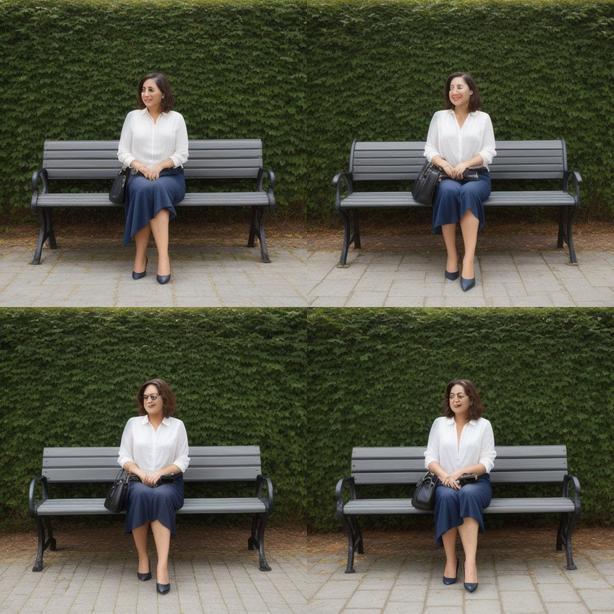} &
        \includegraphics[width=0.18\linewidth]{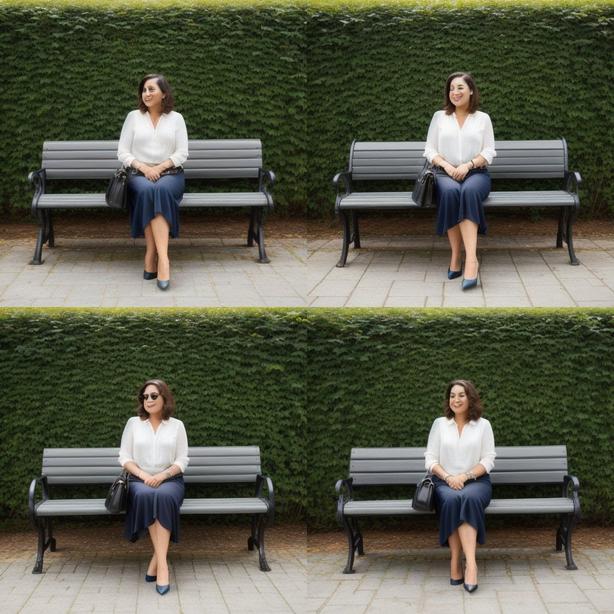} &
        \includegraphics[width=0.18\linewidth]{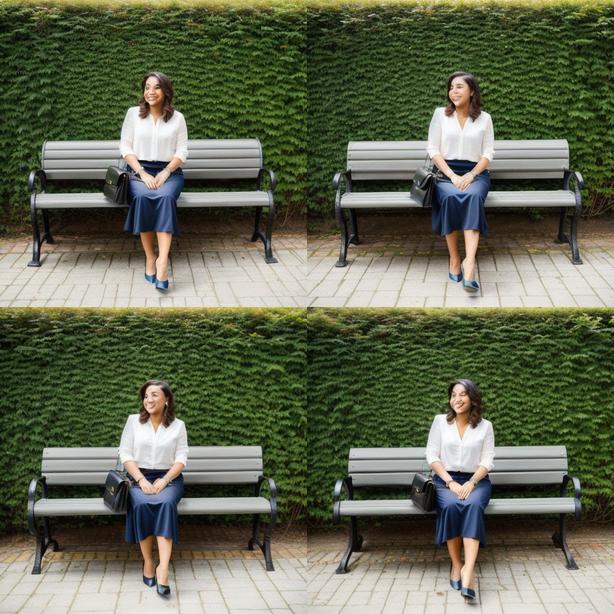}\\

        \raisebox{3.9\height}{\rotatebox[origin=c]{90}{\text{panda}}} &
        \includegraphics[width=0.18\linewidth]{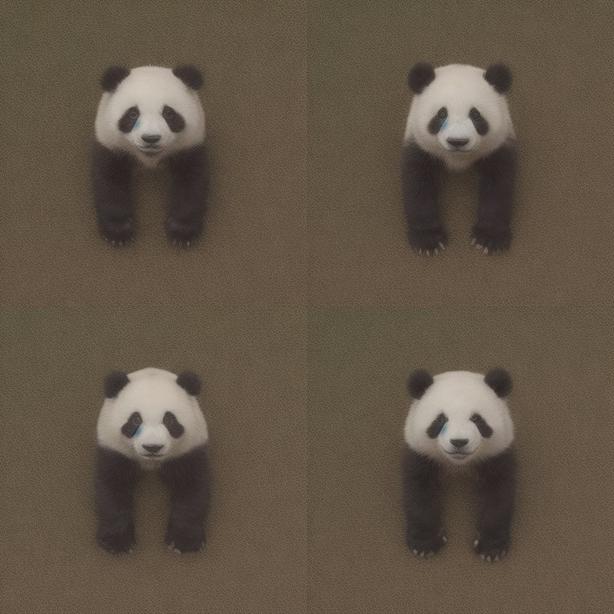} &
        \includegraphics[width=0.18\linewidth]{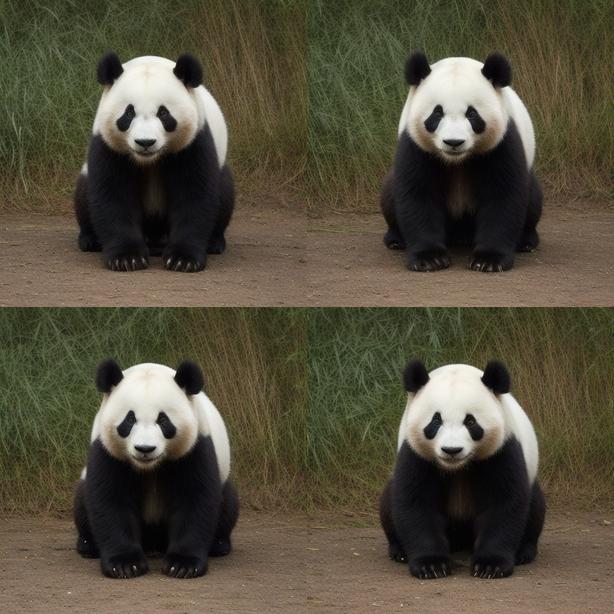} &
        \includegraphics[width=0.18\linewidth]{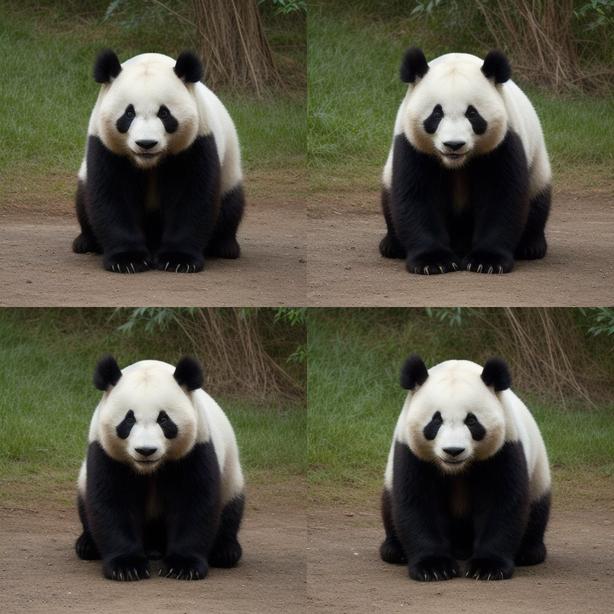} &
        \includegraphics[width=0.18\linewidth]{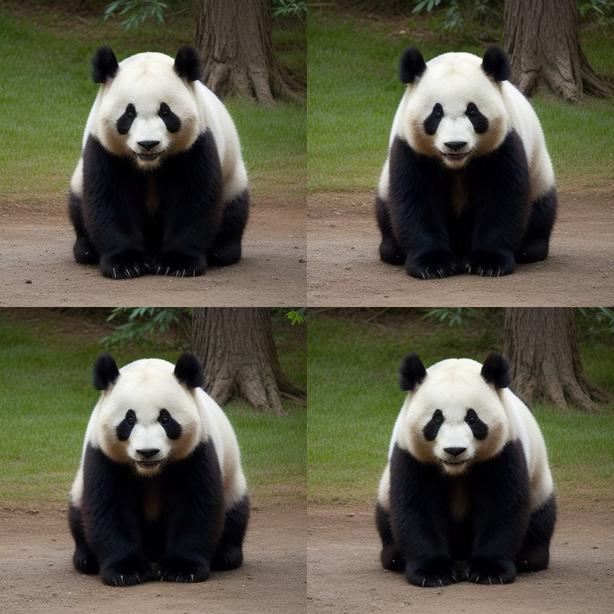} &
        \includegraphics[width=0.18\linewidth]{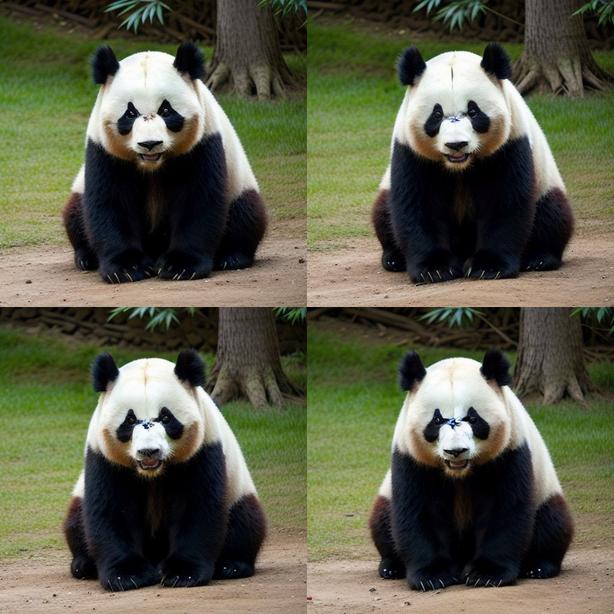}\\

        \raisebox{2.8\height}{\rotatebox[origin=c]{90}{\text{monstera}}} &
        \includegraphics[width=0.18\linewidth]{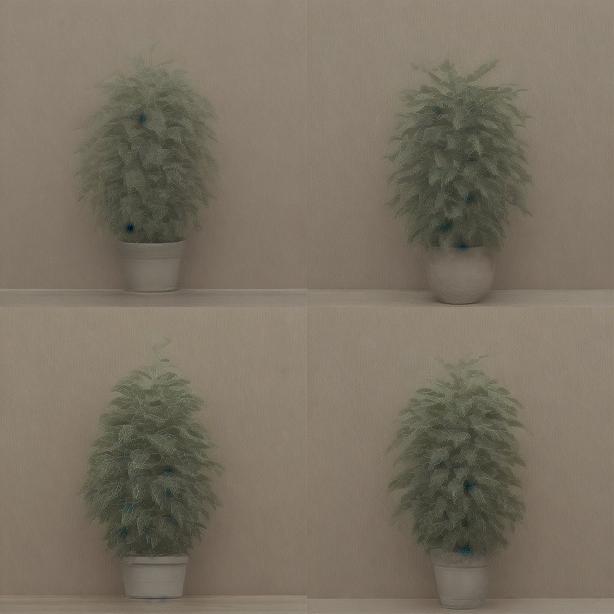} &
        \includegraphics[width=0.18\linewidth]{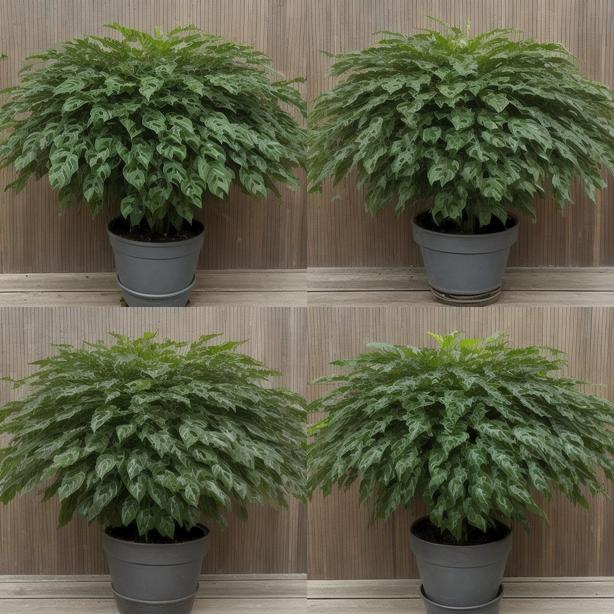} &
        \includegraphics[width=0.18\linewidth]{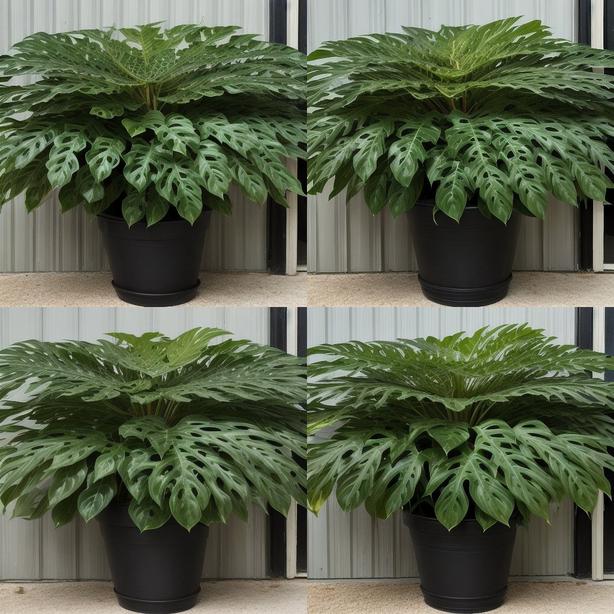} &
        \includegraphics[width=0.18\linewidth]{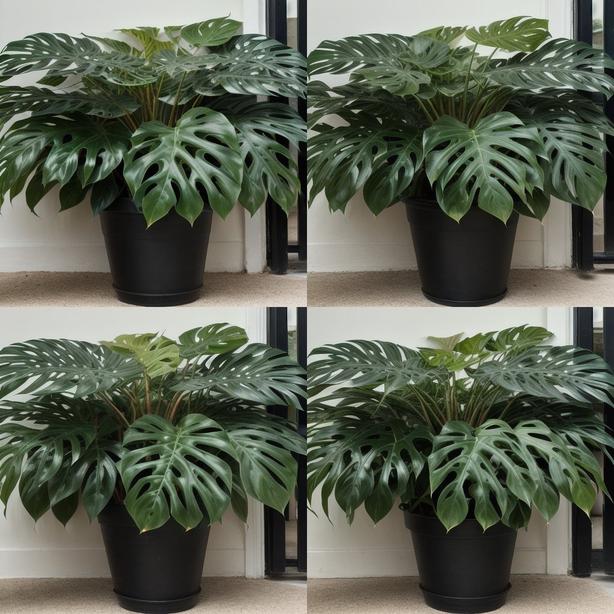} &
        \includegraphics[width=0.18\linewidth]{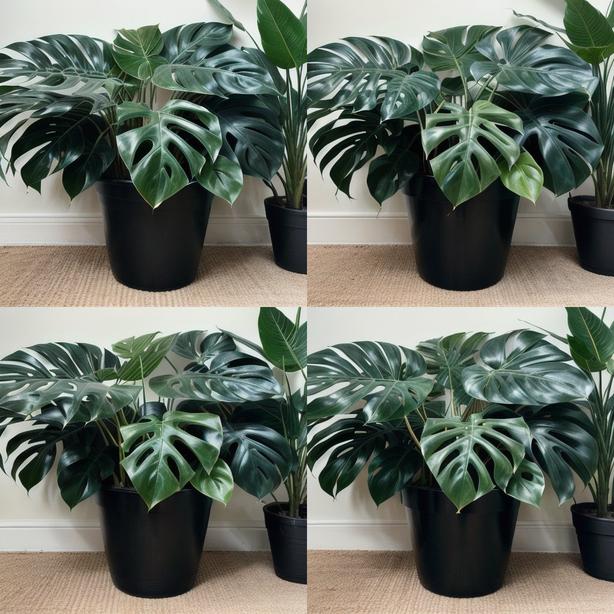}\\
    \end{tabular}
    \vspace{0pt}
    \caption{\textbf{Effects of CFG scale on visual quality.} Low CFG scales fail to generate faithful structures (as seen in \textit{monstera}), while excessively high CFG scales lead to oversaturated colors and unnatural contrast.}
    \label{fig:ab_cfg}
\end{figure*}

We examine the effects of different Classifier-Free Guidance (CFG) scales across various concepts in Figure~\ref{fig:ab_cfg}. Low CFG values (1.0–5.0) produce structurally invalid results, such as distorted \textit{monstera} leaves, while excessively high values (e.g., 12.0) lead to oversaturated images. The optimal CFG, however, varies across concepts: higher values benefit high-variation concepts like \textit{monstera}, whereas lower values suffice for low-variation concepts, such as \textit{zebra}. In this paper, we use a CFG scale of 7.0 for all concepts. 
\begin{figure}[t]
    \centering
    \setlength{\tabcolsep}{1pt}
    \begin{tabular}{cc}
        \includegraphics[width=0.49\linewidth]{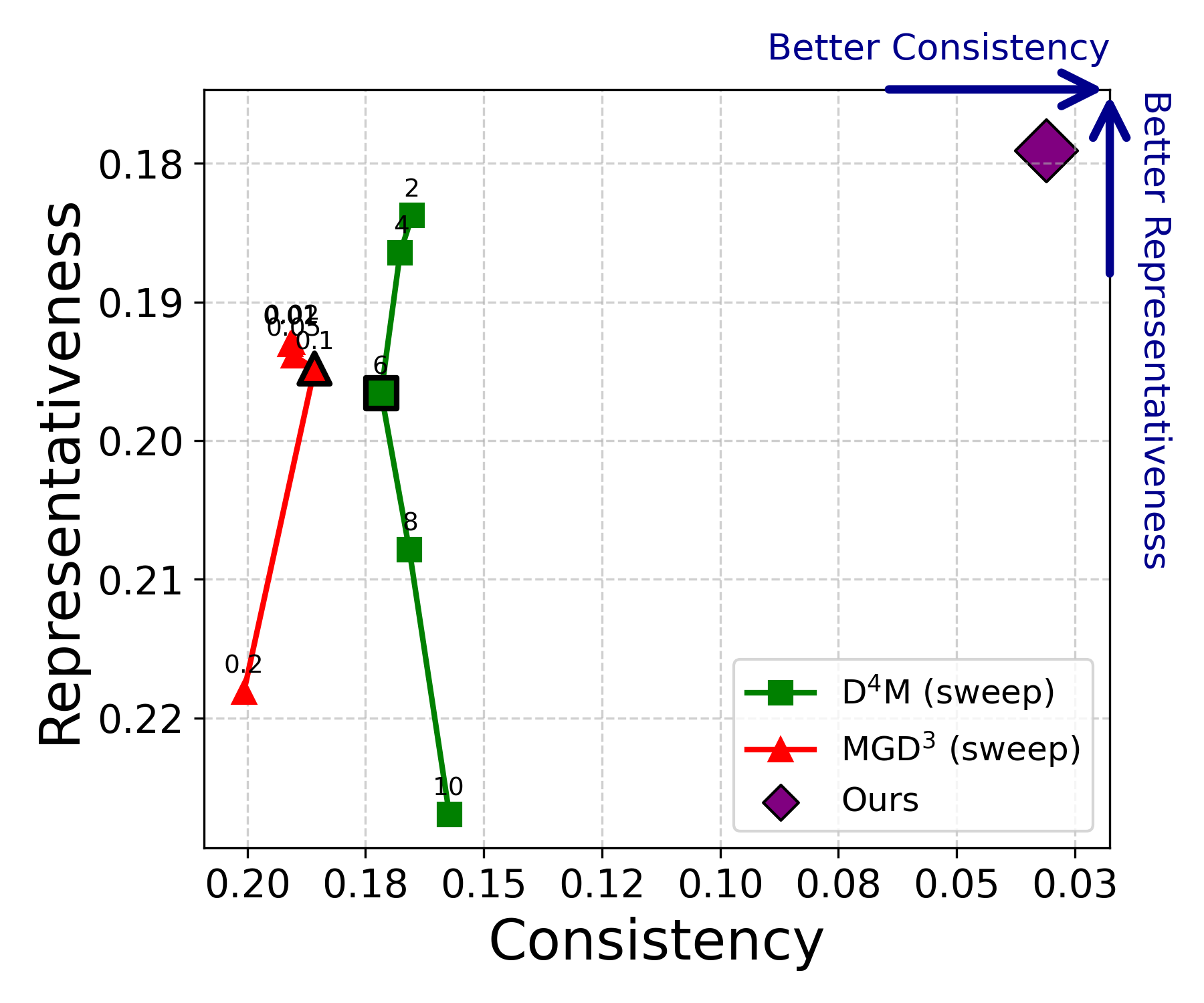} &
        \includegraphics[width=0.49\linewidth]{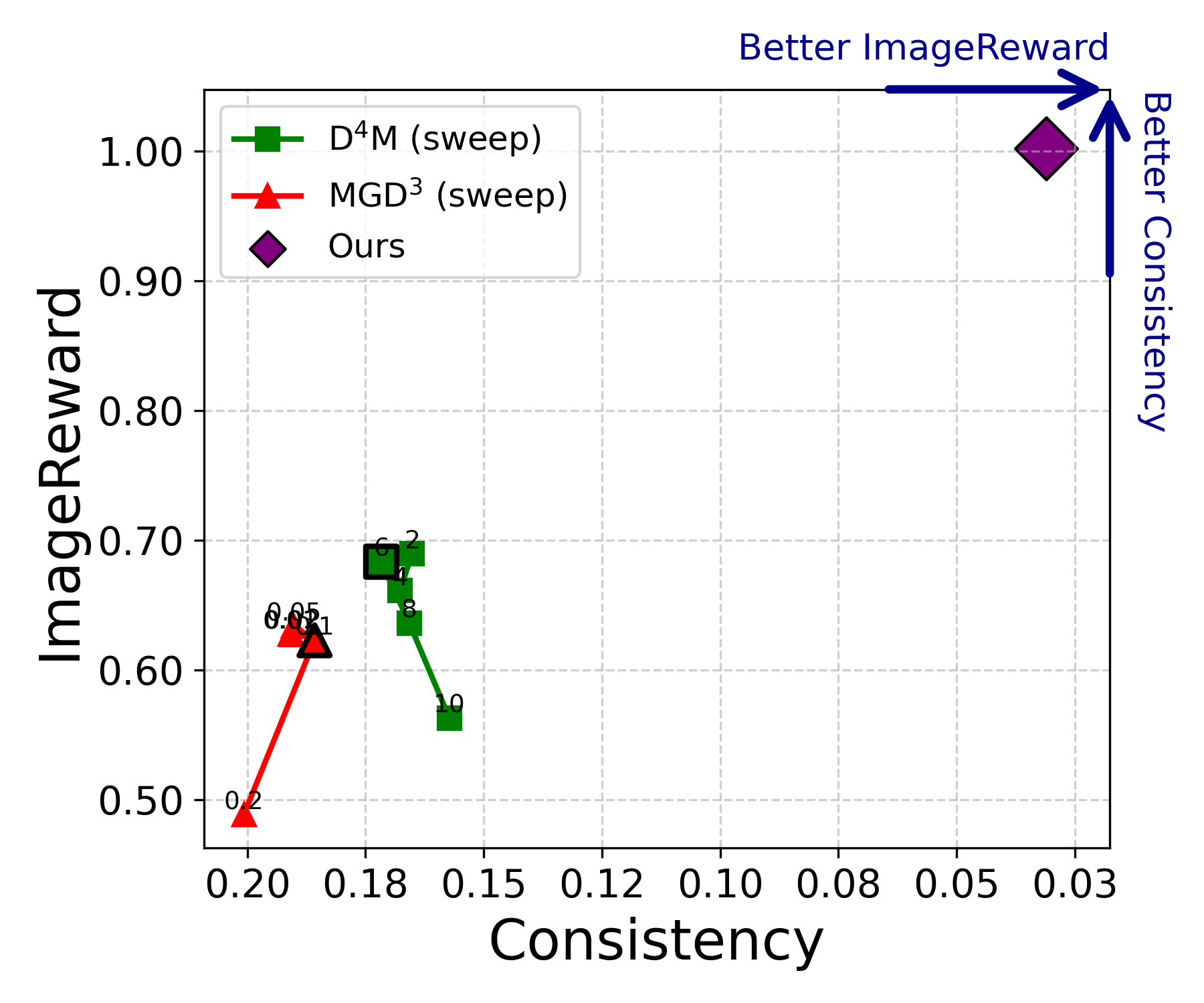}
    \end{tabular}
    \vspace{-10pt}
    \caption{
        \textbf{Quality trade-off across baseline hyperparameters.}
        Across the full range of hyperparameter settings for each baseline, our method consistently achieves superior representativeness, consistency, and visual quality.
    }
    \vspace{-5pt}
    \label{fig:tradeoff_supplement_plots}
\end{figure}
\subsection{Effects of Number of Noisy Latents ($K$)}
\label{sec:effect_num_samples}
\begin{figure*}[ht]
    \centering
    \includegraphics[width=0.95\linewidth]{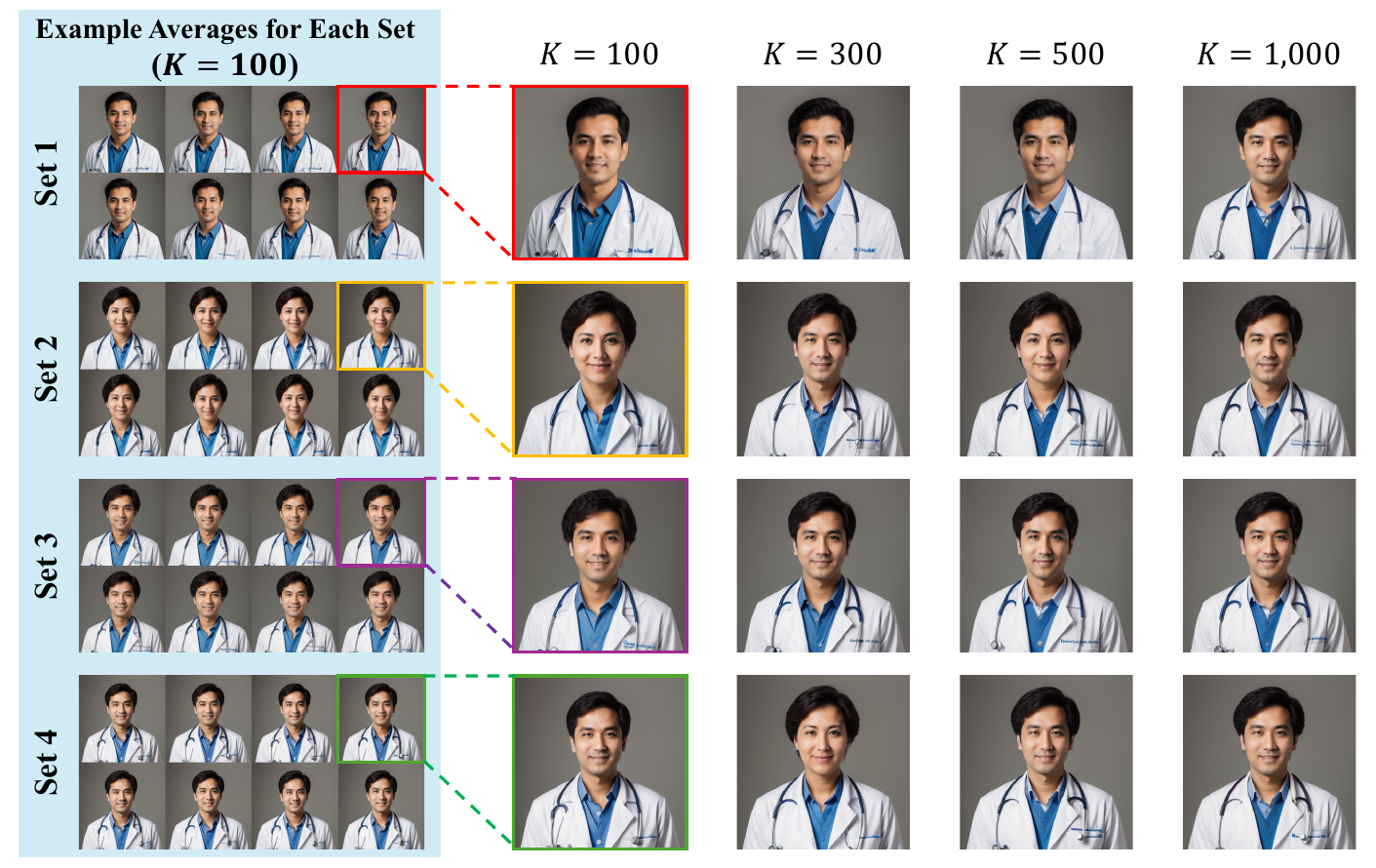}
    \caption{
        \textbf{Effect of the number of noisy latents $K$.}
        In the blue box, we sample four distinct sets of $K=100$ noisy latents and compute their averages. For each set, we show eight example averages (in a 4x2 grid), each computed from a random latent in that set, which converge to similar-looking images \emph{within each set}. However, at $K=100$, noticeable variations are still observed \emph{across the four sets} shown in the images with colored borders. As $K$ increases, averages across distinct sets become more similar, consistent with the statistical behavior of larger sample sizes.
    }
    \label{fig:doctor_num_samples}
\end{figure*}

We analyze how the number of initial noisy latents, $K$, affects the variability of the resulting averages. For each value of $K$, we sample four disjoint sets of $K$ noisy latents, and compute an average image for each set under the same concept prompt. As shown in Figure~\ref{fig:doctor_num_samples}, the variation among the four averages decreases as $K$ increases, consistent with expected statistical behavior.
We find that $K = 1{,}000$ is sufficient for most concepts to appear visually consistent, whereas smaller $K$ often yields noticeable differences across sets.

\begin{figure*}[t]
    \centering
    \renewcommand{\arraystretch}{1.2}
    \setlength{\tabcolsep}{1pt}
    \footnotesize
    \begin{tabular}{c ccc}
        & \shortstack{\textbf{Transformer} \\ \textbf{Block 0}} & \shortstack{\textbf{Transformer} \\ \textbf{Block 13}} & \shortstack{\textbf{Transformer} \\ \textbf{Block 27}} \\
        
        \raisebox{2.9\height}{\rotatebox[origin=c]{90}{\text{baseball player}}} &
        \includegraphics[width=0.3\linewidth]{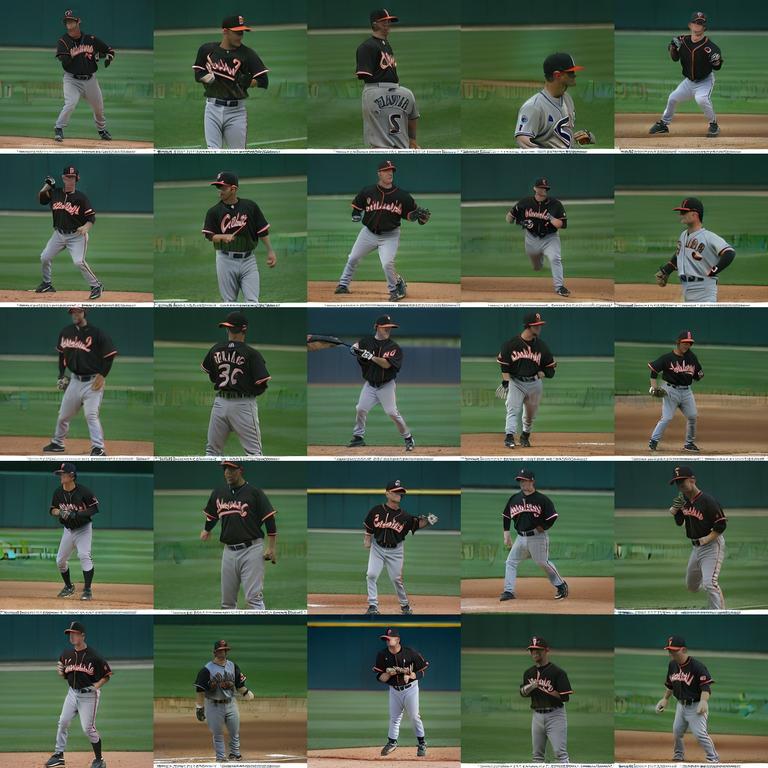} &
        \includegraphics[width=0.3\linewidth]{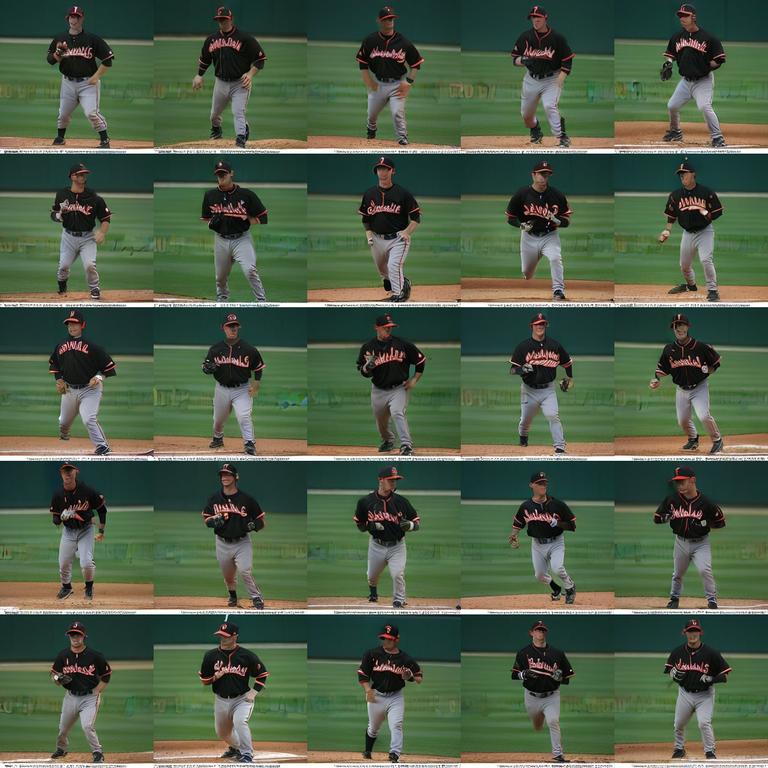} &
        \includegraphics[width=0.3\linewidth]{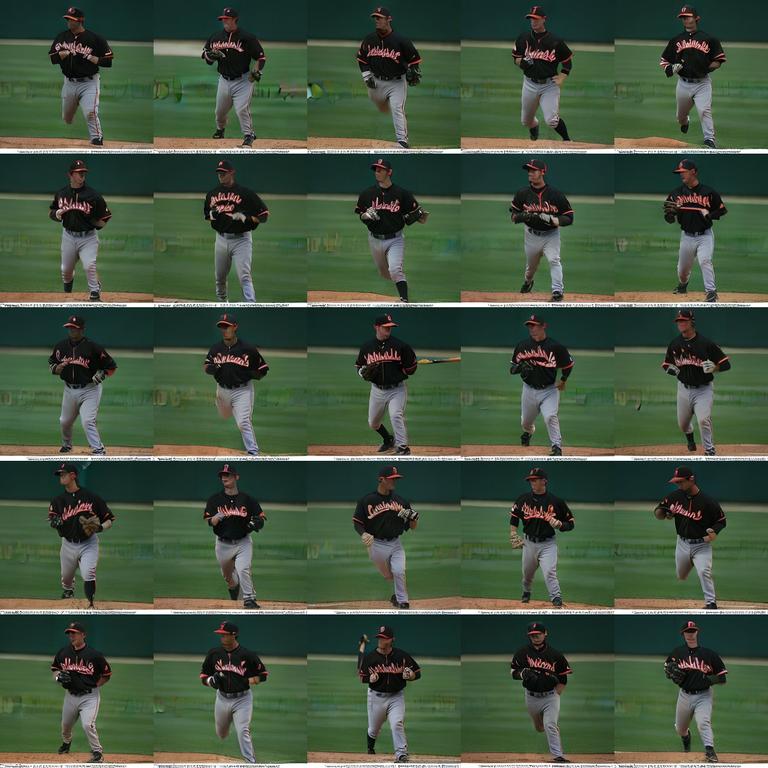}\\

        \raisebox{3.8\height}{\rotatebox[origin=c]{90}{\text{gas helmet}}} &
        \includegraphics[width=0.3\linewidth]{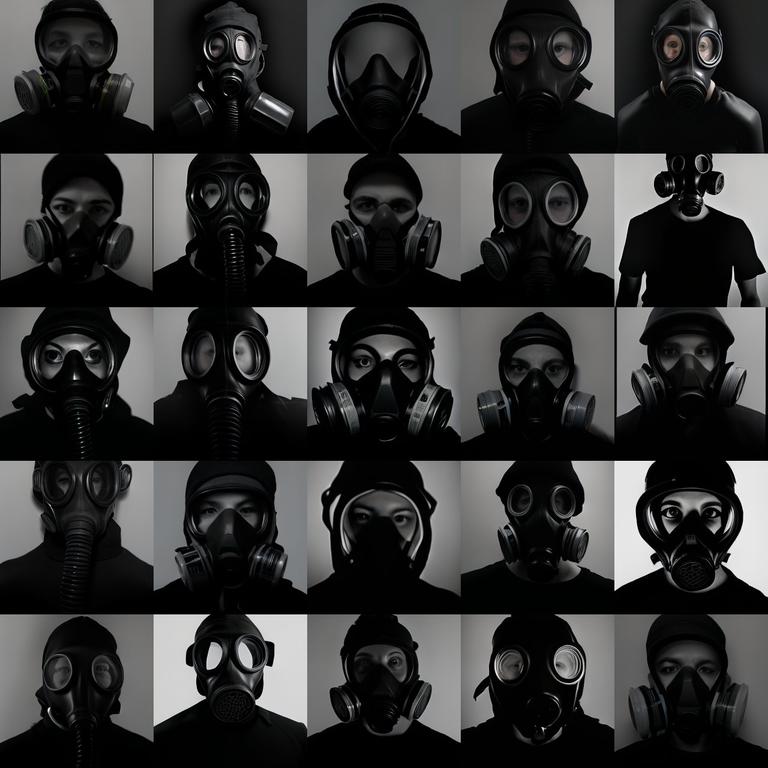} &
        \includegraphics[width=0.3\linewidth]{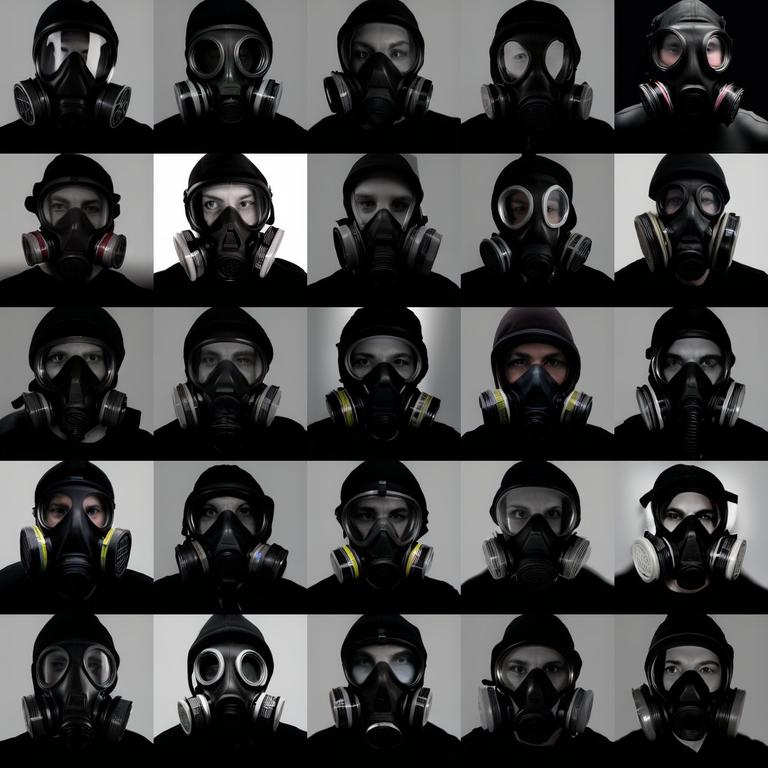} &
        \includegraphics[width=0.3\linewidth]{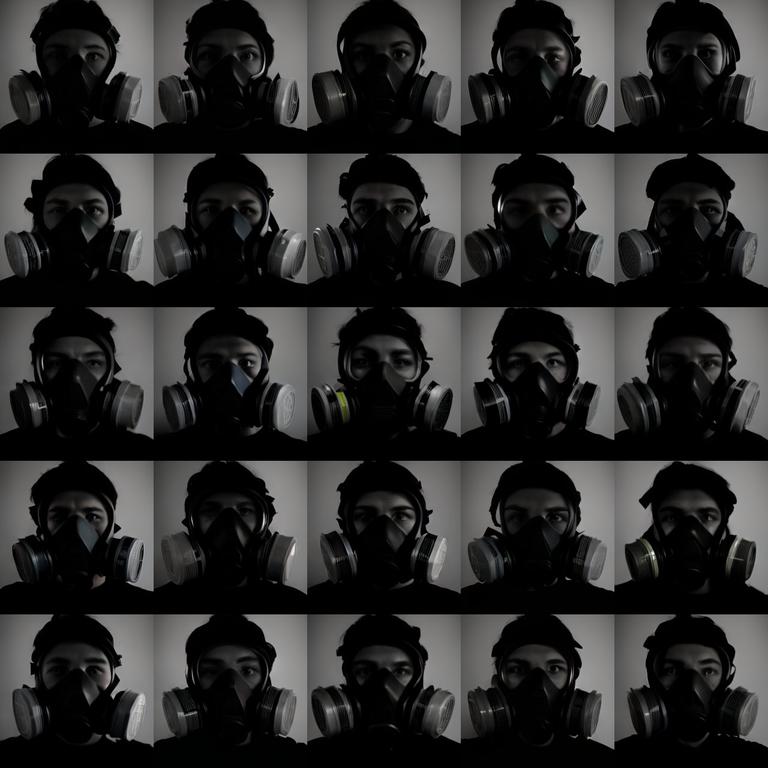}\\

        \raisebox{7\height}{\rotatebox[origin=c]{90}{\text{crane}}} &
        \includegraphics[width=0.3\linewidth]{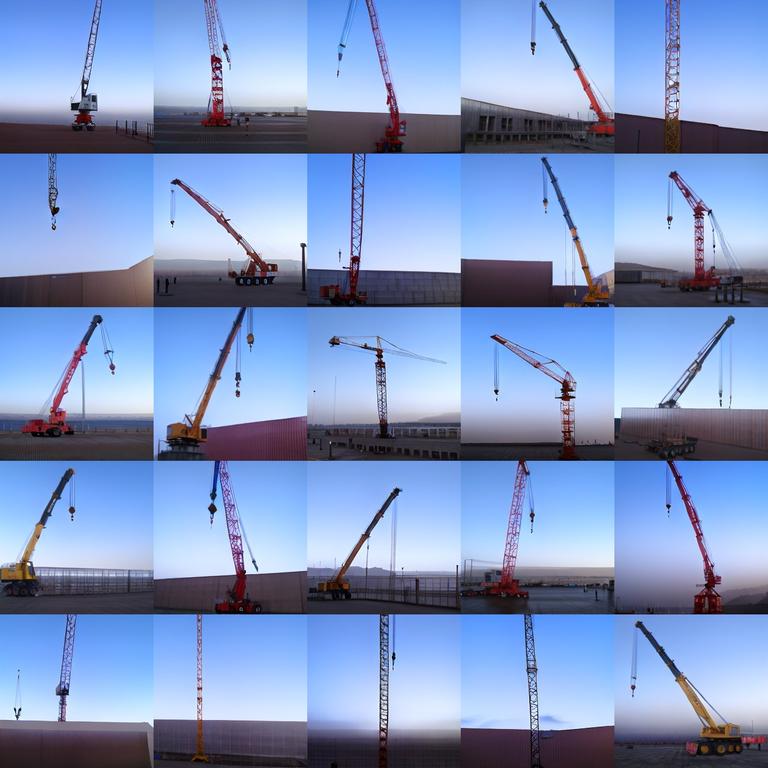} &
        \includegraphics[width=0.3\linewidth]{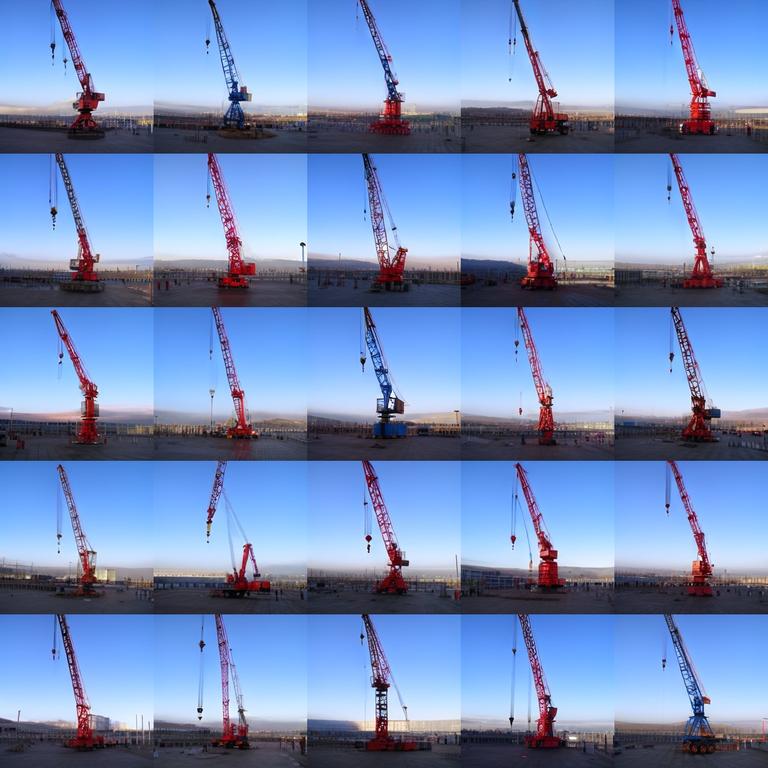} &
        \includegraphics[width=0.3\linewidth]{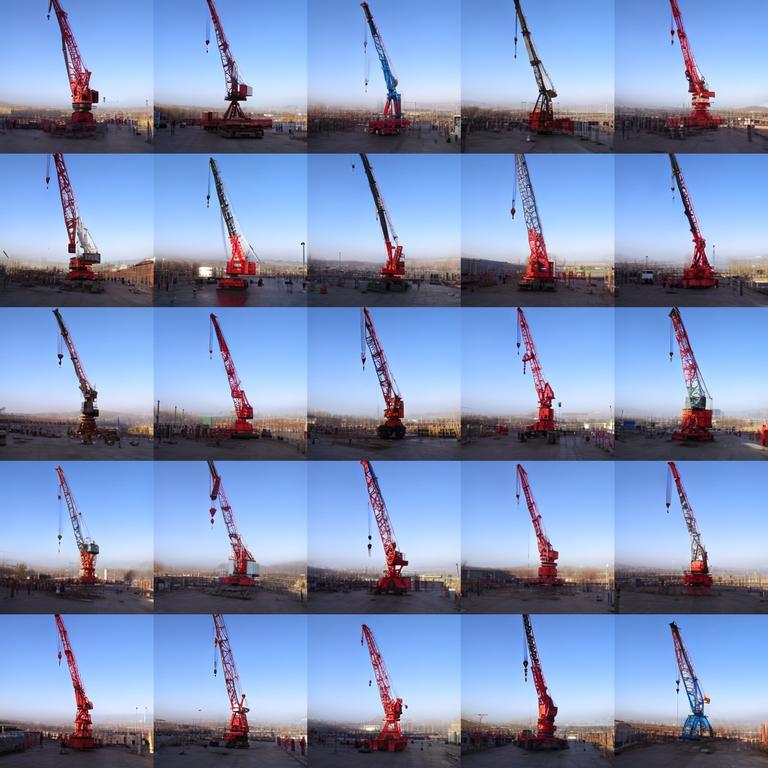}\\

        \raisebox{6.7\height}{\rotatebox[origin=c]{90}{\text{husky}}} &
        \includegraphics[width=0.3\linewidth]{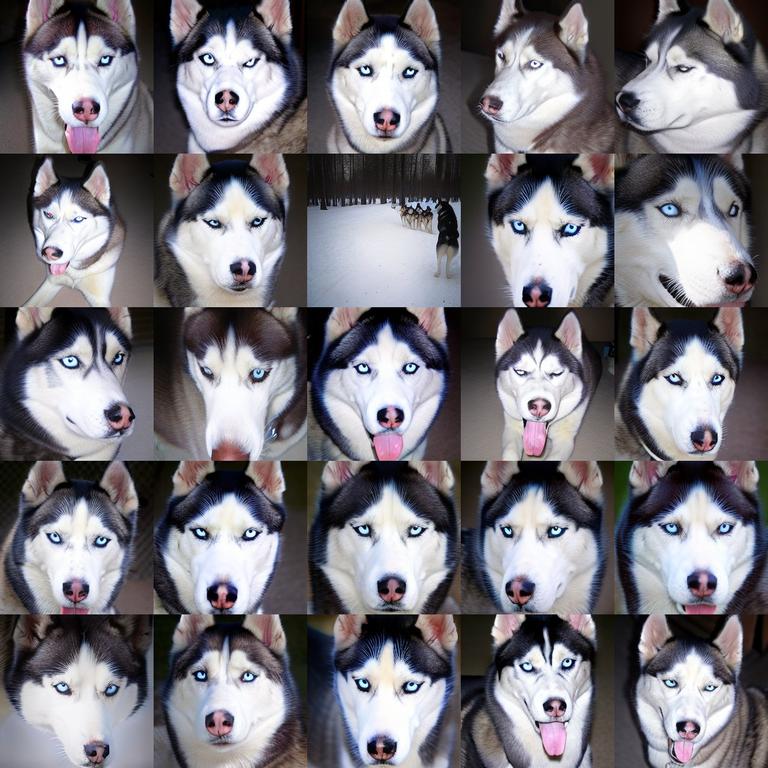} &
        \includegraphics[width=0.3\linewidth]{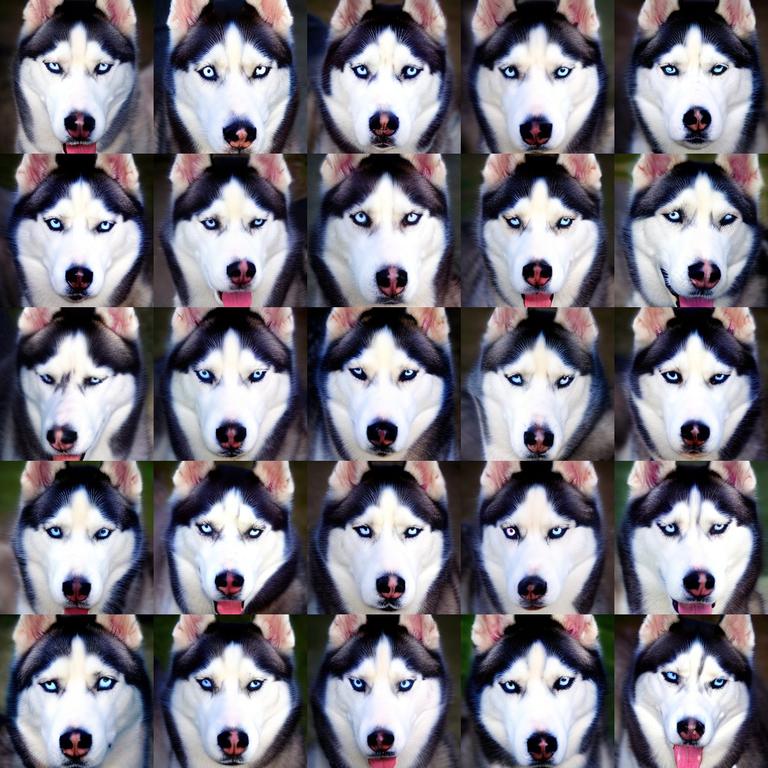} &
        \includegraphics[width=0.3\linewidth]{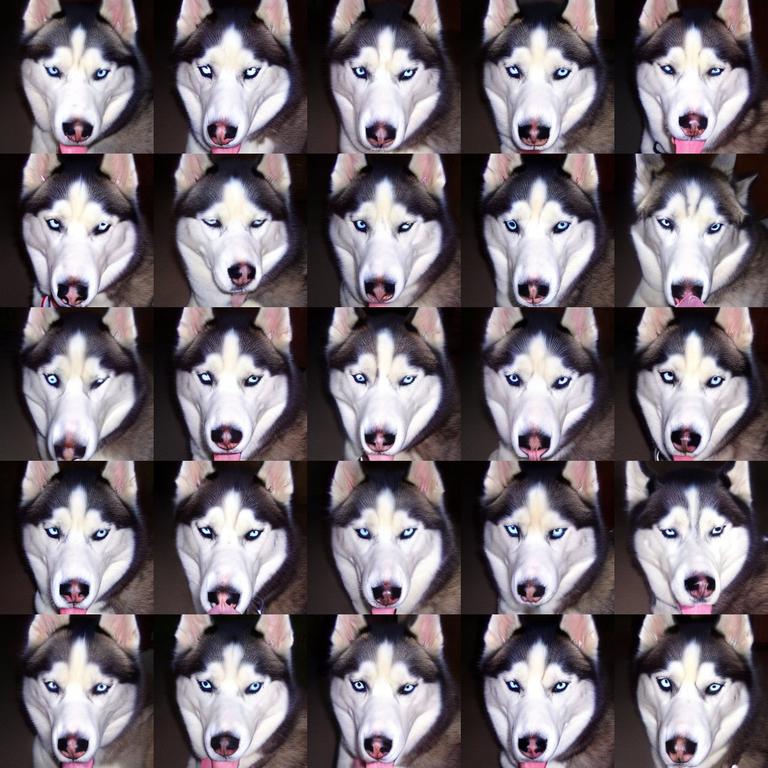}\\
    \end{tabular}
    \vspace{0pt}
    \caption{\textbf{Ablation on DiT transformer block selection for DMA. } See details in Section~\ref{sec:dit_layer}.}
    \label{fig:ab_dit_block}
\end{figure*}

\subsection{DiT Semantic Layer Selection}\label{sec:dit_layer}
In Figure~\ref{fig:ab_dit_block}, we experiment with using the output of different transformer blocks for DMA. Preliminary results indicate that the final block produces the most consistent results. While these findings demonstrate the feasibility of extending DMA to DiT, this block-selection strategy is likely suboptimal, as DiT contains many other components that may provide more meaningful latent representations. A more comprehensive analysis is needed to understand how transformer-based models behave.


\section{Implementation Details of D\texorpdfstring{$^4$}{4}M and MGD\texorpdfstring{$^3$}{3}}\label{sec:d_hyperparams}
\hypertarget{my:d_hyperparams}{}
Both \textbf{D$^4$M}~\cite{su2024d} and \textbf{MGD$^3$}~\cite{chan2025mgd} are dataset-distillation methods that compress a full dataset into a small set of synthetic prototypes. Unlike earlier approaches that train a classifier in the loop, they generate representatives directly from a pre-trained diffusion model, without extra optimization or task-specific losses, and benefit from the model’s generative prior to produce higher-quality results. Both encode the dataset using a VAE, cluster the latents into $w$ clusters (where $w$ corresponds to the number of output prototypes per class (IPC)), and use the cluster centroids as prototype latents for guiding synthesis. \textbf{D$^4$M} adds noise to each centroid and denoises it (as in SDEdit~\cite{meng2021sdedit}), whereas \textbf{MGD$^3$} introduces \emph{Mode Guidance}, using the centroid as an attractor that steers the denoising trajectory at every timestep.


In our experiment in the main paper, we adapt \textbf{D$^4$M} and \textbf{MGD$^3$} to evaluate how well they can produce an \emph{average image} of a diffusion model. Given generated samples from a single class, we treat them as the ``dataset to distill'' and distill them into a single prototype (IPC = 1). We re-implemented both baselines on the same Stable Diffusion model as ours, using the same VAE encoder/decoder, and a 20-step DDIM sampler with classifier-free guidance scale 7.0.

\begin{figure*}[t]
\centering
\setlength{\tabcolsep}{1pt}
\renewcommand{\arraystretch}{1.1}

\begin{tabular}{c ccccc}
& $t=2$ & $t=4$ & $t=6$ & $t=8$ & $t=10$ \\

\raisebox{3.3\height}{\rotatebox[origin=c]{90}{\textbf{Bird}}} &
\includegraphics[width=0.19\linewidth]{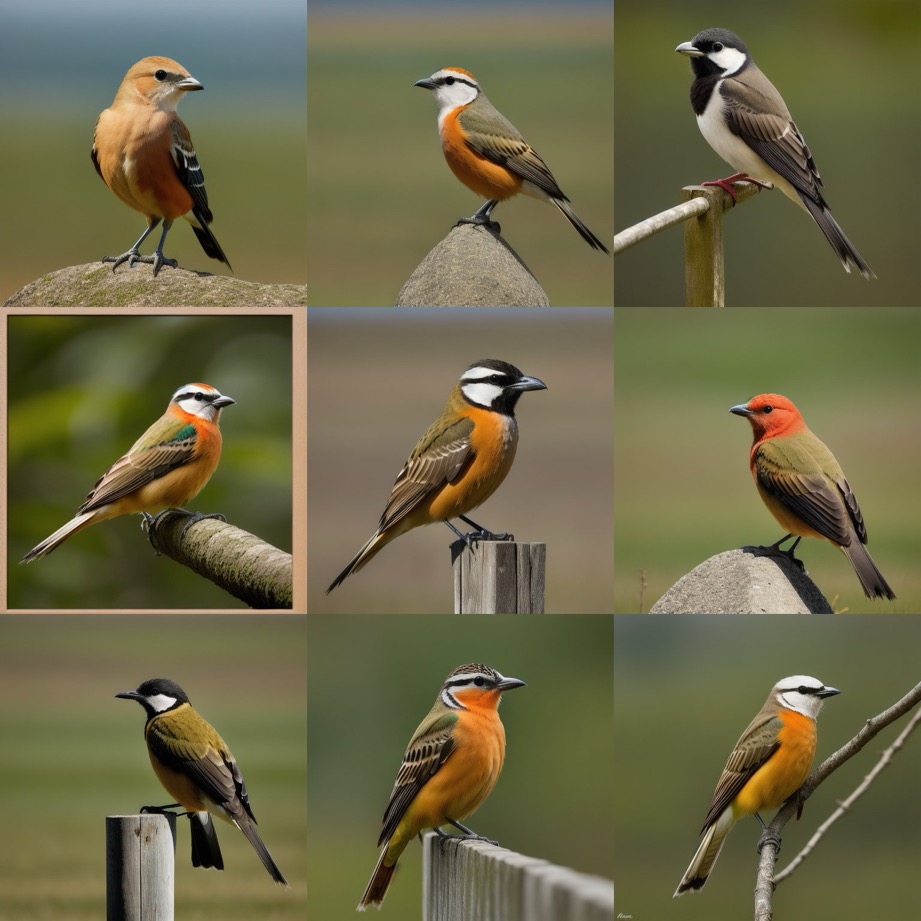} &
\includegraphics[width=0.19\linewidth]{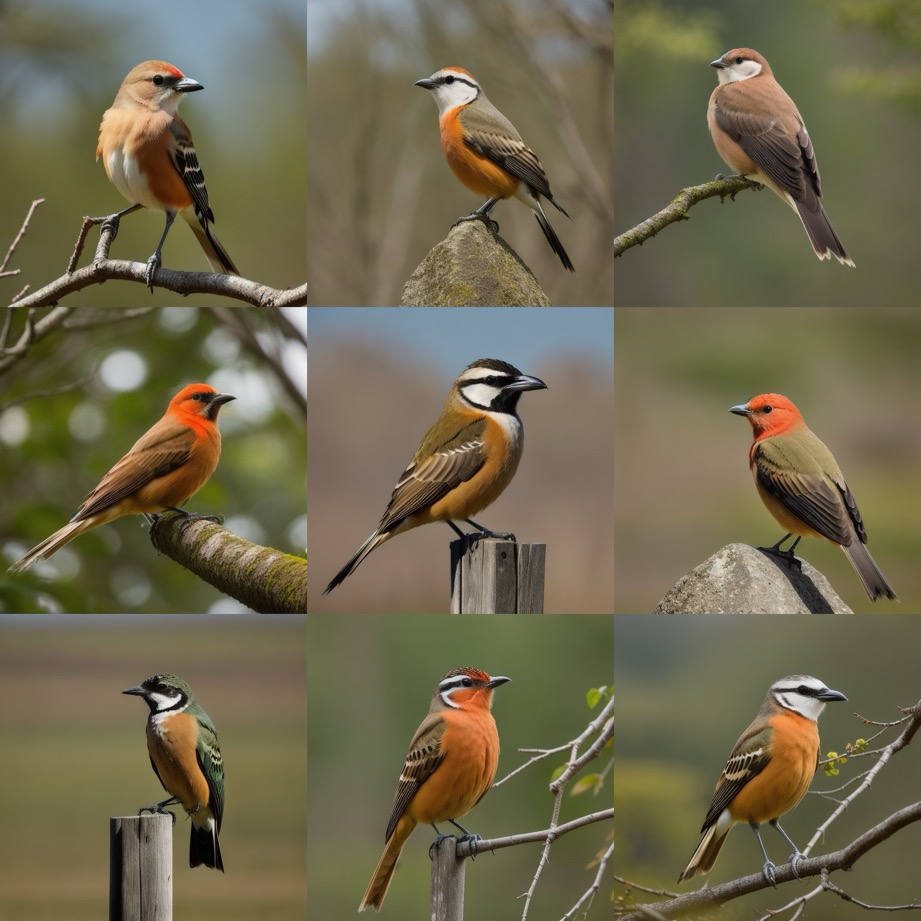} &
\includegraphics[width=0.19\linewidth]{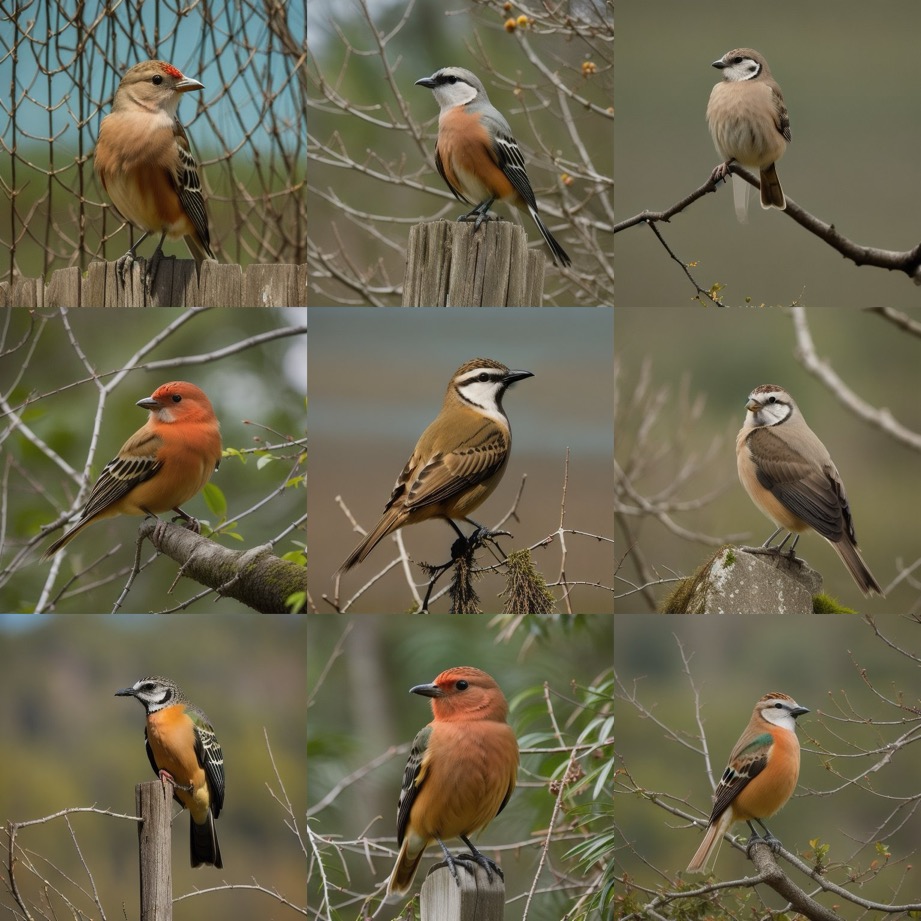} &
\includegraphics[width=0.19\linewidth]{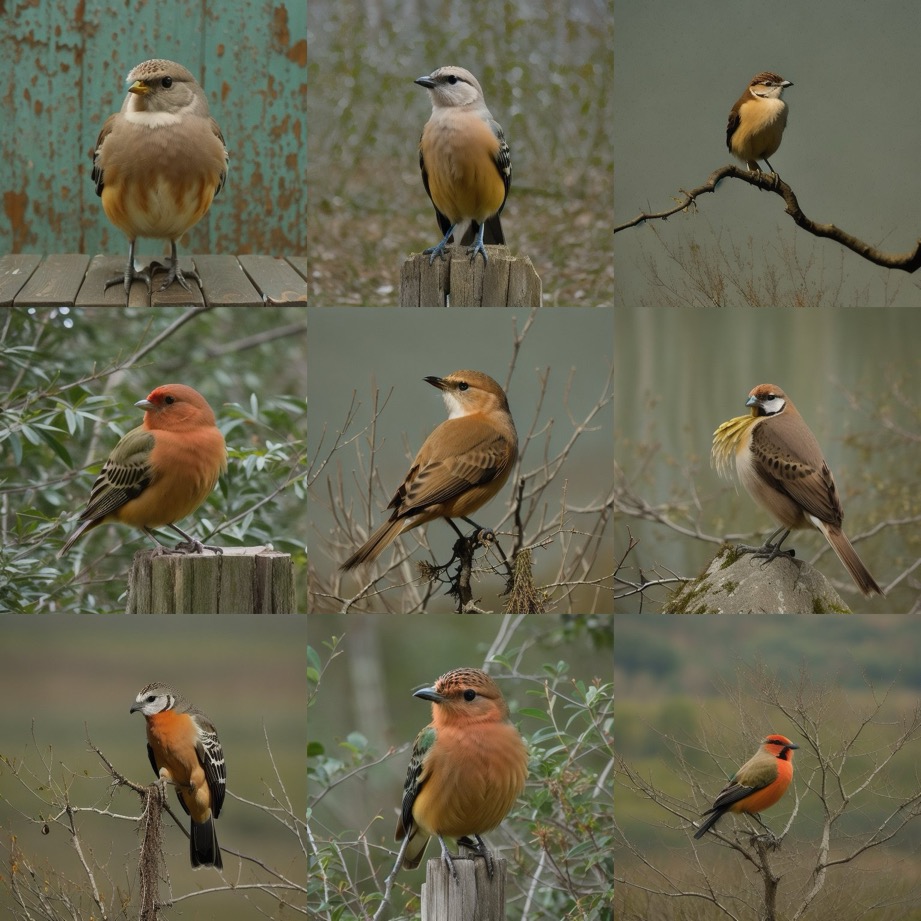} &
\includegraphics[width=0.19\linewidth]{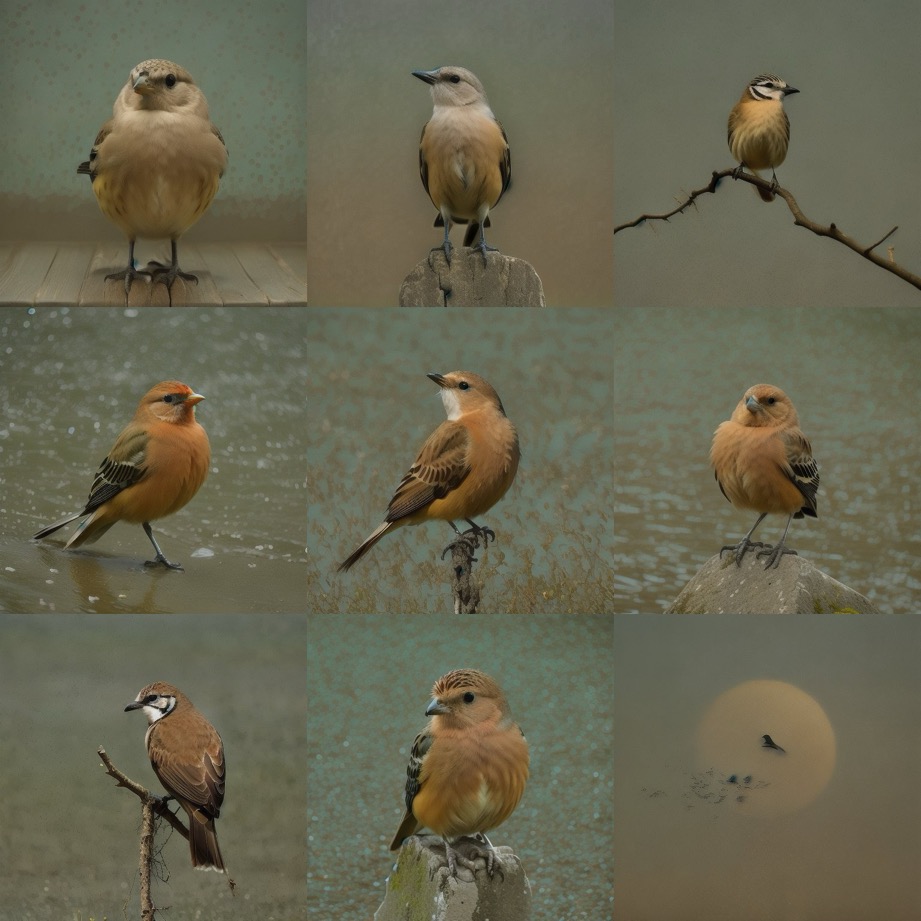}
\\[-2pt]

\raisebox{1.7\height}{\rotatebox[origin=c]{90}{\textbf{Astronaut}}} &
\includegraphics[width=0.19\linewidth]{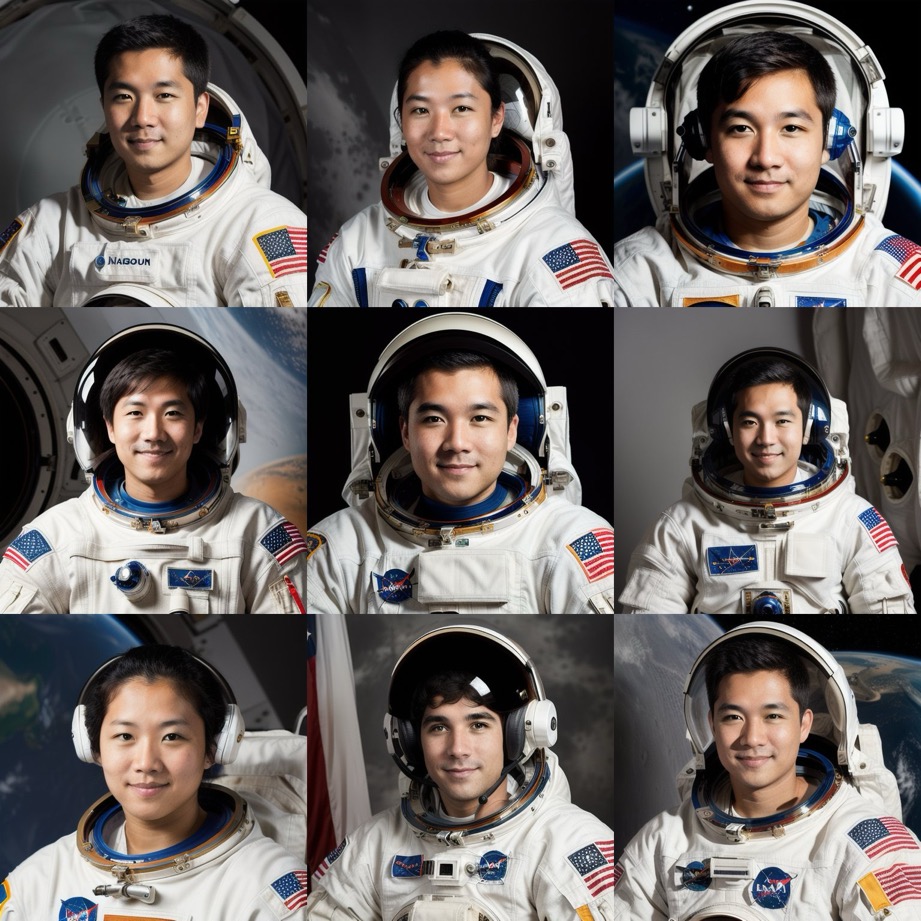} &
\includegraphics[width=0.19\linewidth]{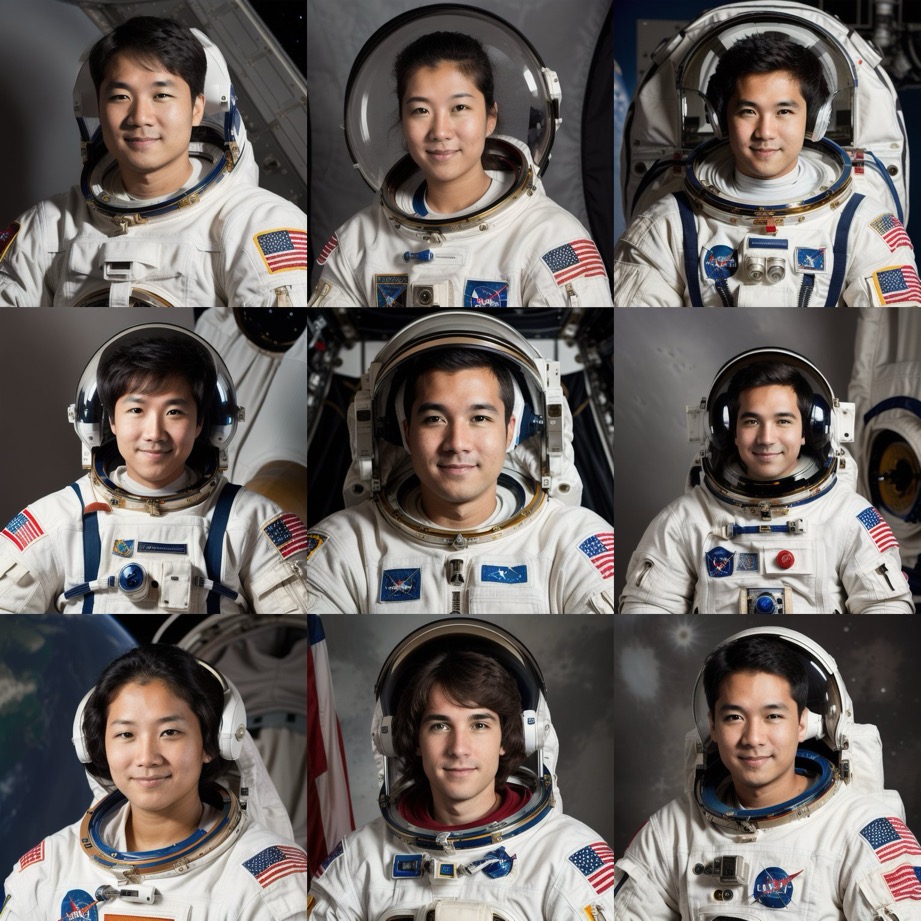} &
\includegraphics[width=0.19\linewidth]{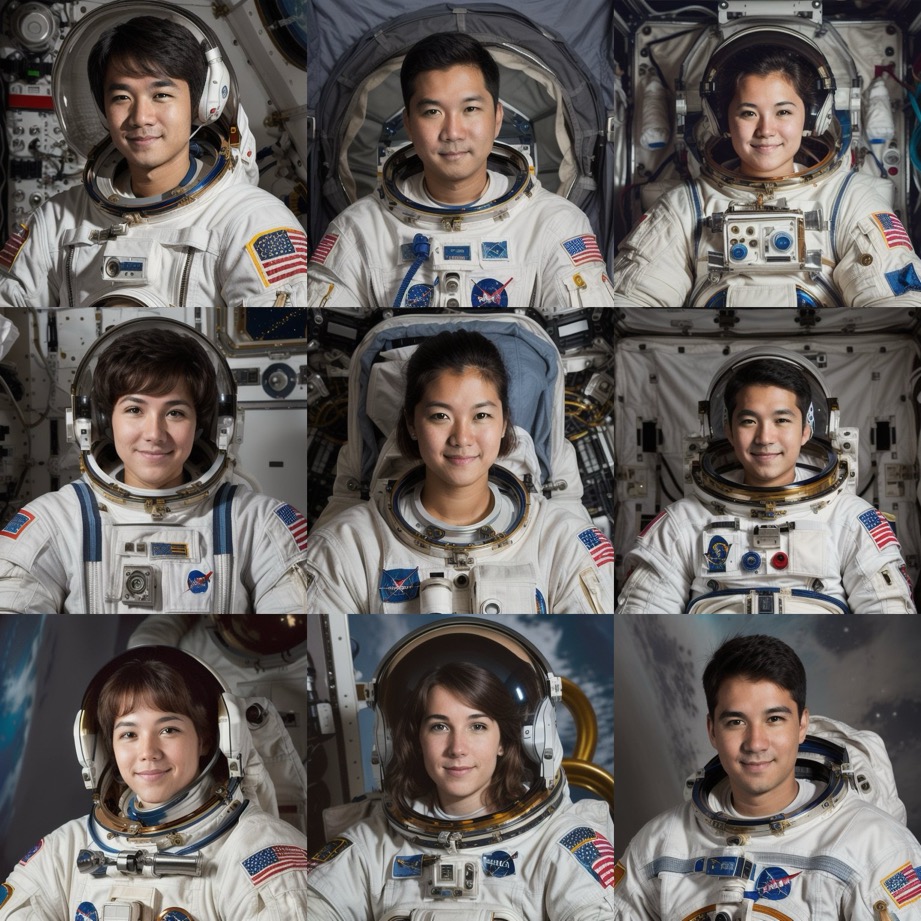} &
\includegraphics[width=0.19\linewidth]{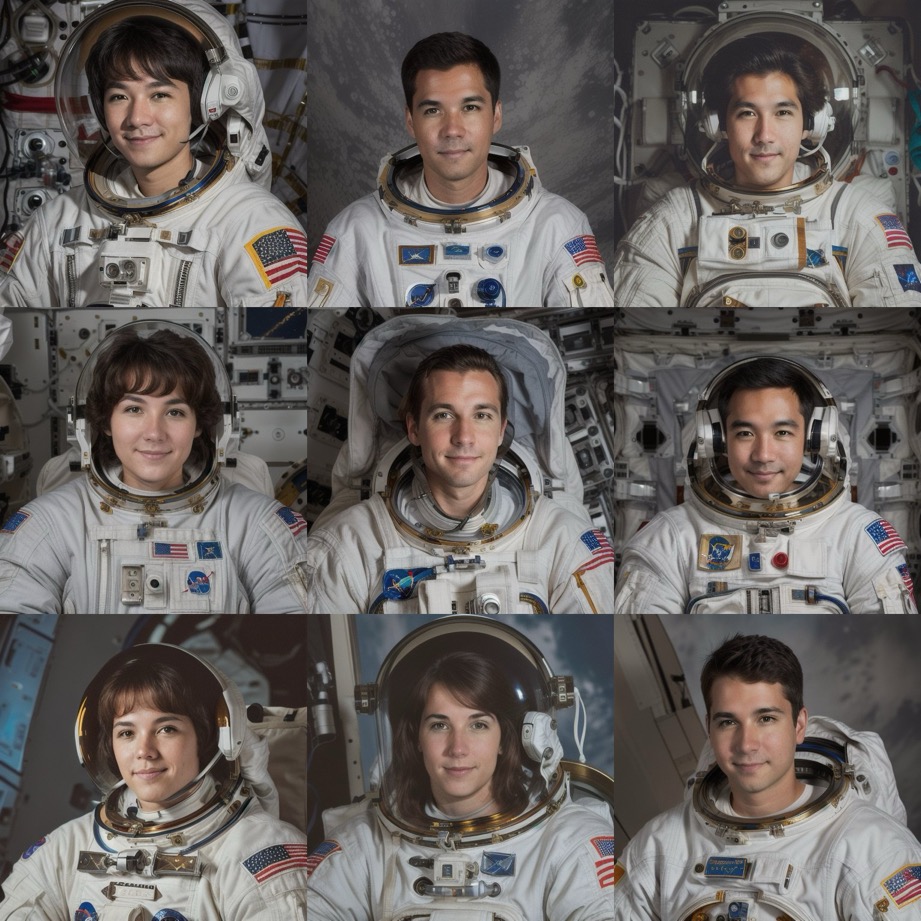} &
\includegraphics[width=0.19\linewidth]{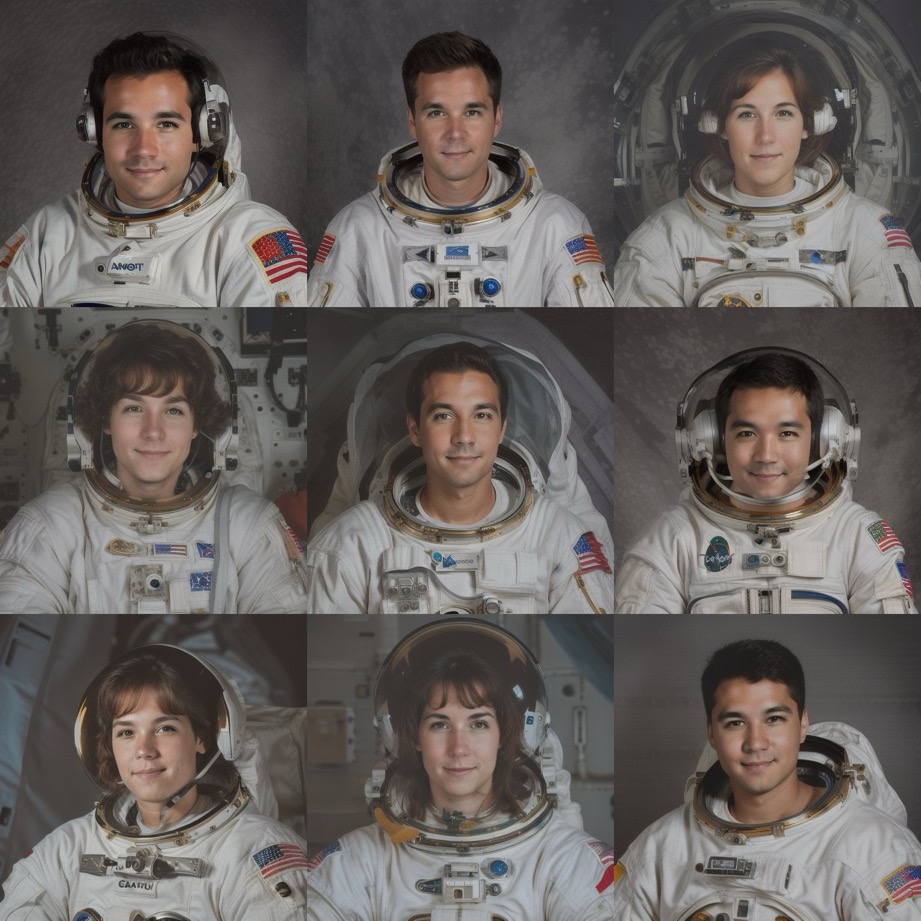}
\\[-2pt]

\raisebox{3.6\height}{\rotatebox[origin=c]{90}{\textbf{Car}}} &
\includegraphics[width=0.19\linewidth]{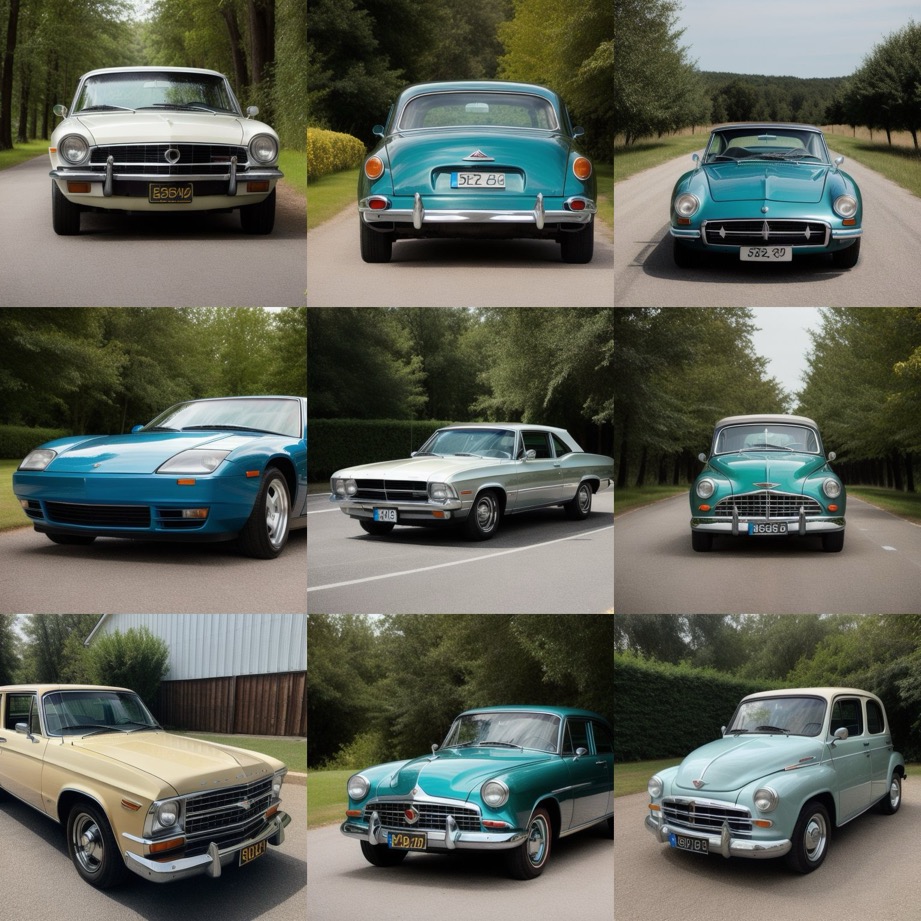} &
\includegraphics[width=0.19\linewidth]{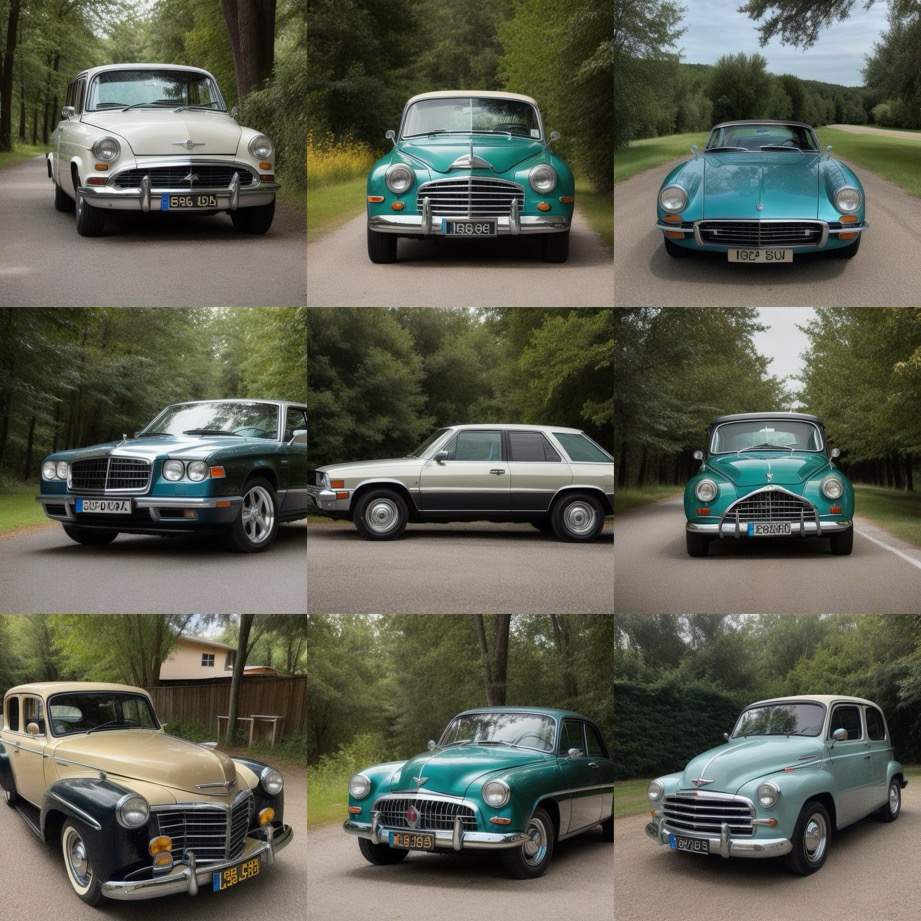} &
\includegraphics[width=0.19\linewidth]{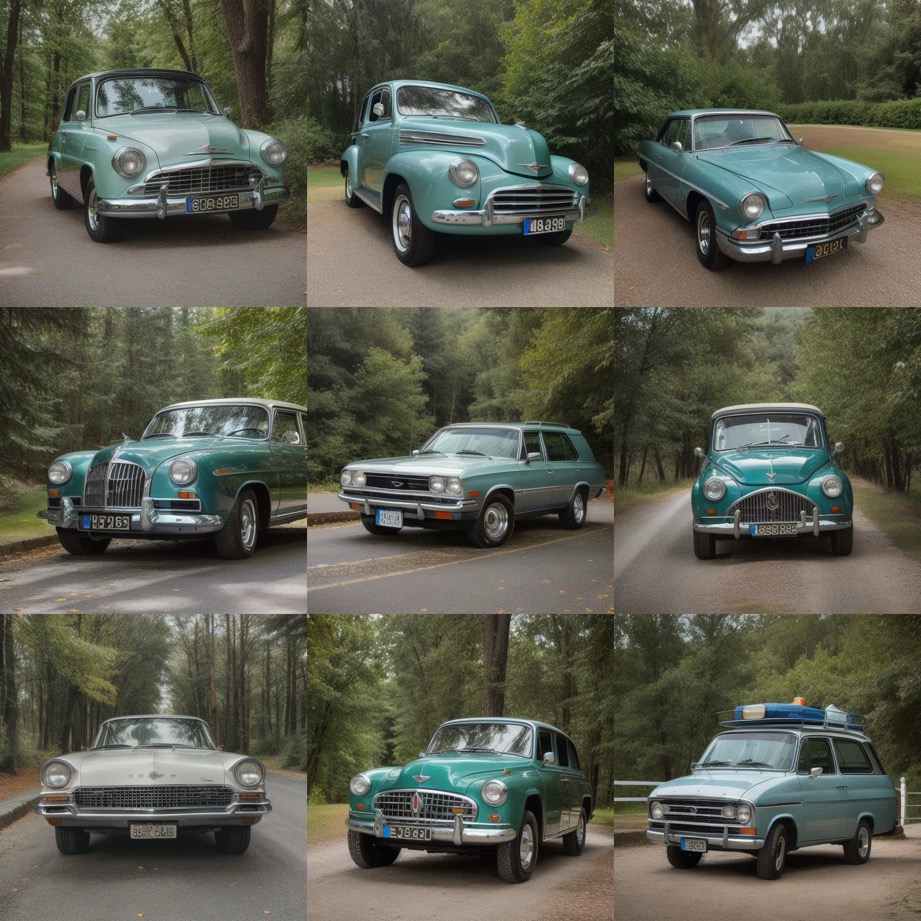} &
\includegraphics[width=0.19\linewidth]{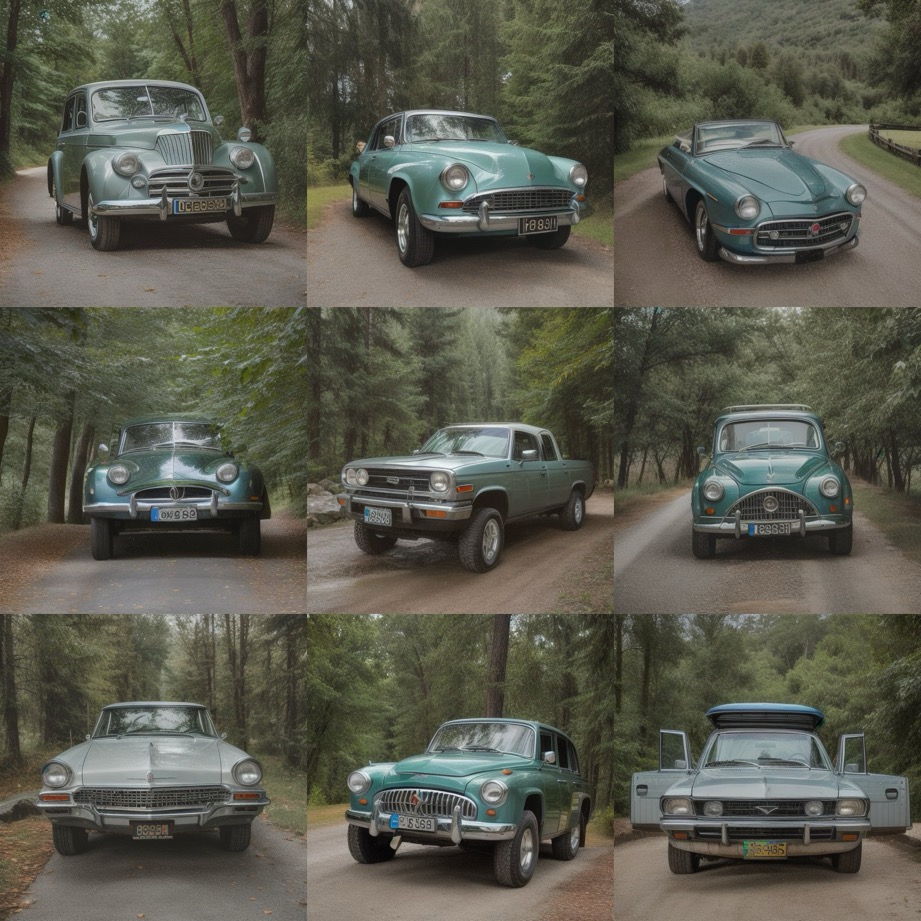} &
\includegraphics[width=0.19\linewidth]{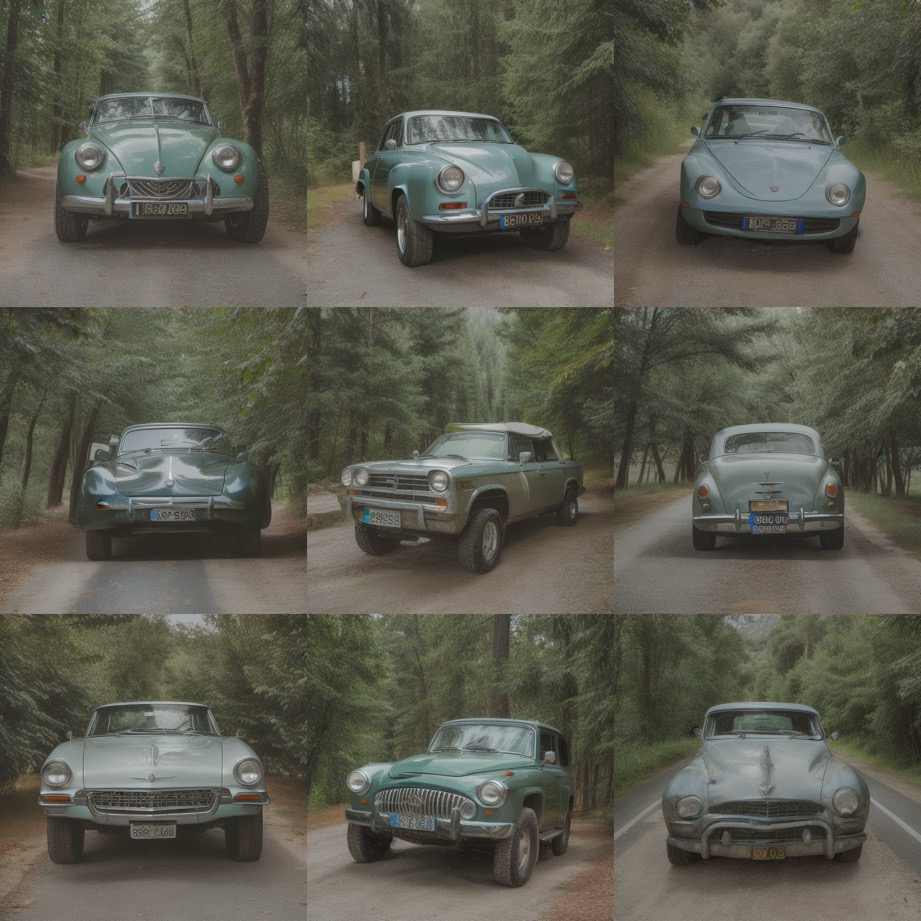}
\\[-2pt]

\raisebox{1.9\height}{\rotatebox[origin=c]{90}{\textbf{Freedom}}} &
\includegraphics[width=0.19\linewidth]{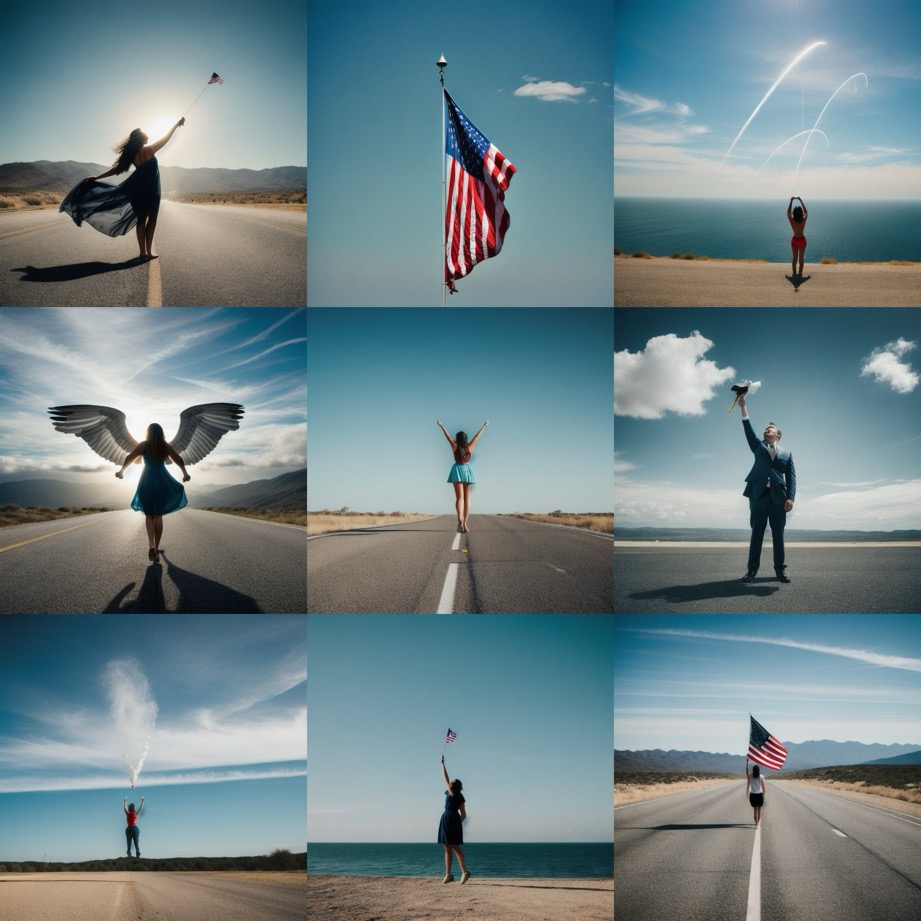} &
\includegraphics[width=0.19\linewidth]{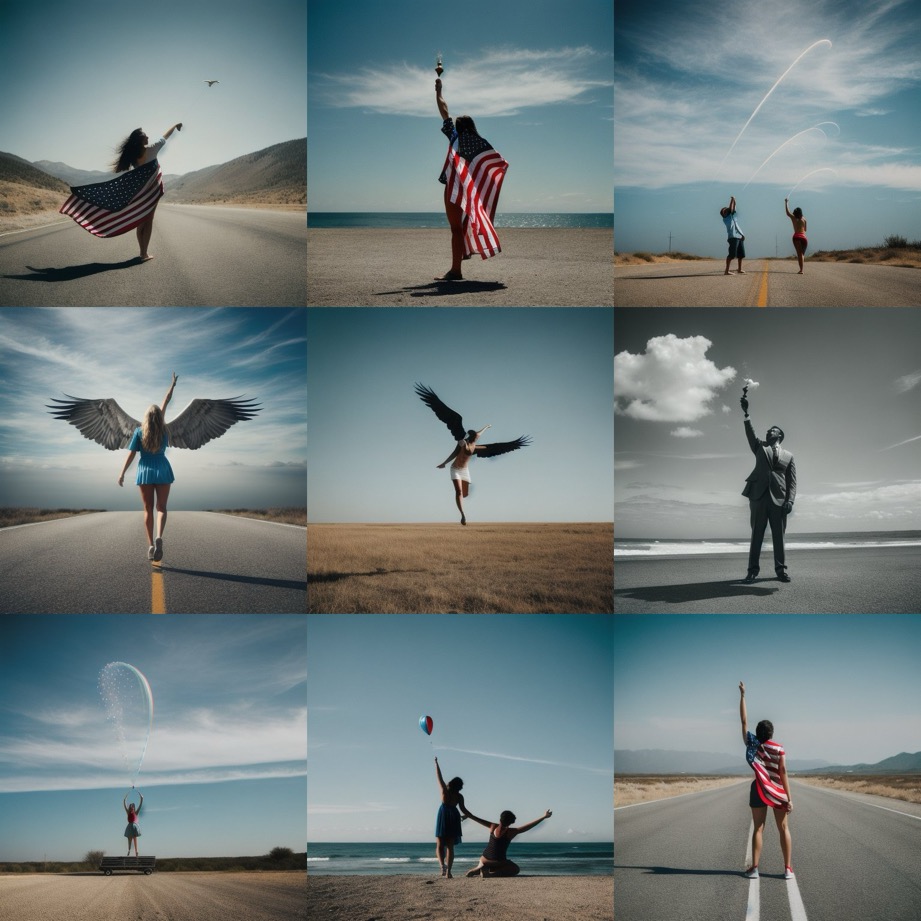} &
\includegraphics[width=0.19\linewidth]{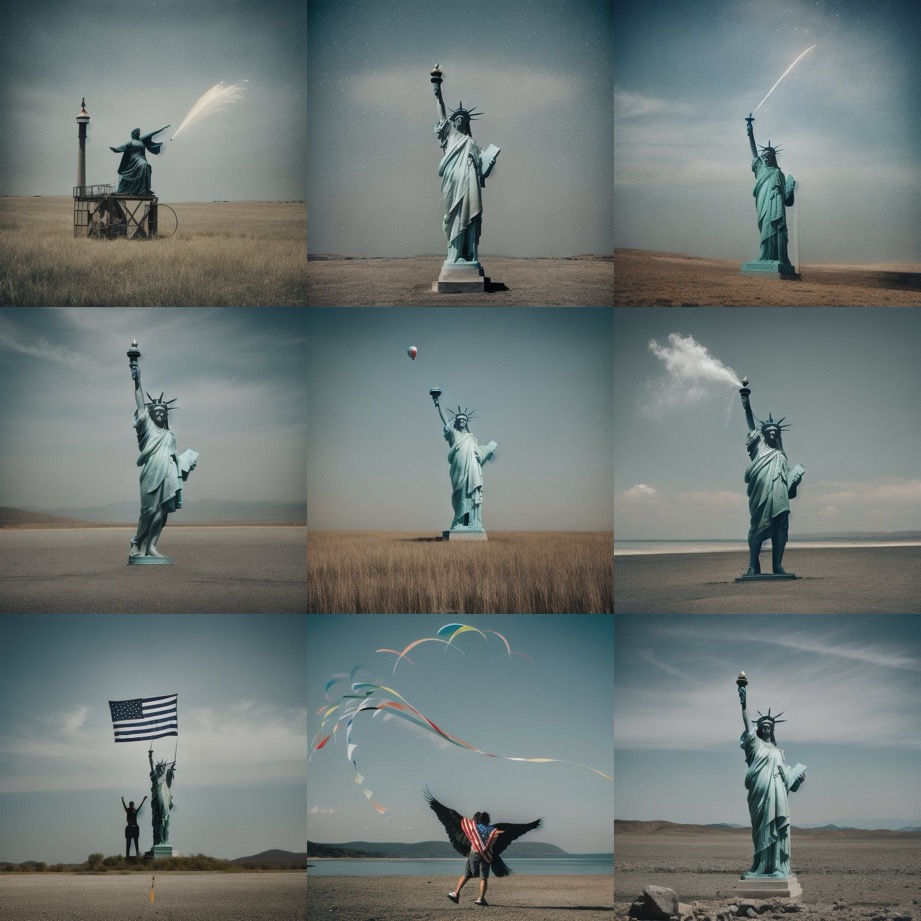} &
\includegraphics[width=0.19\linewidth]{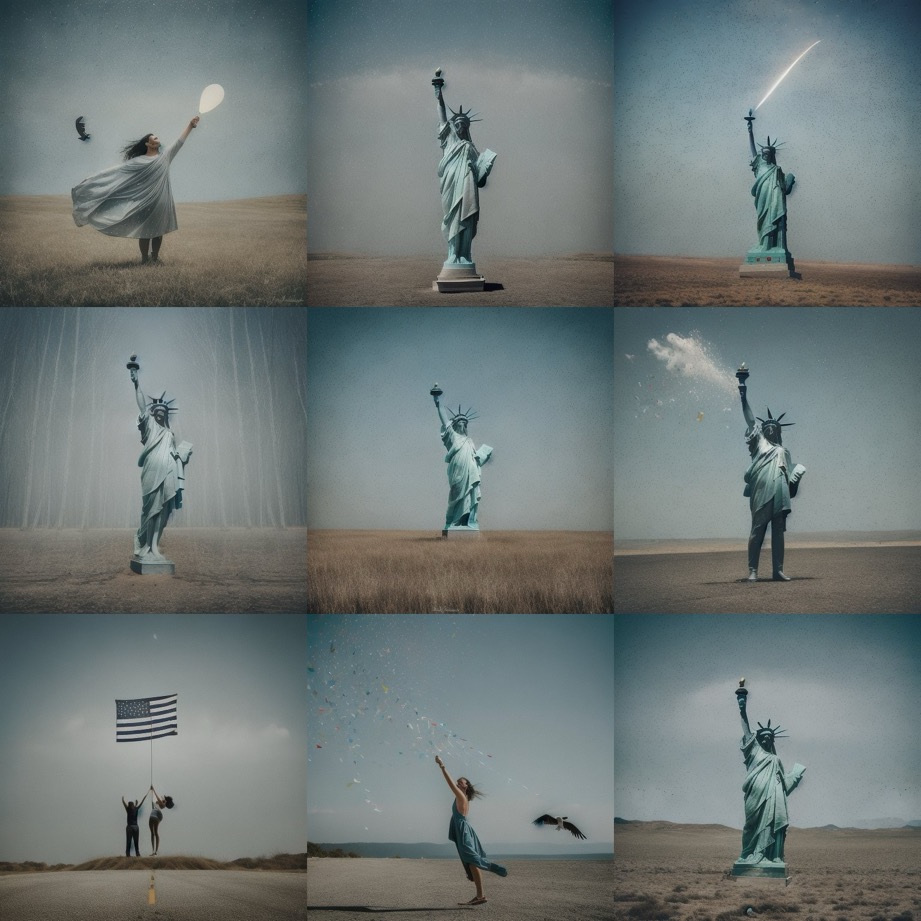} &
\includegraphics[width=0.19\linewidth]{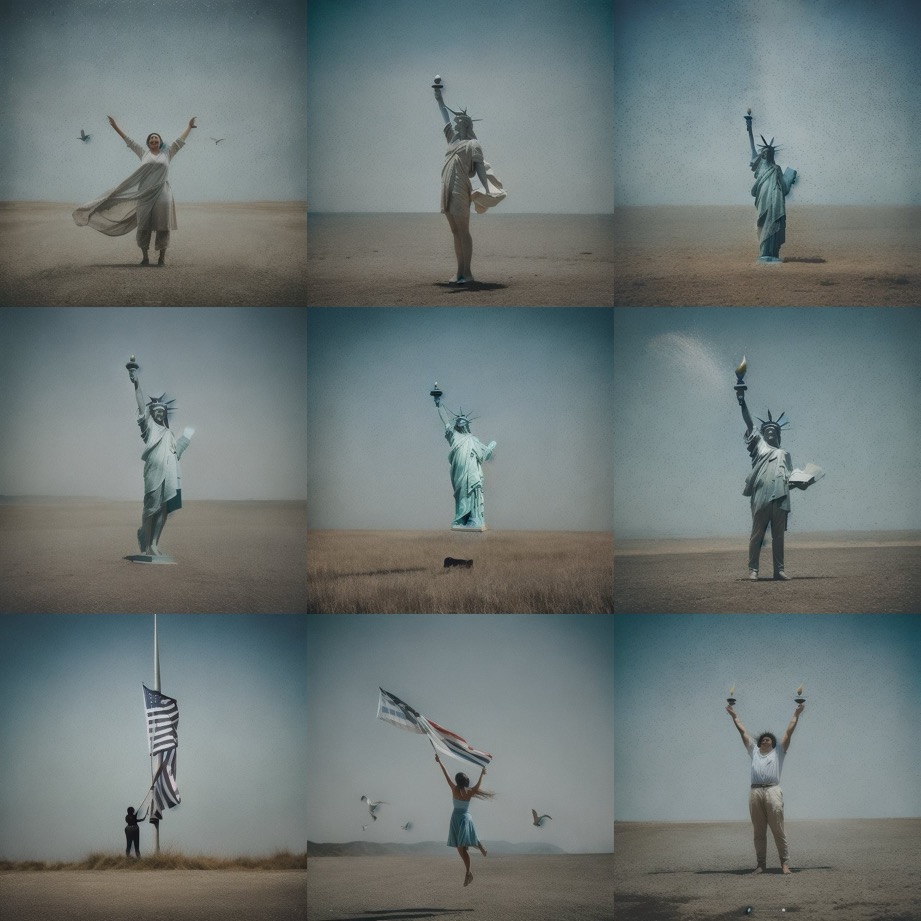}
\\

\end{tabular}

\caption{\textbf{Hyperparameter tuning of D$^4$M for fair comparison.}
We sweep the noise-injection timestep $t \in \{2,4,6,8,10\}$ across four concepts: \textit{bird}, \textit{astronaut}, \textit{car}, and \textit{freedom}. Larger values of~$t$ preserve the averaged latent more strongly but lead to blurrier outputs, while smaller values make the result less consistent.}
\label{fig:d4m_sweep}
\end{figure*}

\boldsubsection{D$^4$M~\cite{su2024d}.}
We implement D$^4$M following Su~\etal~\cite{su2024d}, adapting it to our single-class setting in which all samples belong to the same class. Since we aim to distill the class into one representative image (IPC = 1), the cluster centroid simplifies to the \textbf{average VAE latent} computed over all samples. D$^4$M injects noise into this prototype at diffusion timestep $t$ using the forward diffusion step, followed by a denoising loop to synthesize the final representative image. For a fair comparison, we sweep the noise-injection timestep $t \in \{2,4,6,8,10\}$, where the main-paper configuration $t=6$ corresponds to the recommended SDEdit strength of 0.7. Results are shown in Figure~\ref{fig:d4m_sweep}.

\boldsubsection{MGD$^3$~\cite{chan2025mgd}.} 
We implement MGD$^3$ following Algorithm~1 in Chan-Santiago~\etal~\cite{chan2025mgd}, adapting it to our single-class setting in which all samples belong to the same class. In this case, the estimate prototype $m$ reduces to the \textbf{average VAE latent}. Starting from noisy latents, at each denoising step $t$, we compute the DDIM prediction of the clean latent $\widehat{\mathbf{z}}^{\,(T)}_t$ and form the mode-guidance direction $\mathbf{g}_t = \mathbf{m} - \widehat{\mathbf{z}}^{\,(T)}_t$. The predicted noise is then modified according to Algorithm~1:

\[
\hat{\epsilon} \leftarrow \epsilon_\theta(\textbf{z}, t, c)
\, - \,
\sqrt{1-\bar{\alpha}_t}\,\lambda\,\mathbf{g}_t.
\]

\begin{figure*}[t]
\centering
\setlength{\tabcolsep}{1pt}
\renewcommand{\arraystretch}{1.1}

\begin{tabular}{c ccccc}
& $\lambda=0.01$ & $\lambda=0.02$ & $\lambda=0.05$ & $\lambda=0.1$ & $\lambda=0.2$ \\

\raisebox{3.3\height}{\rotatebox[origin=c]{90}{\textbf{Bird}}} &
\includegraphics[width=0.19\linewidth]{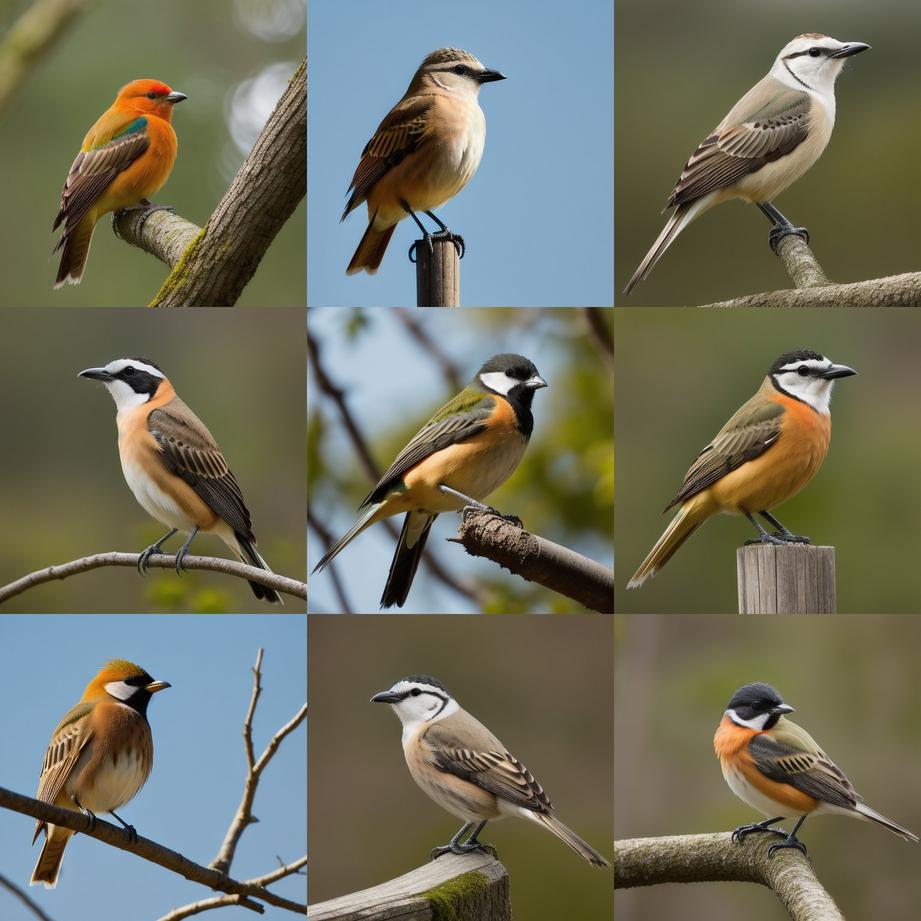} &
\includegraphics[width=0.19\linewidth]{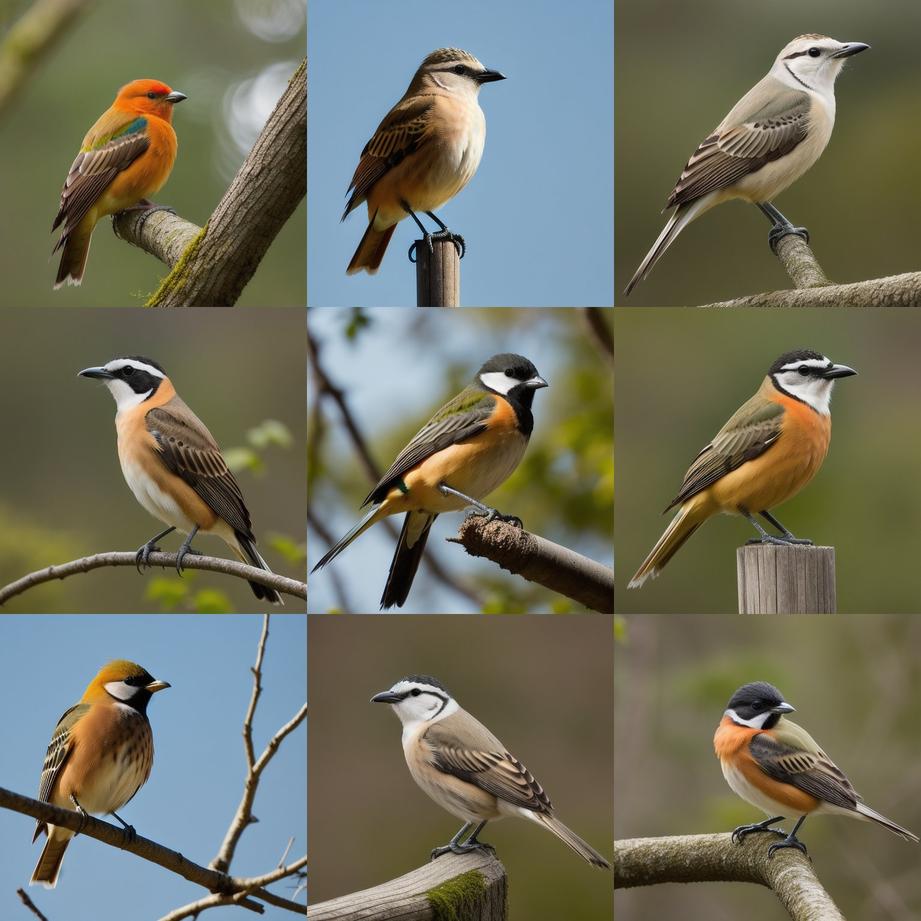} &
\includegraphics[width=0.19\linewidth]{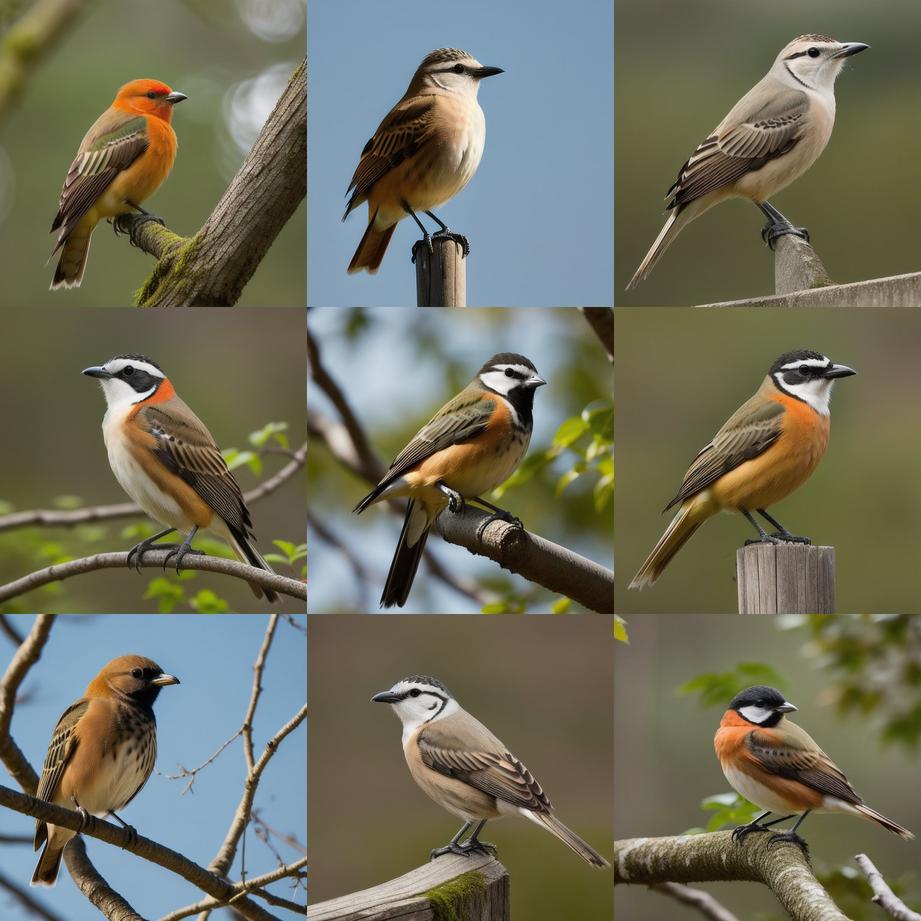} &
\includegraphics[width=0.19\linewidth]{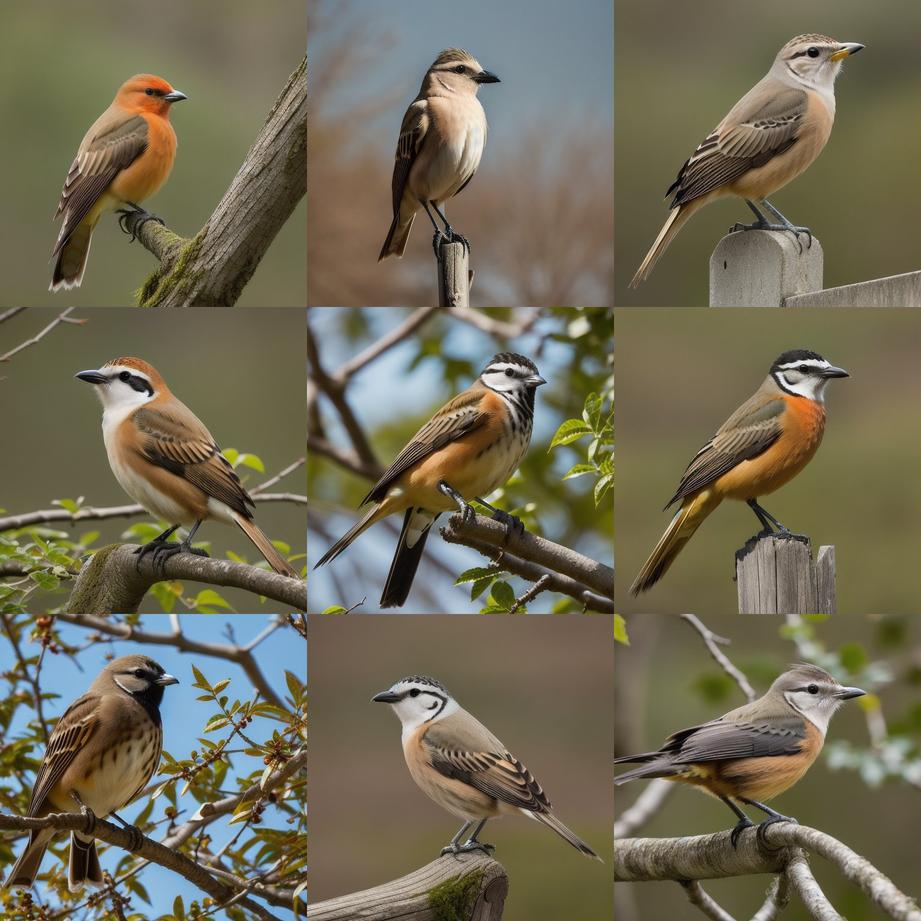} &
\includegraphics[width=0.19\linewidth]{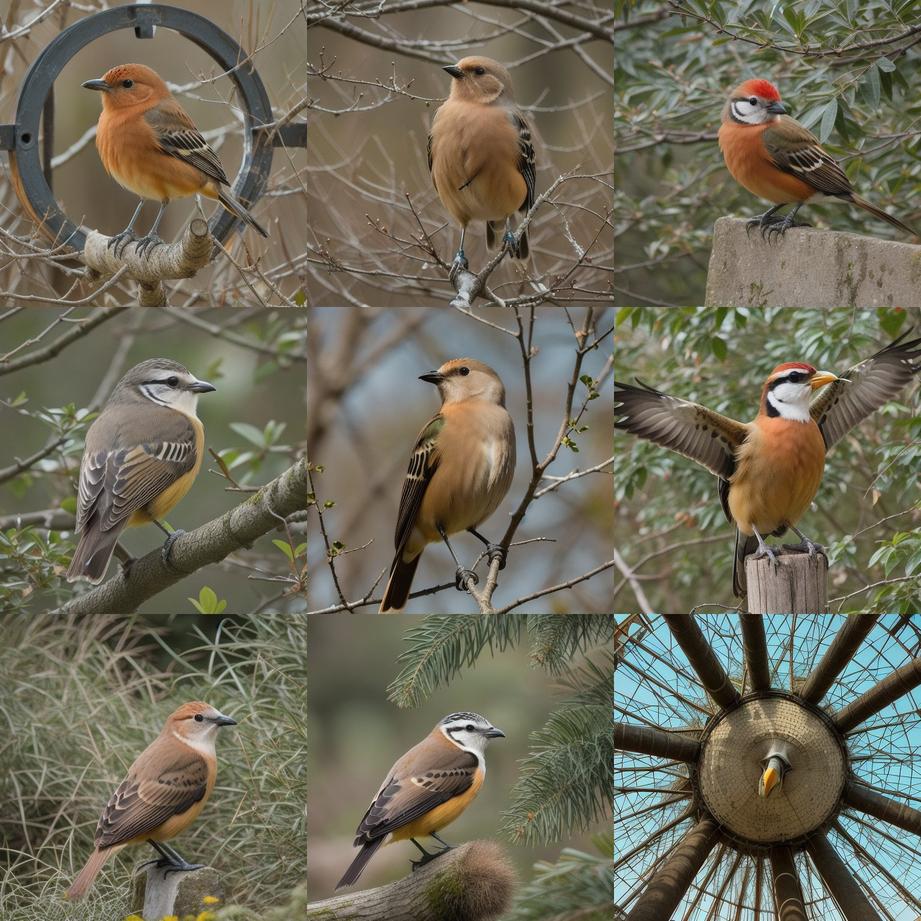}
\\[-2pt]

\raisebox{1.7\height}{\rotatebox[origin=c]{90}{\textbf{Astronaut}}} &
\includegraphics[width=0.19\linewidth]{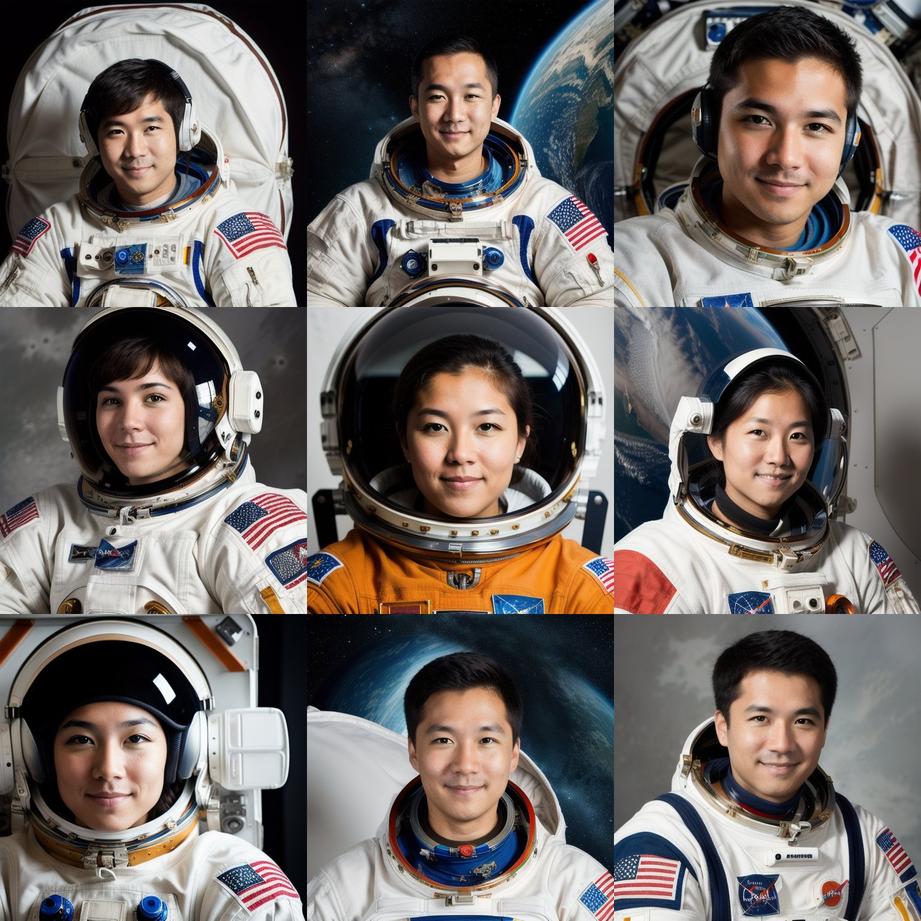} &
\includegraphics[width=0.19\linewidth]{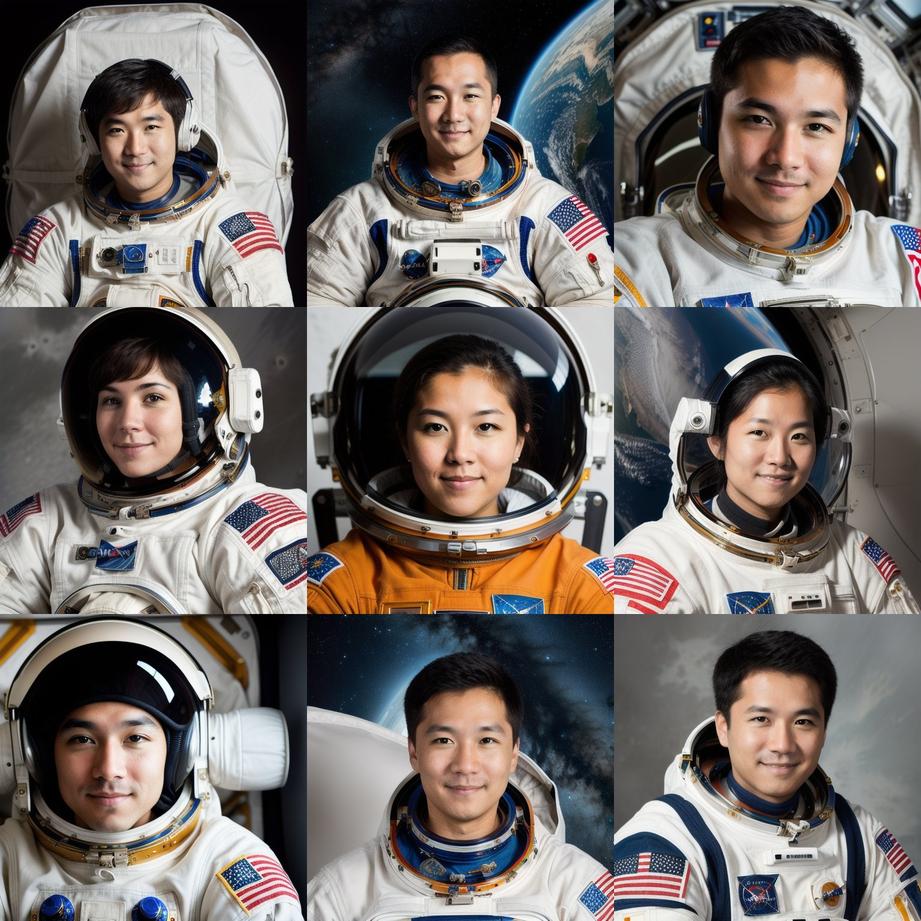} &
\includegraphics[width=0.19\linewidth]{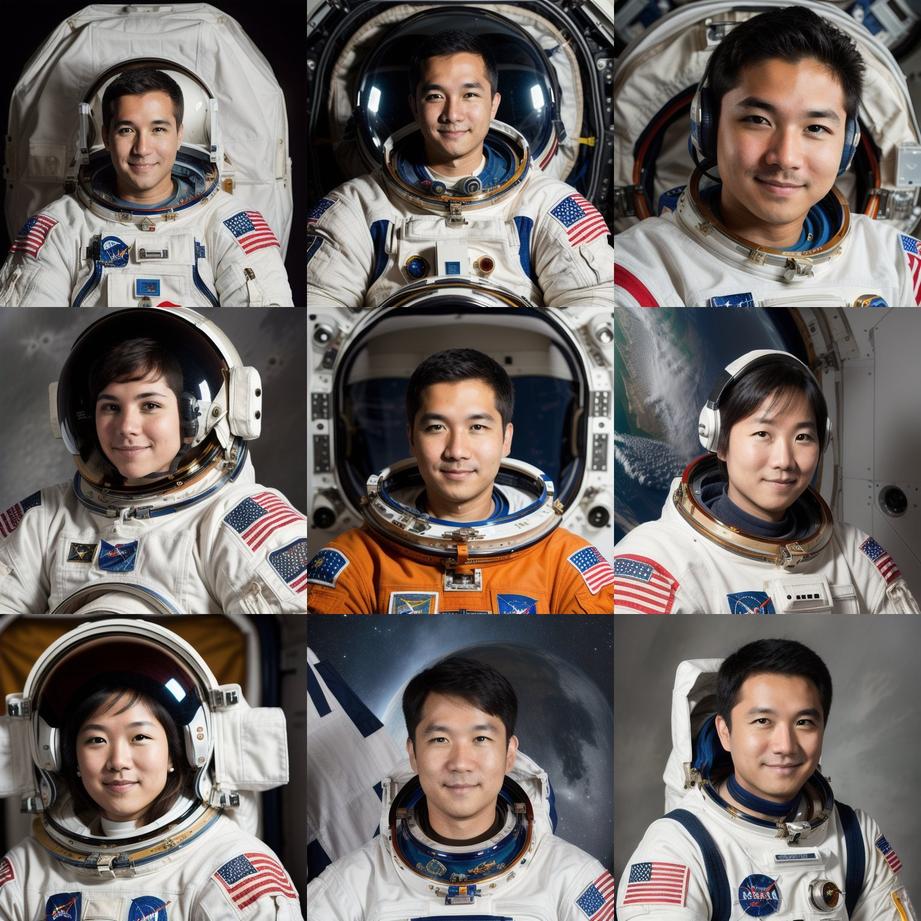} &
\includegraphics[width=0.19\linewidth]{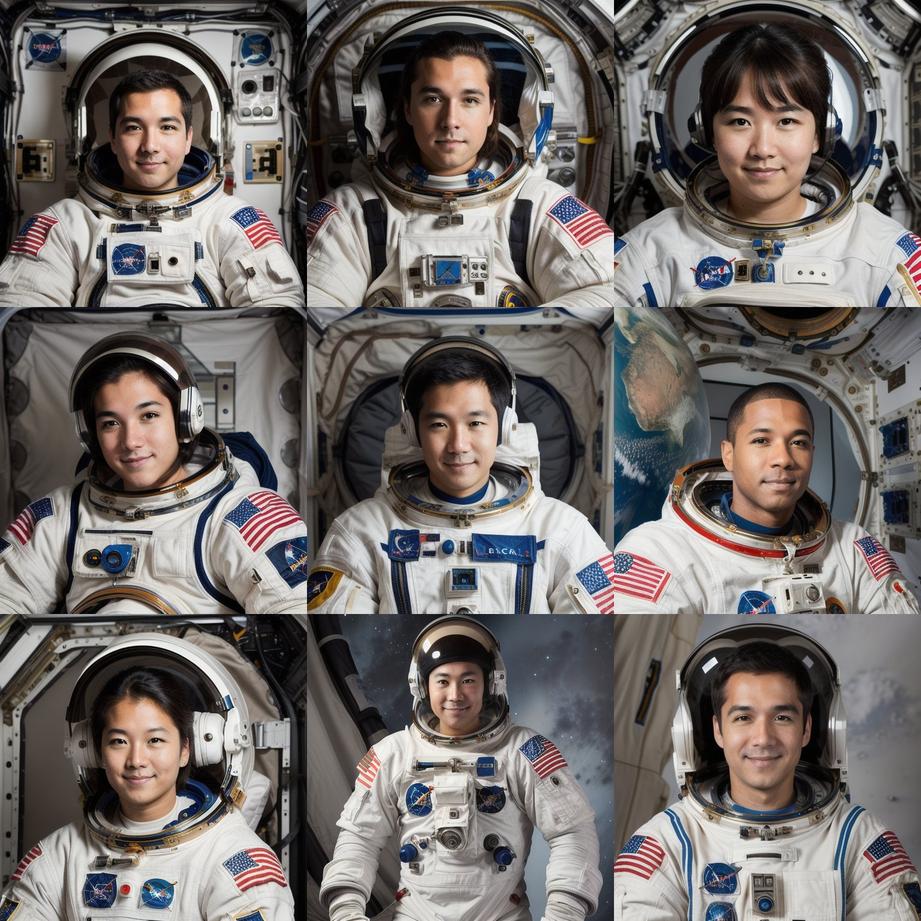} &
\includegraphics[width=0.19\linewidth]{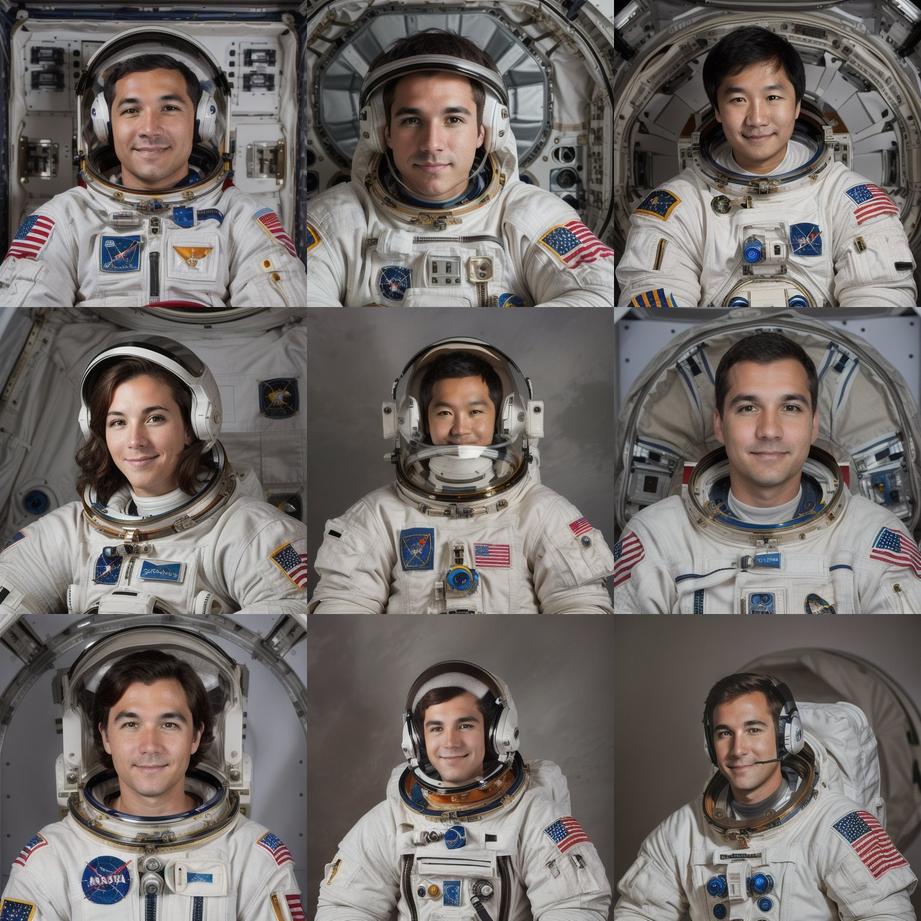}
\\[-2pt]

\raisebox{3.6\height}{\rotatebox[origin=c]{90}{\textbf{Car}}} &
\includegraphics[width=0.19\linewidth]{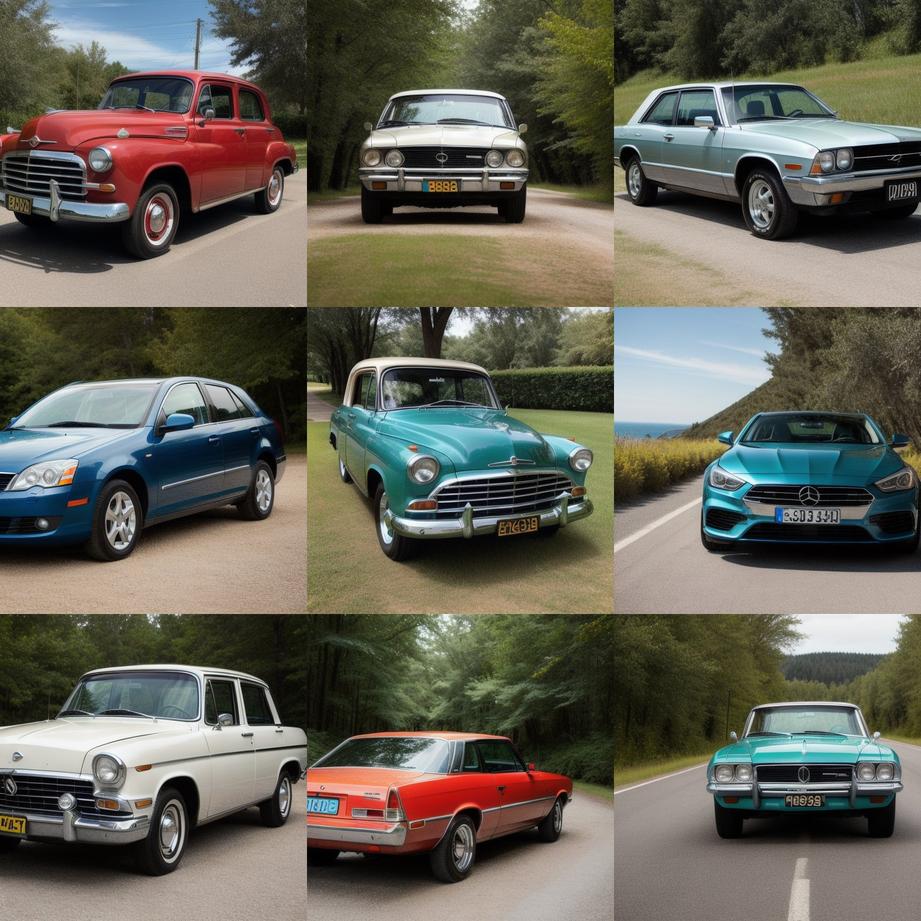} &
\includegraphics[width=0.19\linewidth]{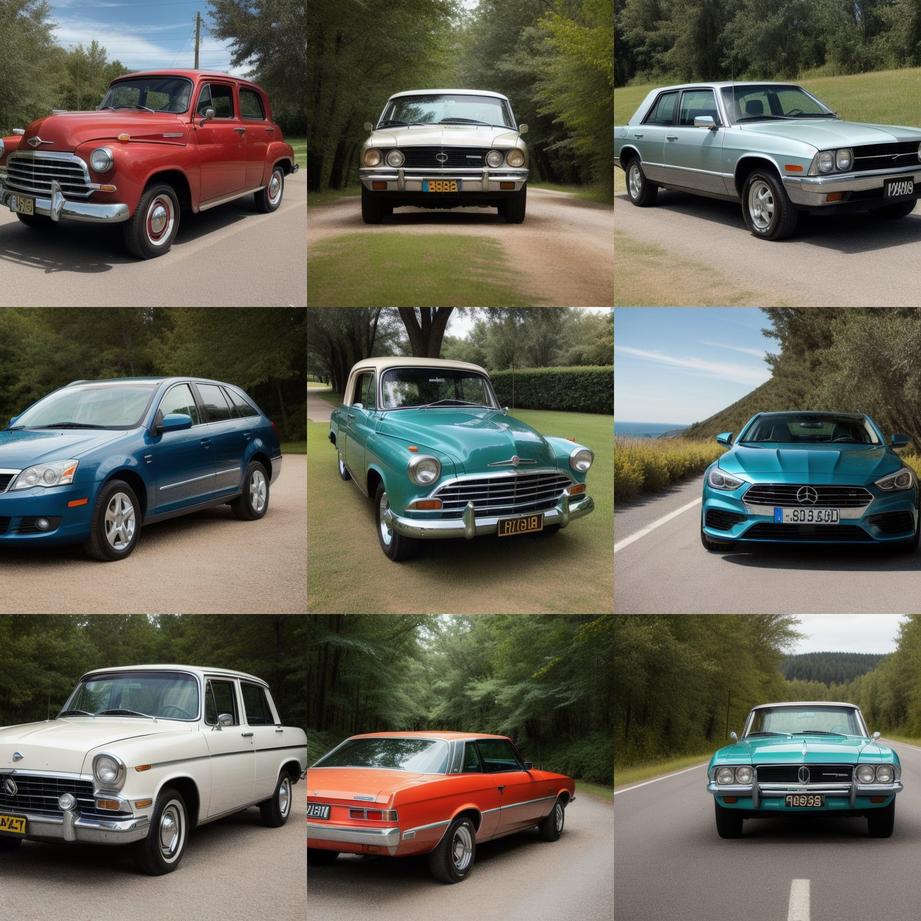} &
\includegraphics[width=0.19\linewidth]{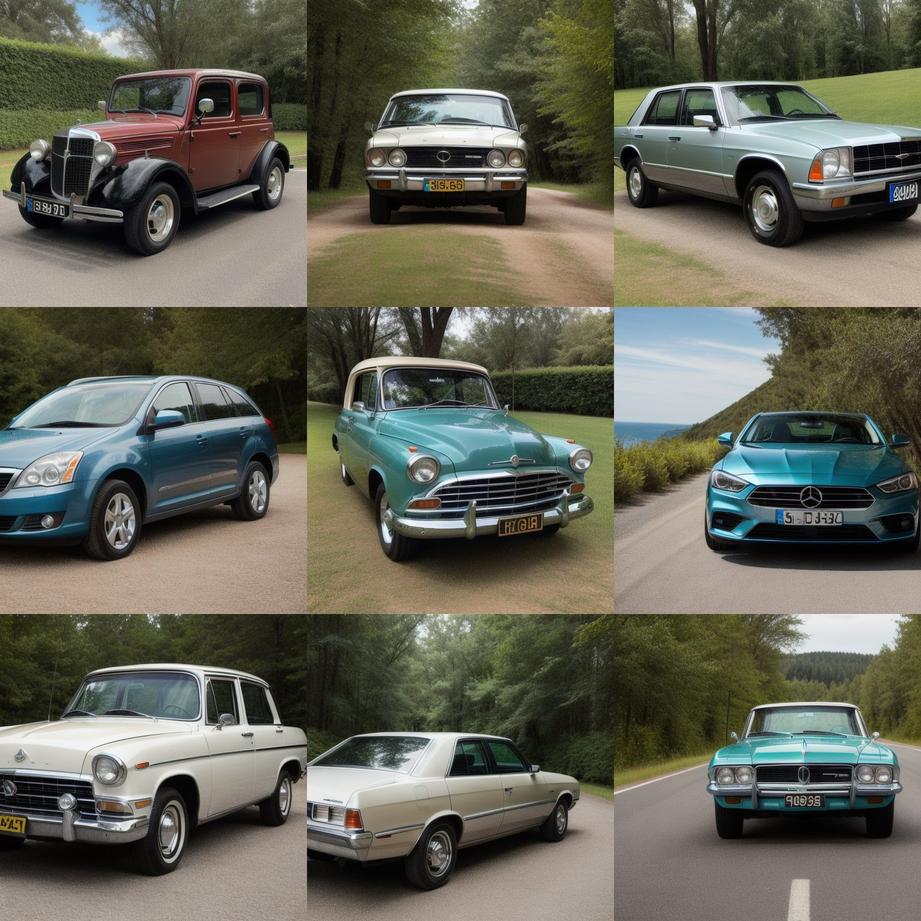} &
\includegraphics[width=0.19\linewidth]{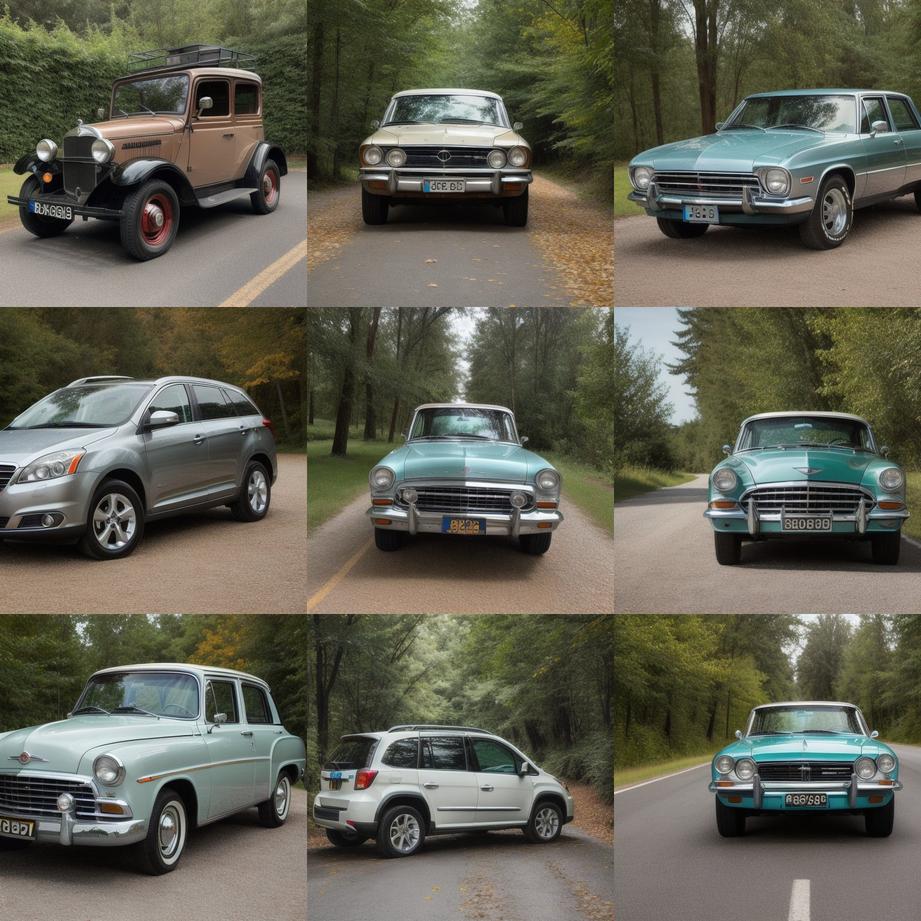} &
\includegraphics[width=0.19\linewidth]{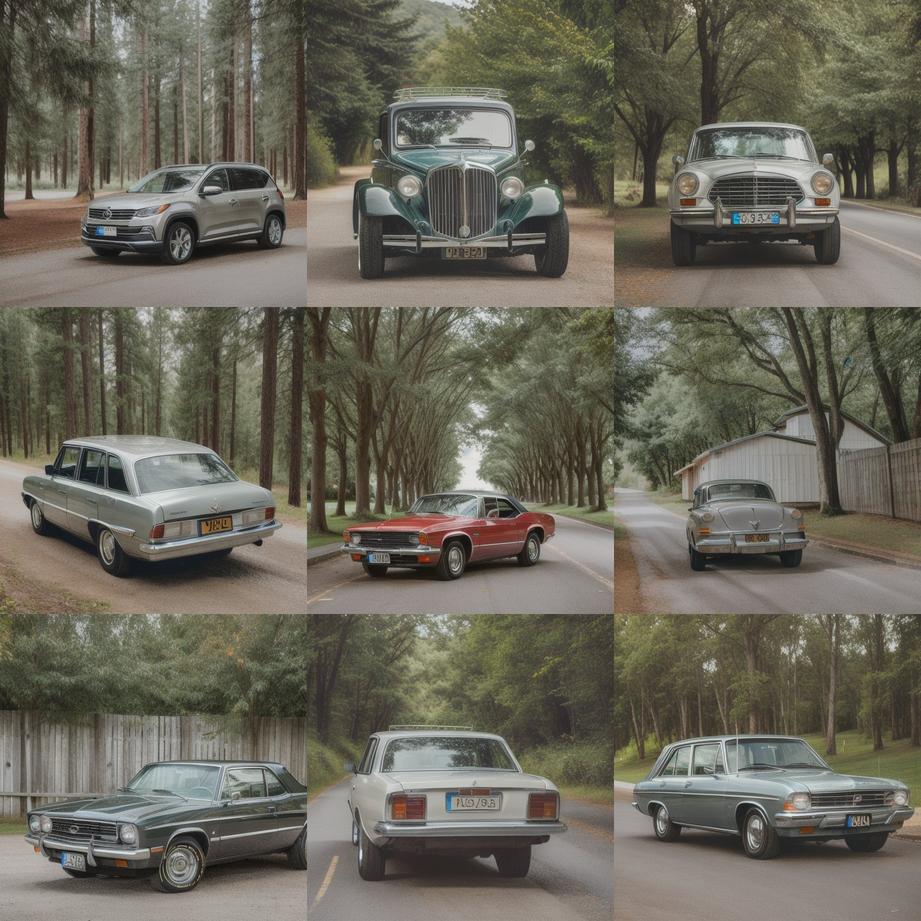}
\\[-2pt]

\raisebox{1.9\height}{\rotatebox[origin=c]{90}{\textbf{Freedom}}} &
\includegraphics[width=0.19\linewidth]{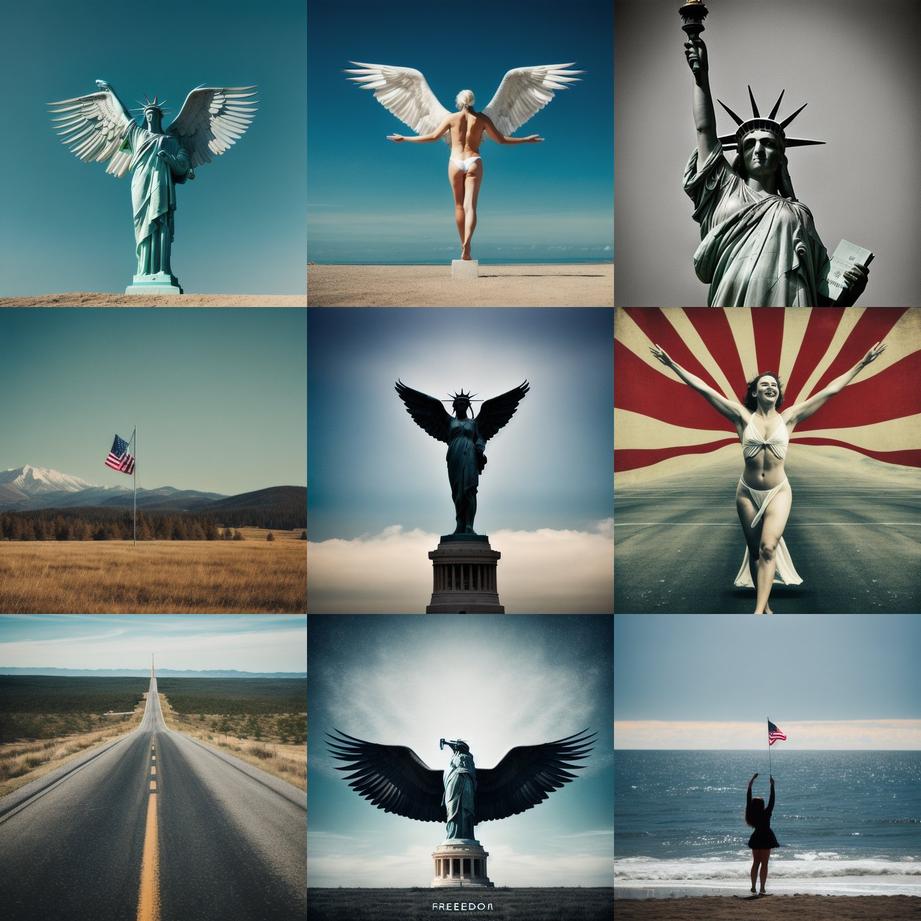} &
\includegraphics[width=0.19\linewidth]{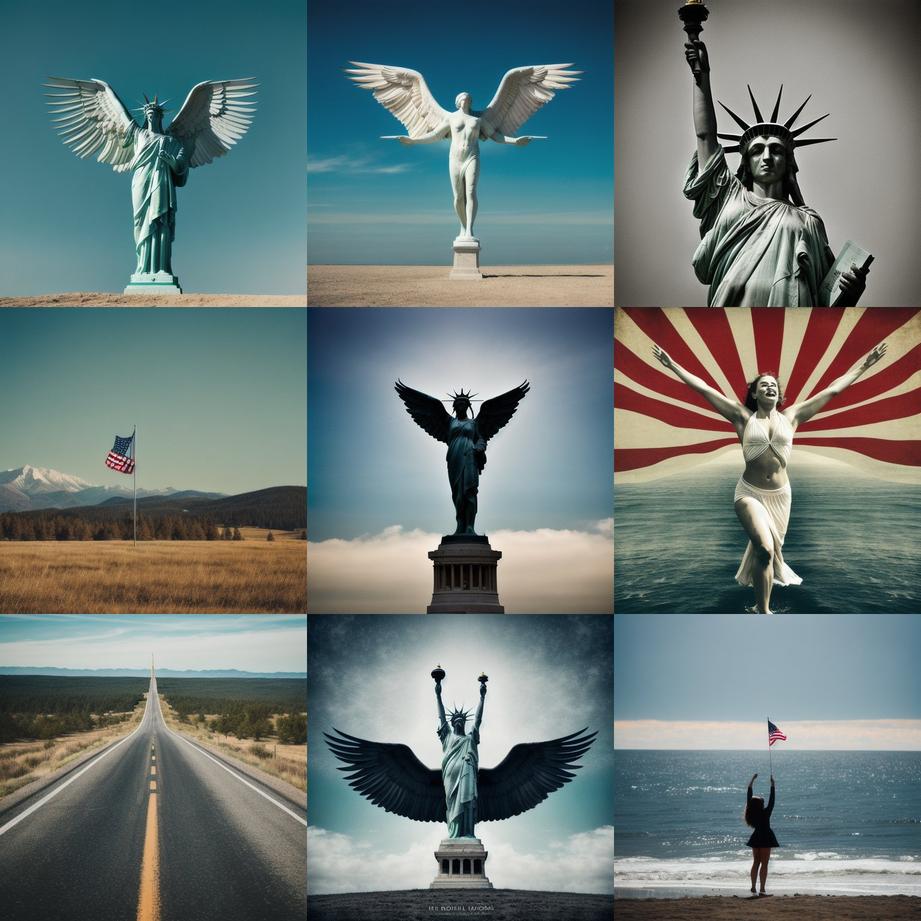} &
\includegraphics[width=0.19\linewidth]{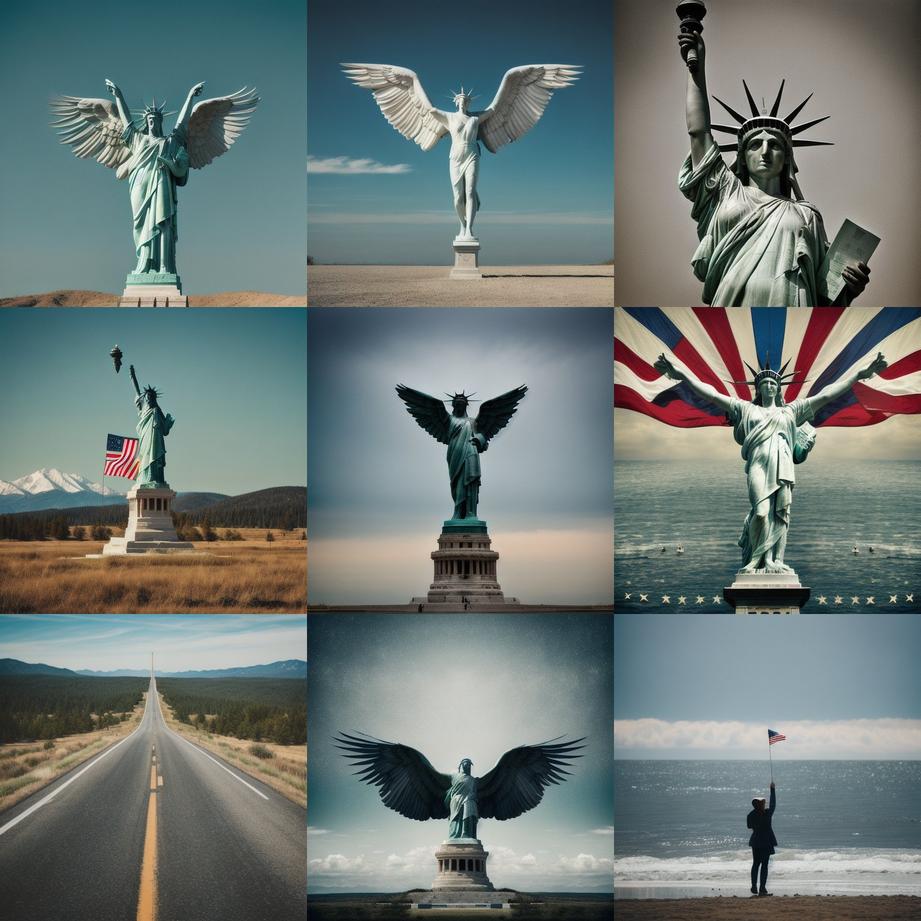} &
\includegraphics[width=0.19\linewidth]{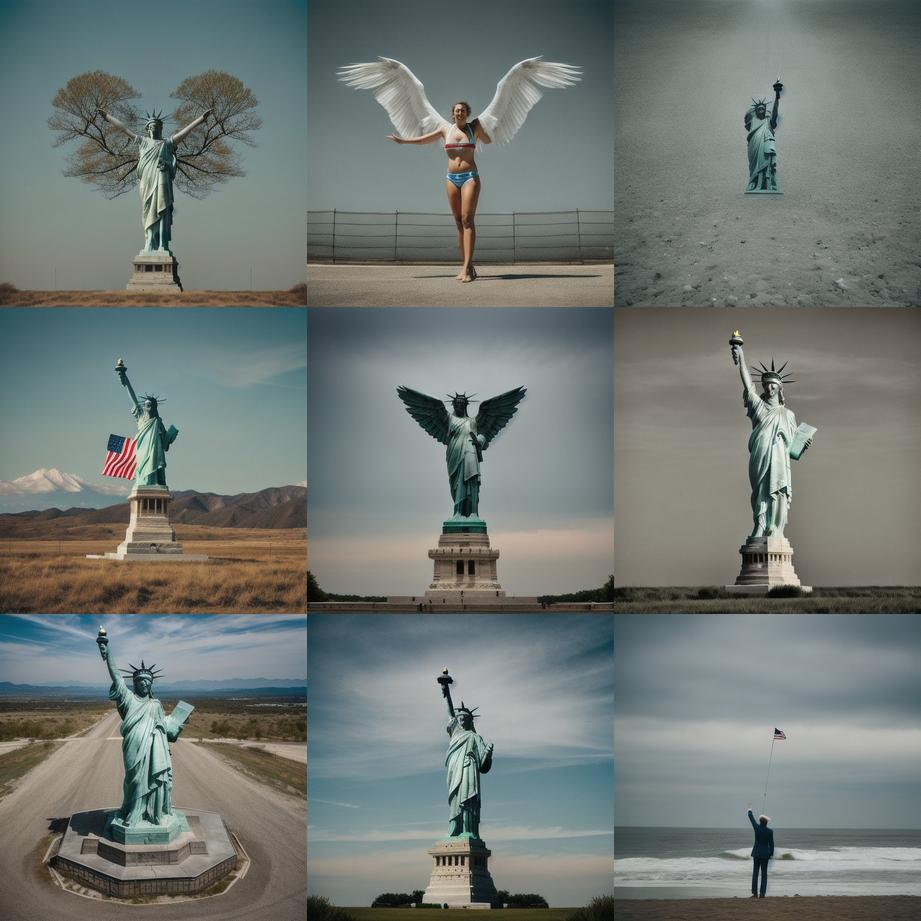} &
\includegraphics[width=0.19\linewidth]{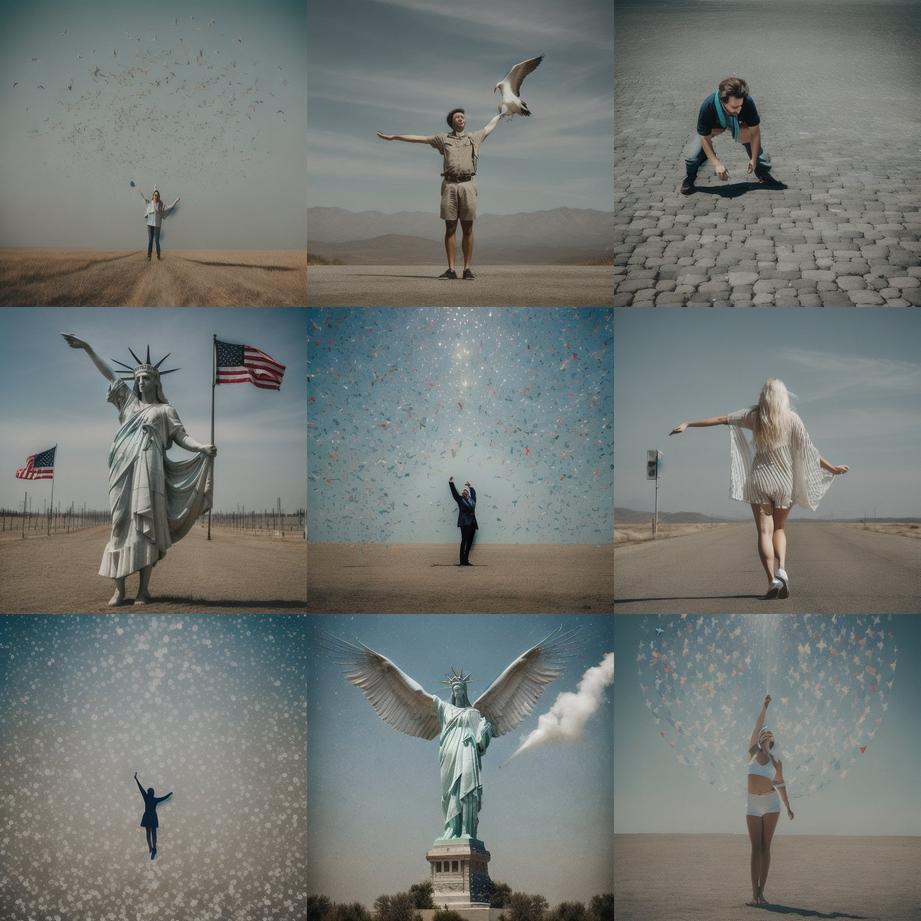}
\\

\end{tabular}

\caption{\textbf{Hyperparameter tuning of MGD$^3$ for fair comparison.}
We sweep the mode-guidance weight $\lambda \in \{0.01, 0.02, 0.05, 0.1, 0.2\}$ across four concepts: \textit{bird}, \textit{astronaut}, \textit{car}, and \textit{freedom}. Larger $\lambda$ increases attraction toward the estimated mode, improving structure but sometimes introducing over-sharpening or hallucinations.}
\label{fig:mgd3_sweep}
\end{figure*}

For a fair comparison, we sweep the guidance weight $\lambda \in \{0.01, 0.02, 0.05, 0.1, 0.2\}$ using 10 guided steps, as recommended in the paper; the main-paper setting uses $\lambda = 0.1$. Results are shown in Figure~\ref{fig:mgd3_sweep}.

\boldsubsection{Comparison.}
Figure~\ref{fig:tradeoff_supplement_plots} offers a more comprehensive comparison by sweeping hyperparameters, yielding curves that show how each baseline's performance varies across the tradeoff space. Our method outperforms both D$^4$M and MGD$^3$ in terms of consistency. While these baselines approach our performance in representativeness at certain hyperparameter settings, those settings result in significantly reduced consistency, demonstrating that our method achieves a superior overall tradeoff.

\section{Additional Results}
\subsection{Quantitative Scores}
We report a detailed breakdown by concept category of the scores from Table~\ref{tab:combined_metrics}. We report representativeness scores in Table~\ref{tab:representativeness_category} and consistency scores in Table~\ref{tab:consistency_category}.
As shown in the tables, our method surpasses the baselines across all categories.

\boldsubsection{DiT quantitative scores.} We evaluate our method on 32 random classes from ImageNet with 1,000 samples per class. Since MGD$^3$ originally supports the DiT architecture, we use it as our baseline and adopt the default hyperparameters from their paper~\cite{chan2025mgd}.  We report the results in Table~\ref{tab:dit}, showing that our method consistently outperforms MGD$^3$ across all metrics.
\begin{table}[htbp]
    \caption{Quantitative comparison on DiT. (CLIP / LPIPS / DreamSim)}
    \centering
    \scriptsize
    \setlength{\tabcolsep}{3pt}
    \renewcommand{\arraystretch}{0.95}
    \begin{tabular}{l|c|c|c}
        \toprule
        Method & Representativeness (↓) & Consistency (↓)& ImageReward (↑) \\
        \midrule
        MGD$^3$ & 0.165 / 0.678 / 0.364 & 0.147 / 0.610 / 0.317 & -0.8151 \\
        DMA (Ours) & 0.151 / 0.626 / 0.318 & 0.050 / 0.192 / 0.093 & -0.7085 \\
        \bottomrule
    \end{tabular}
    
    \label{tab:dit}
\end{table}

\subsection{Qualitative Results}
\begin{figure*}[t]
    \centering
    \renewcommand{\arraystretch}{1.2}
    \setlength{\tabcolsep}{1pt}
    \footnotesize
    \begin{tabular}{c cccc}
        & Asian & Caucasian & African & Middle Easterner \\
        
        \raisebox{2.9\height}{\rotatebox[origin=c]{90}{\textbf{Person}}} &
        \includegraphics[width=0.18\linewidth]{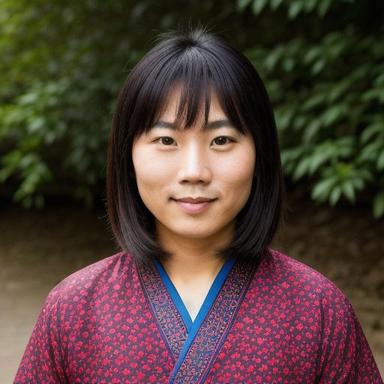} &
        \includegraphics[width=0.18\linewidth]{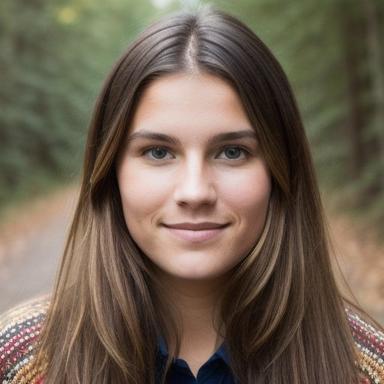} &
        \includegraphics[width=0.18\linewidth]{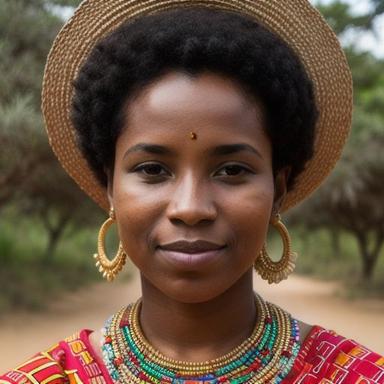} &
        \includegraphics[width=0.18\linewidth]{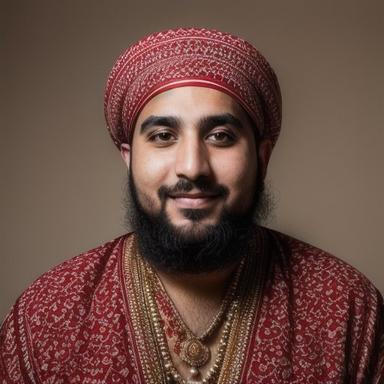}\\

        \raisebox{3.5\height}{\rotatebox[origin=c]{90}{\textbf{House}}} &
        \includegraphics[width=0.18\linewidth]{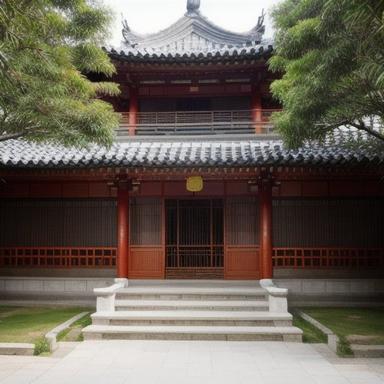} &
        \includegraphics[width=0.18\linewidth]{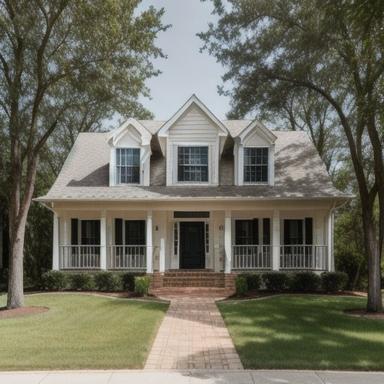} &
        \includegraphics[width=0.18\linewidth]{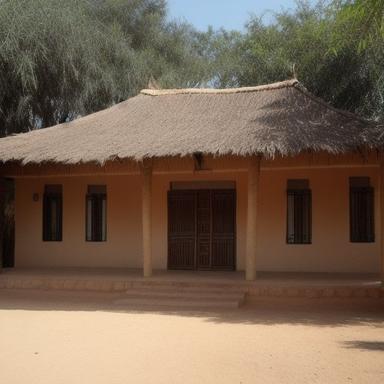} &
        \includegraphics[width=0.18\linewidth]{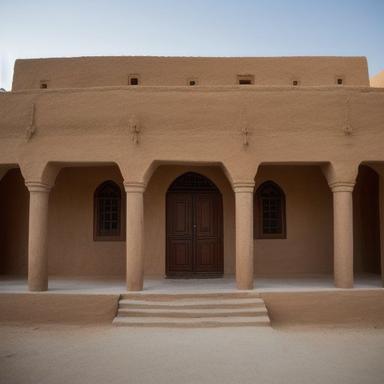}\\

        \raisebox{3.5\height}{\rotatebox[origin=c]{90}{\textbf{Cloth}}} &
        \includegraphics[width=0.18\linewidth]{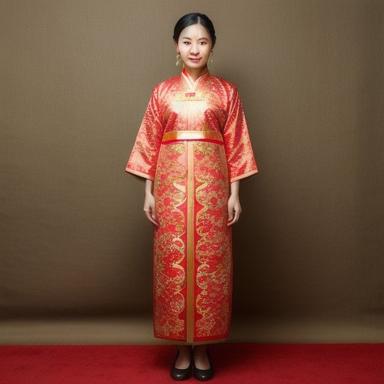} &
        \includegraphics[width=0.18\linewidth]{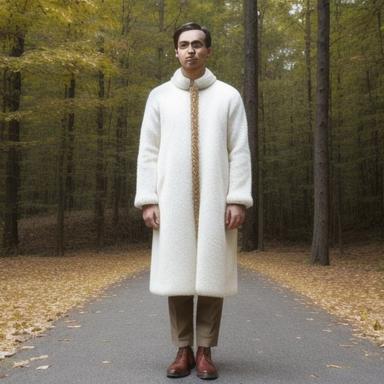} &
        \includegraphics[width=0.18\linewidth]{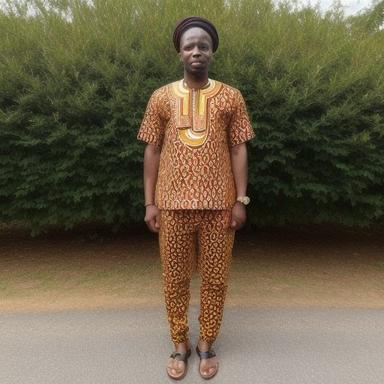} &
        \includegraphics[width=0.18\linewidth]{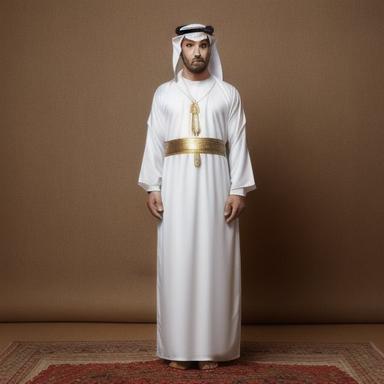}\\
    \end{tabular}
    \vspace{0pt}
    \caption{\textbf{Average images of \textit{person}, \textit{house}, and \textit{cloth} from different ethnicities.}}
    \label{fig:nationality}
\end{figure*}

\begin{figure*}[t]
    \centering
    \renewcommand{\arraystretch}{1.2}
    \setlength{\tabcolsep}{1pt}
    \footnotesize
    \begin{tabular}{cccccc}
        good person & 
        bad person & 
        rich person & 
        poor person & 
        good-looking person &
        bad-looking person\\
        \includegraphics[width=0.14\linewidth]{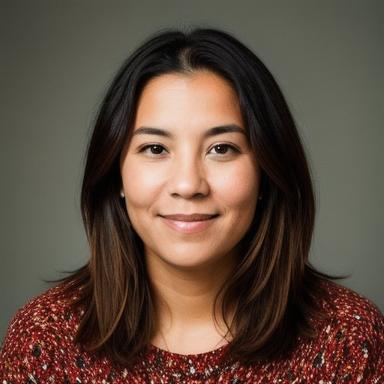} &
        \includegraphics[width=0.14\linewidth]{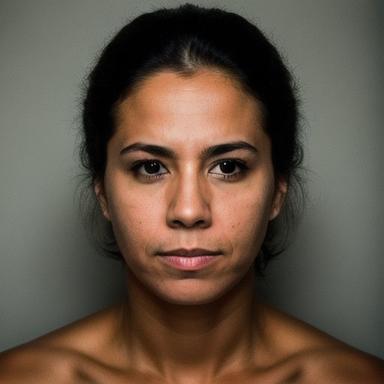} &
        \includegraphics[width=0.14\linewidth]{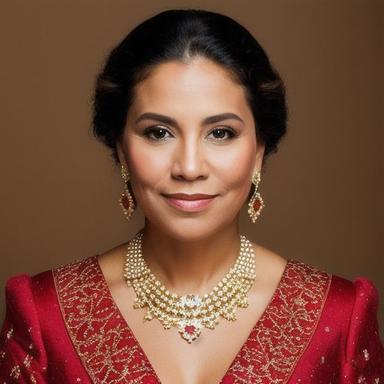} &
        \includegraphics[width=0.14\linewidth]{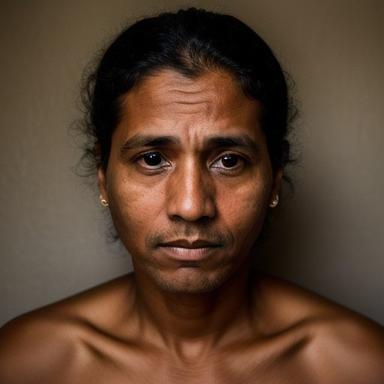} &
        \includegraphics[width=0.14\linewidth]{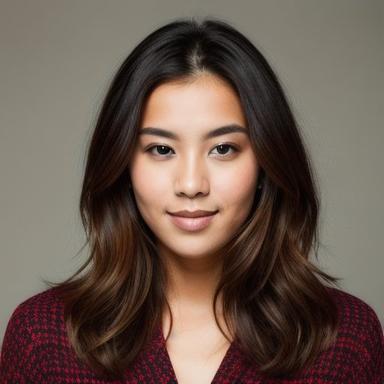} &
        \includegraphics[width=0.14\linewidth]{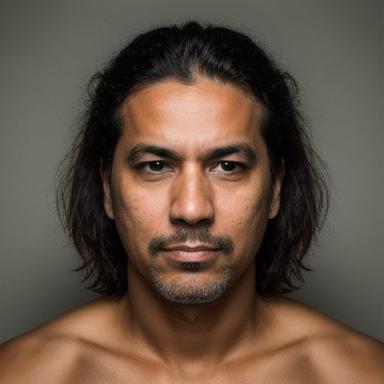}\\

        brave person & 
        curious person & 
        kind person & 
        mysterious person & 
        smart person &
        trustworthy person\\
        \includegraphics[width=0.14\linewidth]{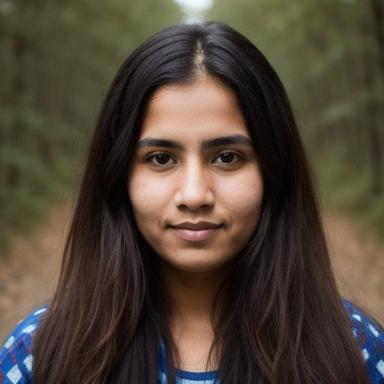} &
        \includegraphics[width=0.14\linewidth]{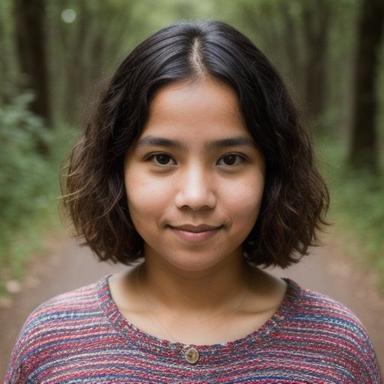} &
        \includegraphics[width=0.14\linewidth]{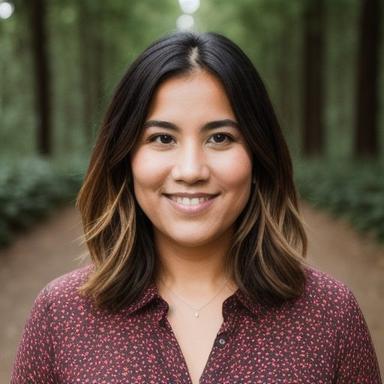} &
        \includegraphics[width=0.14\linewidth]{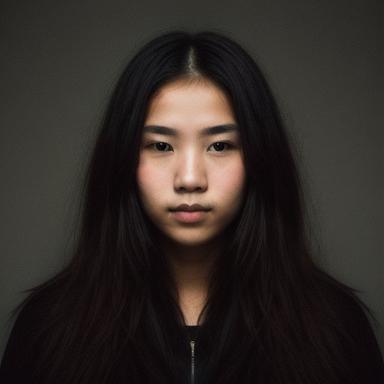} &
        \includegraphics[width=0.14\linewidth]{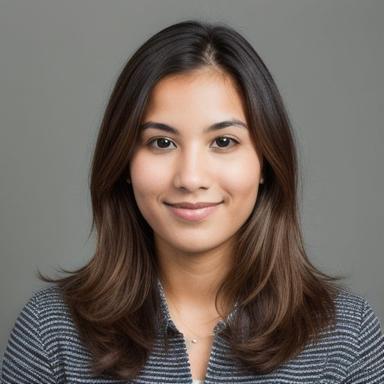} &
        \includegraphics[width=0.14\linewidth]{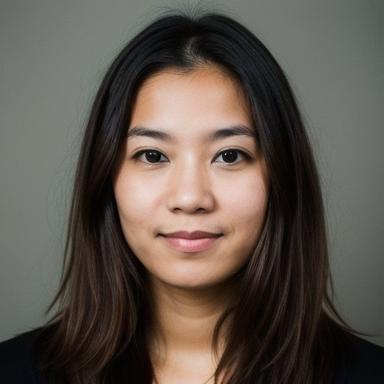}\\
    \end{tabular}
    \vspace{0pt}
    \caption{\textbf{Average images of a \textit{person} with different attributes.}}
    \label{fig:person_adj}
\end{figure*}

\begin{figure*}[t]
    \centering
    \renewcommand{\arraystretch}{1.2}
    \setlength{\tabcolsep}{1pt}
    \footnotesize
    \begin{tabular}{cccc}
        cheap car & 
        luxury car & 
        crowded market & 
        quiet market\\
        \includegraphics[width=0.14\linewidth]{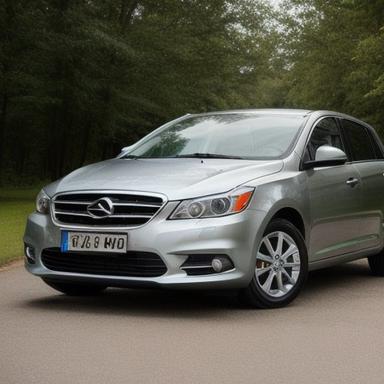} &
        \includegraphics[width=0.14\linewidth]{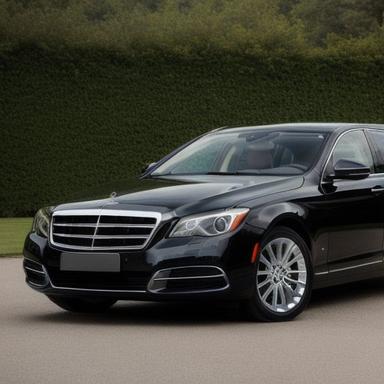} &
        \includegraphics[width=0.14\linewidth]{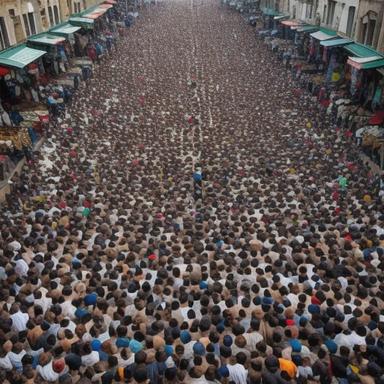} &
        \includegraphics[width=0.14\linewidth]{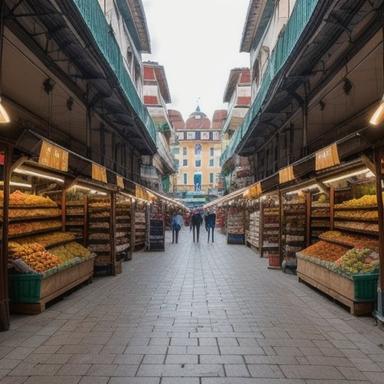} \\

        fast dog & 
        slow dog & 
        new house & 
        old house\\
        \includegraphics[width=0.14\linewidth]{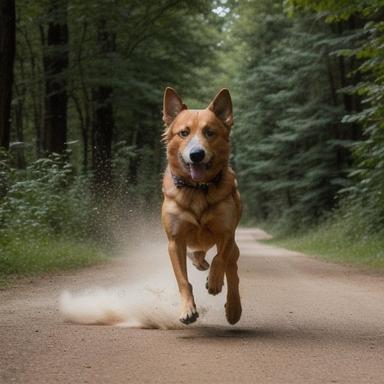} &
        \includegraphics[width=0.14\linewidth]{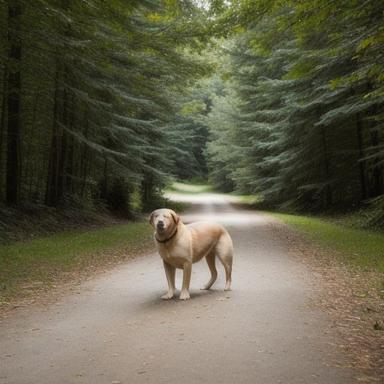} &
        \includegraphics[width=0.14\linewidth]{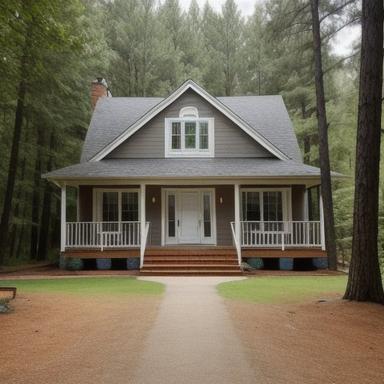} &
        \includegraphics[width=0.14\linewidth]{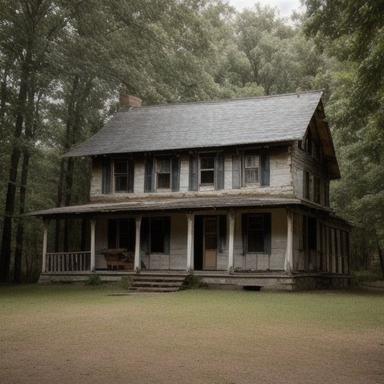} \\

    \end{tabular}
    \vspace{0pt}
    \caption{\textbf{Average images of additional concepts with two different attributes.}}
    \label{fig:others_adj}
\end{figure*}

\begin{figure*}[t]
    \centering
    \renewcommand{\arraystretch}{1.2}
    \setlength{\tabcolsep}{1pt}
    \footnotesize
    \begin{tabular}{cccccc}
        alien & 
        kungfu master & 
        wizard & 
        CEO & 
        fear &
        dance\\
        \includegraphics[width=0.14\linewidth]{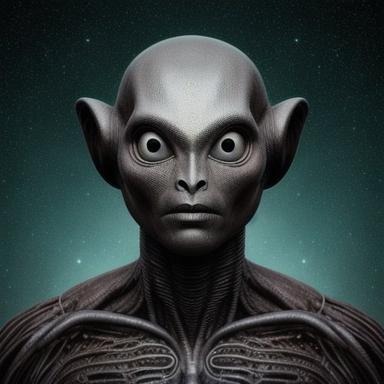} &
        \includegraphics[width=0.14\linewidth]{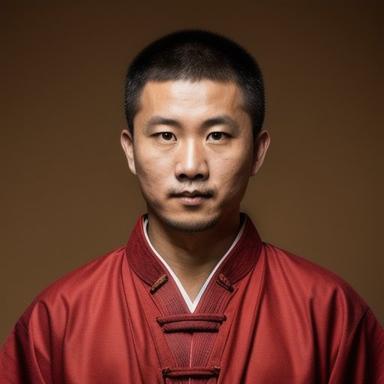} &
        \includegraphics[width=0.14\linewidth]{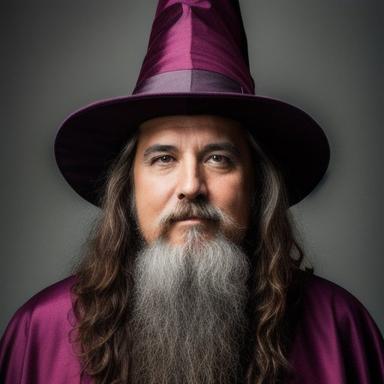} &
        \includegraphics[width=0.14\linewidth]{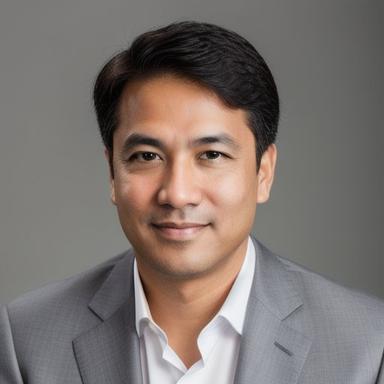} &
        \includegraphics[width=0.14\linewidth]{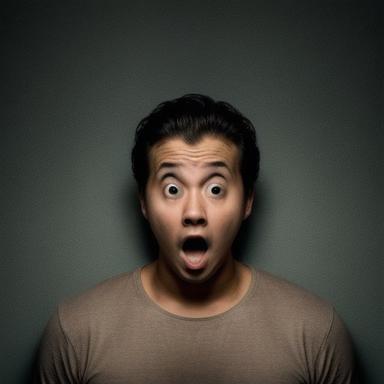} &
        \includegraphics[width=0.14\linewidth]{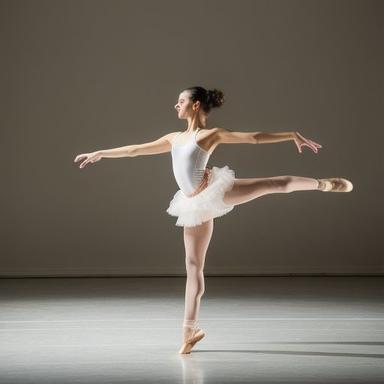}\\

        breakfast & 
        dinner & 
        aurora & 
        ferris wheel & 
        childhood &
        destiny\\
        \includegraphics[width=0.14\linewidth]{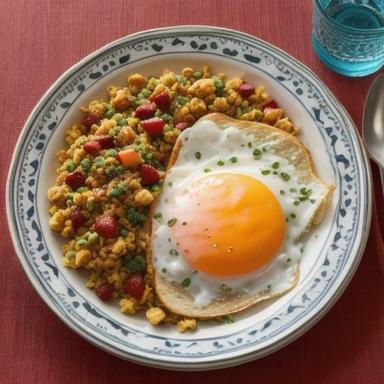} &
        \includegraphics[width=0.14\linewidth]{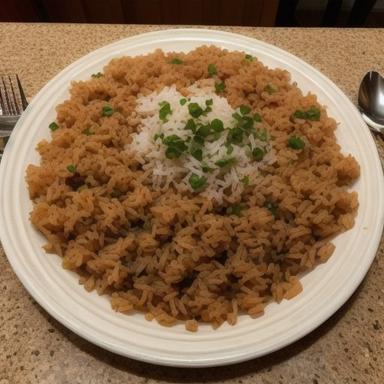} &
        \includegraphics[width=0.14\linewidth]{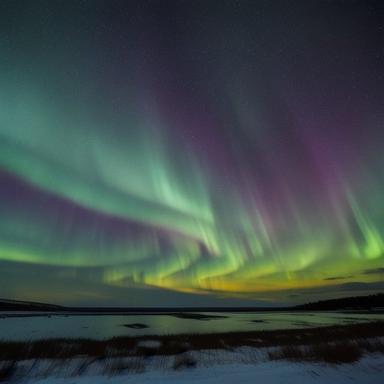} &
        \includegraphics[width=0.14\linewidth]{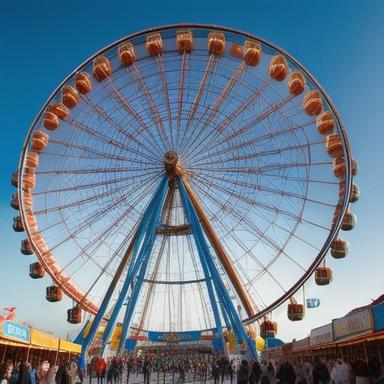} &
        \includegraphics[width=0.14\linewidth]{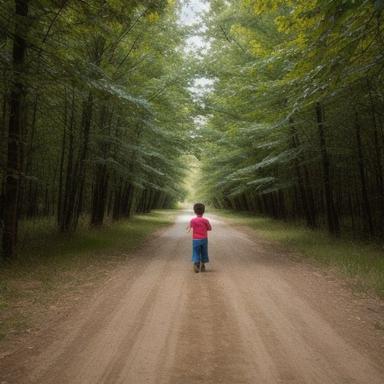} &
        \includegraphics[width=0.14\linewidth]{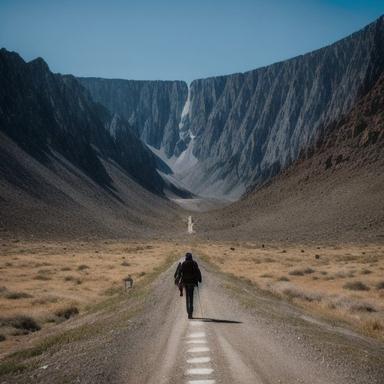}\\

        infinity & 
        mysterious & 
        peace & 
        strength & 
        wabi-sabi &
        war\\
        \includegraphics[width=0.14\linewidth]{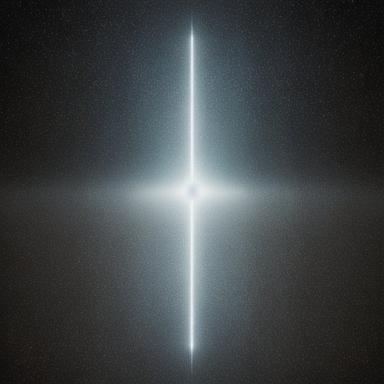} &
        \includegraphics[width=0.14\linewidth]{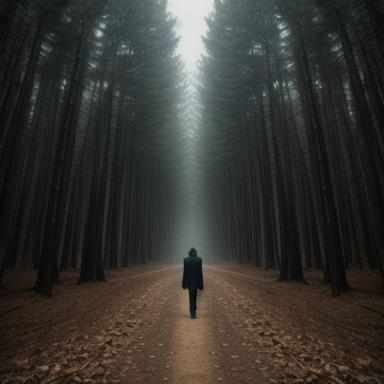} &
        \includegraphics[width=0.14\linewidth]{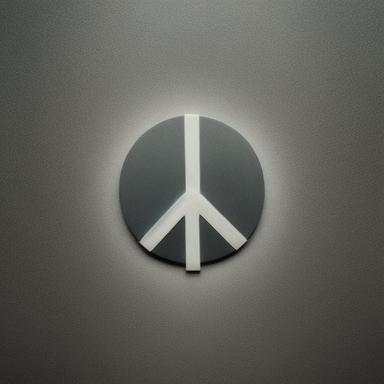} &
        \includegraphics[width=0.14\linewidth]{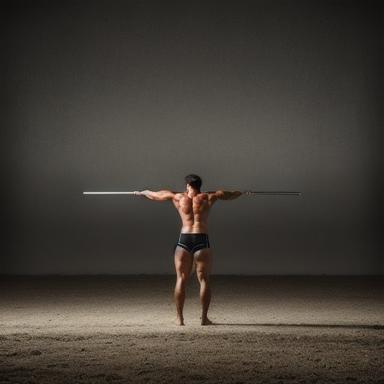} &
        \includegraphics[width=0.14\linewidth]{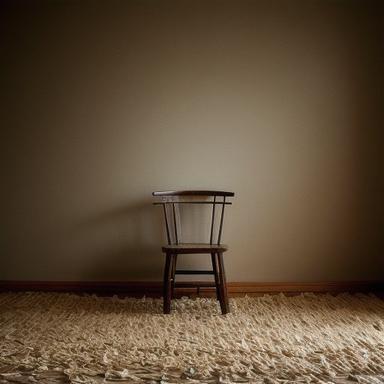} &
        \includegraphics[width=0.14\linewidth]{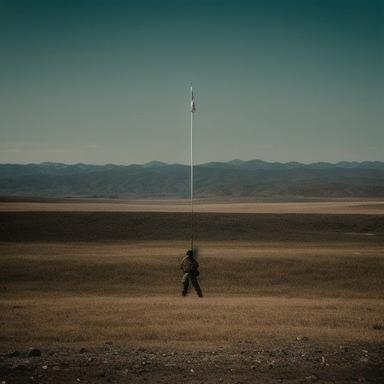}\\
    \end{tabular}
    \vspace{0pt}
    \caption{\textbf{Average images of various concepts.}}
    \label{fig:other_concept}
\end{figure*}

\boldsubsection{Average Images.} Additional results for average images of more concepts are shown in Figure~\ref{fig:nationality}-\ref{fig:other_concept}.

\boldsubsection{Mode Discovery.} Additional results for mode discovery using DMA with LoRA are shown in Figure~\ref{fig:cluster_bicycle}-\ref{fig:cluster_person_pet}.
\def\conceptlist{
    bicycle/Bicycle/color,
    cake/Cake/type,
    jellyfish/Jellyfish/color,
    toy/Toy/shape,
    nurse/Nurse/ethnicity,
    place/Place/environment,
    object/Object/shape,
    building/Building/architecture,
    person_instrument/Person with musical instrument/instrument,
    person_pet/Person with pet/pet
}

\foreach \pair in \conceptlist {%
    \StrBefore{\pair}{/}[\concept]
    \StrBehind{\pair}{/}[\restA]
    \StrBefore{\restA}{/}[\captiontext]
    \StrBehind{\restA}{/}[\grounding]

    \begin{figure*}[t]
    \centering

    \setlength{\tabcolsep}{5pt}
    \begin{tabular}{c|c|c}
        \begin{minipage}{0.23\linewidth}
            \centering
            \includegraphics[width=\linewidth]{images/clustering_fig_supplement/\concept/sample_0.jpg}\\
            \small \textbf{Cluster 1}
        \end{minipage}
        &
        \begin{minipage}{0.23\linewidth}
            \centering
            \includegraphics[width=\linewidth]{images/clustering_fig_supplement/\concept/sample_2.jpg}\\
            \small \textbf{Cluster 2}
        \end{minipage}
        &
        \begin{minipage}{0.23\linewidth}
            \centering
            \includegraphics[width=\linewidth]{images/clustering_fig_supplement/\concept/sample_3.jpg}\\
            \small \textbf{Cluster 3}
        \end{minipage}
        \\[6.5em]

        \begin{minipage}{0.23\linewidth}
            \centering
            \includegraphics[width=0.95\linewidth]{images/clustering_fig_supplement/\concept/lora_0.jpg}\\[0.3em]
            \small \textbf{Average}
        \end{minipage}
        &
        \begin{minipage}{0.23\linewidth}
            \centering
            \includegraphics[width=0.95\linewidth]{images/clustering_fig_supplement/\concept/lora_2.jpg}\\[0.3em]
            \small \textbf{Average}
        \end{minipage}
        &
        \begin{minipage}{0.23\linewidth}
            \centering
            \includegraphics[width=0.95\linewidth]{images/clustering_fig_supplement/\concept/lora_3.jpg}\\[0.3em]
            \small \textbf{Average}
        \end{minipage}
    \end{tabular}

    \caption{\textbf{Cluster averages for the \textit{\captiontext} concept grounded by \textit{\grounding}.}
            Top: Diffusion samples grouped into clusters according to the grounding attribute \textit{\grounding}.
            Bottom: Cluster averages computed by DMA with LoRA.
            These averages highlight how our method captures the dominant visual characteristics within each cluster.}
    \label{fig:cluster_\concept}
    \end{figure*}
}

\begin{table*}[t!]
\centering
\small
\setlength{\tabcolsep}{2pt}
\caption{
\textbf{Consistency (↓)} across concept categories. 
Lower values indicate higher consistency of prototypes within each seed. 
Each category includes representative concepts: 
\textit{Animal} — Bird, Cat, Dog; 
\textit{Person} — Astronaut, Doctor, Firefighter; 
\textit{Object} — Bicycle, Car, TV Monitor; 
\textit{Abstract} — Anger, Freedom, Poverty. 
}
\begin{tabular}{lccc ccc ccc ccc}
\toprule
& \multicolumn{3}{c}{\textbf{Animal}} & \multicolumn{3}{c}{\textbf{Person}} & \multicolumn{3}{c}{\textbf{Object}} & \multicolumn{3}{c}{\textbf{Abstract}} \\
\cmidrule(lr){2-4} \cmidrule(lr){5-7} \cmidrule(lr){8-10} \cmidrule(lr){11-13}
\textbf{Method} & CLIP↓ & DreamSim↓ & LPIPS↓ & CLIP↓ & DreamSim↓ & LPIPS↓ & CLIP↓ & DreamSim↓ & LPIPS↓ & CLIP↓ & DreamSim↓ & LPIPS↓ \\
\midrule
GANgealing~\cite{peebles2022gan} & 0 & 0 & 0 & 0 & 0 & 0 & 0 & 0 & 0 & -- & -- & -- \\
Avg~VAE & 0 & 0 & 0 & 0 & 0 & 0 & 0 & 0 & 0 & 0 & 0 & 0 \\
\midrule
D$^4$M~\cite{su2024d} & 0.123 & 0.191 & 0.615 & 0.127 & 0.174 & 0.510 & 0.193 & 0.308 & 0.627 & 0.229 & 0.425 & 0.534 \\
MGD$^3$~\cite{chan2025mgd} & 0.122 & 0.224 & 0.679 & 0.131 & 0.202 & 0.598 & 0.215 & 0.342 & 0.686 & 0.251 & 0.508 & 0.610 \\
\midrule
DMA (Ours) & 0.015 & 0.021 & 0.138 & 0.022 & 0.019 & 0.118 & 0.040 & 0.039 & 0.158 & 0.047 & 0.049 & 0.101 \\
\bottomrule
\end{tabular}
\label{tab:consistency_category}
\end{table*}

\begin{table*}[t!]
\centering
\small
\setlength{\tabcolsep}{1pt}
\caption{
\textbf{Representativeness (↓)} across concept categories. 
Lower values indicate closer alignment between the prototype and the overall concept distribution.
}
\begin{tabular}{lcccccccccccc}
\toprule
& \multicolumn{3}{c}{\textbf{Animal}} & \multicolumn{3}{c}{\textbf{Person}} & \multicolumn{3}{c}{\textbf{Object}} & \multicolumn{3}{c}{\textbf{Abstract}} \\
\cmidrule(lr){2-4} \cmidrule(lr){5-7} \cmidrule(lr){8-10} \cmidrule(lr){11-13}
\textbf{Method} & CLIP↓ & DreamSim↓ & LPIPS↓ & CLIP↓ & DreamSim↓ & LPIPS↓ & CLIP↓ & DreamSim↓ & LPIPS↓ & CLIP↓ & DreamSim↓ & LPIPS↓ \\
\midrule
GANgealing~\cite{peebles2022gan} & 0.310 & 0.394 & 0.839 & 0.523 & 0.599 & 0.866 & 0.323 & 0.439 & 0.848 & -- & -- & -- \\
Avg~VAE & 0.505 & 0.867 & 0.831 & 0.504 & 0.780 & 0.796 & 0.464 & 0.832 & 0.882 & 0.420 & 0.746 & 0.711 \\
\midrule
D$^4$M~\cite{su2024d} & 0.158 & 0.320 & 0.708 & 0.144 & 0.230 & 0.613 & 0.224 & 0.379 & \textbf{0.714} & 0.260 & 0.522 & 0.652 \\
MGD$^3$~\cite{chan2025mgd} & 0.155 & 0.310 & 0.718 & \textbf{0.140} & 0.230 & 0.636 & 0.223 & 0.375 & 0.724 & 0.262 & 0.543 & 0.671 \\
\midrule
\textbf{DMA (Ours)} & \textbf{0.143} & \textbf{0.307} & \textbf{0.669} & \textbf{0.142} & \textbf{0.208} & \textbf{0.595} & \textbf{0.202} & \textbf{0.368} & \textbf{0.718} & \textbf{0.229} & \textbf{0.479} & \textbf{0.638} \\
\bottomrule
\end{tabular}
\label{tab:representativeness_category}
\end{table*}


\foreach \concept in {astronaut, firefighter, doctor, bird, cat, dog, bicycle, car, tvmonitor} {

\begin{figure*}[t]
\centering

\includegraphics[width=0.9\linewidth]{images/compare_supplement/samples_grid/\concept.jpg}\\[0.3em]
\small Generated Samples\\[1.0em]

\setlength{\tabcolsep}{2pt}
\begin{tabular}{cccc}
\includegraphics[width=0.24\linewidth]{images/compare_supplement/\concept/gangealing.jpg} &
\includegraphics[width=0.24\linewidth]{images/compare_supplement/\concept/sdedit.jpg} &
\includegraphics[width=0.24\linewidth]{images/compare_supplement/\concept/mode_guide.jpg} &
\includegraphics[width=0.24\linewidth]{images/compare_supplement/\concept/ours.jpg}
\\[0.3em]
\small GANgealing~\cite{peebles2022gan} & \small D$^4$M~\cite{su2024d} & \small MGD$^3$~\cite{chan2025mgd} & \small Ours
\end{tabular}

\caption{\textbf{Baseline comparison for the \textit{\concept} concept.}
Top: Generated samples from the diffusion model. Bottom: Average images produced by GANgealing~\cite{peebles2022gan}, D$^4$M~\cite{su2024d}, MGD$^3$~\cite{chan2025mgd}, and our method.}
\label{fig:compare_\concept}
\end{figure*}

}

\section{Potential Negative Impacts}
DMA is designed to reveal how a diffusion model internally represents a concept; any harmful biases present are inherent to the probe model and not a direct result of our method. Nonetheless, there are risks in how DMA representations are interpreted. For example, computing only a single or a few averages may marginalize minority modes. Using these averages as authoritative summaries or to stereotype groups or attributes can be misleading and harmful. To mitigate these concerns, we emphasize that DMA averages reflect only the biases of the specific probe model and should not be interpreted as ground-truth representation of any real population or concept.


\end{document}